\documentclass[10pt,twocolumn,letterpaper]{article}

\usepackage[pagenumbers]{iccv} %

\usepackage{times}
\usepackage{epsfig}
\usepackage{graphicx}
\usepackage{amsmath}
\usepackage{amssymb}
\usepackage{caption}
\usepackage{scalerel,graphicx,xparse}
\usepackage{multirow}
\usepackage{xcolor}
\usepackage{longtable}
\usepackage{adjustbox}
\usepackage{ltablex}
\usepackage{supertabular}
\usepackage{listings}
\usepackage{soul}
\usepackage{color}
\usepackage{tikzsymbols}
\usepackage{graphicx}
\usepackage{adjustbox}
\usepackage{tcolorbox} 
\usepackage{array}
\usepackage{booktabs}
\usepackage{xcolor}  
\usepackage{colortbl}
\usepackage{longtable}

\usepackage{CJKutf8}

\usepackage[english]{babel}

\usepackage{tcolorbox}

\NewDocumentCommand\emojione{}{\scalerel*{\includegraphics[page=1]{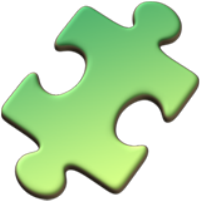}}{\textsc{\scalebox{2.5}{X}}}}

\definecolor{darkblue}{rgb}{0.70, 0.80, 1.0}
\definecolor{lightblue}{rgb}{0.80, 0.90, 1.0}
\definecolor{lightpink}{rgb}{1.0, 0.8, 0.8}
\definecolor{lightgreen}{rgb}{0.8, 1.0, 0.8}

\DeclareRobustCommand{\hlgreen}[1]{{\sethlcolor{lightgreen}\hl{#1}}}
\DeclareRobustCommand{\hlred}[1]{{\sethlcolor{lightpink}\hl{#1}}}

\definecolor{lightgray}{rgb}{0.8, 0.8, 0.8}
\DeclareRobustCommand{\hlgray}[1]{{\sethlcolor{lightgray}\hl{#1}}}

\newcommand{\easy}{\textbf{\textcolor{darkgreen}{easy~\cChangey{2}}}}
\newcommand{\medium}{\textbf{\textcolor{darkyellow}{medium}~\cChangey{0}}}
\newcommand{\hard}{\textbf{\textcolor{red}{hard}~\cChangey{-2}}}

\newcommand{\Easy}{\textbf{\textcolor{darkgreen}{Easy~\cChangey{2}}}}
\newcommand{\Medium}{\textbf{\textcolor{darkyellow}{Medium}~\cChangey{0}}}
\newcommand{\Hard}{\textbf{\textcolor{red}{Hard}~\cChangey{-2}}}

\definecolor{modelcolor}{RGB}{230, 242, 255}
\definecolor{keycolor}{RGB}{245, 235, 255}  %
\DeclareRobustCommand{\hlmodel}[1]{{\sethlcolor{modelcolor}\hl{#1}}}
\DeclareRobustCommand{\hlkey}[1]{{\sethlcolor{keycolor}\hl{#1}}}

\definecolor{darkgreen}{rgb}{0.0, 0.5, 0.0}
\definecolor{darkyellow}{rgb}{0.7, 0.5, 0.0}

\definecolor{iccvblue}{rgb}{0.21,0.49,0.74}
\usepackage[pagebackref,breaklinks,colorlinks,allcolors=iccvblue]{hyperref}

\def\paperID{2060} %
\def\confName{ICCV}
\def\confYear{2025}

\title{\emojione VGRP-Bench: \underline{V}isual \underline{G}rid \underline{R}easoning \underline{P}uzzle \underline{Bench}mark \\ for Large Vision-Language Models}

\author{
    Yufan Ren$^{1}$\footnotemark \quad
    Konstantinos Tertikas$^{2}$\quad
    Shalini Maiti$^{3,4}$ \quad
    Junlin Han$^{3,5}$ \\ 
    Tong Zhang$^1$ \quad 
    Sabine Süsstrunk$^1$ \quad 
    Filippos Kokkinos$^3$ \\
    $^1$School of Computer and Communication Sciences, EPFL \\ $^2$National and Kapodistrian University of Athens \\ $^3$ Meta GenAI \\
    $^4$ University College London \\
    $^5$ University of Oxford 
}

\begin{document}
\twocolumn[{
\begin{center}
    \maketitle
    \vspace{-4mm}  
    \captionsetup{type=figure}
    \includegraphics[width=1.0\textwidth,page=1]{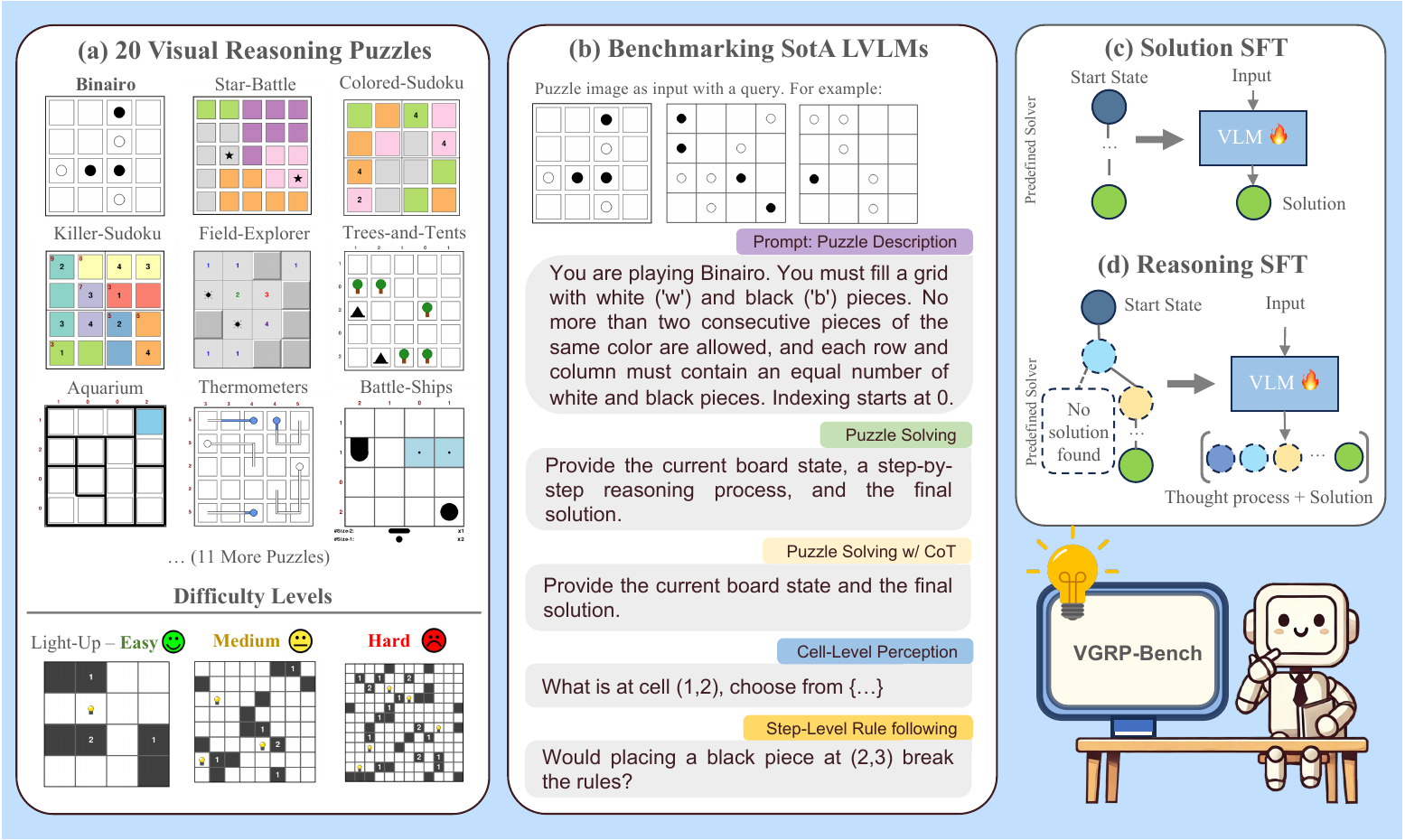} 
\captionof{figure}{\textbf{Benchmark Overview.} 
(a) We present a benchmark for Large Vision-Language Models (LVLMs) consisting of 20 diverse visual grid reasoning puzzles (see supplementary material for complete table of per-puzzle examples and descriptions).
(b) We evaluate state-of-the-art LVLMs, including closed-source models such as GPT-4o~\cite{chatgpt_4o} and Gemini~\cite{team2023gemini}, open-source models like Llama 3.2~\cite{dubey2024llama}, and recently released reasoning models such as Gemini-Thinking, on various aspects, including perception, overall puzzle-solving, and cell-level rule-following.
Additionally, to explore potential approaches for improving LVLMs’ puzzle-solving abilities, 
we examine post-training techniques, including (c) Solution Supervised Fine-Tuning (S-SFT) and (d) Reasoning Supervised Fine-Tuning (R-SFT), where we train on thought trajectories of a predefined solver.
(Best viewed on a screen when zoomed-in) 
}

    \label{fig:teaser_first_page}
\end{center}

}]

\renewcommand{\thefootnote}{$*$}
\footnotetext{Work done at Meta as an intern.}
\renewcommand{\thefootnote}{\arabic{footnote}}

\begin{abstract}
Large Vision-Language Models (LVLMs) struggle with puzzles, which require precise perception, rule comprehension, and logical reasoning. Assessing and enhancing their performance in this domain is crucial, as it reflects their ability to engage in structured reasoning — an essential skill for real-world problem-solving. 
However, existing benchmarks primarily evaluate pre-trained models without additional training or fine-tuning, often lack a dedicated focus on reasoning, and fail to establish a systematic evaluation framework.
To address these limitations, we introduce \textbf{VGRP-Bench}, a Visual Grid Reasoning Puzzle Benchmark featuring 20 diverse puzzles\footnote{Unlike some benchmarks that scrape fixed pre-existing puzzles from various sources, our benchmark supports sampling puzzles with different settings and difficulty levels through hyperparameters.}. 
VGRP-Bench spans multiple difficulty levels, and includes extensive experiments not only on existing chat LVLMs (e.g., GPT-4o), but also on reasoning LVLMs (e.g., Gemini-Thinking). Our results reveal that even the state-of-the-art LVLMs struggle with these puzzles, highlighting fundamental limitations in their puzzle-solving capabilities.
Most importantly, through systematic experiments, we identify and analyze key factors influencing LVLMs' puzzle-solving performance, including the number of clues, grid size, and rule complexity. 
Furthermore, we explore two Supervised Fine-Tuning (SFT) strategies that can be used in post-training: SFT on solutions (S-SFT) and SFT on synthetic reasoning processes (R-SFT). While both methods significantly improve performance on trained puzzles, they exhibit limited generalization to unseen ones.
We will release VGRP-Bench to facilitate further research on LVLMs for complex, real-world problem-solving. Project page: \url{https://yufan-ren.com/subpage/VGRP-Bench/}. 
\end{abstract}
  
\section{Introduction}
\label{sec:intro}

As Large Language Models (LLMs) advance rapidly~\cite{radford2019language,wang2019superglue,hendrycks2020measuring,srivastava2022beyond,cobbe2021training}, researchers are extending their capabilities to multimodal tasks, leading to the rise of Large Vision-Language Models (LVLMs)~\cite{dubey2024llama,yin2024lamm,bai2023qwen,zhu2023minigpt,Metz2024}. While LVLMs demonstrate success in some perception tasks, they often face challenges in strategic planning, especially in visual games that require a combination of perception and multi-step reasoning~\cite{zhang2024ing,paglieri2024balrog,wang2025are}.

Among the visual games, grid-like reasoning puzzles, e.g., Sudoku, Futoshiki, and Thermometers, Fig.~\ref{fig:teaser_first_page}, are renowned for their simple rules yet challenging solutions. They have gained widespread popularity, even being featured in annual world championships~\cite{wpc_wikipedia}. 
Beyond entertainment, grid puzzles also serve as structured reasoning tasks that require logical deduction, constraint satisfaction, and combinatorial search—skills that are fundamental to real-world problem-solving in domains such as robotic path planning~\cite{zhou2023spatial}, automated logistics scheduling~\cite{tang2024automated}, and embodied AI control~\cite{zawalski2024robotic}. Their well-defined rules and inherent complexity make them ideal for testing AI system’s ability to process structured visual information and adhere to logical constraints. Nevertheless, despite their potential as benchmarks for visual reasoning, there are underused for evaluating LVLMs in existing research.

\begin{figure*}[h!] \centering \includegraphics[width=1.0\textwidth, page=1]{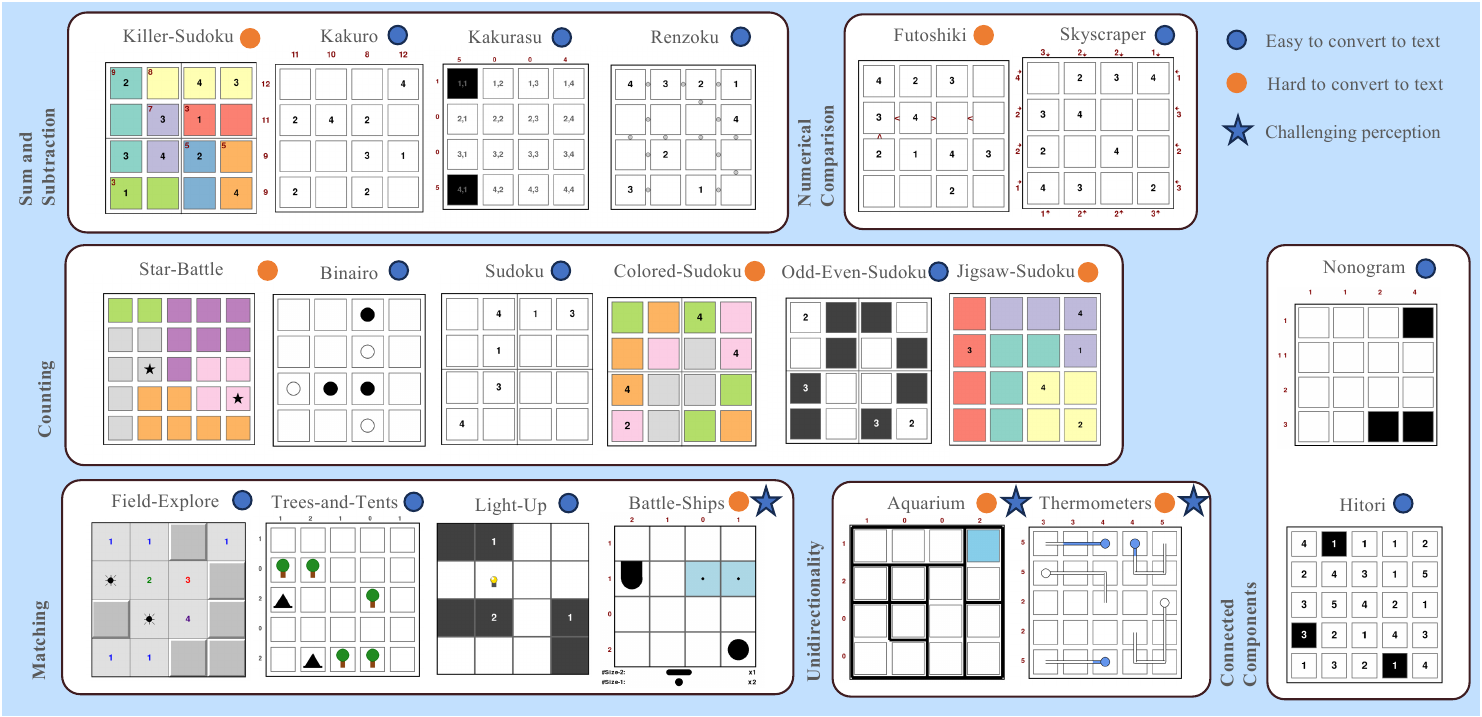} 

\caption{\textbf{Benchmark Games: Primitives and Sample Questions}. we systematically define puzzle primitives, including conditions, constraints, variables, and states, to establish a unified framework for inference and evaluation (left). This benchmark includes tasks designed to evaluate the reasoning, rule-following, and perception capabilities of state-of-the-art LVLMs. (Best viewed on a screen when zoomed in)} 
\label{fig:taxonomy} \end{figure*}

To address this gap, we introduce the \underline{V}isual \underline{G}rid \underline{R}easoning \underline{P}uzzle Benchmark (VGRP-Bench), the largest visual puzzle benchmark to date in terms of puzzle variety and complexity, featuring 20 diverse customizable puzzles that emphasize grid-based visual reasoning and form a taxonomy of rules, attributes, and patterns (Fig.~\ref{fig:taxonomy}). 
We draw inspiration from popular reasoning puzzles~\cite{puzzler_online_puzzles,puzzle_battleships,puzzlemix}, and design this benchmark with different levels of difficulty, \easy, \medium, and \hard, depending on the grid size, the required number of reasoning steps, and the size of the decision space. 
We conduct extensive experiments evaluating state-of-the-art LVLMs, including their reasoning counterparts, Fig.~\ref{fig:benchmark-off-the-shelf-easy-cot}. 
With our benchmark, we assess several aspects of LVLMs including perception, rule adherence, and overall puzzle-solving capabilities. 
To separate reasoning and perception, we additionally provide a text version of all puzzles. 
Through evaluations, we observe that our benchmark poses a huge challenge for most LVLMs, even at the easy level. 
For instance, GPT-4o fails to solve a simple $4\times 4$ Sudoku consistently, even in the text-only version of the game ($<30\%$ solving rate).
We summarize several common failure cases, such as the inability to localize a number on a grid and to correctly keep track of a reasoning process.
Moreover, we investigate factors that might impact an LVLM's performance, such as the difficulty level, the grid size, the number of clues, and the rules involved in a puzzle.

\begin{table}[t]
\centering
\resizebox{1.0\columnwidth}{!}{
\begin{tabular}{|c|c|c|c|c|}
\hline
 & Levels & Fine-Tuning  & \#Puzzles/Games & \#Models \\ \hline
VGRP-Bench & \checkmark & \textbf{$\checkmark$} & \textbf{20}  & \textbf{16}     \\ \hline
ING-VP~\cite{zhang2024ing} & $\times$ & $\times$ & 6 & 15    \\ \hline
BALROG~\cite{paglieri2024balrog} & $\times$  & $\times$ & 6  & 11 \\ \hline
\cite{wang2025are} & $\times$ &$\times$ & 6  & 8   \\ \hline
\end{tabular}
}
\caption{
VGRP-Bench offers a large puzzle collection for LVLM benchmarking, providing a comprehensive evaluation of state-of-the-art LVLMs across different dimensions, such as perception, rule adherence, and overall puzzle-solving, across different difficulty levels. We also investigate post-training strategies to enhance LVLMs' puzzle-solving performance.}
\label{tab:vs_baselines}
\end{table}

Beyond benchmarking off-the-shelf models following other game benchmark papers, we investigate whether post-training techniques can enhance LVLMs' puzzle-solving abilities (Tab.~\ref{tab:vs_baselines}).
Specifically, we explore two post-training strategies, including Solution Supervised Fine-Tuning (S-SFT) and Reasoning SFT (R-SFT).
In S-SFT, we fine-tune LVLMs on final solutions, typically represented as nested lists indicating the board’s final state.
In R-SFT, inspired by human and algorithmic approaches to puzzle solving~\cite{daganzo2018minuet,chi2012techniques} such as step-by-step reasoning and process-of-elimination via rule-based deduction, we construct an SFT dataset by recording a solver’s stepwise reasoning trajectory. We then fine-tune the LVLM on this dataset.
We observe significant improvement in puzzle solving at the easy level, while fine-tuned models still struggle at the medium and hard levels.
Additionally, recognizing the risk of overfitting to the puzzles used for finetuning, we examine the generalization capabilities of models trained with each approach in our benchmark.

In summary, we present a novel, customizable LVLM benchmark tailored for visual reasoning puzzles and conduct a systematic evaluation of LVLMs, as shown in Tab.~\ref{tab:vs_baselines}. Our key contributions are as follows:

\begin{itemize}
    \item We introduce a large LVLM customizable grid-based reasoning benchmark with systematic evaluation protocols structured around a taxonomy of diverse visual clues and rules.
    \item We conduct extensive experiments on state-of-the-art closed-source and open-source LVLMs using our benchmark, including fine-grained evaluations such as cell-level perception and step-wise rule understanding.
    \item We summarize common failure cases of LVLMs in puzzle solving and provide detailed ablation studies on various factors that impact an LVLM's puzzle solving, such as difficulty level, number of clues, and rules involved.
    \item To gain deeper insights into the challenges faced by LVLMs in puzzle solving, we explore two post-training strategies: Solution SFT and Reasoning SFT.
\end{itemize}

\section{Related Works}

\subsection{General LLM/LVLM Benchmarks}

The advanced capabilities of Large Language Models (LLMs)~\cite{achiam2023gpt,anil2023palm,team2023gemini,touvron2023llama} and Large Vision-Language Models (LVLMs)~\cite{liu2024llavanext,liu2023improvedllava,liu2023llava,llama_3_2} have inspired extensive research on benchmarking their capabilities. 
Prominent benchmarks like SuperGLUE~\cite{wang2019superglue}, MMLU~\cite{hendrycks2020measuring}, and BigBench~\cite{srivastava2022beyond}, evaluate general language understanding and multitasking text-based capabilities.
Domain-specific benchmarks evaluate specialized competencies such as coding~\cite{mbxp_athiwaratkun2022,jain2024livecodebench} and mathematics~\cite{hendrycks2021measuring,cobbe2021training}. 
Notable early examples include Science QA~\cite{lu2022learn}, VizWiz~\cite{bigham2010vizwiz}, and VQAv2~\cite{goyal2017making}. Specific domains, such as image captioning, are represented by works such as~\cite{lin2014microsoft}. 
More recent efforts~\cite{zhang2024task}, such as MMBench~\cite{liu2025mmbench}, EMMA~\cite{hao2025can}, and SEED-Bench~\cite{li2023seed}, offer comprehensive evaluations of multimodal reasoning and perception. BLINK~\cite{fu2024blink} focuses on visual perception tasks that humans can solve in an instant. LMEvalKit~\cite{duan2024vlmevalkit} unifies model comparisons across various benchmarks.

Our VGRP-Bench differs from other benchmarks by focusing on reasoning puzzles, a special challenge to LVLMs that requires combining perception and decision making with multi-step reasoning.

\begin{figure}[t] \centering \includegraphics[width=0.50\textwidth, page=1]{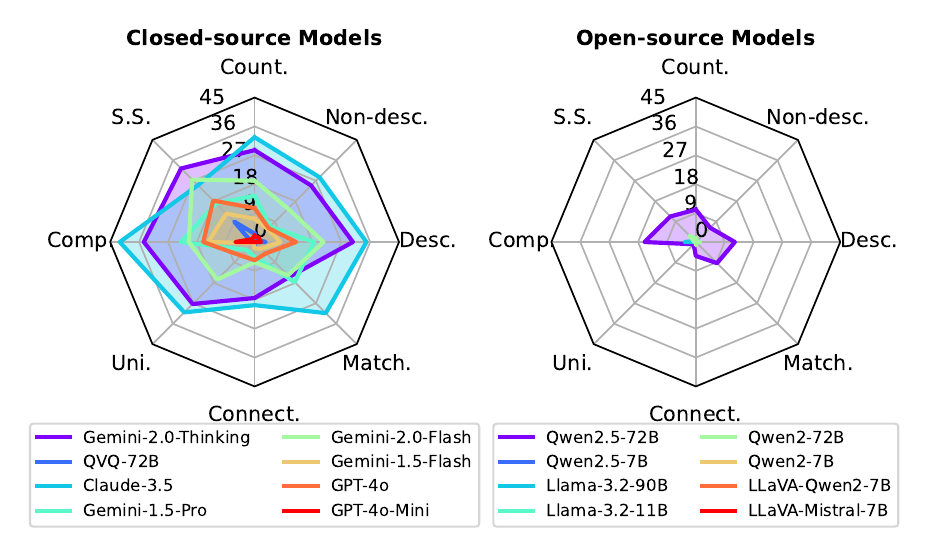} 
\caption{\textbf{Result Summary on \Easy~Level.} Puzzle-solving rate of state-of-the-art chat LVLMs on easy-level puzzles associated with each rule. Please refer to the experiment section for detailed result analysis. Note that this plot's score ranges from 0 to 45\%, instead of 100\%.  (Best viewed on a screen when zoomed in)} 
\label{fig:benchmark-radar} \end{figure}

\subsection{LLM/LVLM Game Benchmarks}

Challenging games have long been regarded as milestones of machine intelligence as exemplified by Deep Blue~\cite{hsu2022behind} and AlphaGo~\cite{silver2016mastering}.
Classical benchmarks, such as Atari~\cite{schrittwieser2020mastering} and the Arcade Learning Environment~\cite{bellemare2013arcade}, have played a crucial role in developing reinforcement learning algorithms and improving agent capabilities.
Given the natural language capabilities of LLMs, researchers have introduced benchmarks where LLM agents interact within game environments~\cite{wu2023smartplay,park2023generative}.
\cite{hong2023metagpt,qian2023communicative,chen2023agentverse,wang2023voyager,tan2024towards} investigate LLMs' performance in agent-based and collaborative game environments, emphasizing interaction and teamwork skills.

Several recent studies benchmark LVLMs on visual games. ING-VP~\cite{zhang2024ing} shows that LVLMs still struggle with easy games. 
\cite{wang2025are} proposes a benchmark with fine-grained evaluation.
BALROG~\cite{paglieri2024balrog} measures LVLM games like MiniHack and NetHack.
\cite{estermann2025puzzles} proposed a puzzle RL environment, and benchmark several RL algorithms.
ZeroBench~\cite{roberts2025zerobench} proposes a benchmark in which current LVLMs struggle to achieve meaningful accuracy.
A concurrent work, \cite{wang2025enigmaeval}, created a visual benchmark by scraping existing puzzles from online sources, resulting in a dataset of 949 instances of puzzles. 

VGRP-Bench distinguishes itself by focusing on reasoning puzzles, employing customizable puzzle generators, and systematically evaluating models from inference to post-training techniques.
\section{VGRP-Bench: The Benchmark}

This section is organized as follows: we first present our benchmark in Sec.~\ref{sec:benchmark}, along with its evaluation protocol in Sec.~\ref{sec:evaluation} and taxonomy in Sec.~\ref{taxonomy}.
In addition to benchmarking off-the-shelf models, we investigate the challenges faced by existing LVLMs in solving visual puzzles and propose strategies to address these limitations.
Specifically, we use two fine-tuning strategies, Solution Supervised Fine-Tuning (S-SFT) and Reasoning SFT (R-SFT), as described in detail in Sec.~\ref{sec::post-training}. 

\subsection{Grid-Like Visual Reasoning Puzzles}
\label{sec:benchmark}

\begin{figure*}[t] 
\centering 
\includegraphics[width=1.0\textwidth,page=2]{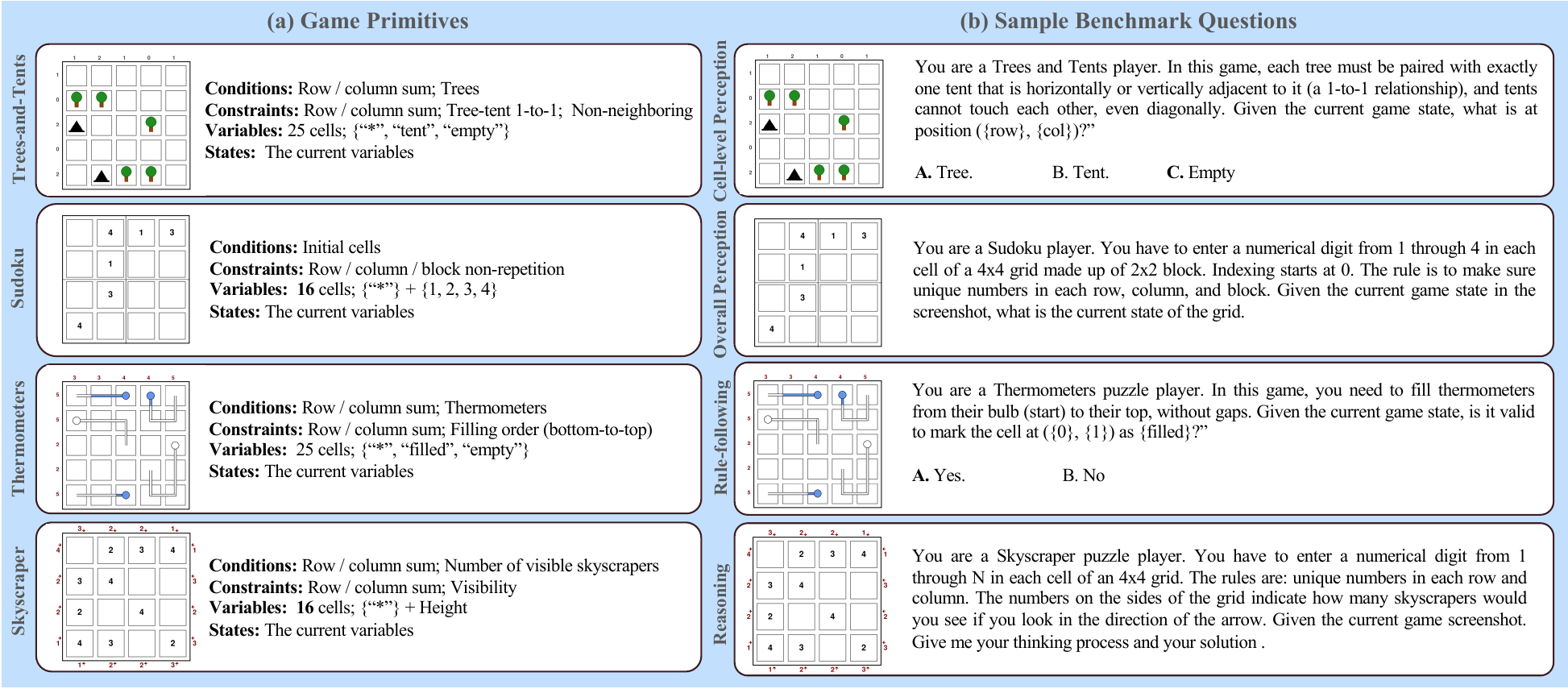} 
\caption{
\textbf{Diverse Rules and Visual Patterns in VGRP-Bench.}
Our benchmark includes a diverse set of rules, such as counting and mathematical calculations, 
and also exhibits diversity in visual patterns, encompassing text, numerical values, and objects such as trees. 
We highlight puzzles that are easy or difficult to convert into text.
}
\label{fig:primitives-result} 
\end{figure*}

\noindent \textbf{Puzzle Selection.} 
To form this benchmark, we select visual puzzle games based on the following criteria: requiring multi-step reasoning for decision-making and rule validation, incorporating a diverse range of visual clues, rules and interaction methods, and ultimately contributing to a structured taxonomy (Fig.~\ref{fig:primitives-result}).
For example, vanilla Sudoku is purely numerical and relies on repetition-based constraints, while Trees-and-Tents demands pattern recognition, relational reasoning between trees and tents, and checking 1-to-1 matching. 
In contrast, Thermometers relies heavily on understanding and applying physical-world rules, e.g., thermometers must be filled starting from their base\footnote{Here, Sudoku serves as an example of puzzles that could be easily converted to text, owing to its widespread popularity, while Trees-and-Tents and Thermometers represent puzzles harder to convert to text.}.

\noindent \textbf{Puzzle Primitives.}
To ensure consistency across different puzzles and facilitate future integration of new ones, we design the benchmark around four core primitives—variables, states, constraints, and conditions—to provide a unified structure, as depicted in Fig.\ref{fig:primitives-result} left.
\noindent \textbf{Variables $\mathcal{V}$ and States $\mathcal{S}$.}
Each puzzle consists of a set of variables, $\mathcal{V} = \{v_i\}_{i=1}^{n}$, representing cells or elements requiring value assignments. 
For example, a $4 \times 4$ \textit{Sudoku} grid comprises 16 variables, with each variable taking a value from the set of possible values $\{1, 2, 3, 4\}$. 
The set of states $\mathcal{S} = \{ s_i \}_{i=1}^{n}$ represents the current value assignments of the variables. 
\noindent \textbf{Constraints.}
Constraints $\mathcal{C} = \{ c_j \}_{j=1}^{m}$ define rules for valid puzzle state configurations. 
For instance, in \textit{Sudoku}, constraints enforce the non-repetition of values in each row, column, and block.
In \textit{Trees and Tents}, constraints enforce a bijective mapping between trees and tents while adhering to row and column sums.
\noindent \textbf{Conditions.}
Conditions correspond to preset values or clues that define the puzzle's starting state. Examples include predefined digits that act as initial clues in \textit{Sudoku} or row and column constraints given as clues in \textit{Thermometers}.

\subsection{Evaluation Protocol}
\label{sec:evaluation}
Our benchmark evaluates LVLM performance across several capabilities, including perception, rule-following, and reasoning tasks at multiple granular levels, and on difficulty levels, as illustrated in the right column of Fig.~\ref{fig:primitives-result}.
Specifically, at the puzzle-solving level, we assess overall perception accuracy and puzzle-solving success rate by evaluating the LVLM's holistic understanding of the board and its ability to generate a correct solution.
Moreover, we provide additional evaluations at finer levels of granularity, including evaluations at the cell and step level. 

\subsection{Puzzle Rule/Capability Taxonomy}

\label{taxonomy}
We create a taxonomy of rule/capabilities required to solve the puzzles in our benchmark, and visualize the prominent ones in Fig.~\ref{fig:primitives-result}, as one puzzle might require multiple capabilities like counting, a basic rule in most puzzles.
For example, Killer-Sudoku, Kakuro, Kakurasu, and Renzoku require mathematical calculations involving addition and subtraction.
Trees-and-Tents, requiring the LVLM to understand bijective matching of trees and tents, is an example the matching rule of associating spatially or semantically relevant components. 
Other rules and capabilities are numerical comparison, understanding procedural order (unidirectionality) and putting connected components together.

\begin{figure*}[t] \centering \includegraphics[width=1.0\textwidth, page=1]{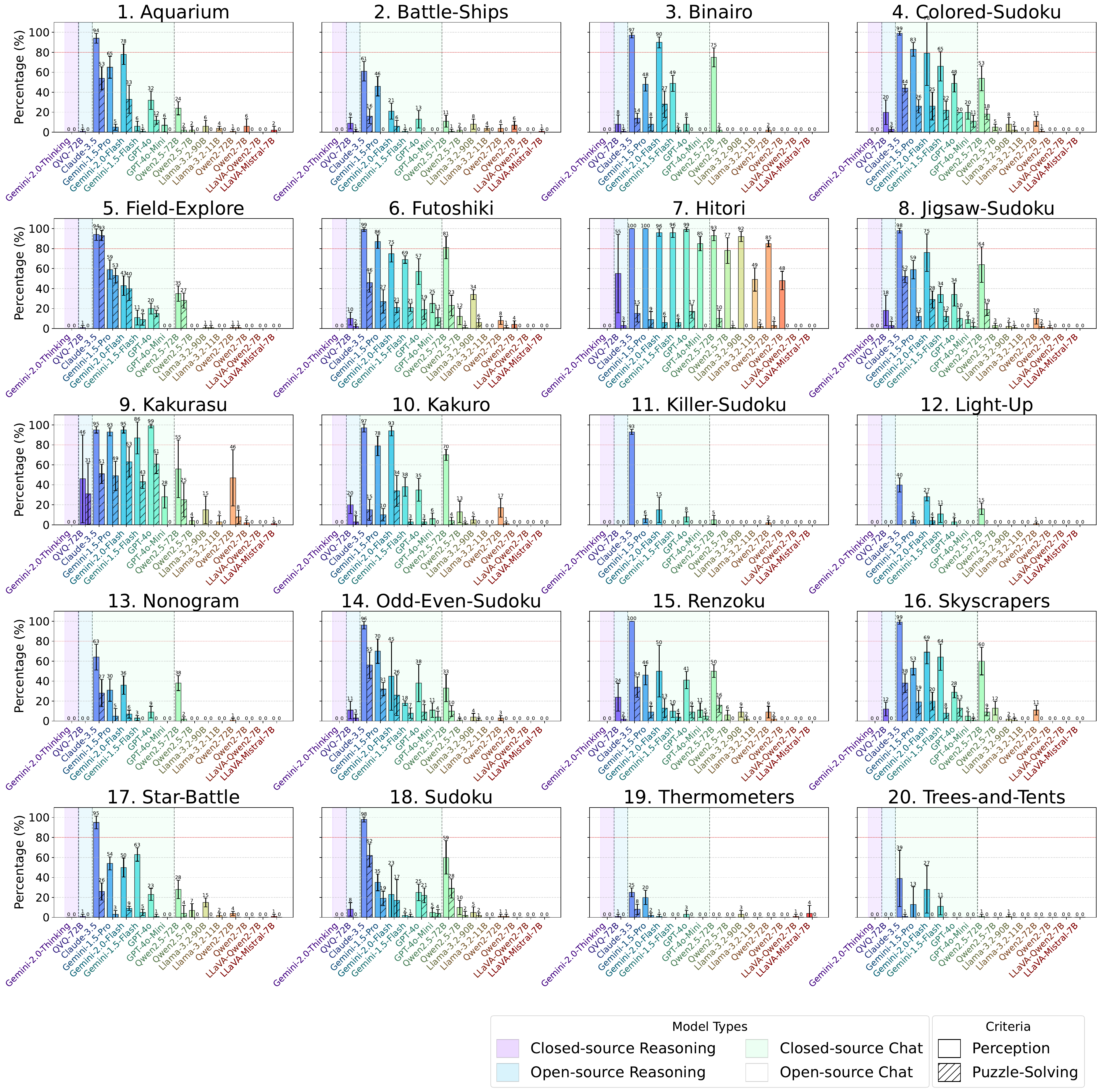} \caption{\textbf{Off-the-Shelf LVLMs on Level-\Easy~with CoT.} We report both correct perception rate and puzzle-solving rate evaluations with closed-source / open-source and reasoning / chat models. Please refer to supplementary for additional evaluations such as finer granularity evaluations and other difficulty levels, e.g., \medium~and \hard. 
(Puzzle-solving in hatched bars and best viewed on a screen when zoomed in)} 
\label{fig:benchmark-off-the-shelf-easy-cot} 
\end{figure*}

\subsection{Post-Training Techniques}
\label{sec::post-training}
Beyond assessing off-the-shelf LVLMs, we would like to take a step further to explore potential approaches to boost their performance. 
In this subsection, we utilize two post-training methods to tune a pretrained LVLM, i.e., Solution Supervised Fine-Tuning (S-SFT) and Reasoning Supervised Fine-Tuning (R-SFT).

\label{sec:SSFT_RSFT}
\noindent \textbf{S-SFT.} A baseline is to use Supervised Fine-Tuning.
Here, we adopt two strategies. First, we adopt a naive SFT for supervision of the LVLM to generate solutions. 
More specifically, we first convert the solution into a JSON-formatted text file, `` \{``answer": [[1, 2, 3, 4], [3, 4, 1, 2] , [2, 1, 4, 3], [4, 3, 2, 1]]\}". 
During training, we provide a text puzzle description as prompt and a screenshot of the puzzle as input. 
Then we use the predefined solution as supervision for the model.

\noindent \textbf{R-SFT.} We introduce a SFT data creation method specific for puzzle solving. 
Inspired by human and algorithmic puzzle solving that feature step-by-step reasoning and per-cell rule violation checking, we propose to conduct supervised Fine-Tuning (SFT) on synthetic trajectories. 
In this way, we would like to supervise LVLMs to imitate step-by-step reasoning, in a similar manner to how a predefined solver solves these puzzles.
To generate thought trajectories, we define the reasoning process as a trajectory through states. 
\textbf{A Trajectory}, $\mathcal{T} = \{s_i\}_{i=1}^{T}$, encodes key intermediate states encountered during puzzle solving. Each state $s_t$ captures variable assignments and potential values for unassigned variables. 
To avoid the inefficiency of starting from a random cell, Depth-First Search (DFS) with process-of-elimination is employed, enabling systematic exploration and backtracking upon failure states.
For instance, in a 4×4 Sudoku with 12 missing values, a random start often leads to excessive branching, producing trajectories that exceed the model’s output window.

\section{Experiments}
\label{sec:experiments}

\subsection{Implementation Details}

\noindent We benchmark several state-of-the-art LVLMs. For accessibility purposes, we include both closed-source and open-source models like Gemini-Pro~\cite{team2023gemini} and LlaVA-OneVision-7B~\cite{li2024llava} respectively. To assess different types of models, we include both chat LVLMs and reasoning LVLMs\footnote{In the reasoning model category, we include Gemini-2.0-Thinking and Qwen-QVQ, as other reasoning models are either lacking vision capabilities, e.g., DeepSeek~\cite{deepseekai2025deepseekr1incentivizingreasoningcapability}, or only accessible to \href{https://platform.openai.com/docs/guides/rate-limits}{high-tier users}. Due to the rate limit in Gemini-2.0-Thinking, we only evaluate puzzle-solving with chain-of-thought prompting.}. 
For evaluation, we launch 5 independent inference runs, with each run containing 20 instances, resulting in a total of 100 samples. We report the overall mean correctness and standard deviation across all sample runs. 
For post-training, we use Llama 3.2 Vision Instruct as the base model and conduct training on a single node equipped with 8 A100 GPUs.
We ensure that the training and test splits contain no overlapping puzzles in terms of input or solution.
Please refer to supplementary for more implementation details. 
\subsection{Off-the-Shelf LVLMs Evaluation}
\label{sec:experiment-results}
\label{sec:discussion}

\noindent 
We present the overall perception and puzzle-solving results in Fig.~\ref{fig:benchmark-off-the-shelf-easy-cot}, where all LVLMs struggle with puzzle-solving, achieving a success rate below 80\%. 
Additional granularity and evaluation results are discussed below, and the complete evaluation on all puzzles can be found in the supplementary material.
More specifically, \textbf{regarding perception, most closed-source models, except for Claude, achieve less than 50\% accuracy. Among open-source models, Qwen2.5-72B performs the best.}
Hitori exhibits the highest perception accuracy among all puzzles, suggesting that LVLMs struggle with grids containing missing cells.
Secondly, \textbf{in terms of puzzle-solving, though all models struggle, closed-source models generally outperform open-source ones}. We also observe that larger models tend to perform better; for example, GPT-4o outperforms GPT-4o-mini.
For reasoning models, we find that Gemini-2.0-Thinking performs well, whereas Qwen-QVQ underperforms compared to Qwen2.5-72B, potentially because Qwen-QVQ is a preview version.

\begin{figure}[h] \centering \includegraphics[width=0.45\textwidth, page=1]{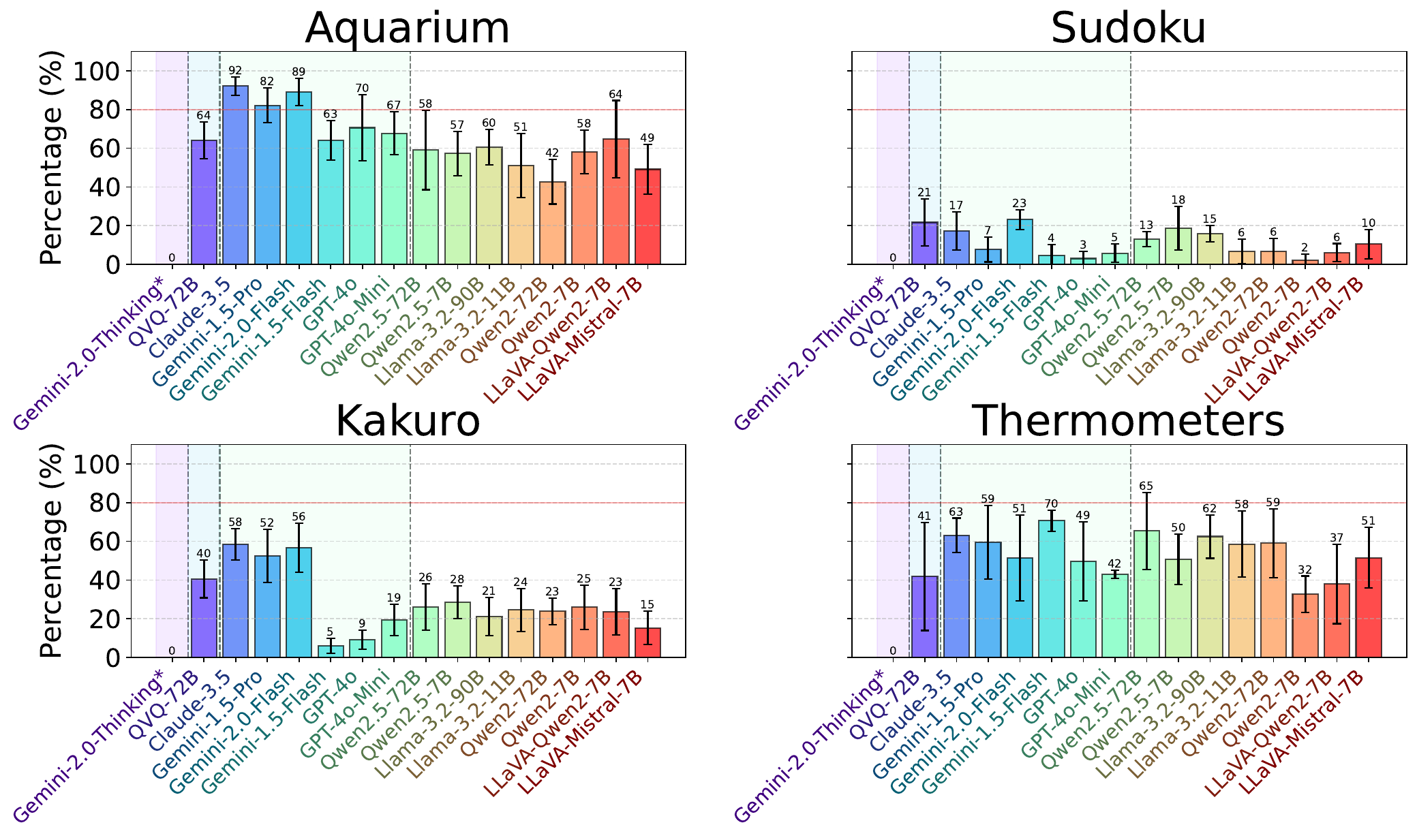} 
\caption{\textbf{Cell-level Perception Accuracy at Level-\Easy}. (Best viewed on a screen when zoomed in)} 
\label{fig:benchmark-main-cell-at} \end{figure}

\noindent \textbf{Cell-Level Evaluation.} 
We provide cell-level perception evaluation in Fig.~\ref{fig:benchmark-main-cell-at}. Similar to overall perception, closed-source models—particularly Claude and Gemini 2.0-Flash—generally achieve the highest performance.
Interestingly, we notice cases when querying the LVLM for the entire board yields the correct answer, whereas querying a specific cell results in an incorrect response.
This phenomenon mirrors previously observed failures in LVLMs, such as their struggles with counting tasks like "How many R’s are in the word Strawberry"~\cite{xu2024llm}.

\begin{figure}[h] \centering \includegraphics[width=0.45\textwidth, page=1]{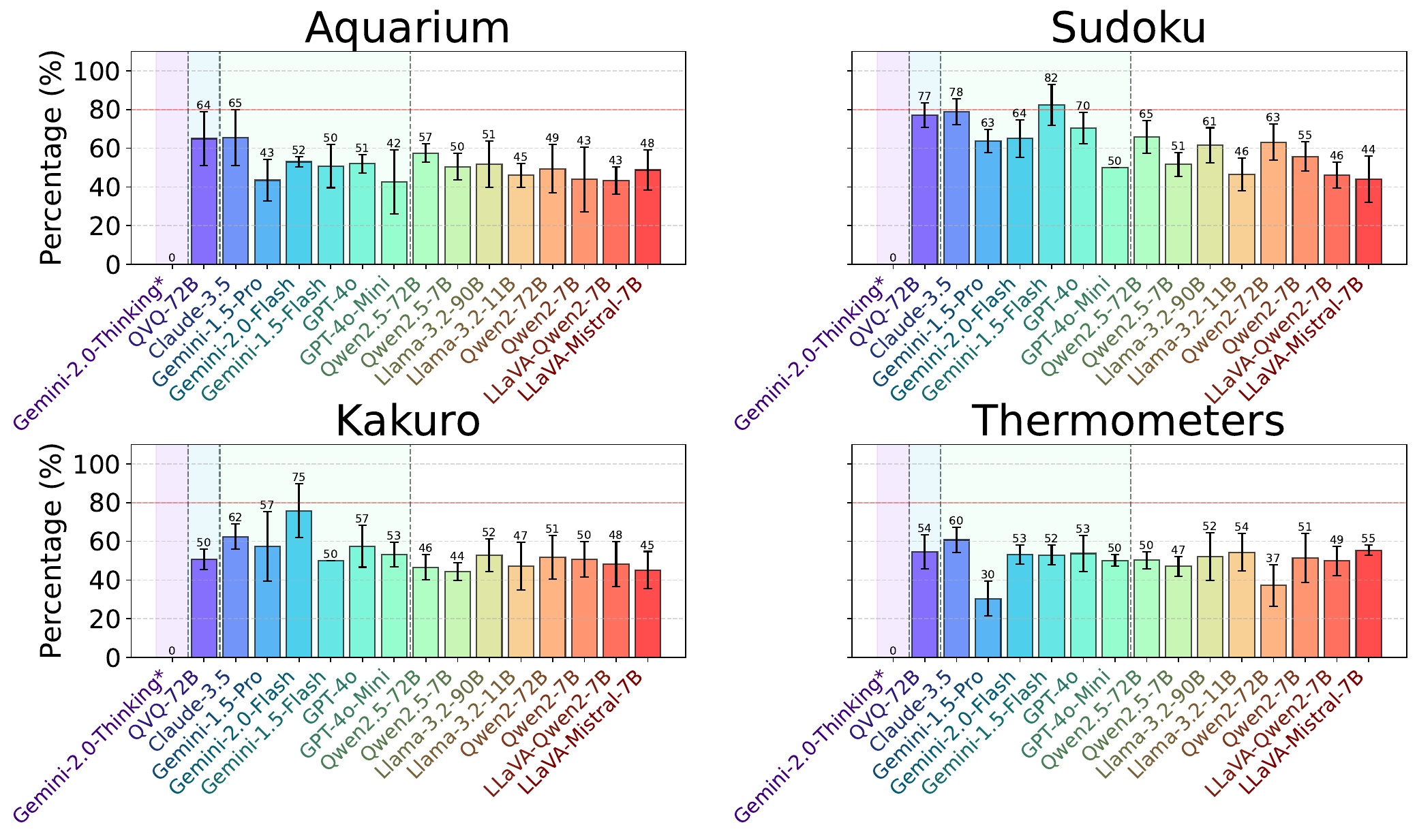} 
\caption{\textbf{Step-Level Rule-Following Accuracy at Level-\Easy.} (Best viewed on a screen when zoomed in)} 
\label{fig:benchmark-main-valid} \end{figure}

\noindent \textbf{Step-Level Rule-Following Evaluation.}
Claude consistently achieves the highest performance, whereas LlaVA performs the worst among all models.
Among the four puzzles shown in Fig.~\ref{fig:benchmark-main-valid}, Sudoku attains the highest accuracy, aligning with the intuition that it is a widely recognized puzzle with relatively simple and well-defined rules compared to the others.

\begin{figure}[h] \centering \includegraphics[width=0.45\textwidth, page=1]{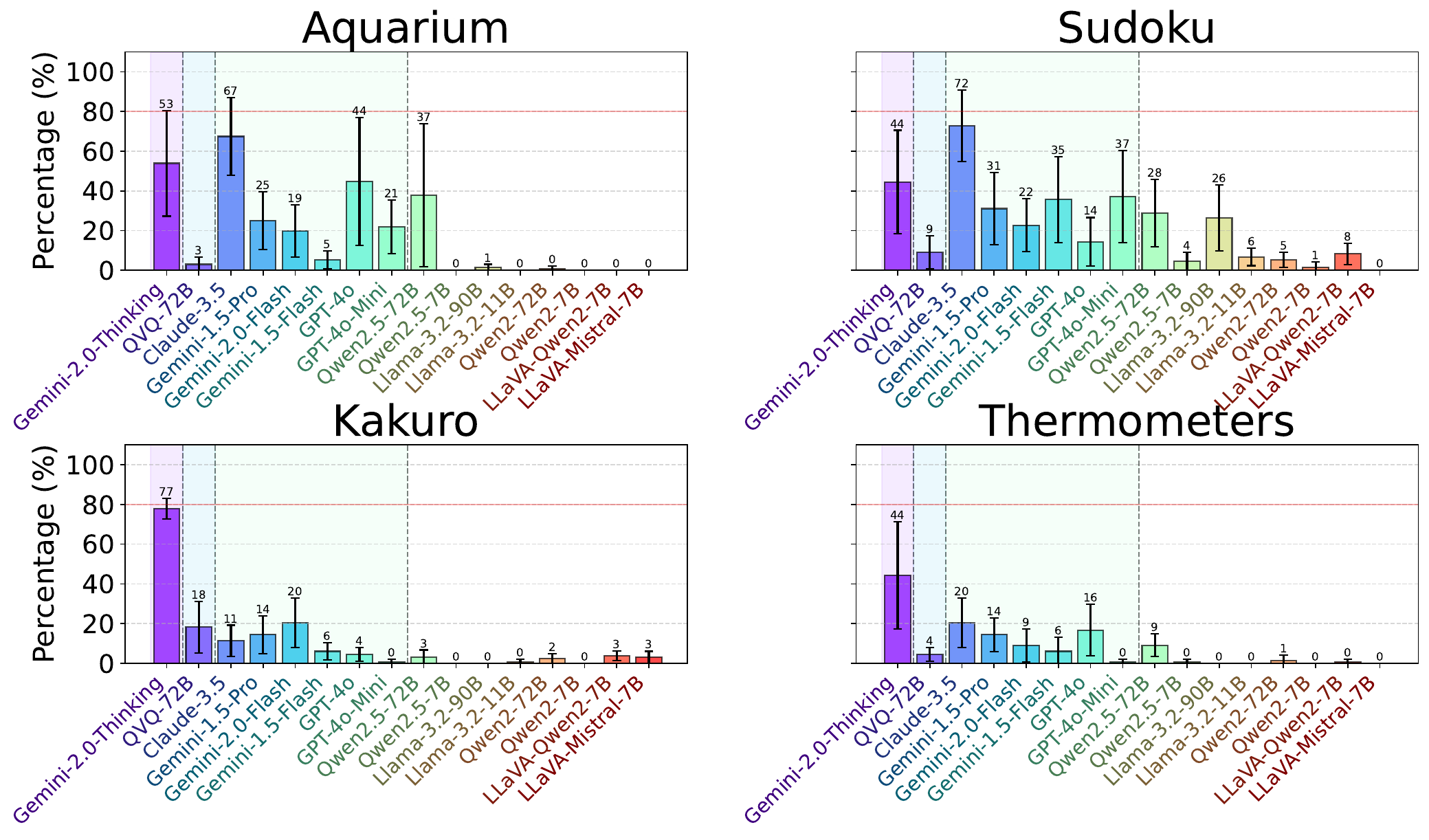} \caption{
\textbf{Performance of Text Version Puzzles on Level-\Easy}. 
For the text version of puzzles, the puzzle-solving rate increases significantly compared to the vision-based setting, highlighting the challenge of visual perception in our benchmark.
(Best viewed on a screen when zoomed in)} 
\label{fig:benchmark-text-version} 
\end{figure}

\noindent \textbf{Text Puzzles Evaluation.}
To understand the reasoning challenges in the text domain, we present the results of off-the-shelf models using text input in Fig.~\ref{fig:benchmark-text-version}. Notably, while this setting eliminates vision-related losses, the puzzles remain challenging for LVLMs.

\noindent \textbf{Puzzle Taxonomy Analysis.}
The diversity of puzzles and rule types in our benchmark enables analysis through the lens of puzzle taxonomy, making it a key differentiator from other existing benchmarks.
Each category includes at least two puzzles. For example, both Field-Explore and Trees-and-Tents require matching and pairing components. We present results aggregated by puzzle taxonomy in Fig.~\ref{fig:benchmark-radar}.

\begin{figure}[h] \centering \includegraphics[width=0.5\textwidth, page=1]{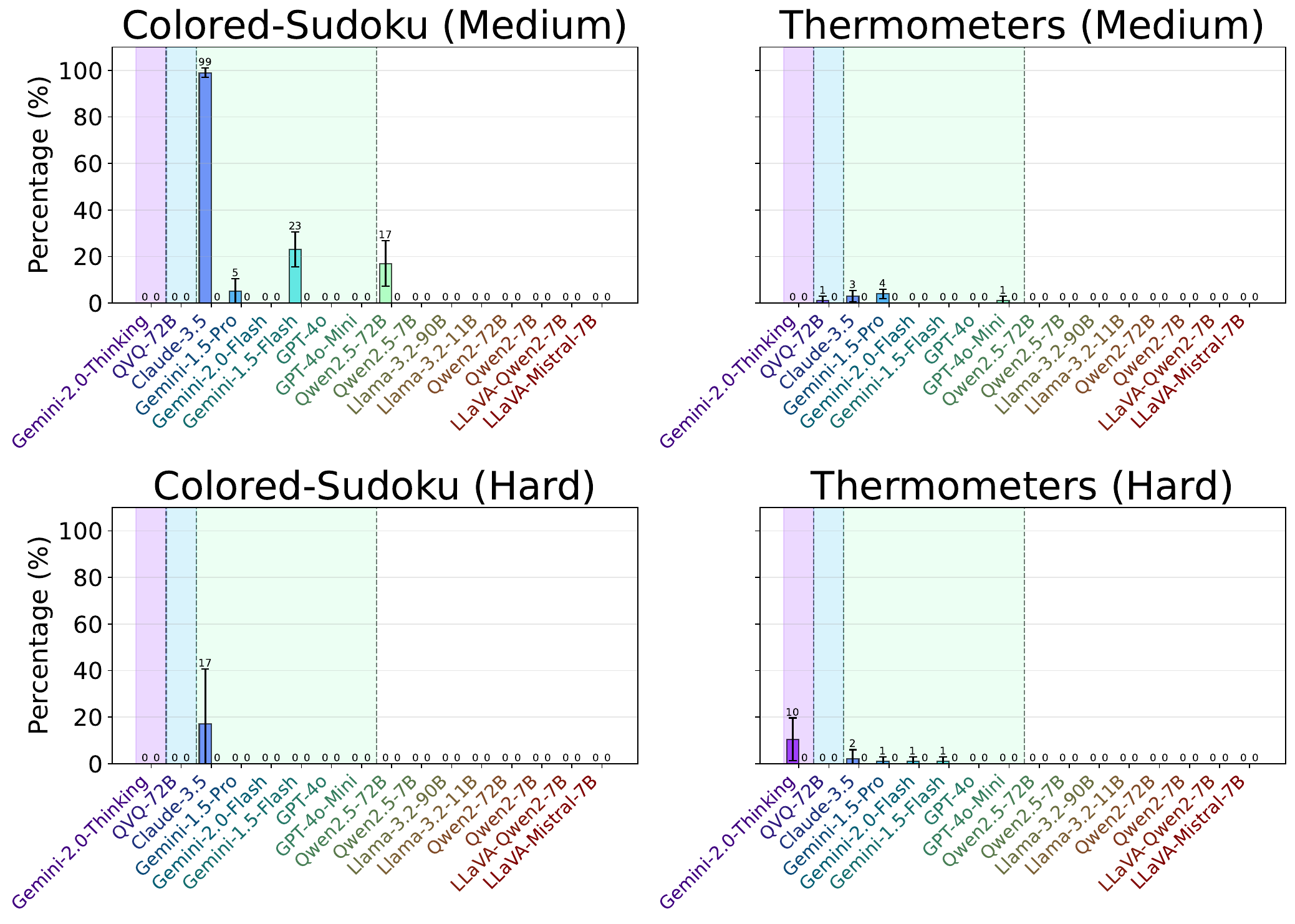} \caption{\textbf{Off-the-Shelf LVLMs on Level-\Medium~(top row) and \Hard~(bottom row) with CoT.} 
(Best viewed on a screen when zoomed in)} 
\label{fig:benchmark-off-the-shelf-medium-hard-main} 
\end{figure}

\noindent \textbf{Effect of Difficulty Level.} 
As difficulty increases, reflected in larger grids and more steps required to complete the puzzle—accuracy declines in both perception and puzzle-solving (Fig.~\ref{fig:benchmark-off-the-shelf-medium-hard-main}).
Notably, at the medium difficulty level with Thermometers, all LVLMs achieve a perception accuracy below 5\% and fail to solve the puzzles completely.
Performance further deteriorates at the hard difficulty level, indicating significant limitations in handling complex puzzles.

\begin{figure}[h] \centering \includegraphics[width=0.5\textwidth, page=1]{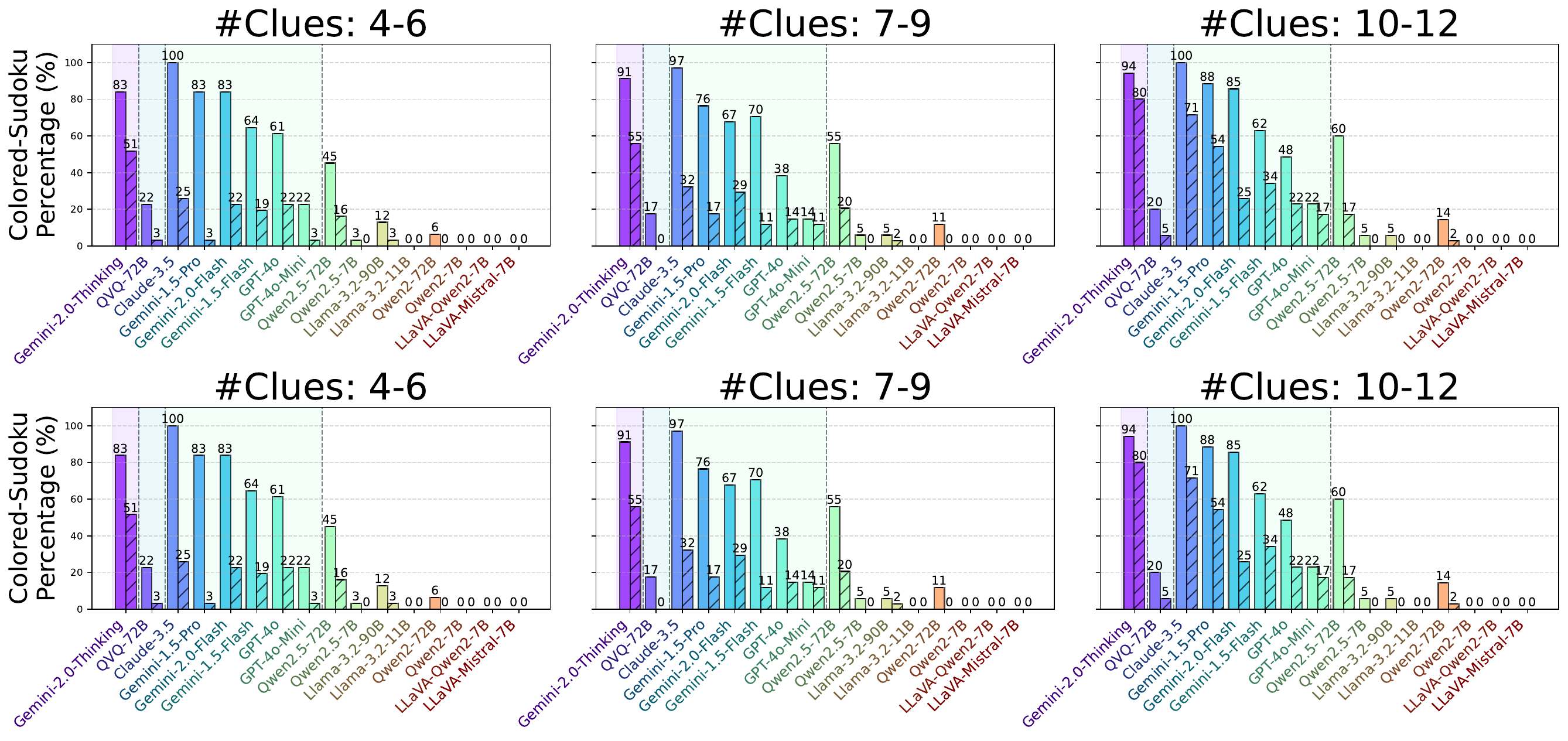} 
\caption{\textbf{Results with Different Number of Clues on Level-\Easy.} When more clues are provided (to the right), puzzles become easier, resulting in a higher puzzle-solving rate. (Best viewed on a screen when zoomed in)} 
\label{fig:benchmark-main-clue-number} \end{figure}

\noindent \textbf{Effect of Clue Number.} 
Intuitively, providing more clues simplifies the puzzles, leading to improved performance.
This trend is evident in Fig.~\ref{fig:benchmark-main-clue-number}, where we also observe a corresponding increase in perception accuracy.

\noindent \textbf{Common Failure Patterns.}
Off-the-shelf chat models exhibit several common failure cases. For instance, chat LVLMs often struggle to localize values on a grid, misinterpreting sequences like [*, 2, *, ] as [, *, 2, *]. Additionally, they frequently misunderstand the roles of different components, such as mistaking a cage clue for a board number in Killer Sudoku, and they tend to repeat responses. Extensive sample outputs and common failure cases are provided in the supplementary material.

\subsection{Post-Training Evaluation}
\noindent
We compare the pre-trained Llama 3.2 model with its fine-tuned versions after S-SFT and R-SFT in Fig.~\ref{fig:benchmark-reasoning-finetuning-2x2}, with additional details provided in the supplementary material.
First, \textbf{we observe that both S-SFT and R-SFT significantly enhance performance}, as the pre-trained model initially fails to produce any correct answers. This suggests that generalization to new puzzle settings is feasible.
Comparing S-SFT and R-SFT, their effectiveness varies across puzzles: S-SFT outperforms R-SFT in some cases, whereas R-SFT excels in others such as Aquarium.
We hypothesize that this is because R-SFT receives more supervision but is also more susceptible to compounding errors in long reasoning trajectories. We provide an evaluation on cross-puzzle generalization in the supplementary material. 

\begin{figure}[h] \centering \includegraphics[width=0.5\textwidth, page=1]{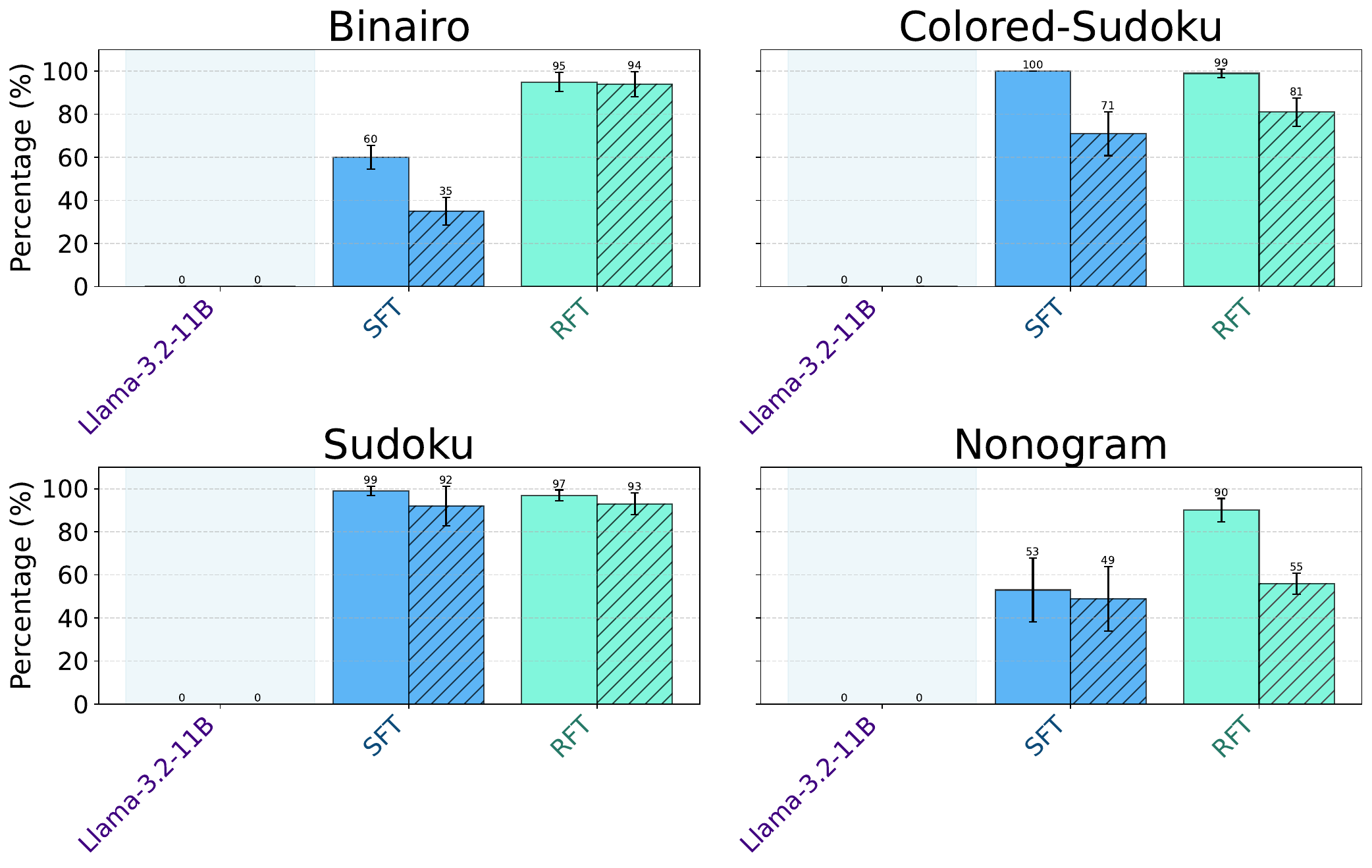}
\caption{
\textbf{Comparing S-SFT and R-SFT on Level-\Easy.} 
Both S-SFT and R-SFT significantly improve the pretrained model's performance in perception and puzzle-solving, with R-SFT achieves slightly better results in a few puzzles such as Binairo, while being lower in puzzles like Field-Explore. 
(Puzzle-solving in hatched and best viewed on a screen when zoomed in)} 
\label{fig:benchmark-reasoning-finetuning-2x2} \end{figure}
\section{Limitations and Future Work}

Due to the high computational cost of fine-tuning large models (e.g., 70B parameter models), our SFT experiments are limited to smaller 11B models.
Future research could explore inference-time strategies, including Monte Carlo Tree Search~\cite{silver2016mastering}. 
Another promising direction is to enhance puzzle-solving performance by integrating RL with outcome-based reward models. We report preliminary findings in the supplementary material. 

\section{Conclusion}
In this work, we have introduced VGRP-Bench, a large visual grid puzzle benchmark with various setting, including difficulty levels and diversified puzzle rules, and systematic evaluation. We evaluated off-the-shelf LVLMs on our VGRP-Bench showing their inability of puzzle solving. Furthermore, we explore post-training for improving LVLM performance, revealing significant improvement on the trained puzzle but also a lack of generalization to unseen ones. We hope this benchmark inspires future research and advances LVLM studies for complex, real-world tasks. 

{
    \small
    \bibliographystyle{ieeenat_fullname}
    \bibliography{main}

\begin{thebibliography}{69}
\providecommand{\natexlab}[1]{#1}
\providecommand{\url}[1]{\texttt{#1}}
\expandafter\ifx\csname urlstyle\endcsname\relax
  \providecommand{\doi}[1]{doi: #1}\else
  \providecommand{\doi}{doi: \begingroup \urlstyle{rm}\Url}\fi

\bibitem[Achiam et~al.(2023)Achiam, Adler, Agarwal, Ahmad, Akkaya, Aleman, Almeida, Altenschmidt, Altman, Anadkat, et~al.]{achiam2023gpt}
Josh Achiam, Steven Adler, Sandhini Agarwal, Lama Ahmad, Ilge Akkaya, Florencia~Leoni Aleman, Diogo Almeida, Janko Altenschmidt, Sam Altman, Shyamal Anadkat, et~al.
\newblock Gpt-4 technical report.
\newblock \emph{arXiv preprint arXiv:2303.08774}, 2023.

\bibitem[Anil et~al.(2023)Anil, Dai, Firat, Johnson, Lepikhin, Passos, Shakeri, Taropa, Bailey, Chen, et~al.]{anil2023palm}
Rohan Anil, Andrew~M Dai, Orhan Firat, Melvin Johnson, Dmitry Lepikhin, Alexandre Passos, Siamak Shakeri, Emanuel Taropa, Paige Bailey, Zhifeng Chen, et~al.
\newblock Palm 2 technical report.
\newblock \emph{arXiv preprint arXiv:2305.10403}, 2023.

\bibitem[Athiwaratkun et~al.(2022)Athiwaratkun, Gouda, Wang, Li, Tian, Tan, Ahmad, Wang, Sun, Shang, Gonugondla, Ding, Kumar, Fulton, Farahani, Jain, Giaquinto, Qian, Ramanathan, Nallapati, Ray, Bhatia, Sengupta, Roth, and Xiang]{mbxp_athiwaratkun2022}
Ben Athiwaratkun, Sanjay~Krishna Gouda, Zijian Wang, Xiaopeng Li, Yuchen Tian, Ming Tan, Wasi~Uddin Ahmad, Shiqi Wang, Qing Sun, Mingyue Shang, Sujan~Kumar Gonugondla, Hantian Ding, Varun Kumar, Nathan Fulton, Arash Farahani, Siddhartha Jain, Robert Giaquinto, Haifeng Qian, Murali~Krishna Ramanathan, Ramesh Nallapati, Baishakhi Ray, Parminder Bhatia, Sudipta Sengupta, Dan Roth, and Bing Xiang.
\newblock Multi-lingual evaluation of code generation models.
\newblock 2022.

\bibitem[Baccianella(2024)]{Baccianella_JSON_Repair_2024}
Stefano Baccianella.
\newblock {JSON Repair - A python module to repair invalid JSON, commonly used to parse the output of LLMs}, 2024.

\bibitem[Bai et~al.(2023)Bai, Bai, Yang, Wang, Tan, Wang, Lin, Zhou, and Zhou]{bai2023qwen}
Jinze Bai, Shuai Bai, Shusheng Yang, Shijie Wang, Sinan Tan, Peng Wang, Junyang Lin, Chang Zhou, and Jingren Zhou.
\newblock Qwen-vl: A frontier large vision-language model with versatile abilities.
\newblock \emph{arXiv preprint arXiv:2308.12966}, 2023.

\bibitem[Becker and Soatto(2024)]{becker2024cycles}
Evan Becker and Stefano Soatto.
\newblock Cycles of thought: Measuring llm confidence through stable explanations.
\newblock \emph{arXiv preprint arXiv:2406.03441}, 2024.

\bibitem[Bellemare et~al.(2013)Bellemare, Naddaf, Veness, and Bowling]{bellemare2013arcade}
Marc~G Bellemare, Yavar Naddaf, Joel Veness, and Michael Bowling.
\newblock The arcade learning environment: An evaluation platform for general agents.
\newblock \emph{Journal of Artificial Intelligence Research}, 47:\penalty0 253--279, 2013.

\bibitem[Bigham et~al.(2010)Bigham, Jayant, Ji, Little, Miller, Miller, Miller, Tatarowicz, White, White, et~al.]{bigham2010vizwiz}
Jeffrey~P Bigham, Chandrika Jayant, Hanjie Ji, Greg Little, Andrew Miller, Robert~C Miller, Robin Miller, Aubrey Tatarowicz, Brandyn White, Samual White, et~al.
\newblock Vizwiz: nearly real-time answers to visual questions.
\newblock In \emph{Proceedings of the 23nd annual ACM symposium on User interface software and technology}, pages 333--342, 2010.

\bibitem[Chen et~al.(2023)Chen, Su, Zuo, Yang, Yuan, Qian, Chan, Qin, Lu, Xie, et~al.]{chen2023agentverse}
Weize Chen, Yusheng Su, Jingwei Zuo, Cheng Yang, Chenfei Yuan, Chen Qian, Chi-Min Chan, Yujia Qin, Yaxi Lu, Ruobing Xie, et~al.
\newblock Agentverse: Facilitating multi-agent collaboration and exploring emergent behaviors in agents.
\newblock \emph{arXiv preprint arXiv:2308.10848}, 2\penalty0 (4):\penalty0 6, 2023.

\bibitem[Chi and Lange(2012)]{chi2012techniques}
Eric~C Chi and Kenneth Lange.
\newblock Techniques for solving sudoku puzzles.
\newblock \emph{arXiv preprint arXiv:1203.2295}, 2012.

\bibitem[Chu et~al.(2025)Chu, Zhai, Yang, Tong, Xie, Schuurmans, Le, Levine, and Ma]{chu2025sft}
Tianzhe Chu, Yuexiang Zhai, Jihan Yang, Shengbang Tong, Saining Xie, Dale Schuurmans, Quoc~V Le, Sergey Levine, and Yi Ma.
\newblock Sft memorizes, rl generalizes: A comparative study of foundation model post-training.
\newblock \emph{arXiv preprint arXiv:2501.17161}, 2025.

\bibitem[Cobbe et~al.(2021)Cobbe, Kosaraju, Bavarian, Chen, Jun, Kaiser, Plappert, Tworek, Hilton, Nakano, et~al.]{cobbe2021training}
Karl Cobbe, Vineet Kosaraju, Mohammad Bavarian, Mark Chen, Heewoo Jun, Lukasz Kaiser, Matthias Plappert, Jerry Tworek, Jacob Hilton, Reiichiro Nakano, et~al.
\newblock Training verifiers to solve math word problems.
\newblock \emph{arXiv preprint arXiv:2110.14168}, 2021.

\bibitem[Daganzo(2018)]{daganzo2018minuet}
Carlos~F Daganzo.
\newblock Minuet: A method to solve sudoku puzzles by hand.
\newblock \emph{arXiv preprint arXiv:1812.06778}, 2018.

\bibitem[DeepSeek-AI et~al.(2025)DeepSeek-AI, Guo, Yang, Zhang, Song, Zhang, Xu, Zhu, Ma, Wang, Bi, Zhang, Yu, Wu, Wu, Gou, Shao, Li, Gao, Liu, Xue, Wang, Wu, Feng, Lu, Zhao, Deng, Zhang, Ruan, Dai, Chen, Ji, Li, Lin, Dai, Luo, Hao, Chen, Li, Zhang, Bao, Xu, Wang, Ding, Xin, Gao, Qu, Li, Guo, Li, Wang, Chen, Yuan, Qiu, Li, Cai, Ni, Liang, Chen, Dong, Hu, Gao, Guan, Huang, Yu, Wang, Zhang, Zhao, Wang, Zhang, Xu, Xia, Zhang, Zhang, Tang, Li, Wang, Li, Tian, Huang, Zhang, Wang, Chen, Du, Ge, Zhang, Pan, Wang, Chen, Jin, Chen, Lu, Zhou, Chen, Ye, Wang, Yu, Zhou, Pan, Li, Zhou, Wu, Ye, Yun, Pei, Sun, Wang, Zeng, Zhao, Liu, Liang, Gao, Yu, Zhang, Xiao, An, Liu, Wang, Chen, Nie, Cheng, Liu, Xie, Liu, Yang, Li, Su, Lin, Li, Jin, Shen, Chen, Sun, Wang, Song, Zhou, Wang, Shan, Li, Wang, Wei, Zhang, Xu, Li, Zhao, Sun, Wang, Yu, Zhang, Shi, Xiong, He, Piao, Wang, Tan, Ma, Liu, Guo, Ou, Wang, Gong, Zou, He, Xiong, Luo, You, Liu, Zhou, Zhu, Xu, Huang, Li, Zheng, Zhu, Ma, Tang, Zha, Yan, Ren, Ren, Sha, Fu, Xu, Xie, Zhang,
  Hao, Ma, Yan, Wu, Gu, Zhu, Liu, Li, Xie, Song, Pan, Huang, Xu, Zhang, and Zhang]{deepseekai2025deepseekr1incentivizingreasoningcapability}
DeepSeek-AI, Daya Guo, Dejian Yang, Haowei Zhang, Junxiao Song, Ruoyu Zhang, Runxin Xu, Qihao Zhu, Shirong Ma, Peiyi Wang, Xiao Bi, Xiaokang Zhang, Xingkai Yu, Yu Wu, Z.~F. Wu, Zhibin Gou, Zhihong Shao, Zhuoshu Li, Ziyi Gao, Aixin Liu, Bing Xue, Bingxuan Wang, Bochao Wu, Bei Feng, Chengda Lu, Chenggang Zhao, Chengqi Deng, Chenyu Zhang, Chong Ruan, Damai Dai, Deli Chen, Dongjie Ji, Erhang Li, Fangyun Lin, Fucong Dai, Fuli Luo, Guangbo Hao, Guanting Chen, Guowei Li, H. Zhang, Han Bao, Hanwei Xu, Haocheng Wang, Honghui Ding, Huajian Xin, Huazuo Gao, Hui Qu, Hui Li, Jianzhong Guo, Jiashi Li, Jiawei Wang, Jingchang Chen, Jingyang Yuan, Junjie Qiu, Junlong Li, J.~L. Cai, Jiaqi Ni, Jian Liang, Jin Chen, Kai Dong, Kai Hu, Kaige Gao, Kang Guan, Kexin Huang, Kuai Yu, Lean Wang, Lecong Zhang, Liang Zhao, Litong Wang, Liyue Zhang, Lei Xu, Leyi Xia, Mingchuan Zhang, Minghua Zhang, Minghui Tang, Meng Li, Miaojun Wang, Mingming Li, Ning Tian, Panpan Huang, Peng Zhang, Qiancheng Wang, Qinyu Chen, Qiushi Du, Ruiqi Ge, Ruisong
  Zhang, Ruizhe Pan, Runji Wang, R.~J. Chen, R.~L. Jin, Ruyi Chen, Shanghao Lu, Shangyan Zhou, Shanhuang Chen, Shengfeng Ye, Shiyu Wang, Shuiping Yu, Shunfeng Zhou, Shuting Pan, S.~S. Li, Shuang Zhou, Shaoqing Wu, Shengfeng Ye, Tao Yun, Tian Pei, Tianyu Sun, T. Wang, Wangding Zeng, Wanjia Zhao, Wen Liu, Wenfeng Liang, Wenjun Gao, Wenqin Yu, Wentao Zhang, W.~L. Xiao, Wei An, Xiaodong Liu, Xiaohan Wang, Xiaokang Chen, Xiaotao Nie, Xin Cheng, Xin Liu, Xin Xie, Xingchao Liu, Xinyu Yang, Xinyuan Li, Xuecheng Su, Xuheng Lin, X.~Q. Li, Xiangyue Jin, Xiaojin Shen, Xiaosha Chen, Xiaowen Sun, Xiaoxiang Wang, Xinnan Song, Xinyi Zhou, Xianzu Wang, Xinxia Shan, Y.~K. Li, Y.~Q. Wang, Y.~X. Wei, Yang Zhang, Yanhong Xu, Yao Li, Yao Zhao, Yaofeng Sun, Yaohui Wang, Yi Yu, Yichao Zhang, Yifan Shi, Yiliang Xiong, Ying He, Yishi Piao, Yisong Wang, Yixuan Tan, Yiyang Ma, Yiyuan Liu, Yongqiang Guo, Yuan Ou, Yuduan Wang, Yue Gong, Yuheng Zou, Yujia He, Yunfan Xiong, Yuxiang Luo, Yuxiang You, Yuxuan Liu, Yuyang Zhou, Y.~X. Zhu,
  Yanhong Xu, Yanping Huang, Yaohui Li, Yi Zheng, Yuchen Zhu, Yunxian Ma, Ying Tang, Yukun Zha, Yuting Yan, Z.~Z. Ren, Zehui Ren, Zhangli Sha, Zhe Fu, Zhean Xu, Zhenda Xie, Zhengyan Zhang, Zhewen Hao, Zhicheng Ma, Zhigang Yan, Zhiyu Wu, Zihui Gu, Zijia Zhu, Zijun Liu, Zilin Li, Ziwei Xie, Ziyang Song, Zizheng Pan, Zhen Huang, Zhipeng Xu, Zhongyu Zhang, and Zhen Zhang.
\newblock Deepseek-r1: Incentivizing reasoning capability in llms via reinforcement learning, 2025.

\bibitem[Duan et~al.(2024)Duan, Yang, Qiao, Fang, Chen, Liu, Dong, Zang, Zhang, Wang, et~al.]{duan2024vlmevalkit}
Haodong Duan, Junming Yang, Yuxuan Qiao, Xinyu Fang, Lin Chen, Yuan Liu, Xiaoyi Dong, Yuhang Zang, Pan Zhang, Jiaqi Wang, et~al.
\newblock Vlmevalkit: An open-source toolkit for evaluating large multi-modality models.
\newblock In \emph{ACMMM}, pages 11198--11201, 2024.

\bibitem[Dubey et~al.(2024)Dubey, Jauhri, Pandey, Kadian, Al-Dahle, Letman, Mathur, Schelten, Yang, Fan, et~al.]{dubey2024llama}
Abhimanyu Dubey, Abhinav Jauhri, Abhinav Pandey, Abhishek Kadian, Ahmad Al-Dahle, Aiesha Letman, Akhil Mathur, Alan Schelten, Amy Yang, Angela Fan, et~al.
\newblock The llama 3 herd of models.
\newblock \emph{arXiv preprint arXiv:2407.21783}, 2024.

\bibitem[Estermann et~al.(2025)Estermann, Lanzend{\"o}rfer, Niedermayr, and Wattenhofer]{estermann2025puzzles}
Benjamin Estermann, Luca Lanzend{\"o}rfer, Yannick Niedermayr, and Roger Wattenhofer.
\newblock Puzzles: A benchmark for neural algorithmic reasoning.
\newblock \emph{NIPS}, 37:\penalty0 127059--127098, 2025.

\bibitem[Fu et~al.(2024)Fu, Hu, Li, Feng, Wang, Lin, Roth, Smith, Ma, and Krishna]{fu2024blink}
Xingyu Fu, Yushi Hu, Bangzheng Li, Yu Feng, Haoyu Wang, Xudong Lin, Dan Roth, Noah~A Smith, Wei-Chiu Ma, and Ranjay Krishna.
\newblock Blink: Multimodal large language models can see but not perceive.
\newblock \emph{arXiv preprint arXiv:2404.12390}, 2024.

\bibitem[Goyal et~al.(2017)Goyal, Khot, Summers-Stay, Batra, and Parikh]{goyal2017making}
Yash Goyal, Tejas Khot, Douglas Summers-Stay, Dhruv Batra, and Devi Parikh.
\newblock Making the v in vqa matter: Elevating the role of image understanding in visual question answering.
\newblock In \emph{CVPR}, pages 6904--6913, 2017.

\bibitem[Hao et~al.(2025)Hao, Gu, Wang, Li, Yang, Wang, and Cheng]{hao2025can}
Yunzhuo Hao, Jiawei Gu, Huichen~Will Wang, Linjie Li, Zhengyuan Yang, Lijuan Wang, and Yu Cheng.
\newblock Can mllms reason in multimodality? emma: An enhanced multimodal reasoning benchmark.
\newblock \emph{arXiv preprint arXiv:2501.05444}, 2025.

\bibitem[Hendrycks et~al.(2020)Hendrycks, Burns, Basart, Zou, Mazeika, Song, and Steinhardt]{hendrycks2020measuring}
Dan Hendrycks, Collin Burns, Steven Basart, Andy Zou, Mantas Mazeika, Dawn Song, and Jacob Steinhardt.
\newblock Measuring massive multitask language understanding.
\newblock \emph{arXiv preprint arXiv:2009.03300}, 2020.

\bibitem[Hendrycks et~al.(2021)Hendrycks, Burns, Kadavath, Arora, Basart, Tang, Song, and Steinhardt]{hendrycks2021measuring}
Dan Hendrycks, Collin Burns, Saurav Kadavath, Akul Arora, Steven Basart, Eric Tang, Dawn Song, and Jacob Steinhardt.
\newblock Measuring mathematical problem solving with the math dataset.
\newblock \emph{arXiv preprint arXiv:2103.03874}, 2021.

\bibitem[Hong et~al.(2023)Hong, Zheng, Chen, Cheng, Wang, Zhang, Wang, Yau, Lin, Zhou, et~al.]{hong2023metagpt}
Sirui Hong, Xiawu Zheng, Jonathan Chen, Yuheng Cheng, Jinlin Wang, Ceyao Zhang, Zili Wang, Steven Ka~Shing Yau, Zijuan Lin, Liyang Zhou, et~al.
\newblock Metagpt: Meta programming for multi-agent collaborative framework.
\newblock \emph{arXiv preprint arXiv:2308.00352}, 2023.

\bibitem[Hsu(2022)]{hsu2022behind}
Feng-Hsiung Hsu.
\newblock \emph{Behind Deep Blue: Building the computer that defeated the world chess champion}.
\newblock Princeton University Press, 2022.

\bibitem[Kamoi et~al.(2024)Kamoi, Zhang, Zhang, Han, and Zhang]{kamoi2024can}
Ryo Kamoi, Yusen Zhang, Nan Zhang, Jiawei Han, and Rui Zhang.
\newblock When can llms actually correct their own mistakes? a critical survey of self-correction of llms.
\newblock \emph{Transactions of the Association for Computational Linguistics}, 12:\penalty0 1417--1440, 2024.

\bibitem[Kwon et~al.(2023)Kwon, Li, Zhuang, Sheng, Zheng, Yu, Gonzalez, Zhang, and Stoica]{kwon2023efficient}
Woosuk Kwon, Zhuohan Li, Siyuan Zhuang, Ying Sheng, Lianmin Zheng, Cody~Hao Yu, Joseph~E. Gonzalez, Hao Zhang, and Ion Stoica.
\newblock Efficient memory management for large language model serving with pagedattention.
\newblock In \emph{Proceedings of the ACM SIGOPS 29th Symposium on Operating Systems Principles}, 2023.

\bibitem[Li et~al.(2023)Li, Wang, Wang, Ge, Ge, and Shan]{li2023seed}
Bohao Li, Rui Wang, Guangzhi Wang, Yuying Ge, Yixiao Ge, and Ying Shan.
\newblock Seed-bench: Benchmarking multimodal llms with generative comprehension.
\newblock \emph{arXiv preprint arXiv:2307.16125}, 2023.

\bibitem[Li et~al.(2024)Li, Zhang, Guo, Zhang, Li, Zhang, Zhang, Zhang, Li, Liu, et~al.]{li2024llava}
Bo Li, Yuanhan Zhang, Dong Guo, Renrui Zhang, Feng Li, Hao Zhang, Kaichen Zhang, Peiyuan Zhang, Yanwei Li, Ziwei Liu, et~al.
\newblock Llava-onevision: Easy visual task transfer.
\newblock \emph{arXiv preprint arXiv:2408.03326}, 2024.

\bibitem[Lin et~al.(2014)Lin, Maire, Belongie, Hays, Perona, Ramanan, Doll{\'a}r, and Zitnick]{lin2014microsoft}
Tsung-Yi Lin, Michael Maire, Serge Belongie, James Hays, Pietro Perona, Deva Ramanan, Piotr Doll{\'a}r, and C~Lawrence Zitnick.
\newblock Microsoft coco: Common objects in context.
\newblock In \emph{ECCV}, pages 740--755. Springer, 2014.

\bibitem[Liu et~al.(2023{\natexlab{a}})Liu, Li, Li, and Lee]{liu2023improvedllava}
Haotian Liu, Chunyuan Li, Yuheng Li, and Yong~Jae Lee.
\newblock Improved baselines with visual instruction tuning, 2023{\natexlab{a}}.

\bibitem[Liu et~al.(2023{\natexlab{b}})Liu, Li, Wu, and Lee]{liu2023llava}
Haotian Liu, Chunyuan Li, Qingyang Wu, and Yong~Jae Lee.
\newblock Visual instruction tuning, 2023{\natexlab{b}}.

\bibitem[Liu et~al.(2024)Liu, Li, Li, Li, Zhang, Shen, and Lee]{liu2024llavanext}
Haotian Liu, Chunyuan Li, Yuheng Li, Bo Li, Yuanhan Zhang, Sheng Shen, and Yong~Jae Lee.
\newblock Llava-next: Improved reasoning, ocr, and world knowledge, 2024.

\bibitem[Liu et~al.(2025)Liu, Duan, Zhang, Li, Zhang, Zhao, Yuan, Wang, He, Liu, et~al.]{liu2025mmbench}
Yuan Liu, Haodong Duan, Yuanhan Zhang, Bo Li, Songyang Zhang, Wangbo Zhao, Yike Yuan, Jiaqi Wang, Conghui He, Ziwei Liu, et~al.
\newblock Mmbench: Is your multi-modal model an all-around player?
\newblock In \emph{ECCV}, pages 216--233. Springer, 2025.

\bibitem[Lu et~al.(2022)Lu, Mishra, Xia, Qiu, Chang, Zhu, Tafjord, Clark, and Kalyan]{lu2022learn}
Pan Lu, Swaroop Mishra, Tanglin Xia, Liang Qiu, Kai-Wei Chang, Song-Chun Zhu, Oyvind Tafjord, Peter Clark, and Ashwin Kalyan.
\newblock Learn to explain: Multimodal reasoning via thought chains for science question answering.
\newblock \emph{NIPS}, 35:\penalty0 2507--2521, 2022.

\bibitem[{Meta AI Research}(2024)]{llama_3_2}
{Meta AI Research}.
\newblock {Llama 3.2: Revolutionizing edge AI and vision with open, customizable models}.
\newblock Technical report, Meta AI, 2024.
\newblock [Online; accessed 10-Jan-2025].

\bibitem[Metz(2024)]{Metz2024}
Cade Metz.
\newblock Openai unveils new ai model with advanced math and science capabilities.
\newblock \emph{The New York Times}, 2024.

\bibitem[Naman et~al.(2024)Naman, King, Alex, Wen-Ding, Fanjia, Tianjun, Sida, Armando, Koushik, and Ion]{jain2024livecodebench}
Jain Naman, Han King, Gu Alex, Li Wen-Ding, Yan Fanjia, Zhang Tianjun, Wang Sida, Solar-Lezama Armando, Sen Koushik, and Stoica Ion.
\newblock Livecodebench: Holistic and contamination free evaluation of large language models for code.
\newblock \emph{arXiv preprint}, 2024.

\bibitem[OpenAI(2024)]{chatgpt_4o}
OpenAI.
\newblock Hello gpt-4o, 2024.
\newblock Accessed: 2024-12-20.

\bibitem[Paglieri et~al.(2024)Paglieri, Cupia{\l}, Coward, Piterbarg, Wolczyk, Khan, Pignatelli, Kuci{\'n}ski, Pinto, Fergus, et~al.]{paglieri2024balrog}
Davide Paglieri, Bart{\l}omiej Cupia{\l}, Samuel Coward, Ulyana Piterbarg, Maciej Wolczyk, Akbir Khan, Eduardo Pignatelli, {\L}ukasz Kuci{\'n}ski, Lerrel Pinto, Rob Fergus, et~al.
\newblock Balrog: Benchmarking agentic llm and vlm reasoning on games.
\newblock \emph{arXiv preprint arXiv:2411.13543}, 2024.

\bibitem[Park et~al.(2023)Park, O'Brien, Cai, Morris, Liang, and Bernstein]{park2023generative}
Joon~Sung Park, Joseph O'Brien, Carrie~Jun Cai, Meredith~Ringel Morris, Percy Liang, and Michael~S Bernstein.
\newblock Generative agents: Interactive simulacra of human behavior.
\newblock In \emph{Proceedings of the 36th annual acm symposium on user interface software and technology}, pages 1--22, 2023.

\bibitem[P{\'e}rez et~al.(2003)P{\'e}rez, Gangnet, and Blake]{perez2003poisson}
Patrick P{\'e}rez, Michel Gangnet, and Andrew Blake.
\newblock Poisson image editing.
\newblock \emph{TOG}, 22\penalty0 (3):\penalty0 313--318, 2003.

\bibitem[{Puzzle Battleships}()]{puzzle_battleships}
{Puzzle Battleships}.
\newblock Battleships - online puzzle game.
\newblock \url{https://www.puzzle-battleships.com/}.
\newblock Accessed: 2025-01-17.

\bibitem[{Puzzlemix}()]{puzzlemix}
{Puzzlemix}.
\newblock Free puzzles to play online.
\newblock \url{https://www.puzzlemix.com/menu.php}.
\newblock Accessed: 2025-01-17.

\bibitem[{Puzzler Media}()]{puzzler_online_puzzles}
{Puzzler Media}.
\newblock Online puzzles, brain teasers and games.
\newblock \url{https://www.puzzler.com/online-puzzles}.
\newblock Accessed: 2025-01-17.

\bibitem[Qian et~al.(2023)Qian, Cong, Yang, Chen, Su, Xu, Liu, and Sun]{qian2023communicative}
Chen Qian, Xin Cong, Cheng Yang, Weize Chen, Yusheng Su, Juyuan Xu, Zhiyuan Liu, and Maosong Sun.
\newblock Communicative agents for software development.
\newblock \emph{arXiv preprint arXiv:2307.07924}, 6\penalty0 (3), 2023.

\bibitem[Radford et~al.(2019)Radford, Wu, Child, Luan, Amodei, Sutskever, et~al.]{radford2019language}
Alec Radford, Jeffrey Wu, Rewon Child, David Luan, Dario Amodei, Ilya Sutskever, et~al.
\newblock Language models are unsupervised multitask learners.
\newblock \emph{OpenAI blog}, 1\penalty0 (8):\penalty0 9, 2019.

\bibitem[Roberts et~al.(2025)Roberts, Taesiri, Sharma, Gupta, Roberts, Croitoru, Bogolin, Tang, Langer, Raina, et~al.]{roberts2025zerobench}
Jonathan Roberts, Mohammad~Reza Taesiri, Ansh Sharma, Akash Gupta, Samuel Roberts, Ioana Croitoru, Simion-Vlad Bogolin, Jialu Tang, Florian Langer, Vyas Raina, et~al.
\newblock Zerobench: An impossible visual benchmark for contemporary large multimodal models.
\newblock \emph{arXiv preprint arXiv:2502.09696}, 2025.

\bibitem[Schrittwieser et~al.(2020)Schrittwieser, Antonoglou, Hubert, Simonyan, Sifre, Schmitt, Guez, Lockhart, Hassabis, Graepel, et~al.]{schrittwieser2020mastering}
Julian Schrittwieser, Ioannis Antonoglou, Thomas Hubert, Karen Simonyan, Laurent Sifre, Simon Schmitt, Arthur Guez, Edward Lockhart, Demis Hassabis, Thore Graepel, et~al.
\newblock Mastering atari, go, chess and shogi by planning with a learned model.
\newblock \emph{Nature}, 588\penalty0 (7839):\penalty0 604--609, 2020.

\bibitem[Silver et~al.(2016)Silver, Huang, Maddison, Guez, Sifre, van~den Driessche, Schrittwieser, Antonoglou, Panneershelvam, Lanctot, et~al.]{silver2016mastering}
David Silver, Aja Huang, Chris~J Maddison, Arthur Guez, Laurent Sifre, George van~den Driessche, Julian Schrittwieser, Ioannis Antonoglou, Veda Panneershelvam, Marc Lanctot, et~al.
\newblock Mastering the game of go with deep neural networks and tree search.
\newblock \emph{Nature}, 529\penalty0 (7587):\penalty0 484--489, 2016.

\bibitem[Srivastava et~al.(2022)Srivastava, Rastogi, Rao, Shoeb, Abid, Fisch, Brown, Santoro, Gupta, Garriga-Alonso, et~al.]{srivastava2022beyond}
Aarohi Srivastava, Abhinav Rastogi, Abhishek Rao, Abu Awal~Md Shoeb, Abubakar Abid, Adam Fisch, Adam~R Brown, Adam Santoro, Aditya Gupta, Adri{\`a} Garriga-Alonso, et~al.
\newblock Beyond the imitation game: Quantifying and extrapolating the capabilities of language models.
\newblock \emph{arXiv preprint arXiv:2206.04615}, 2022.

\bibitem[Tan et~al.(2024)Tan, Ding, Zhang, Li, Zhou, Yue, Xia, Jiang, Zheng, Xu, et~al.]{tan2024towards}
Weihao Tan, Ziluo Ding, Wentao Zhang, Boyu Li, Bohan Zhou, Junpeng Yue, Haochong Xia, Jiechuan Jiang, Longtao Zheng, Xinrun Xu, et~al.
\newblock Towards general computer control: A multimodal agent for red dead redemption ii as a case study.
\newblock In \emph{ICLR 2024 Workshop on Large Language Model (LLM) Agents}, 2024.

\bibitem[Tang et~al.(2024)Tang, Leong, Shaik, and Lau]{tang2024automated}
Paul~Mingzheng Tang, Kenji Kah~Hoe Leong, Nowshad Shaik, and Hoong~Chuin Lau.
\newblock Automated conversion of static to dynamic scheduler via natural language.
\newblock \emph{arXiv preprint arXiv:2405.06697}, 2024.

\bibitem[Team et~al.(2023)Team, Anil, Borgeaud, Alayrac, Yu, Soricut, Schalkwyk, Dai, Hauth, Millican, et~al.]{team2023gemini}
Gemini Team, Rohan Anil, Sebastian Borgeaud, Jean-Baptiste Alayrac, Jiahui Yu, Radu Soricut, Johan Schalkwyk, Andrew~M Dai, Anja Hauth, Katie Millican, et~al.
\newblock Gemini: a family of highly capable multimodal models.
\newblock \emph{arXiv preprint arXiv:2312.11805}, 2023.

\bibitem[Touvron et~al.(2023)Touvron, Lavril, Izacard, Martinet, Lachaux, Lacroix, Rozi{\`e}re, Goyal, Hambro, Azhar, et~al.]{touvron2023llama}
Hugo Touvron, Thibaut Lavril, Gautier Izacard, Xavier Martinet, Marie-Anne Lachaux, Timoth{\'e}e Lacroix, Baptiste Rozi{\`e}re, Naman Goyal, Eric Hambro, Faisal Azhar, et~al.
\newblock Llama: Open and efficient foundation language models.
\newblock \emph{arXiv preprint arXiv:2302.13971}, 2023.

\bibitem[Wang et~al.(2019)Wang, Pruksachatkun, Nangia, Singh, Michael, Hill, Levy, and Bowman]{wang2019superglue}
Alex Wang, Yada Pruksachatkun, Nikita Nangia, Amanpreet Singh, Julian Michael, Felix Hill, Omer Levy, and Samuel Bowman.
\newblock Superglue: A stickier benchmark for general-purpose language understanding systems.
\newblock \emph{NIPS}, 32, 2019.

\bibitem[Wang et~al.(2025{\natexlab{a}})Wang, Lee, Menghini, Mols, Doughty, Khoja, Lynch, Hendryx, Yue, and Hendrycks]{wang2025enigmaeval}
Clinton~J Wang, Dean Lee, Cristina Menghini, Johannes Mols, Jack Doughty, Adam Khoja, Jayson Lynch, Sean Hendryx, Summer Yue, and Dan Hendrycks.
\newblock Enigmaeval: A benchmark of long multimodal reasoning challenges.
\newblock \emph{arXiv preprint arXiv:2502.08859}, 2025{\natexlab{a}}.

\bibitem[Wang et~al.(2023)Wang, Xie, Jiang, Mandlekar, Xiao, Zhu, Fan, and Anandkumar]{wang2023voyager}
Guanzhi Wang, Yuqi Xie, Yunfan Jiang, Ajay Mandlekar, Chaowei Xiao, Yuke Zhu, Linxi Fan, and Anima Anandkumar.
\newblock Voyager: An open-ended embodied agent with large language models.
\newblock \emph{arXiv preprint arXiv:2305.16291}, 2023.

\bibitem[Wang et~al.(2024)Wang, Li, Lian, Ma, Song, and Wei]{wang2024mitigating}
Weichuan Wang, Zhaoyi Li, Defu Lian, Chen Ma, Linqi Song, and Ying Wei.
\newblock Mitigating the language mismatch and repetition issues in llm-based machine translation via model editing.
\newblock \emph{arXiv preprint arXiv:2410.07054}, 2024.

\bibitem[Wang et~al.(2025{\natexlab{b}})Wang, Zhuang, and Wu]{wang2025are}
Xinyu Wang, Bohan Zhuang, and Qi Wu.
\newblock Are large vision language models good game players?
\newblock In \emph{The Thirteenth International Conference on Learning Representations}, 2025{\natexlab{b}}.

\bibitem[{Wikipedia contributors}(2024)]{wpc_wikipedia}
{Wikipedia contributors}.
\newblock World puzzle championship --- {W}ikipedia{,} the free encyclopedia.
\newblock \url{https://en.wikipedia.org/wiki/World_Puzzle_Championship}, 2024.
\newblock [Online; accessed 6-Dec-2024].

\bibitem[Wu et~al.(2023)Wu, Tang, Mitchell, and Li]{wu2023smartplay}
Yue Wu, Xuan Tang, Tom~M Mitchell, and Yuanzhi Li.
\newblock Smartplay: A benchmark for llms as intelligent agents.
\newblock \emph{arXiv preprint arXiv:2310.01557}, 2023.

\bibitem[Xu and Ma(2024)]{xu2024llm}
Nan Xu and Xuezhe Ma.
\newblock Llm the genius paradox: A linguistic and math expert's struggle with simple word-based counting problems.
\newblock \emph{arXiv preprint arXiv:2410.14166}, 2024.

\bibitem[Yin et~al.(2024)Yin, Wang, Cao, Shi, Liu, Li, Huang, Wang, Sheng, Bai, et~al.]{yin2024lamm}
Zhenfei Yin, Jiong Wang, Jianjian Cao, Zhelun Shi, Dingning Liu, Mukai Li, Xiaoshui Huang, Zhiyong Wang, Lu Sheng, Lei Bai, et~al.
\newblock Lamm: Language-assisted multi-modal instruction-tuning dataset, framework, and benchmark.
\newblock \emph{NIPS}, 36, 2024.

\bibitem[Zawalski et~al.(2024)Zawalski, Chen, Pertsch, Mees, Finn, and Levine]{zawalski2024robotic}
Micha{\l} Zawalski, William Chen, Karl Pertsch, Oier Mees, Chelsea Finn, and Sergey Levine.
\newblock Robotic control via embodied chain-of-thought reasoning.
\newblock \emph{arXiv preprint arXiv:2407.08693}, 2024.

\bibitem[Zhai et~al.(2024)Zhai, Bai, Lin, Pan, Tong, Zhou, Suhr, Xie, LeCun, Ma, and Levine]{zhai2024finetuning}
Yuexiang Zhai, Hao Bai, Zipeng Lin, Jiayi Pan, Shengbang Tong, Yifei Zhou, Alane Suhr, Saining Xie, Yann LeCun, Yi Ma, and Sergey Levine.
\newblock Fine-tuning large vision-language models as decision-making agents via reinforcement learning.
\newblock In \emph{NIPS}, 2024.

\bibitem[Zhang et~al.(2024{\natexlab{a}})Zhang, Guo, Guo, Cao, Huang, Liu, and Zhang]{zhang2024ing}
Haoran Zhang, Hangyu Guo, Shuyue Guo, Meng Cao, Wenhao Huang, Jiaheng Liu, and Ge Zhang.
\newblock Ing-vp: Mllms cannot play easy vision-based games yet.
\newblock \emph{arXiv preprint arXiv:2410.06555}, 2024{\natexlab{a}}.

\bibitem[Zhang et~al.(2024{\natexlab{b}})Zhang, Huang, Ma, Michel, He, Gupta, Ma, Farhadi, Kembhavi, and Krishna]{zhang2024task}
Jieyu Zhang, Weikai Huang, Zixian Ma, Oscar Michel, Dong He, Tanmay Gupta, Wei-Chiu Ma, Ali Farhadi, Aniruddha Kembhavi, and Ranjay Krishna.
\newblock Task me anything.
\newblock \emph{arXiv preprint arXiv:2406.11775}, 2024{\natexlab{b}}.

\bibitem[Zhou et~al.(2023)Zhou, Schubert, Toussaint, and Oguz]{zhou2023spatial}
Hongyou Zhou, Ingmar Schubert, Marc Toussaint, and Ozgur~S Oguz.
\newblock Spatial reasoning via deep vision models for robotic sequential manipulation.
\newblock In \emph{IROS}, pages 11328--11335. IEEE, 2023.

\bibitem[Zhu et~al.(2023)Zhu, Chen, Shen, Li, and Elhoseiny]{zhu2023minigpt}
Deyao Zhu, Jun Chen, Xiaoqian Shen, Xiang Li, and Mohamed Elhoseiny.
\newblock Minigpt-4: Enhancing vision-language understanding with advanced large language models.
\newblock \emph{arXiv preprint arXiv:2304.10592}, 2023.

\end{thebibliography}
}

\clearpage
\setcounter{page}{1}

\onecolumn

\tableofcontents

\clearpage

\section{Versions of Models Used}
\label{sec:model-version}
Since off-the-shelf models may exist in multiple versions even under the same name, we provide the specific version numbers for the models used, where applicable, in the list below.

\begin{itemize}
    \item Gemini 2.0 Flash Thinking: gemini-2.0-flash-thinking-exp-01-21
    \item QVQ-72B: Qwen/QVQ-72B-Preview
    \item Claude 3.5: claude-3-5-sonnet-20241022 
    \item Gemini 2.0 Flash: gemini-2.0-flash
    \item Gemini 1.5 Pro: gemini-1.5-pro
    \item Gemini 1.5 Flash: gemini-1.5-flash
    \item GPT-4o: gpt-4o-2024-08-06
    \item GPT-4o-mini: gpt-4o-mini-2024-07-18
    \item Qwen2.5-VL-72B-it: Qwen/Qwen2.5-VL-72B-Instruct
    \item Qwen2.5-VL-7B-it: Qwen/Qwen2.5-VL-7B-Instruct
    \item Llama 3.2 90B it: Llama-3.2-90B-Vision-Instruct
    \item Llama 3.2 11B it: Llama-3.2-11B-Vision-Instruct
    \item Qwen2-VL-7B-it: Qwen/Qwen2-VL-7B-Instruct
    \item Qwen2-VL-72B-it: Qwen/Qwen2-VL-72B-Instruct
    \item llava-onevision-7B: llava-hf/llava-onevision-qwen2-7b-ov-hf
    \item llava-mistral-7B: llava-hf/llava-v1.6-mistral-7b-hf
\end{itemize}

\section{Additional Evaluation Details}
\label{sec:additional-evaluation-details}
We observed that many off-the-shelf models—particularly open-source, small-scale ones such as Llama 3.2 11B—do not strictly adhere to the required output format. For example, their outputs frequently omit closing quotes or inconsistently alternate between single and double quotes, rendering them unsuitable for JSON parsing. Furthermore, these LVLMs often fail to follow instructions for representing empty cells. Instead of using the designated symbols (e.g., "0", "", ".", or "\_") to denote an empty cell, they sometimes default to using "*". In the chain-of-thought (CoT) setting, the final answer may not appear at the end of the sentence, further complicating extraction via regular expressions. Since our primary focus is on assessing puzzle-solving capabilities, rather than outright rejecting non-compliant responses, we have implemented two targeted post-processing strategies to address these issues.
First, we employ the json-repair package~\cite{Baccianella_JSON_Repair_2024} to repair broken JSON outputs, addressing issues such as incorrect escape characters and inconsistent usage of single and double quotes. Second, we utilize an LLM—specifically GPT-4o—as an output formatter to standardize the outputs into a unified format.

\section{Additional Implementation details}
We provide each model's version id in Supp.\ref{sec:model-version}. 
For the closed-source models, we directly use their official API calls to query. For the open-source models, we deploy them locally using the vLLM framework\cite{kwon2023efficient} before querying.
For a fair comparison, none of the closed-source LVLMs dynamically construct a solver to solve the puzzles. 
Instead, they rely solely on vision-language modeling capabilities to solve puzzles.
For dataset creation, we compile a dataset of 100,000 puzzles per puzzle type and split it into training and validation sets, ensuring there are no duplicated conditions or solutions within or across splits. 
Note that we do not enforce unique solutions for the games, as doing so would significantly reduce the number of possible conditions and increase computational overhead during game generation. 
For both puzzle-solution and reasoning fine-tuning, we use LLaMa 3.2 Vision Instruct as the base model and employ one node with 8 A100 GPUs, using a batch size of 4 on each GPU.
Training for 5 epochs with the \textit{llama-recipe} code base and default fine-tuning settings for LVLMs takes approximately 24 hours. %
The training process follows a causal modeling paradigm, masking the input question to focus supervision solely on the answer predictions. 
Number of clues, considering that puzzles typically become more challenging with fewer clues. For reveal-based games, such as Sudoku, at the easy level we randomly select 25\% to 75\% of cells as hints. For the medium level, we randomly select 25\% to 50\% as clues. For the hard level, we randomly select 15\% to 40\% as clues.

\section{Image Augmentations}

In our framework, we strategically employ two light-weight augmentation techniques to enhance model generalization across diverse visual inputs: (1) randomized affine transforms for geometric variation simulation, and (2) Poisson blending~\cite{perez2003poisson} for seamless texture integration.This streamlined augmentation approach effectively bridges the domain gap between training samples and real-world scenarios while preserving the pre-trained model's inherent visual understanding capabilities. As demonstrated in Table~\ref{tab:poisson-blending}, our method achieves realistic image transformations with significantly fewer artifacts compared to conventional augmentation strategies.

\begin{table}[h]
    \centering
    \begin{adjustbox}{max width=0.7\textwidth}
    \begin{tabular}{|c|c|c|}
    \hline
    \multicolumn{3}{|c|}{\textbf{Qualitative Comparison of Model Performance on Sudoku}} \\ \hline
    \includegraphics[width=40mm]{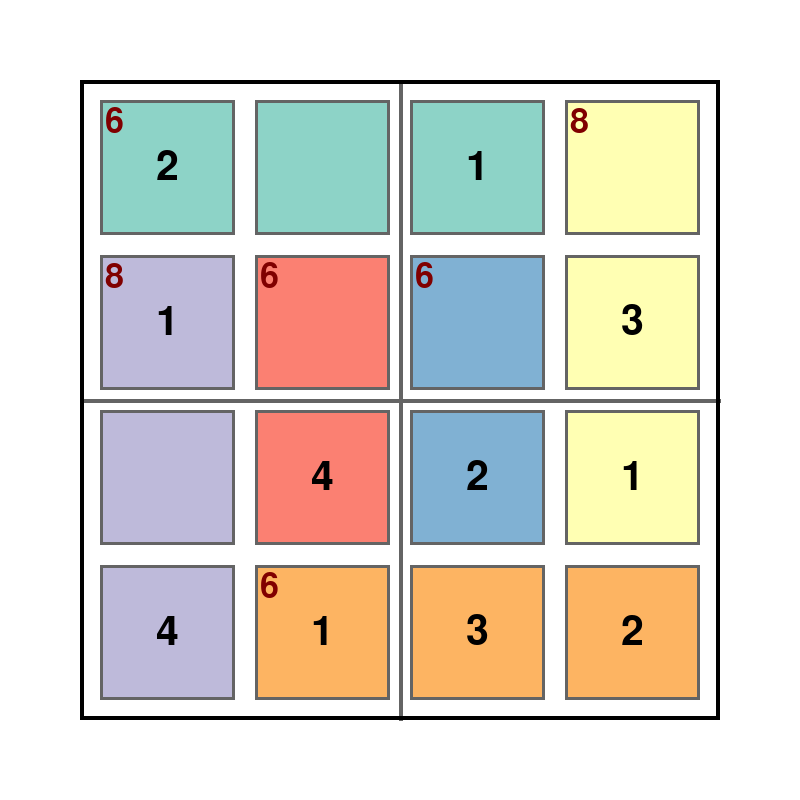} & 
    \includegraphics[width=40mm]{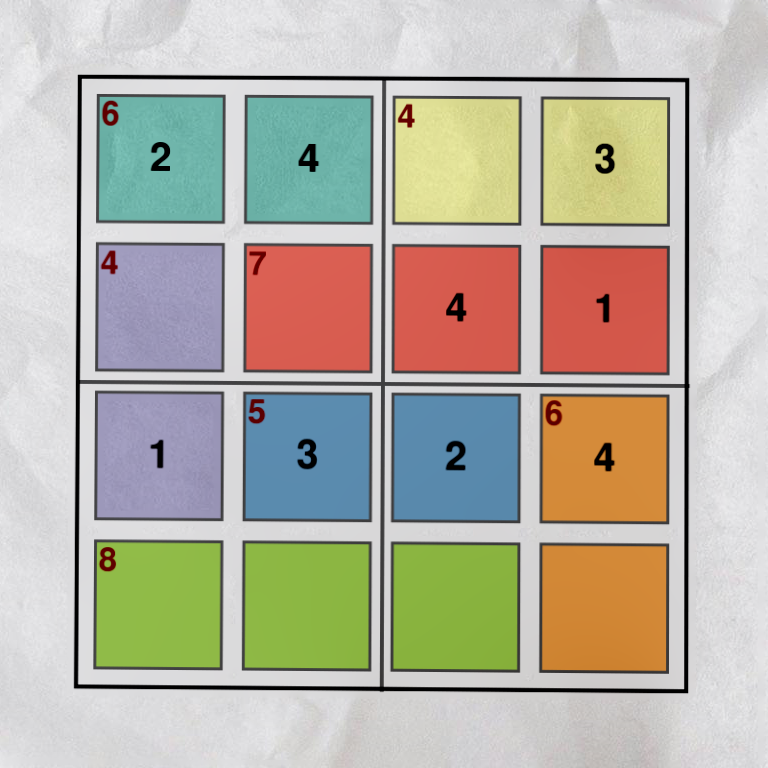} & 
    \includegraphics[width=40mm]{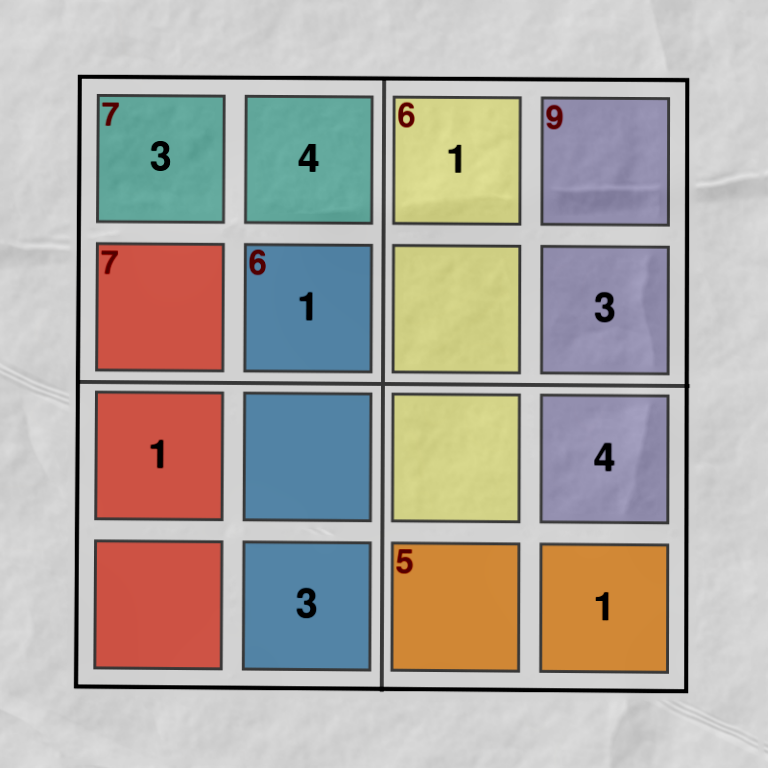} \\ \hline
    \end{tabular}
    \end{adjustbox}
    \caption{Examples of Image Augmentations}
    \label{tab:poisson-blending}
\end{table}

\section{Common Mistakes of LVLMs in Puzzle Solving}
\label{example-output-and-common-mistakes}

\subsection{Common Mistakes in LVLMs' Perception}
\label{common-mistakes-perception}

\noindent \textbf{Grid Layout Misunderstanding.} As shown in 
Tab.~\ref{supp-tab-sudoku-vision-easy}, 
GPT-4o fails to accurately perceive the Sudoku board layout—for instance, it erroneously places a ``2" in the first row where the cell should remain empty. Although LVLMs generally recognize numerical values, they often misplace them; for example, they might confuse a sequence intended as [*, 2, *, ] with [, *, 2, *]. Notably, performance improves as more hints are provided, as observed in Sudoku tasks, suggesting that LVLMs struggle to detect board gridlines, which are essential for deducing the correct board organization.

\noindent \textbf{Misinterpretation of Component Roles.} In visual puzzles, distinct components serve unique functions. For example, in Killer-Sudoku, the grid incorporates both cell numbers and cage sums. However, LVLMs frequently misinterpret cage sums as cell values, even when differences in font size and color are present.

\noindent \textbf{Difficulty Understanding Complex Visual Components.} 
In puzzles like Thermometers and Battle-Ships, we notice that LVLMs sometimes fail to comprehend complex visual components. For example, LVLMs may erroneously perceive an empty cell in Thermometers as filled, and they often struggle to discern the distinct roles of different ship segments in Battle-Ships.

\noindent \textbf{Rejection of answering. } 
We noticed that sometimes a model responds with a message such as ``I'm sorry, I can't view or process images directly. Could you please describe the puzzle to me in text form?” to avoid answering, even though it sometimes does provide an answer. We hypothesize that the criteria for providing or rejecting an answer may depend on the model's internal confidence level~\cite{becker2024cycles}.

\subsection{Common Mistakes in Pre-Trained LVLMs' Puzzle-Solving}
\label{common-mistakes-reasoning}

\noindent \textbf{Cell-by-Cell Solving without Prioritizing.} 
LVLMs often solve puzzles in a strictly sequential manner, overlooking constraints that emerge in later steps and failing to exploit the fact that some cells, with fewer possible options, are easier to resolve. For example, as demonstrated by Gemini-2.0-Flash in Tab.~\ref{supp-sudoku-qualitative-text}, the LVLM does not begin with the final cell in the second row, which has only one option.

\noindent \textbf{Failure to Store Previous Actions.} LVLMs frequently fail to update their understanding of the puzzle state based on previous actions. An example is an LVLM places a number at one cell. However, when making later decisions, it just ignored the previous actions it take. 

\noindent \textbf{Inability to Detect Error and Backtrack~\cite{kamoi2024can}.} 
As is shown in the GPT-4o-Mini example in text version sudoku in Tab.~\ref{supp-sudoku-qualitative-text}, the LVLM finally outputs [2,1,4,2] as the last row, where a clear violation of duplication of number ``2" exist. However, GPT-4o-Mini replies it as an answer. Another example is Qwen2.5-72B and Qwen2.5-7B in the same table, we noticed that model outputs a possibly valid board but does not obey the initial condition. 

\noindent \textbf{Repetitive Generation.} Repetitive output is a common issue~\cite{wang2024mitigating}, especially in the LLaVA and Qwen families of models. For example, as shown in Tab.~\ref{supp-tab-sudoku-vision-easy}, the model generates excessively repetitive sequences, such as repeating ``perception" followed by more than 50 ``[*]" symbols. We also notice the repetitive generation issue frequently in the reasoning models QVQ, that it reason many steps until running out of the maximum context length, as shown in Tab.~\ref{supp-tab-sudoku-vision-easy}.

\section{Generalization Capability of SFT Models}
\label{evaluation-generalization}

\begin{figure}[h]
\center
\includegraphics[width=0.4\textwidth, page=1]{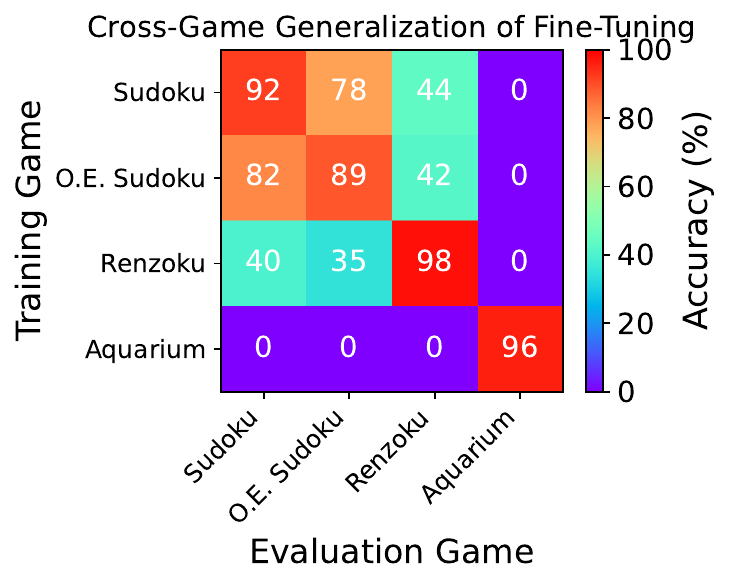} \caption{\textbf{Generalization Evaluation of SFT Models on the Level \Easy.}
(Best viewed on a screen when zoomed in)} \label{fig:generalization} 
\end{figure}

Here, we analyze how an LVLM with SFT on one puzzle generalizes to another. In Fig.\ref{fig:generalization}, we use four puzzles—Sudoku, Odd-Even Sudoku, Renzoku, and Aquarium—with increasingly different rules. Notably, despite the fact that the Llama model trained on different puzzles performs well on the same puzzle, their performance on other puzzles is significantly lower. When the rules are similar, e.g., Sudoku and Odd-Even Sudoku, the puzzle-solving rate under generalization remains high, while evaluating a Sudoku-trained model on Aquarium yields a zero success rate. We notice similar situation for both S-SFT and R-SFT. Some recent work has also discussed the generalization limitations of Supervised Fine-Tuning~\cite{chu2025sft,zhai2024finetuning}.

\section{Reinforcement Learning with Text Input}

PPO stabilizes training by incorporating a clipping mechanism into the policy ratio, which prevents excessive deviations from the previous policy, where \( r_t = \frac{\pi_{\theta}(a_t | s_t)}{\pi_{\theta_{\text{old}}}(a_t | s_t)} \) is the policy ratio.

\begin{equation}
\mathbb{E} \left[ \min \left( r_t \hat{A}_t, \operatorname{clip}(r_t, 1 - \epsilon, 1 + \epsilon) \hat{A}_t \right) \right],
\end{equation}

GRPO extends this framework by optimizing over groups of trajectories and regularizing updates with an explicit KL penalty, where \( r_{i,t} \) compares token probabilities under the updated and old policies, conditioned on \( q \) and prior outputs.

\begin{equation}
\begin{aligned}
&\mathbb{E} \Bigg[ \frac{1}{G} \sum_{i=1}^{G} \frac{1}{|o_i|} \sum_{t=1}^{|o_i|} 
\min \left( r_{i,t} \hat{A}_{i,t}, \operatorname{clip}(r_{i,t}, 1 \!-\! \epsilon, 1 \!+\! \epsilon) \hat{A}_{i,t} \right) 
\Bigg] \\
&- \beta D_{\text{KL}}(\pi_{\theta} \| \pi_{\text{ref}})
\end{aligned}
\end{equation}

To ensure stable training, GRPO calculate the advantage by normalizing reward with all rollouts: 

\begin{equation}
\hat{A}_{i,t} = \widetilde{r}_i = \frac{r_i - \operatorname{mean}(\mathbf{r})}{\operatorname{std}(\mathbf{r})}.
\end{equation}

The total reward consists of:
success reward $r_{\text{succ}}$ for generating the correct solution, and format reward $r_{\text{fmt}}$ for producing a structured, extractable output, where $\lambda_{\text{succ}}$ and $\lambda_{\text{fmt}}$ balance the two components. We provide the training loss curve on page \pageref{supp-RL}. 

\begin{equation}
r_i = \lambda_{\text{succ}} r_{\text{succ}} + \lambda_{\text{fmt}} r_{\text{fmt}},
\end{equation}

\begin{table*}[p]
    \centering

    \refstepcounter{section}
    \phantomsection         
    \addcontentsline{toc}{section}{Visualizations} 
    \addcontentsline{toc}{subsection}{Per-Puzzle Examples of Easy, Medium, and Hard Levels} 
    
    \renewcommand{\arraystretch}{1.5} 
    \resizebox{0.9\textwidth}{!}{%
    \begin{tabular}{|c|c|c|c|}
    \hline

    1. Aquarium & 2. Battle-Ships & 3. Binairo & 4. Colored-Sudoku \\ \hline

    {\includegraphics[height=55mm]{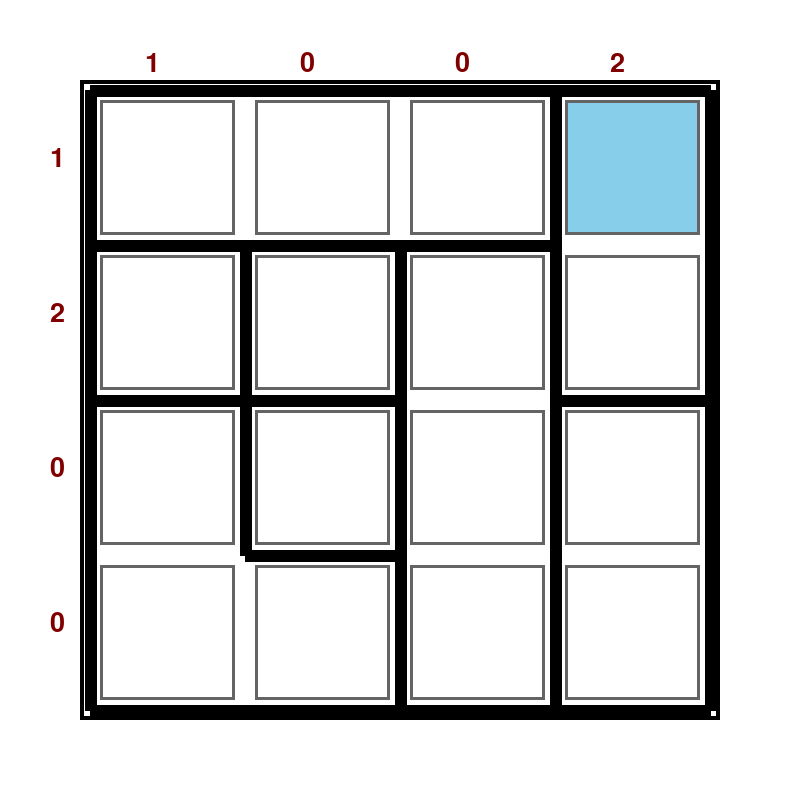}} & {\includegraphics[height=55mm]{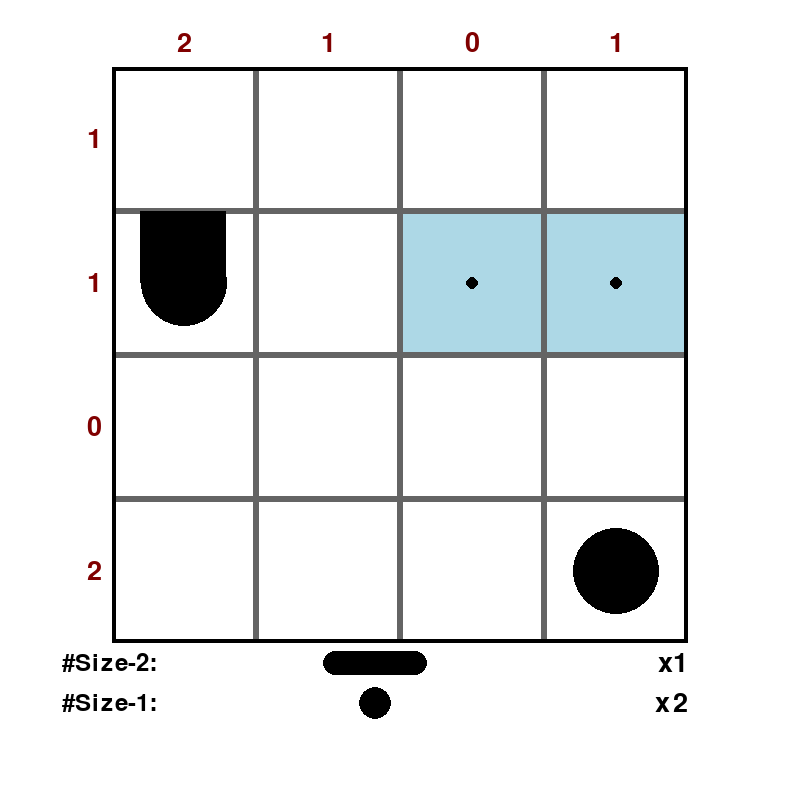}} & {\includegraphics[height=55mm]{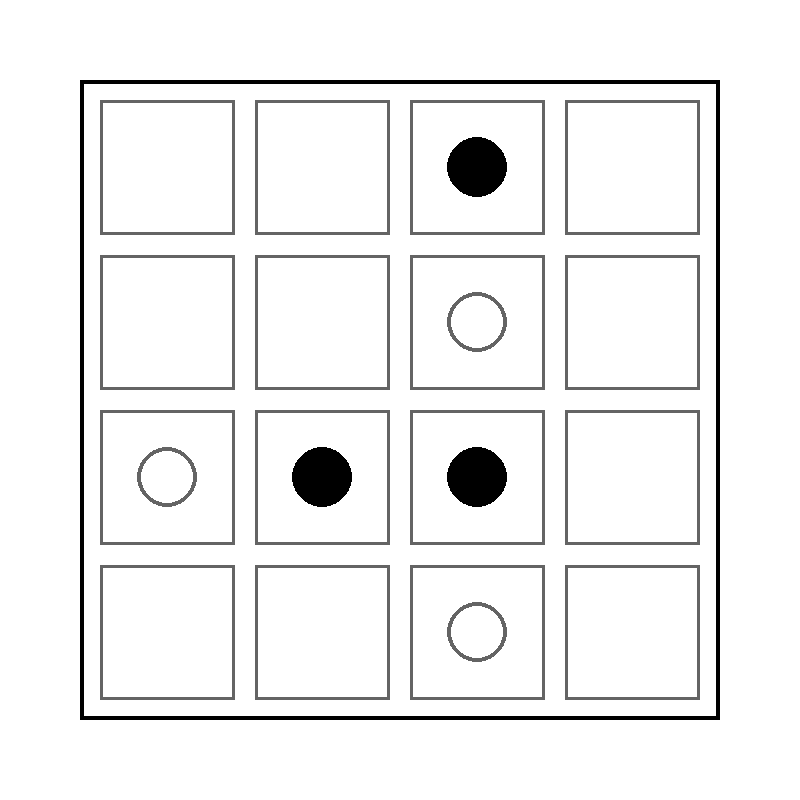}} & {\includegraphics[height=55mm]{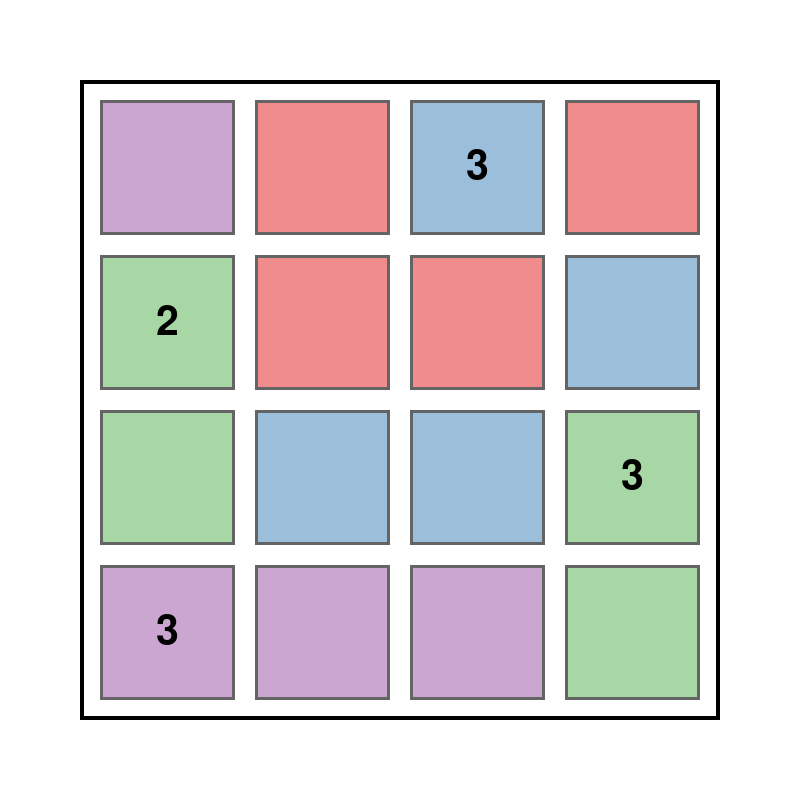}} \\ \hline

    5. Field-Explore & 6. Futoshiki & 7. Hitori & 8. Jigsaw-Sudoku \\ \hline

    {\includegraphics[height=55mm]{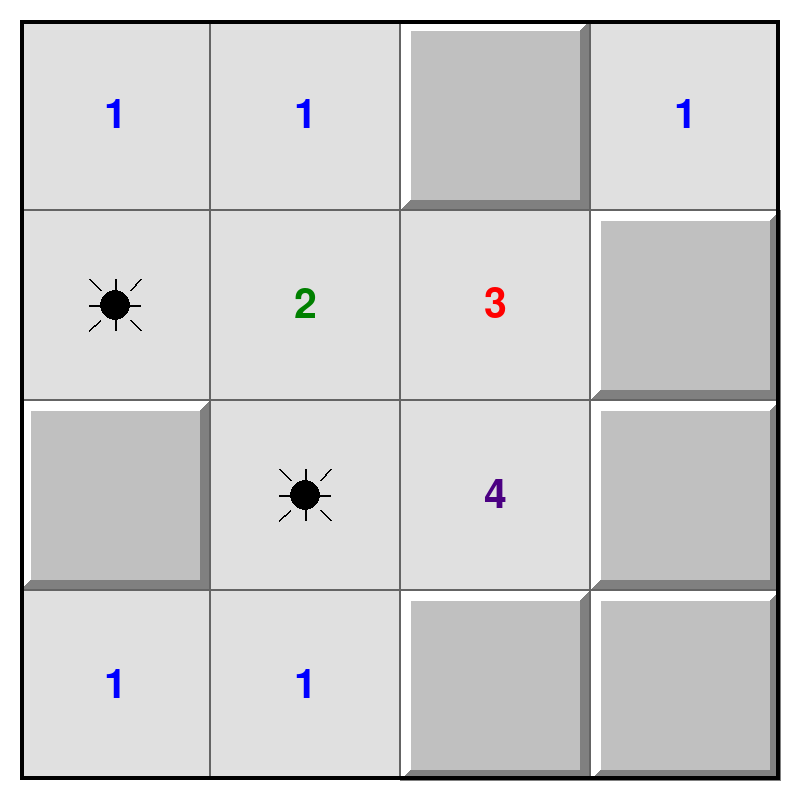}} & {\includegraphics[height=55mm]{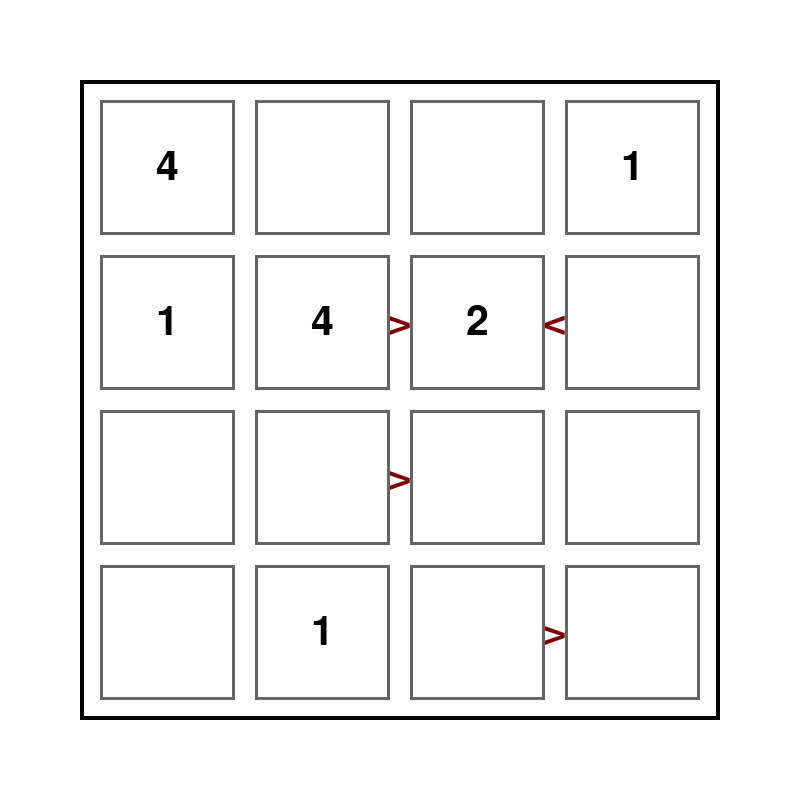}} & {\includegraphics[height=55mm]{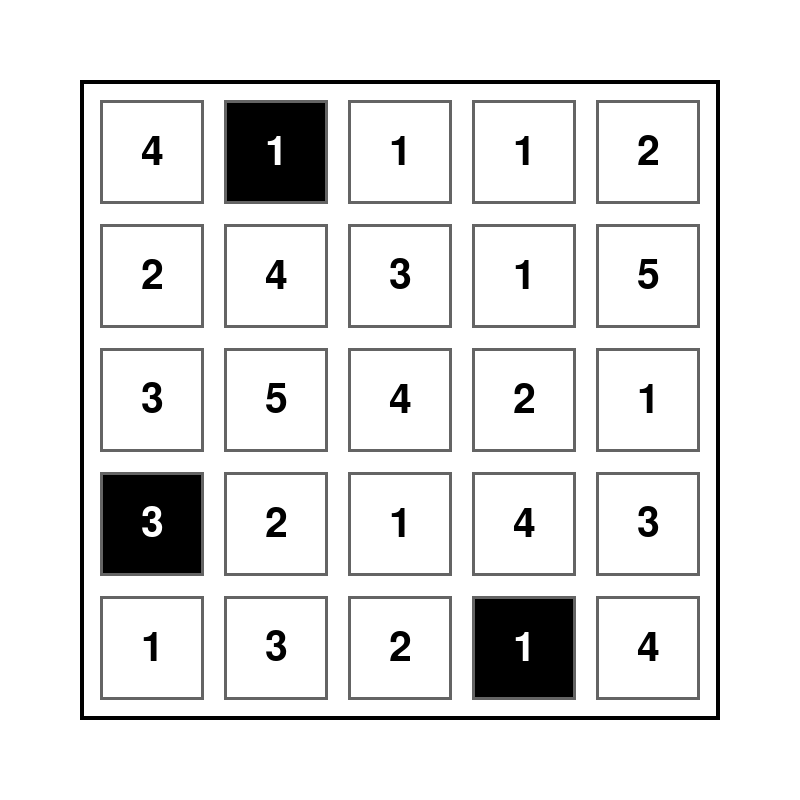}} & {\includegraphics[height=55mm]{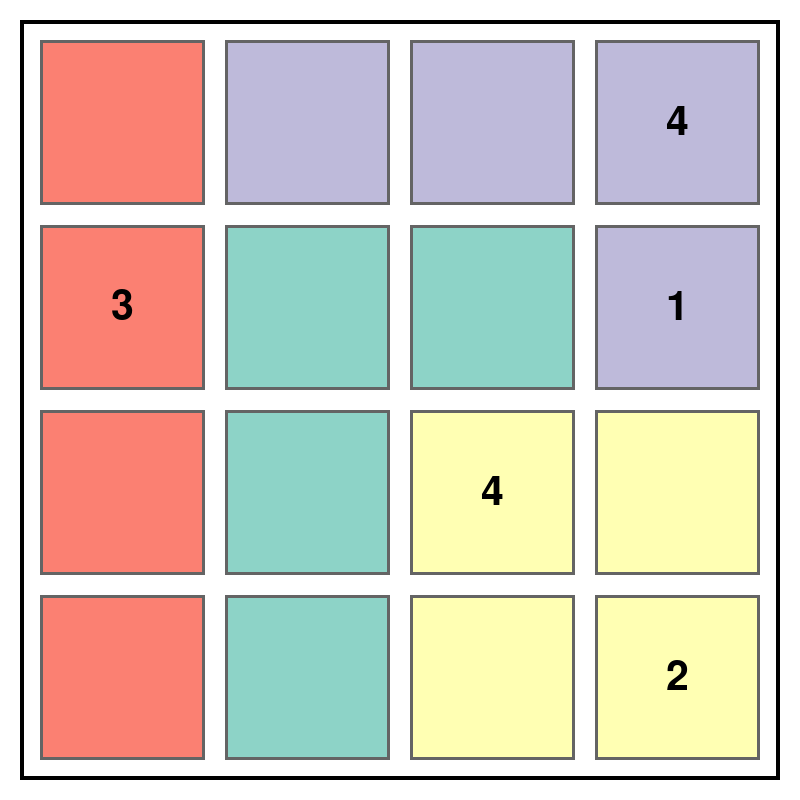}
    } \\ \hline

    9. Kakurasu & 10. Kakuro & 11. Killer-Sudoku & 12. Light-Up \\ \hline

    {\includegraphics[height=55mm]{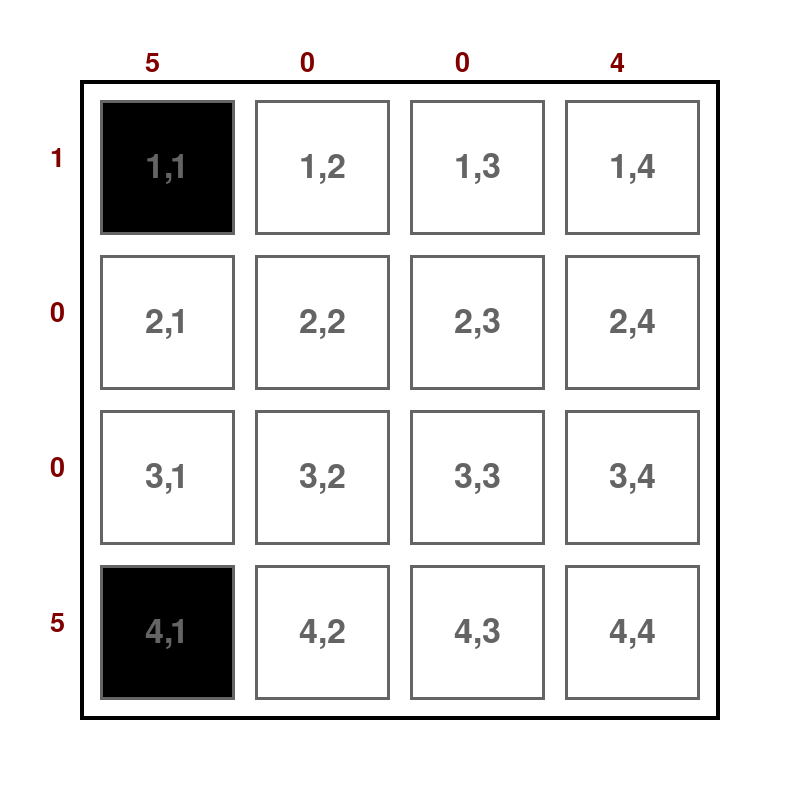}} & {\includegraphics[height=55mm]{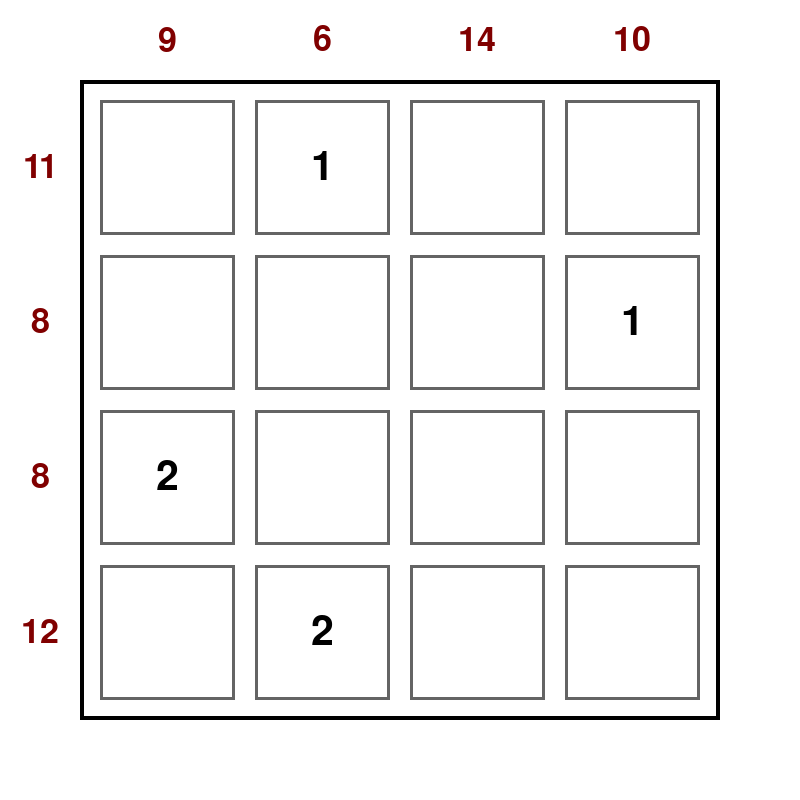}} & {\includegraphics[height=55mm]{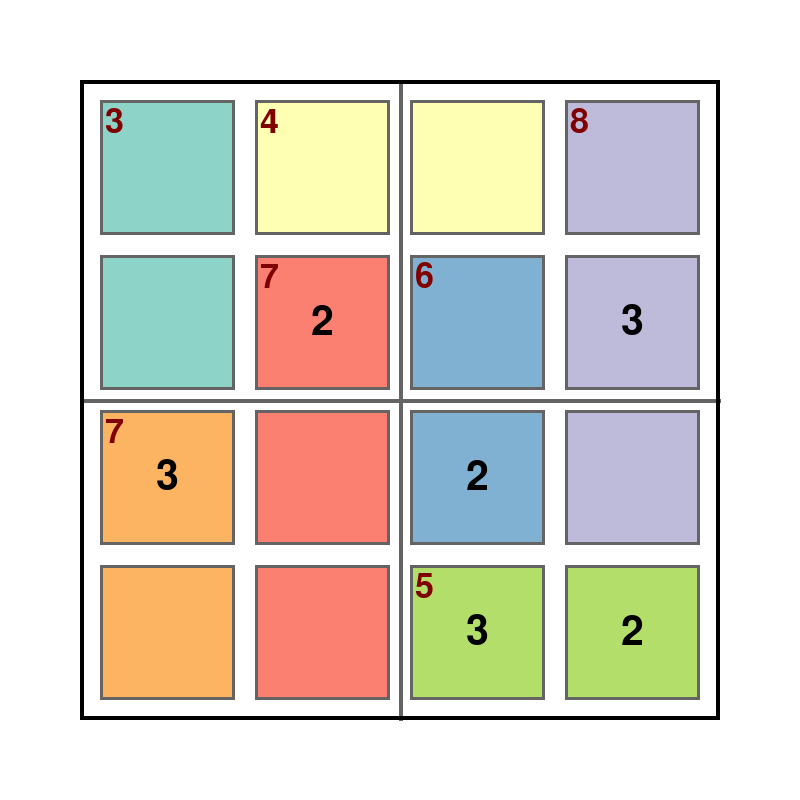}} & {\includegraphics[height=55mm]{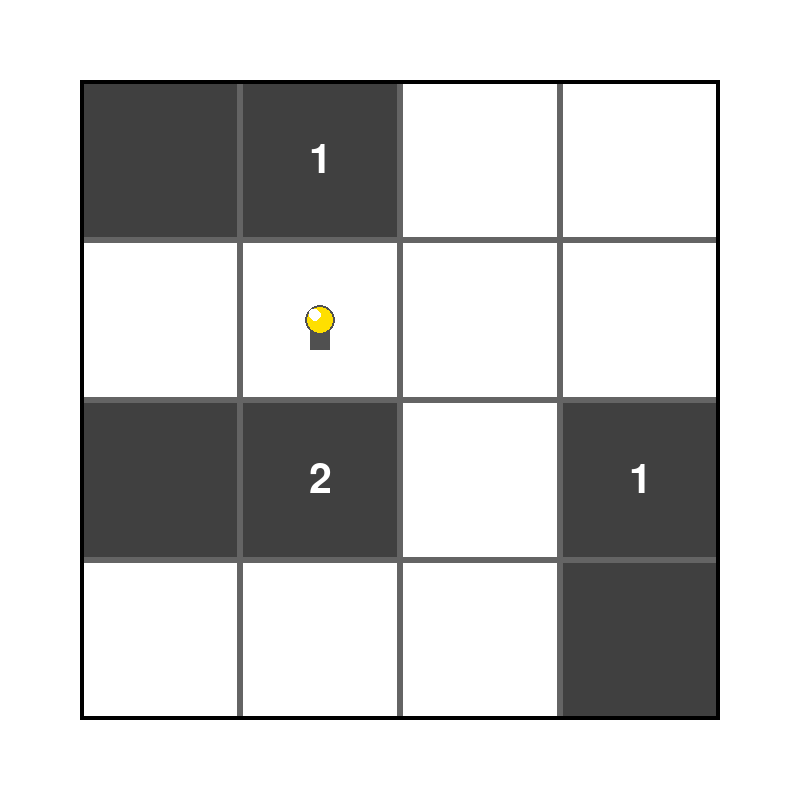}} \\ \hline
    
    13. Nonogram & 14. Odd-Even-Sudoku & 15. Renzoku & 16. Skyscraper \\ \hline

    {\includegraphics[height=55mm]{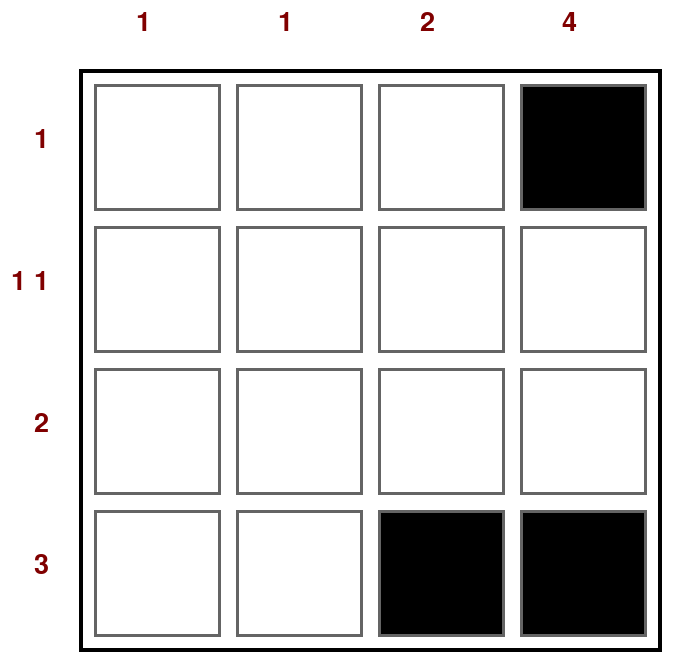}} & {\includegraphics[height=55mm]{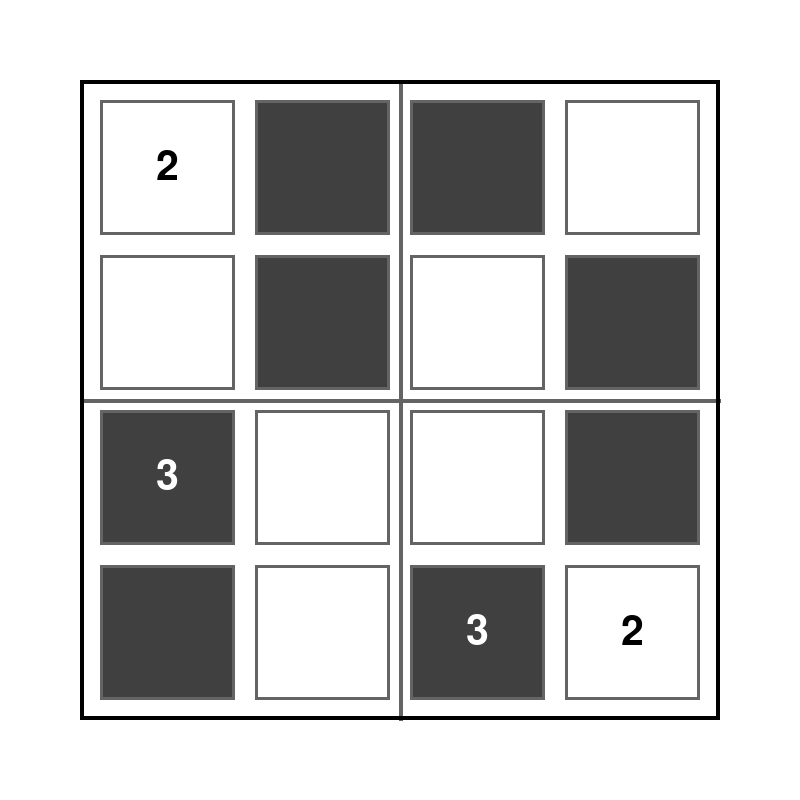}} & {\includegraphics[height=55mm]{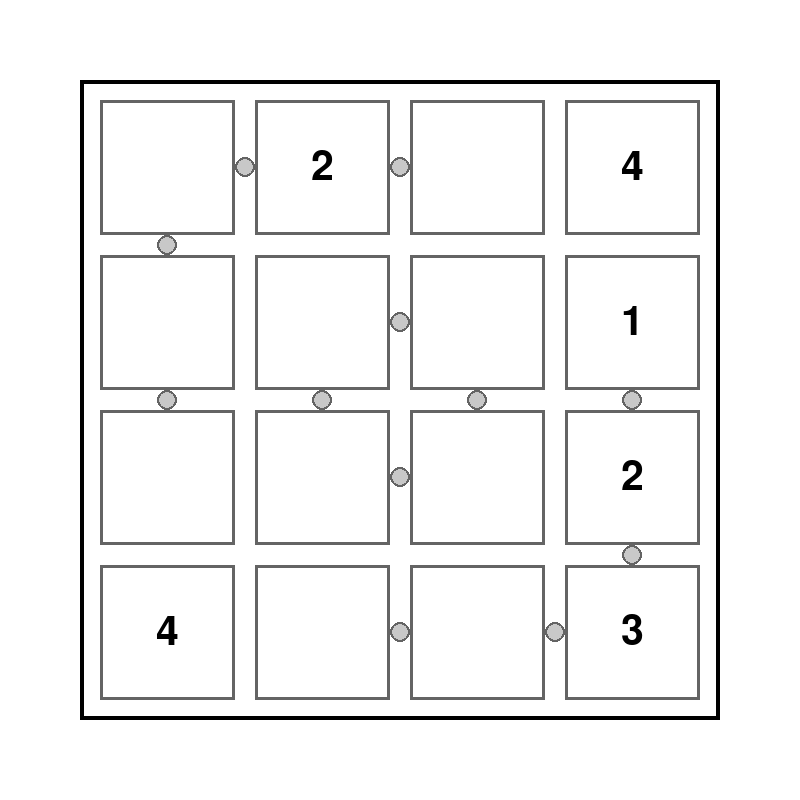}} & {\includegraphics[height=55mm]{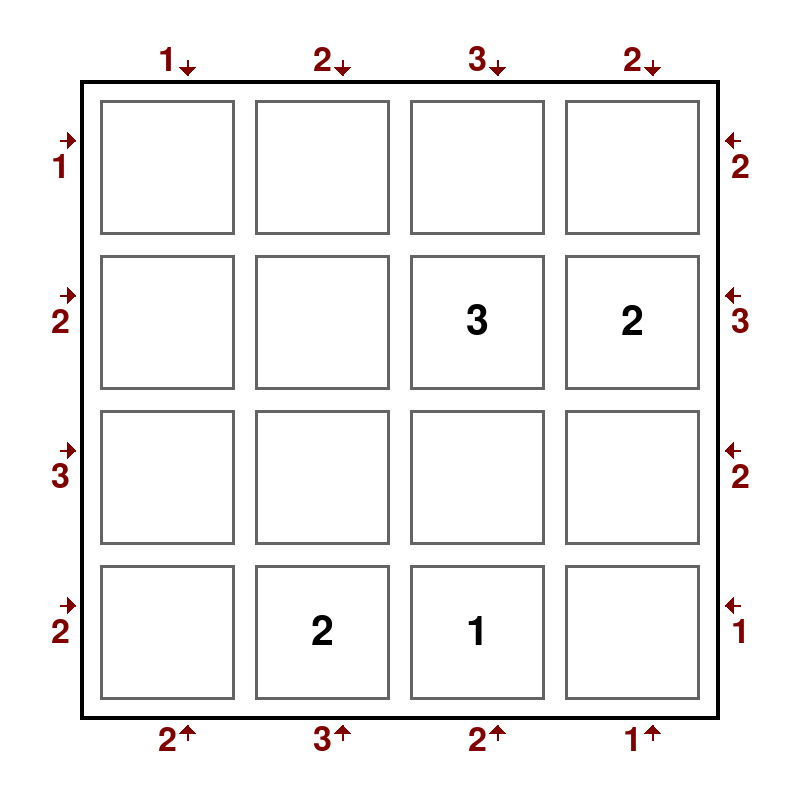}} \\ \hline

    17. Star-Battle & 18. Sudoku & 19. Thermometers & 20. Trees-and-Tents \\ \hline

    {\includegraphics[height=55mm]{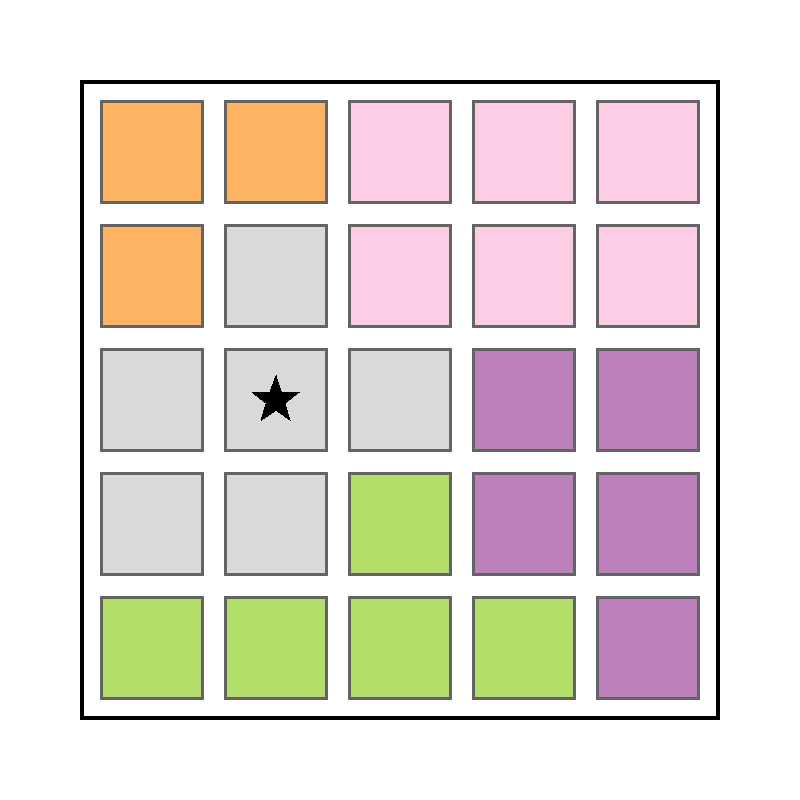}} & {\includegraphics[height=55mm]{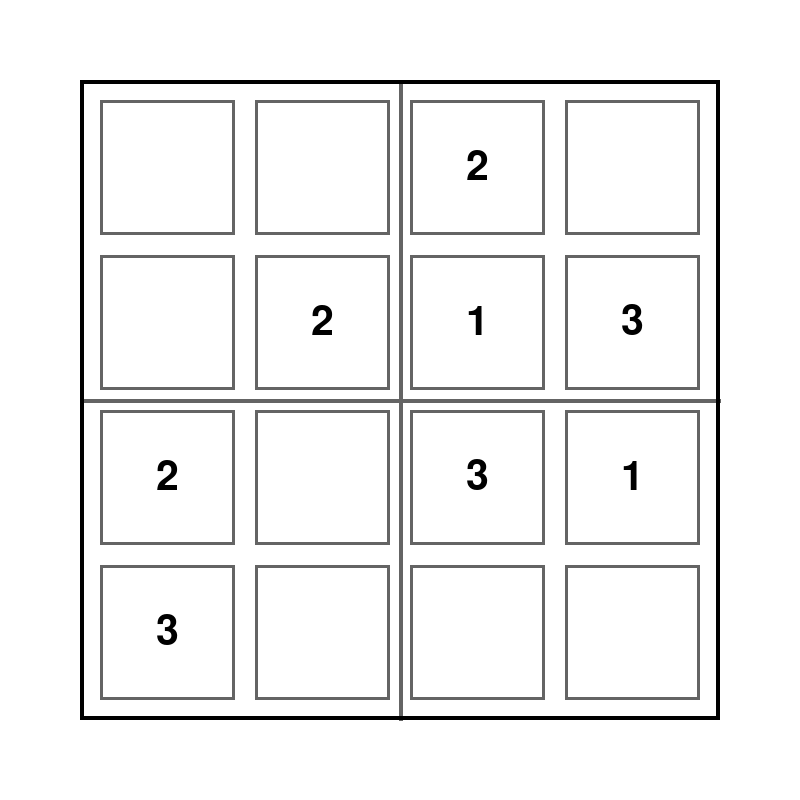}} & {\includegraphics[height=55mm]{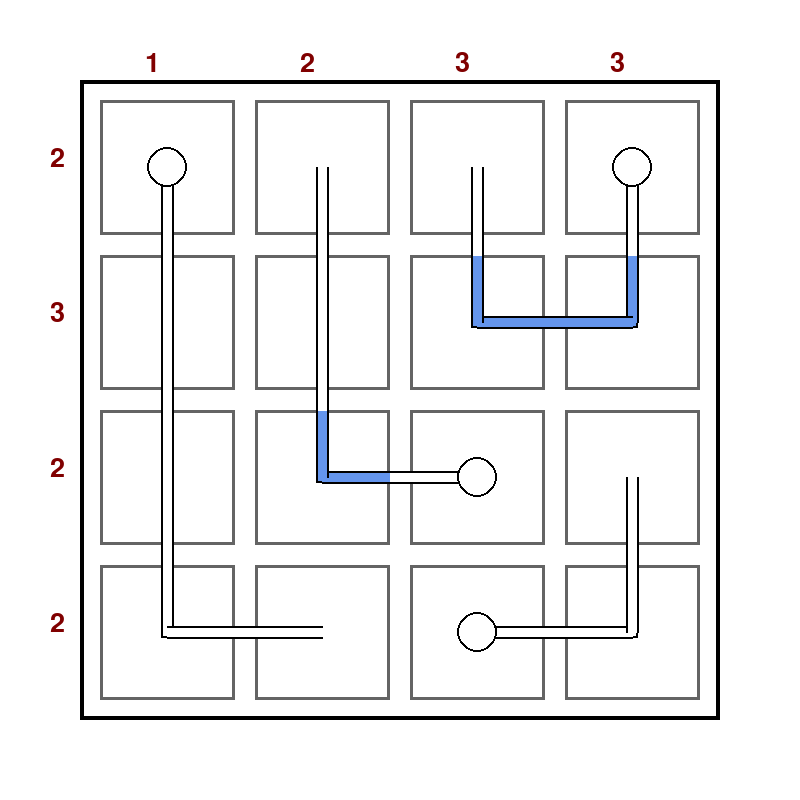}} & {\includegraphics[height=55mm]{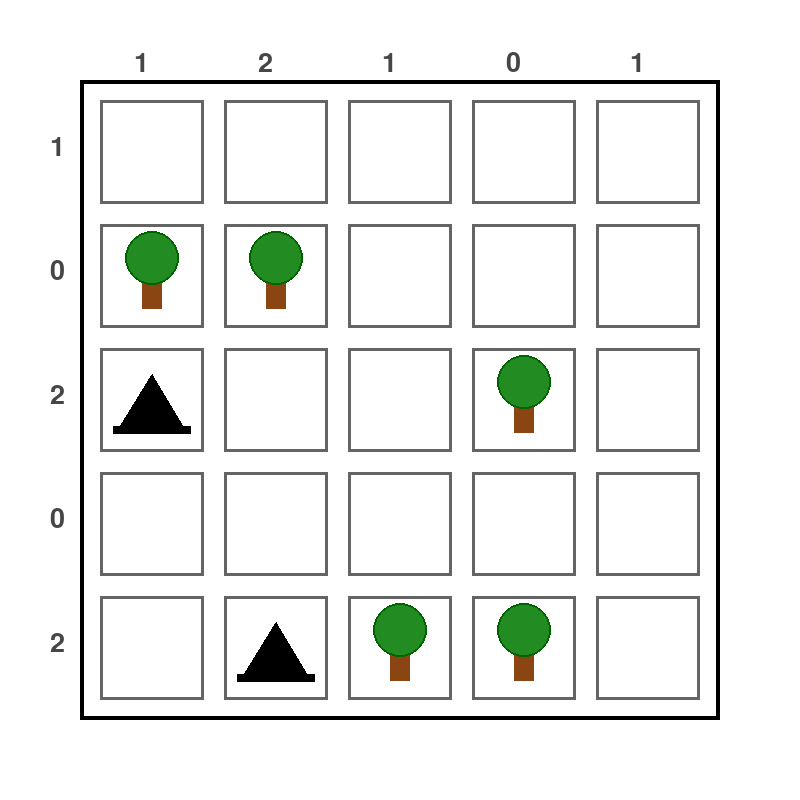}} \\ \hline

    \end{tabular}%
    }
    \caption{Per-Puzzle Sample Screenshots of Level \Easy}
    \label{tab:per-game-query-easy-level}
\end{table*}

\begin{table*}[p]
    \centering
    \renewcommand{\arraystretch}{1.5} 
    \resizebox{0.9\textwidth}{!}{%
    \begin{tabular}{|c|c|c|c|}
    \hline

    1. Aquarium & 2. Battle-Ships & 3. Binairo & 4. Colored-Sudoku \\ \hline

    {\includegraphics[height=55mm]{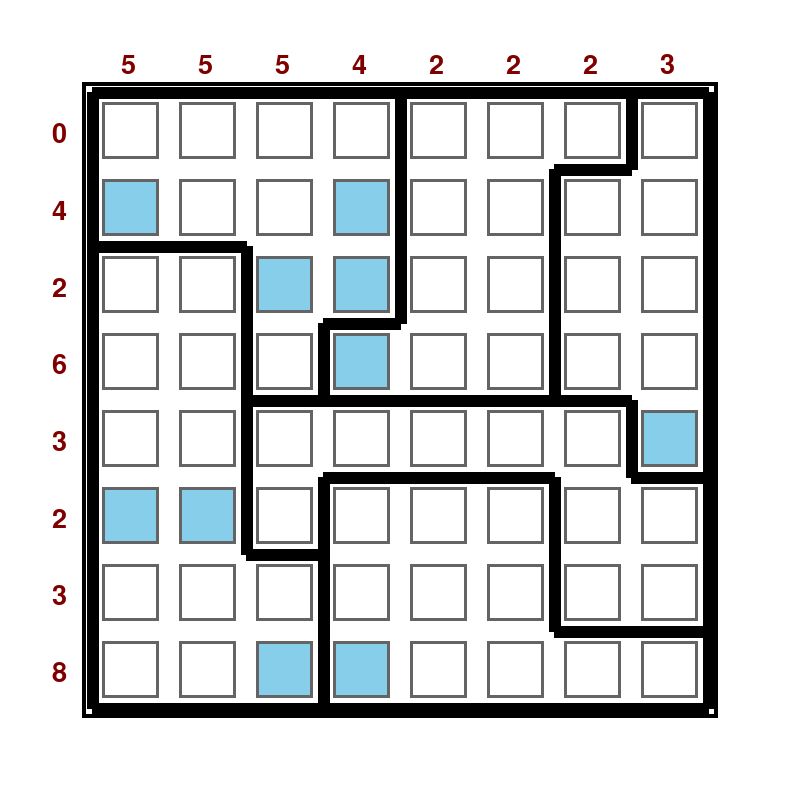}} & {\includegraphics[height=55mm]{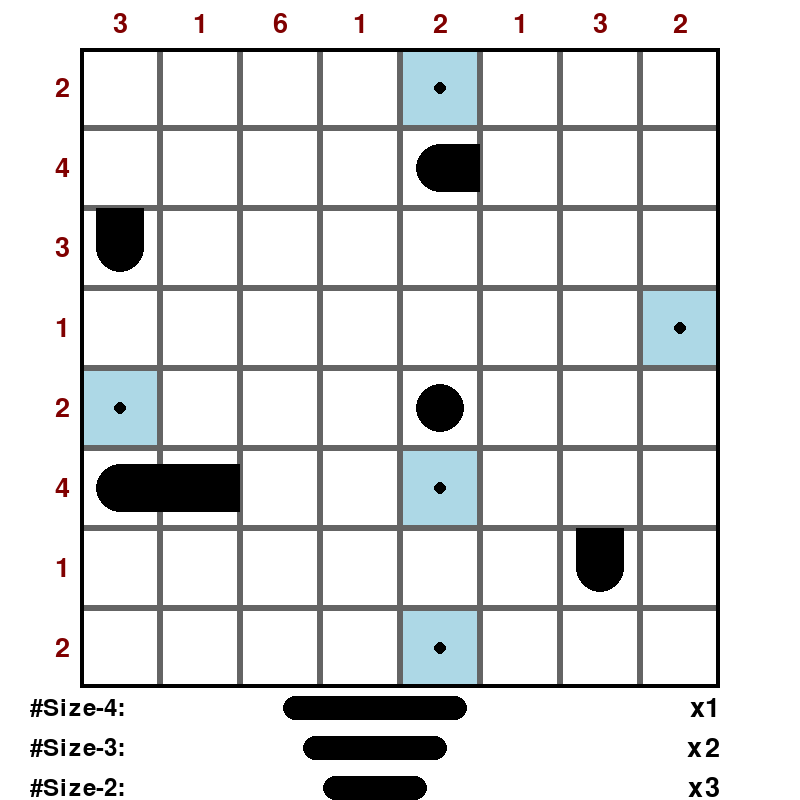}} & {\includegraphics[height=55mm]{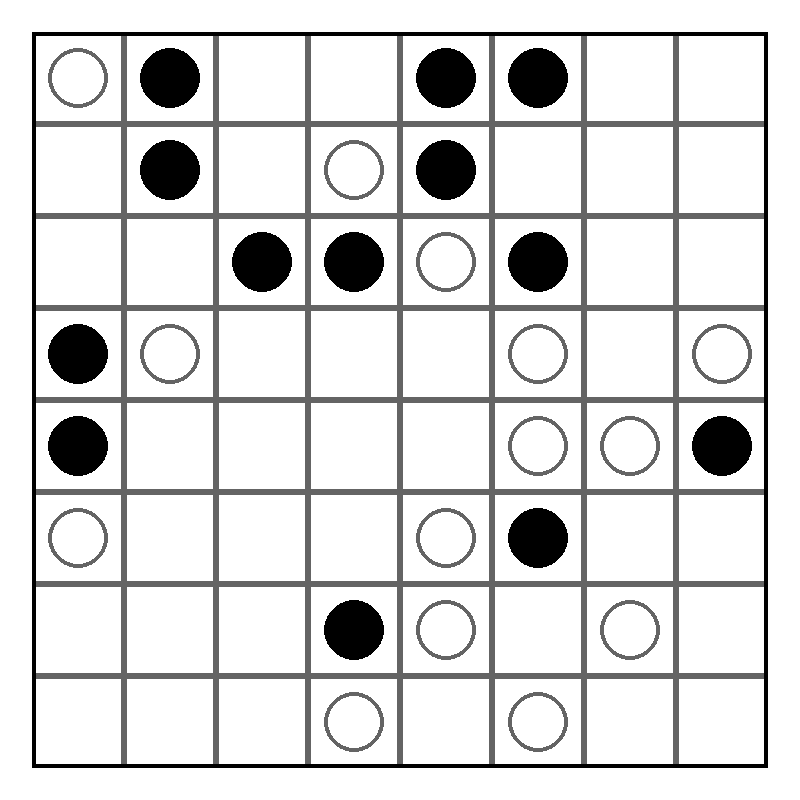}} & {\includegraphics[height=55mm]{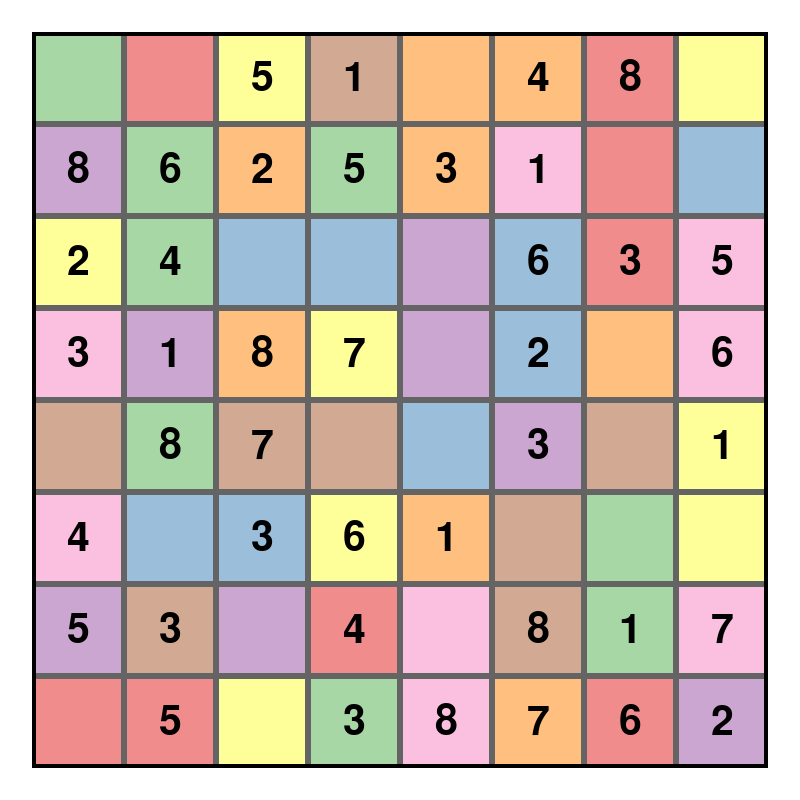}} \\ \hline
    
    5. Field-Explore & 6. Futoshiki & 7. Hitori & 8. Jigsaw-Sudoku \\ \hline

    {\includegraphics[height=55mm]{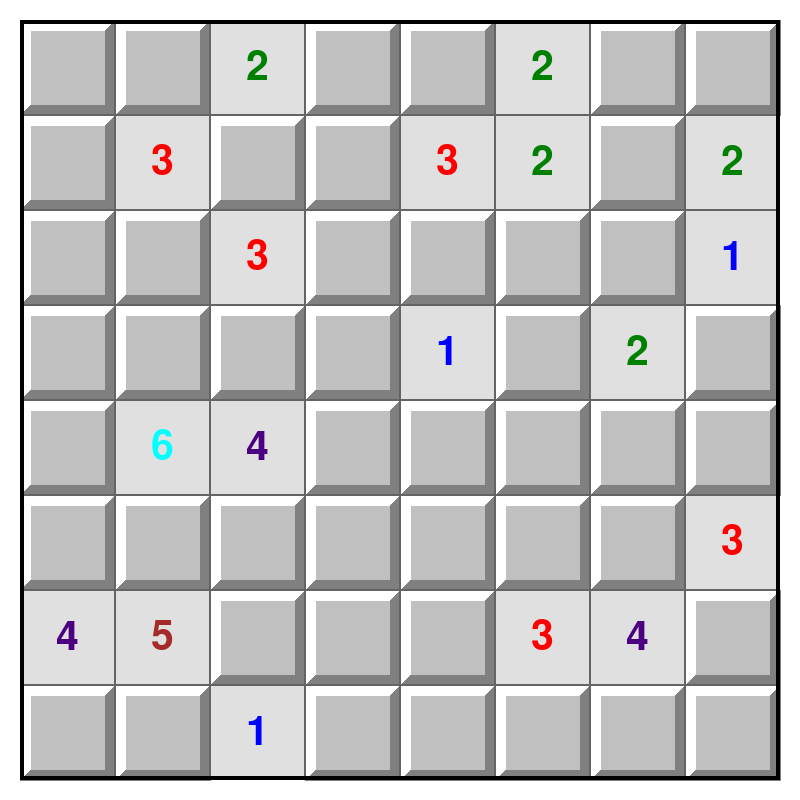}} & {\includegraphics[height=55mm]{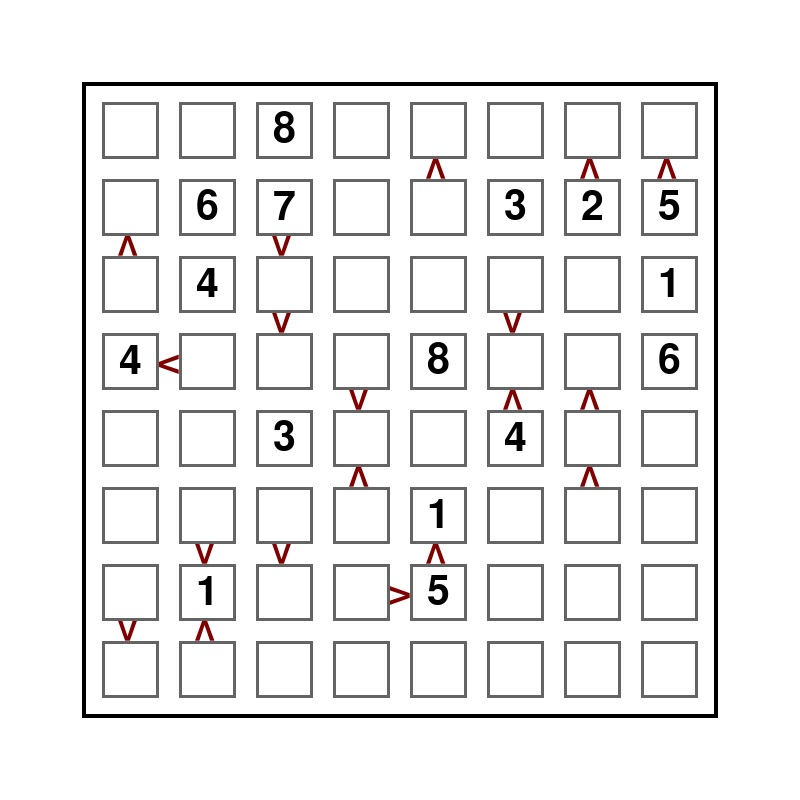}} & {\includegraphics[height=55mm]{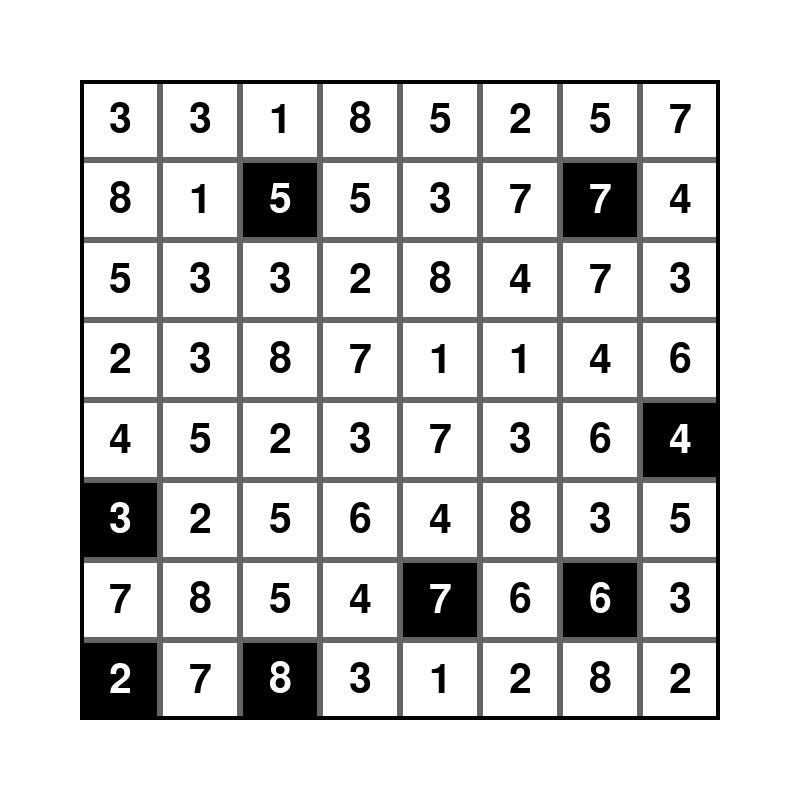}} & {\includegraphics[height=55mm]{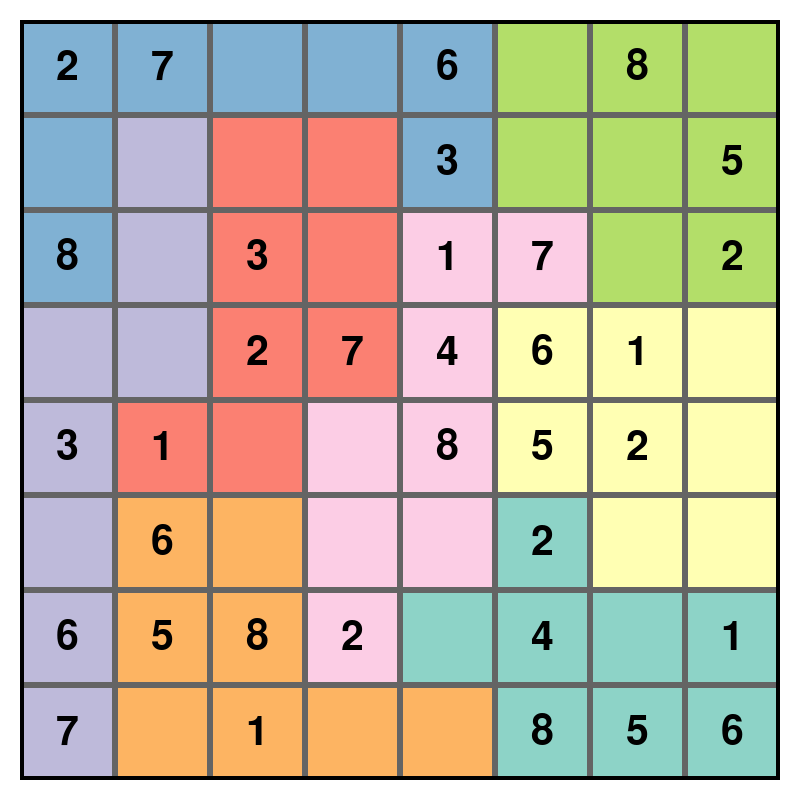}
    } \\ \hline

    9. Kakurasu & 10. Kakuro & 11. Killer-Sudoku & 12. Light-Up \\ \hline

    {\includegraphics[height=55mm]{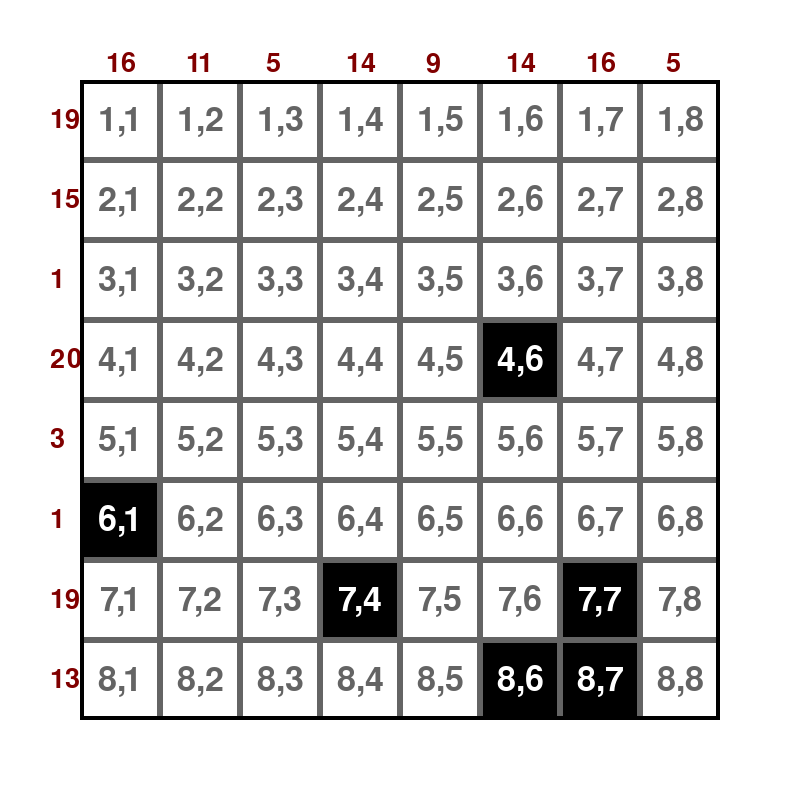}} & {\includegraphics[height=55mm]{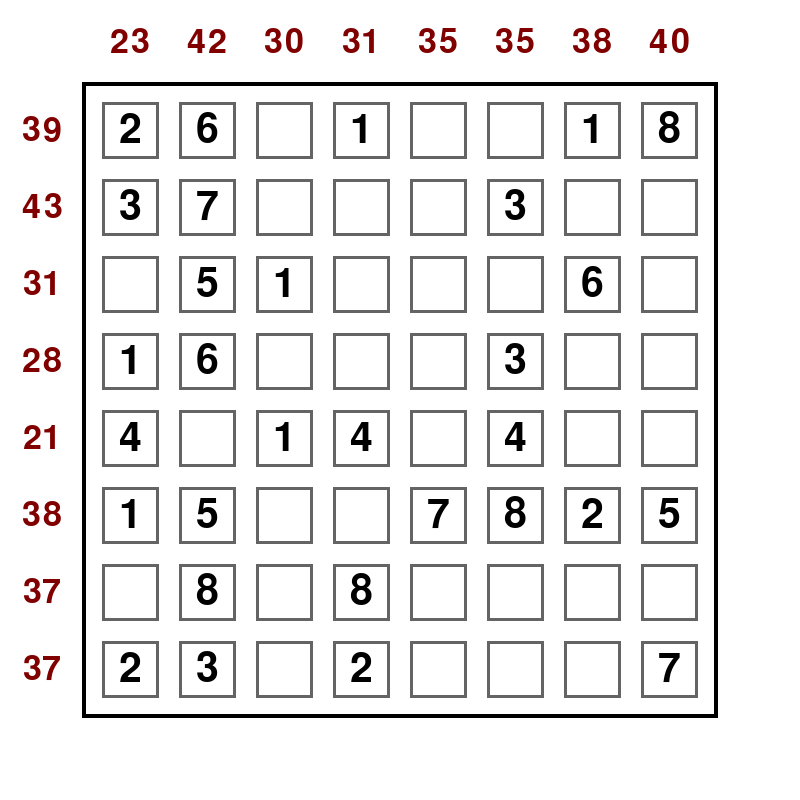}} & {\includegraphics[height=55mm]{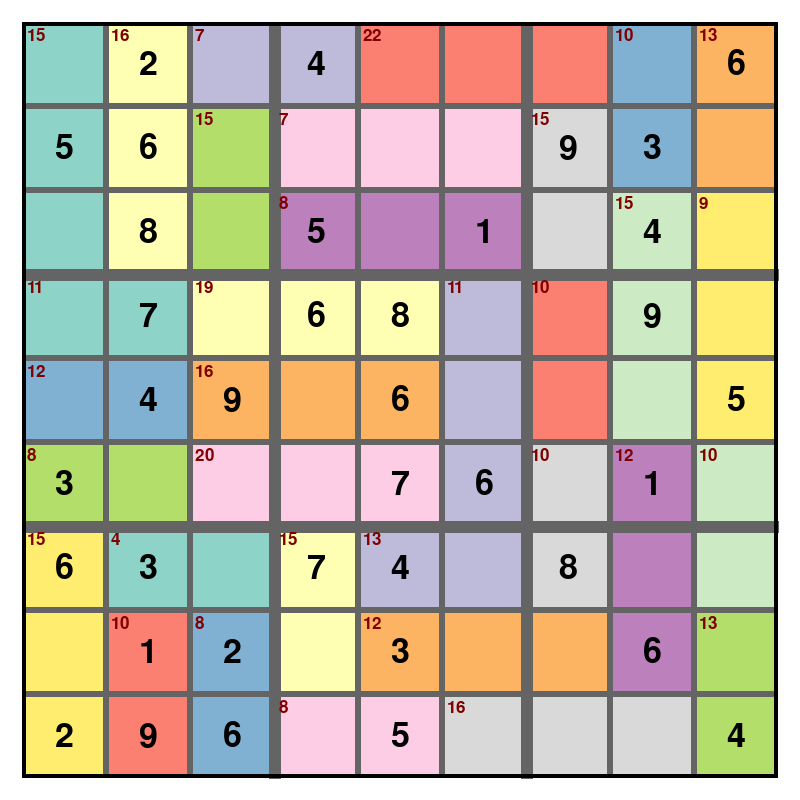}} & {\includegraphics[height=55mm]{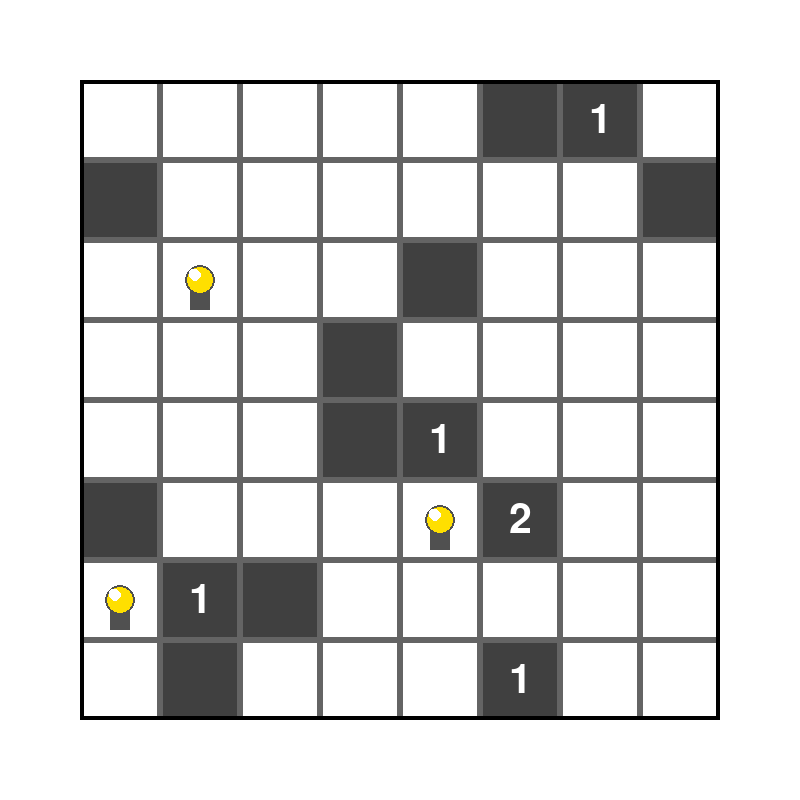}} \\ \hline
    
    13. Nonogram & 14. Odd-Even-Sudoku & 15. Renzoku & 16. Skyscraper \\ \hline

    {\includegraphics[height=55mm]{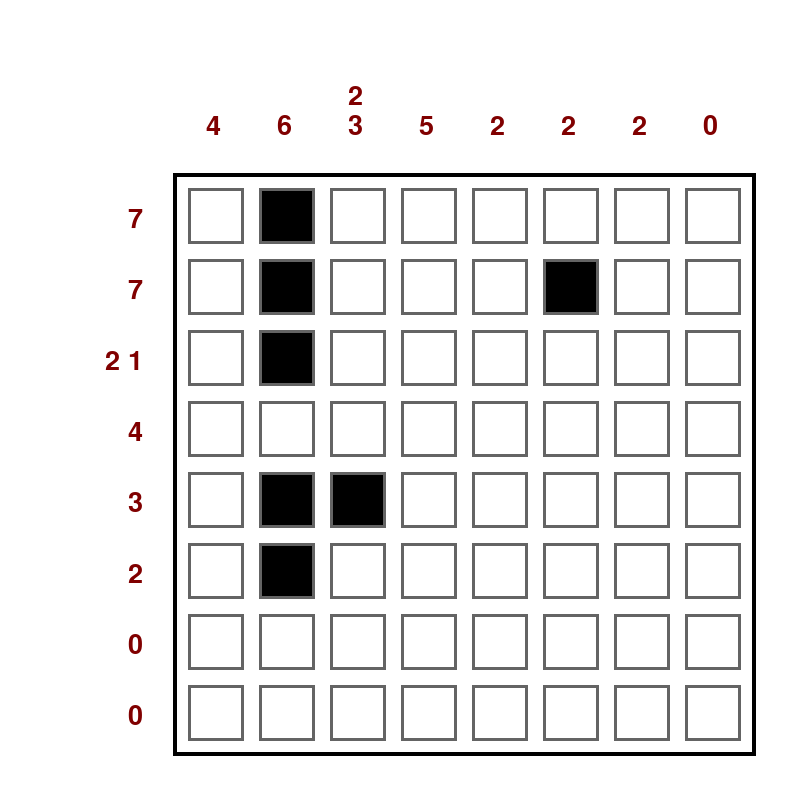}} & {\includegraphics[height=55mm]{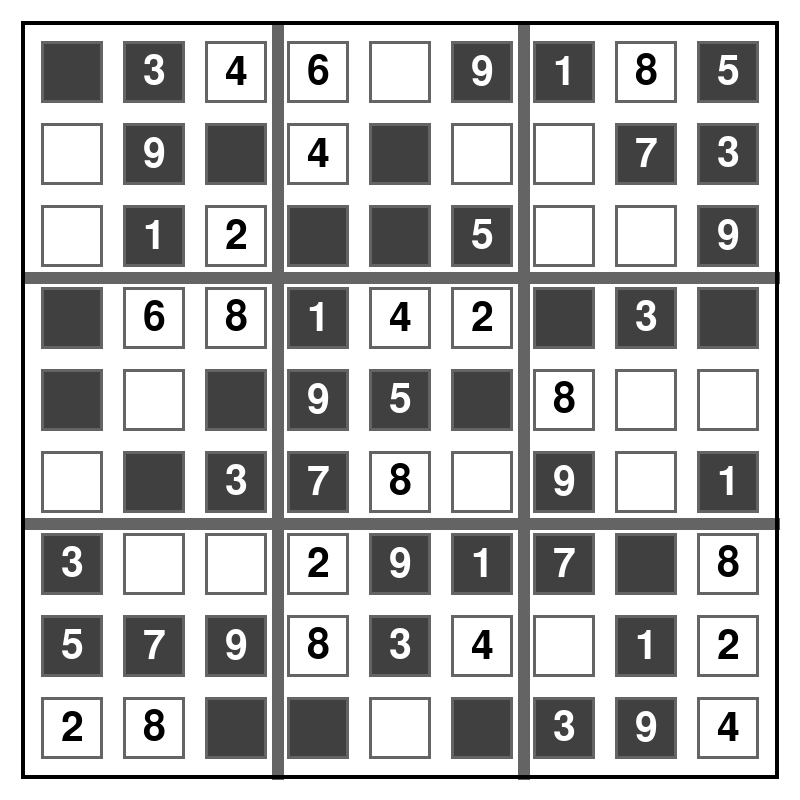}} & {\includegraphics[height=55mm]{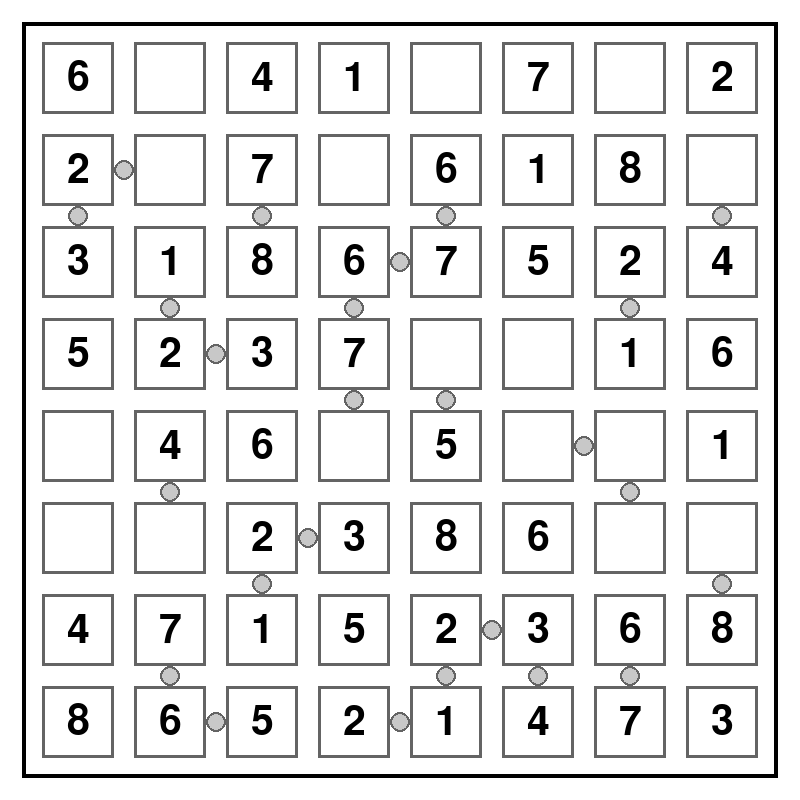}} & {\includegraphics[height=55mm]{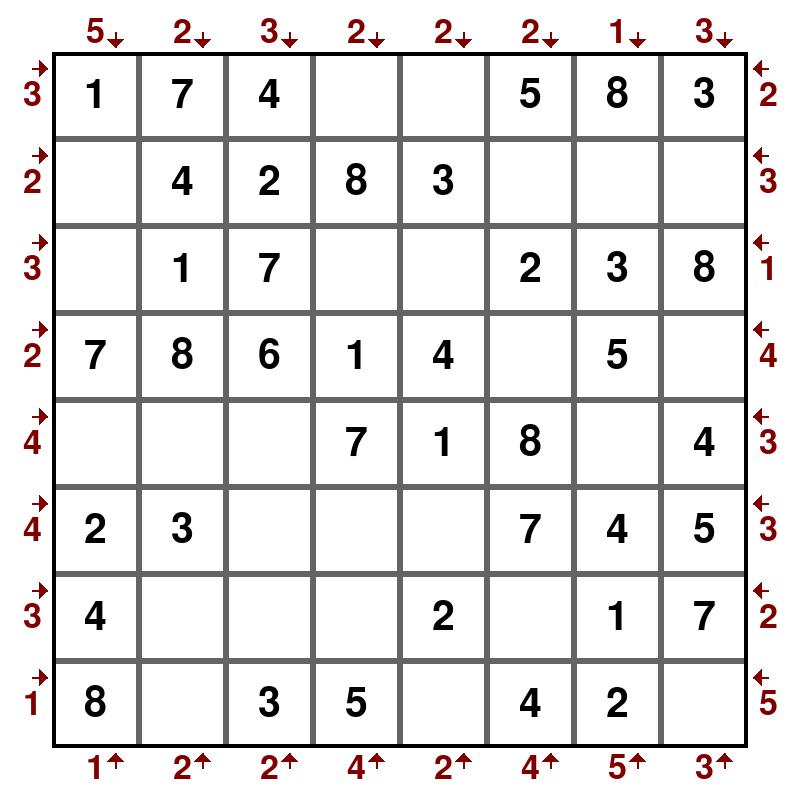}} \\ \hline

    17. Star-Battle & 18. Sudoku & 19. Thermometers & 20. Trees-and-Tents \\ \hline

    {\includegraphics[height=55mm]{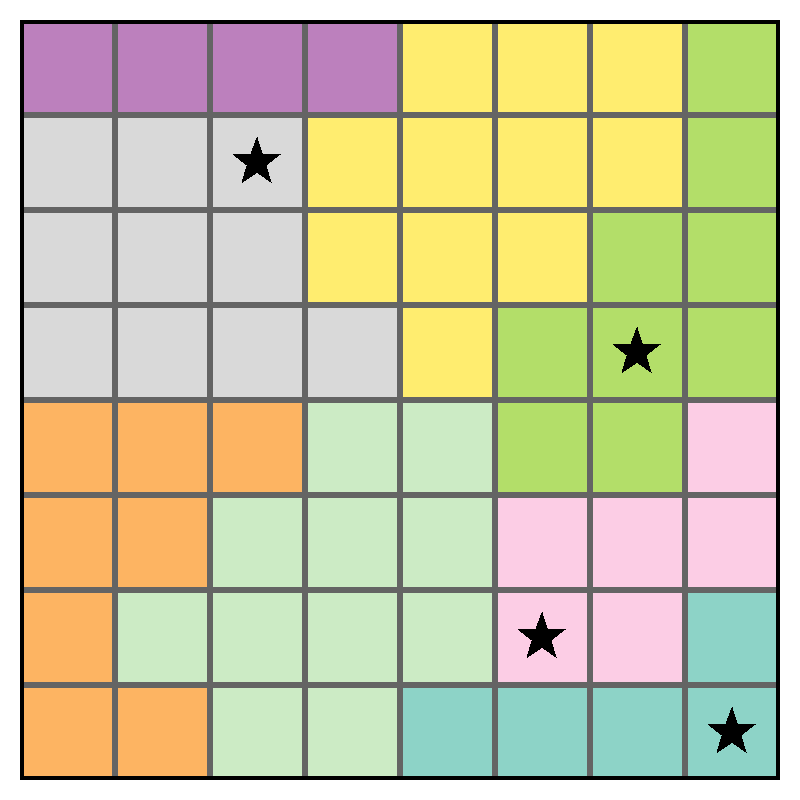}} & {\includegraphics[height=55mm]{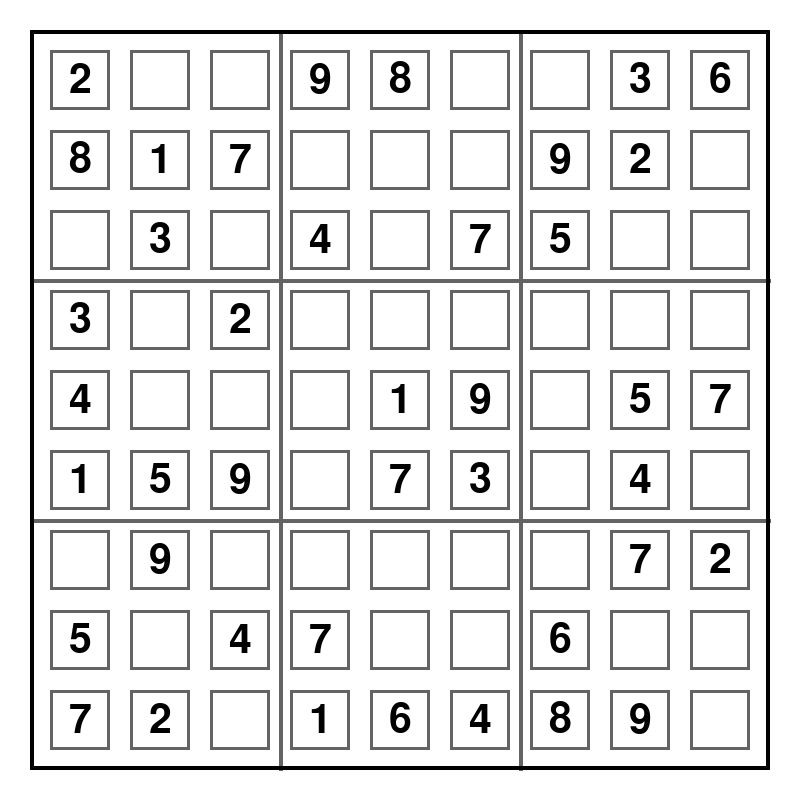}} & {\includegraphics[height=55mm]{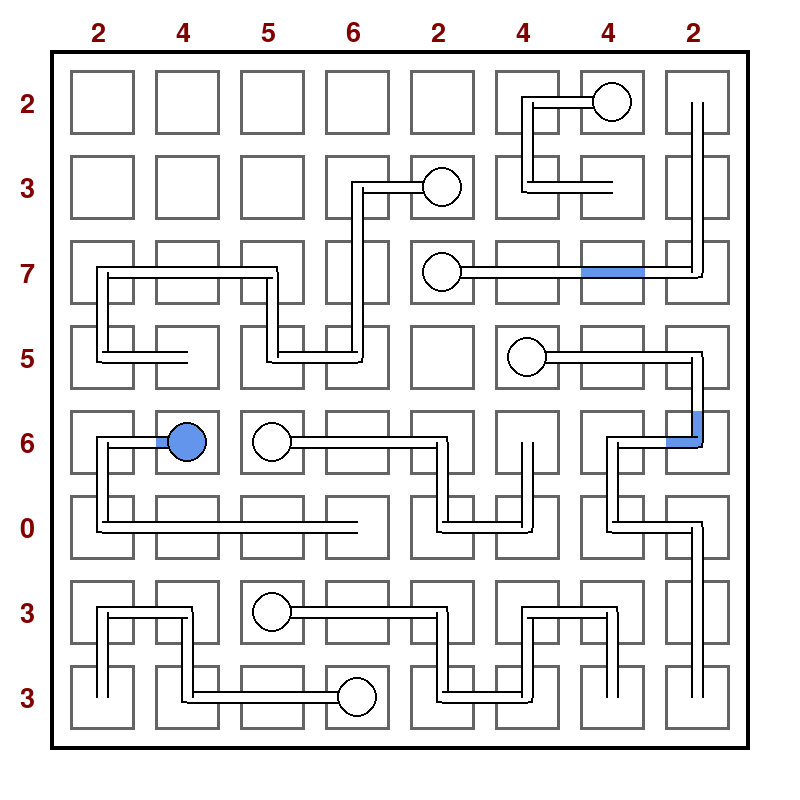}} & {\includegraphics[height=55mm]{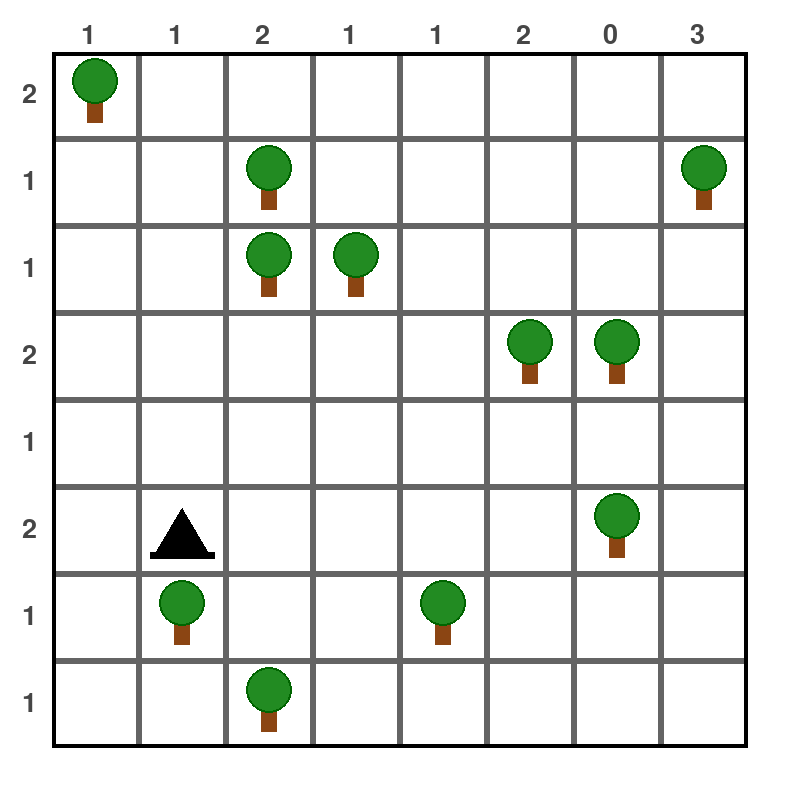}} \\ \hline

    \end{tabular}%
    }
    \caption{Per-Puzzle Sample Screenshots of Level~\Medium}
    \label{tab:per-game-query-medium-level}
\end{table*}

\begin{table*}[t]
    \centering
    \renewcommand{\arraystretch}{1.5} 
    \resizebox{0.9\textwidth}{!}{%
    \begin{tabular}{|c|c|c|c|}
    \hline

    1. Aquarium & 2. Battle-Ships & 3. Binairo & 4. Colored-Sudoku \\ \hline

    {\includegraphics[height=55mm]{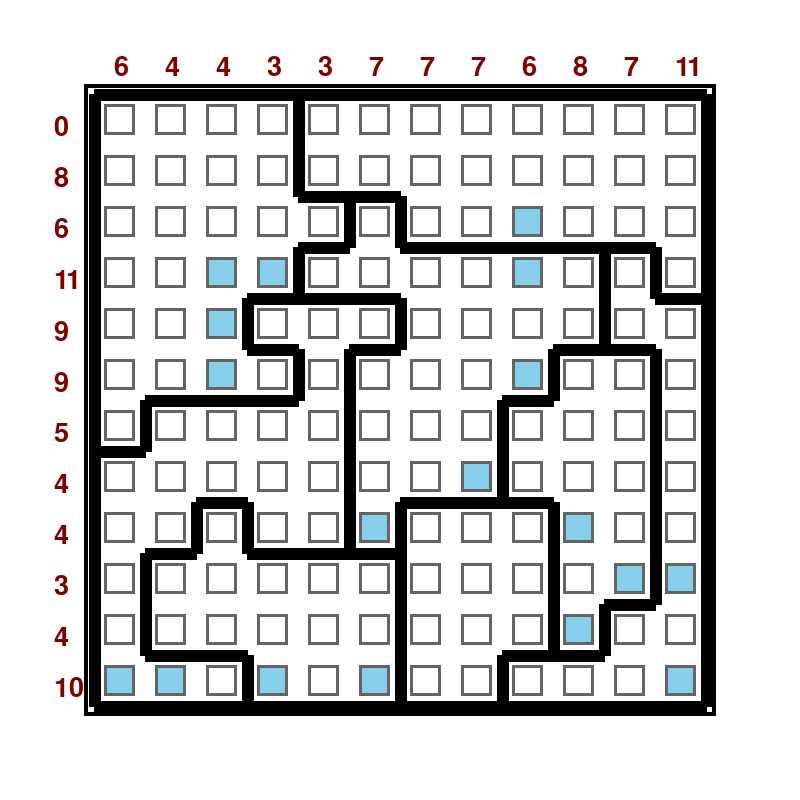}} & {\includegraphics[height=55mm]{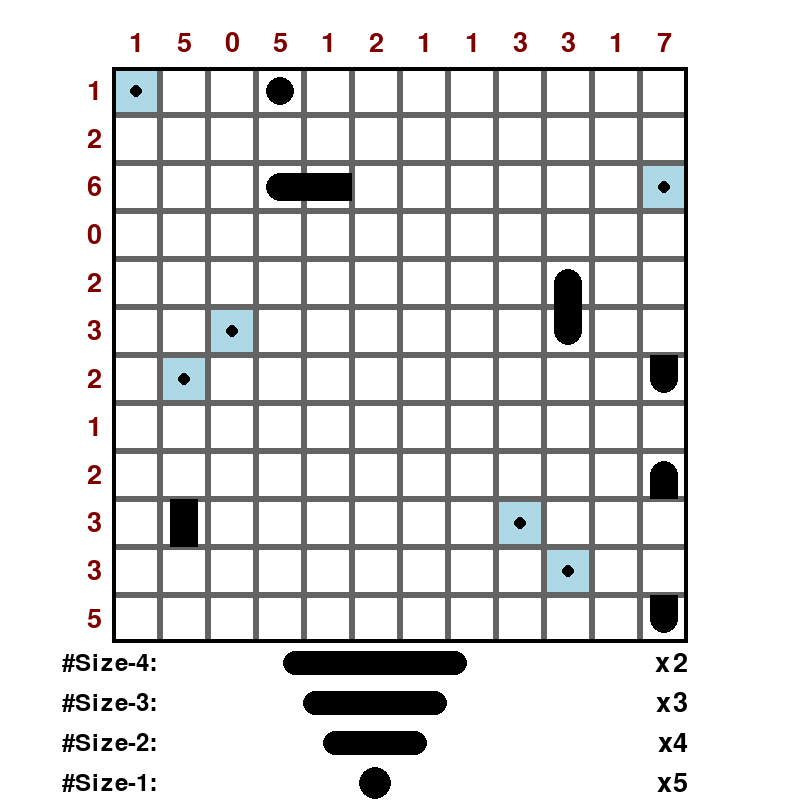}} & {\includegraphics[height=55mm]{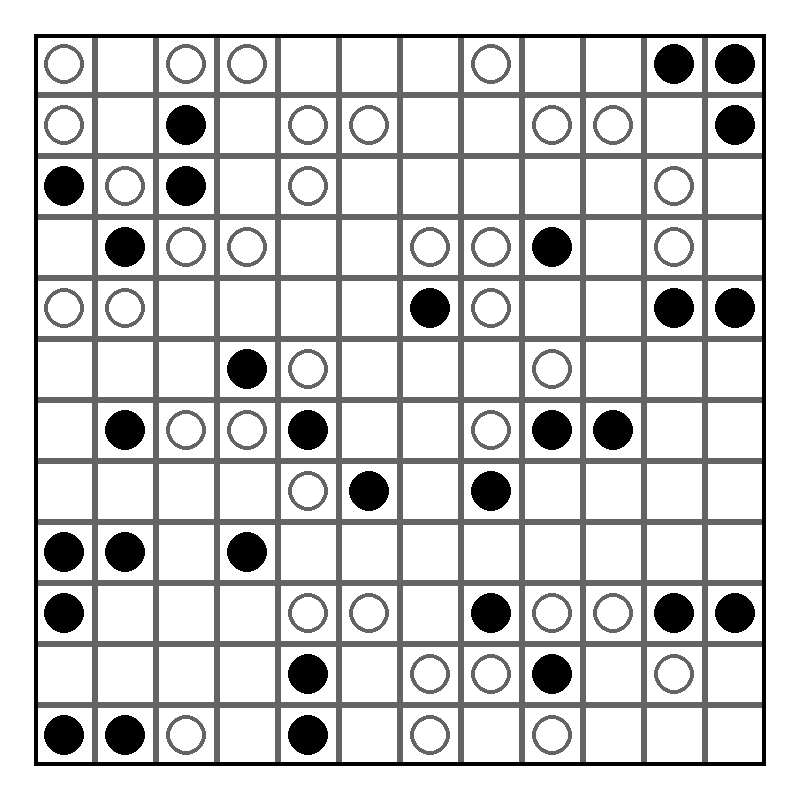}} & {\includegraphics[height=55mm]{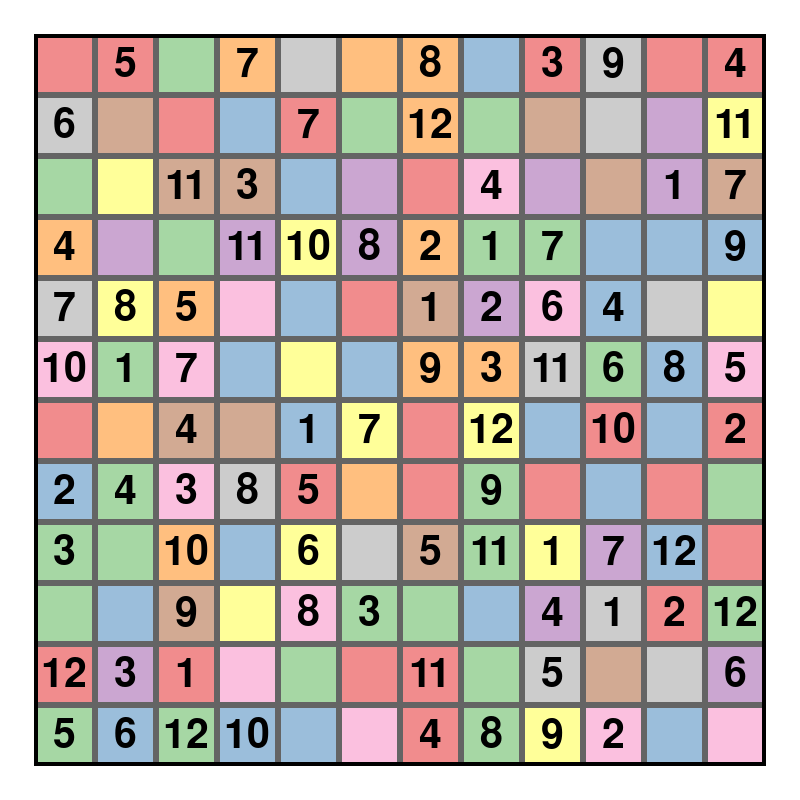}} \\ \hline

    5. Field-Explore & 6. Futoshiki & 7. Hitori & 8. Jigsaw-Sudoku \\ \hline

    {\includegraphics[height=55mm]{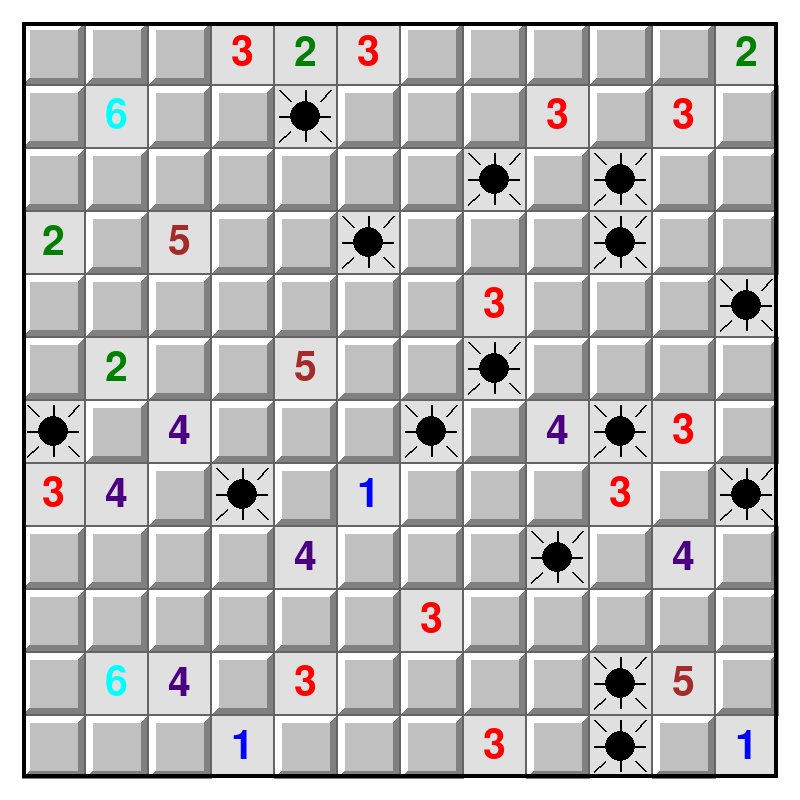}} & {\includegraphics[height=55mm]{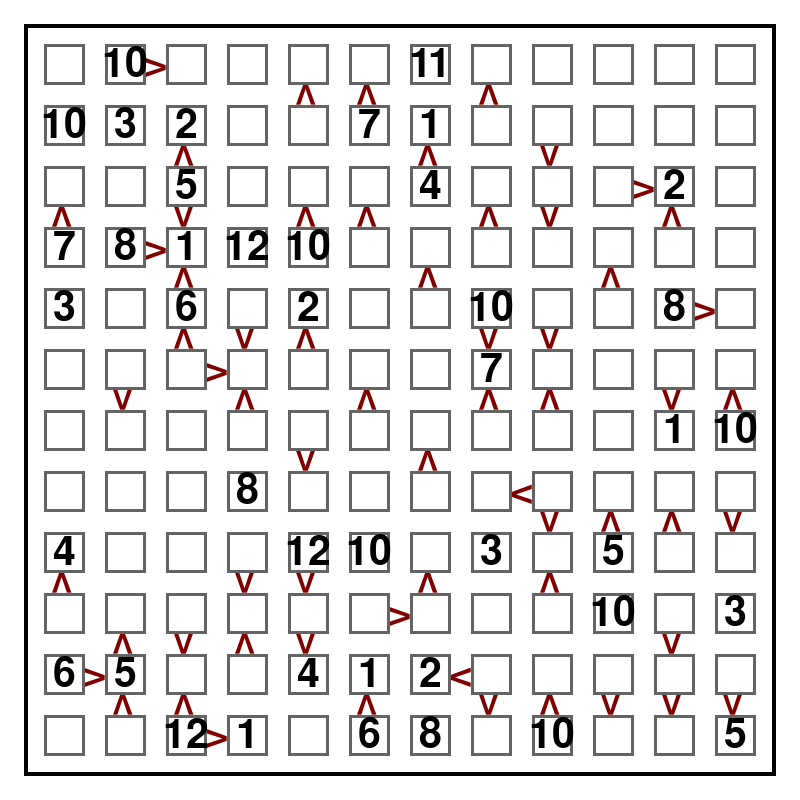}} & {\includegraphics[height=55mm]{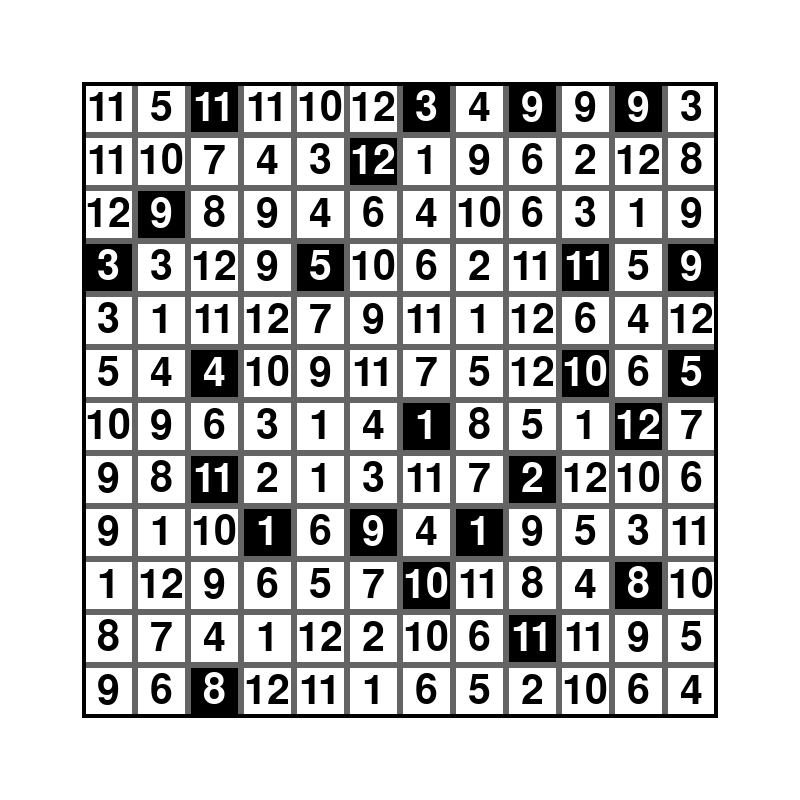}} & {\includegraphics[height=55mm]{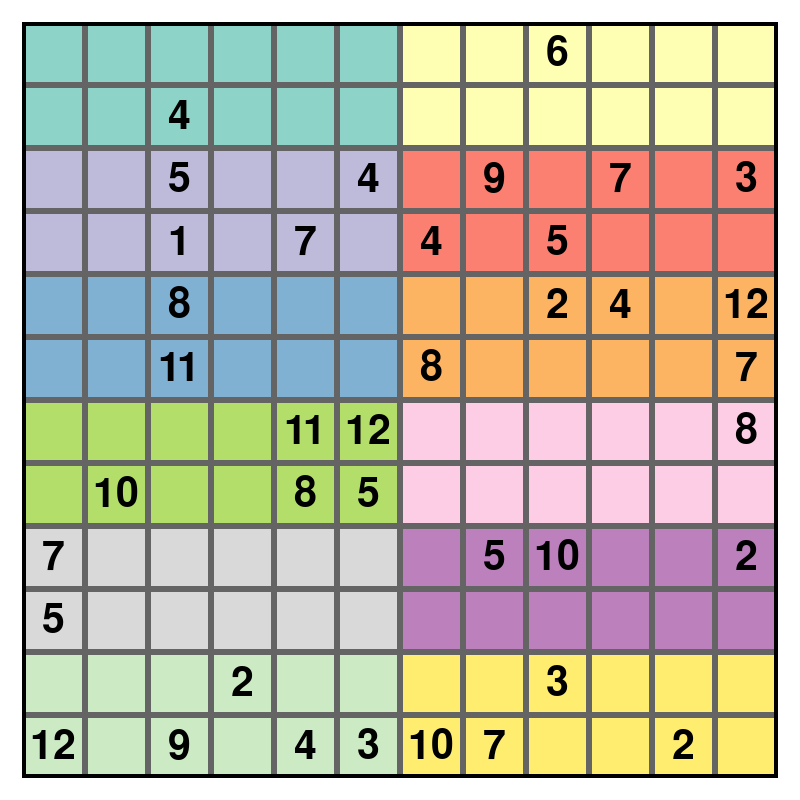}    
    } \\ \hline

    9. Kakurasu & 10. Kakuro & 11. Killer-Sudoku & 12. Light-Up \\ \hline

    {\includegraphics[height=55mm]{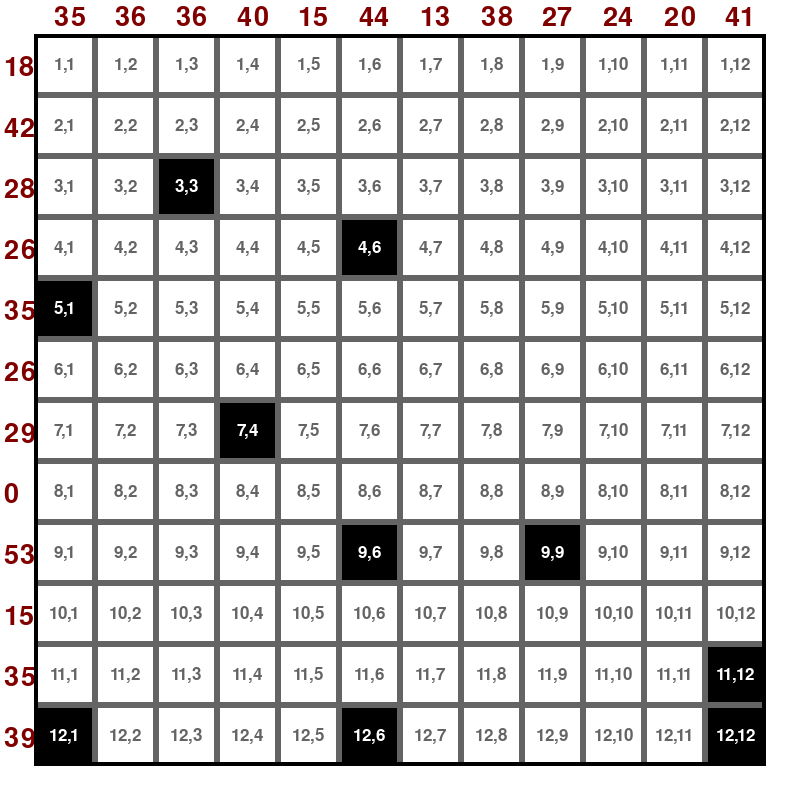}} & {\includegraphics[height=55mm]{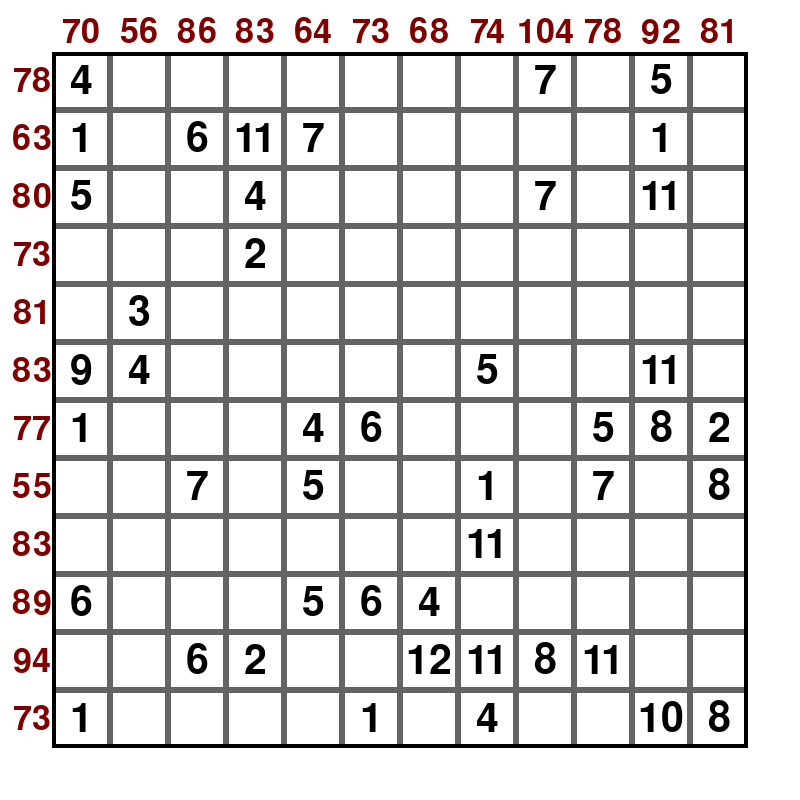}} & {} & {\includegraphics[height=55mm]{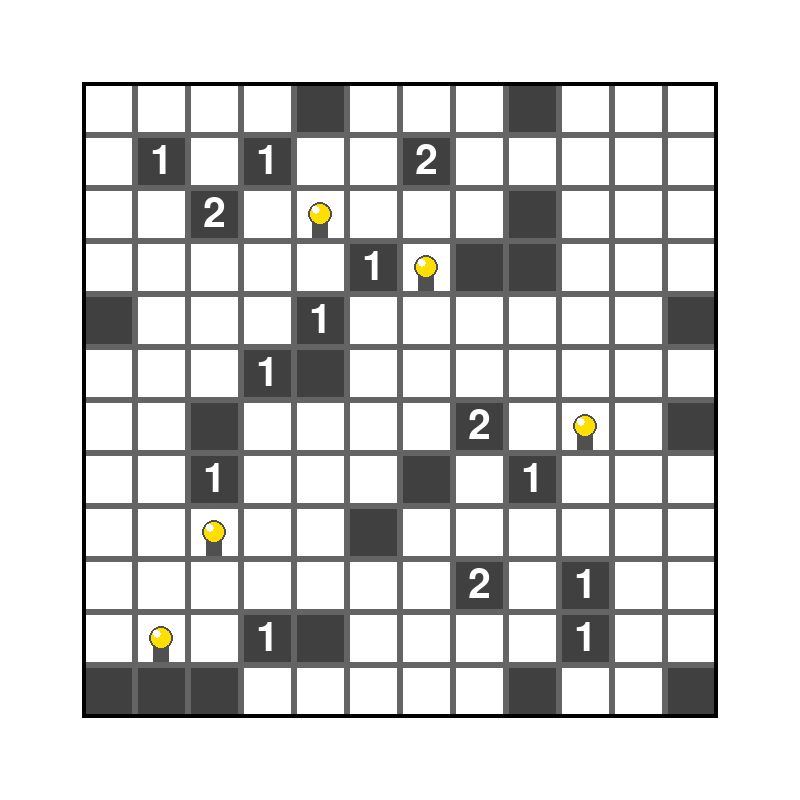}} \\ \hline
    
    13. Nonogram & 14. Odd-Even-Sudoku & 15. Renzoku & 16. Skyscraper \\ \hline

    {\includegraphics[height=55mm]{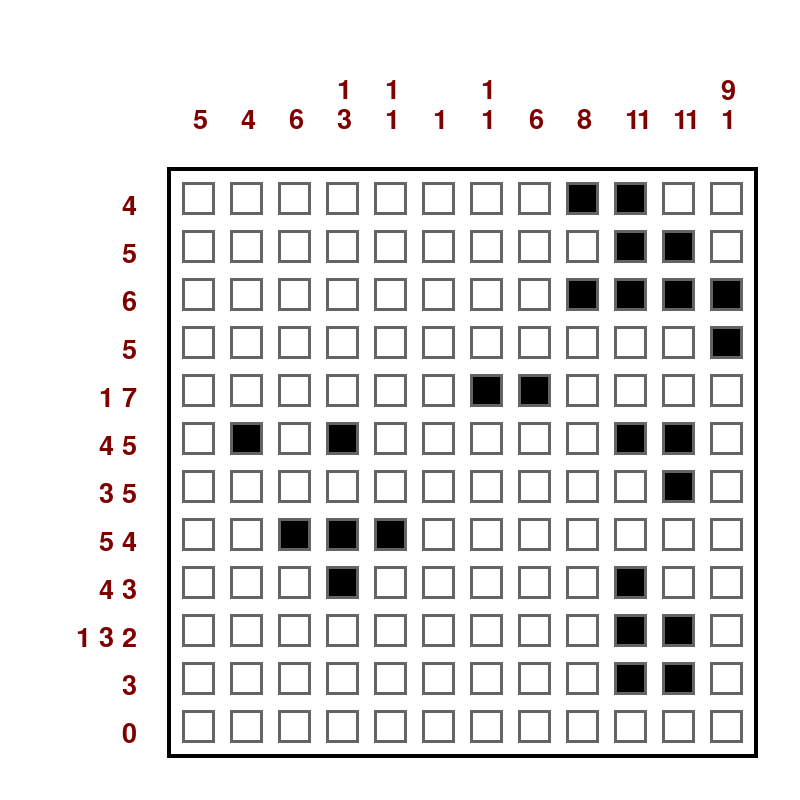}} & {} & {\includegraphics[height=55mm]{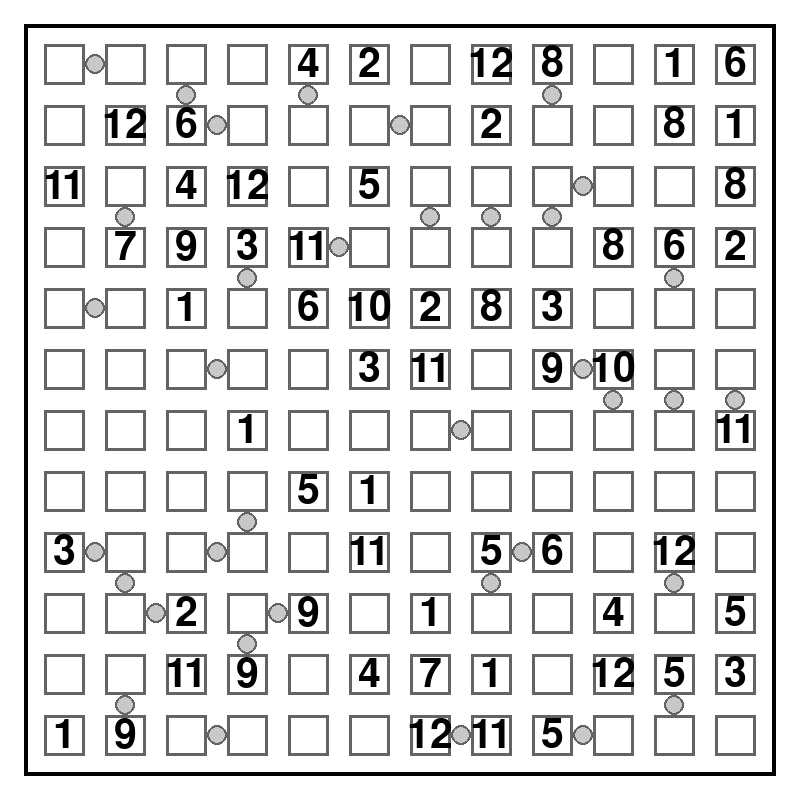}} & {\includegraphics[height=55mm]{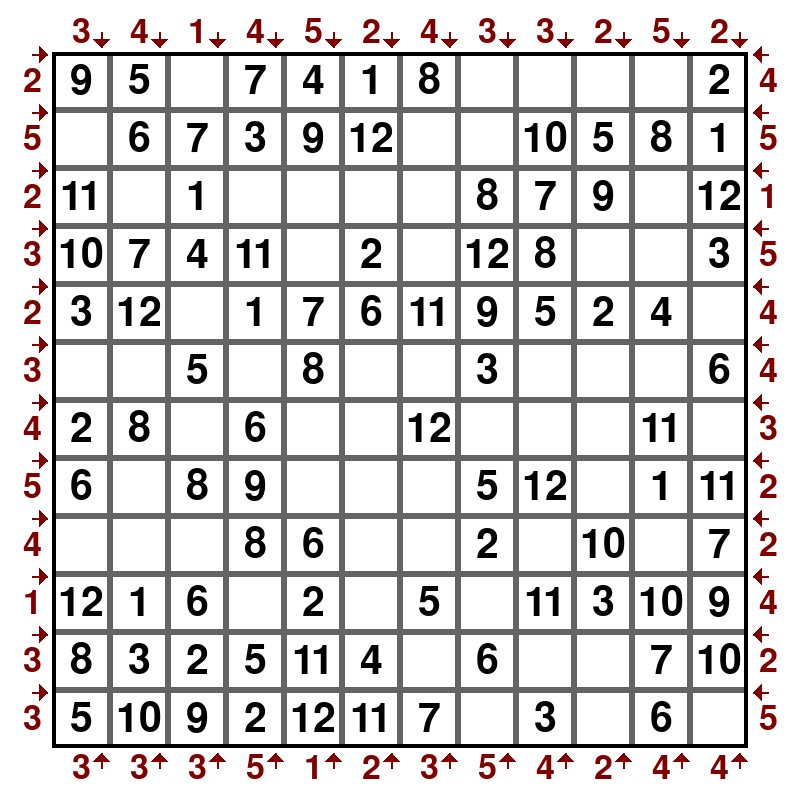}} \\ \hline

    17. Star-Battle & 18. Sudoku & 19. Thermometers & 20. Trees-and-Tents \\ \hline

    {\includegraphics[height=55mm]{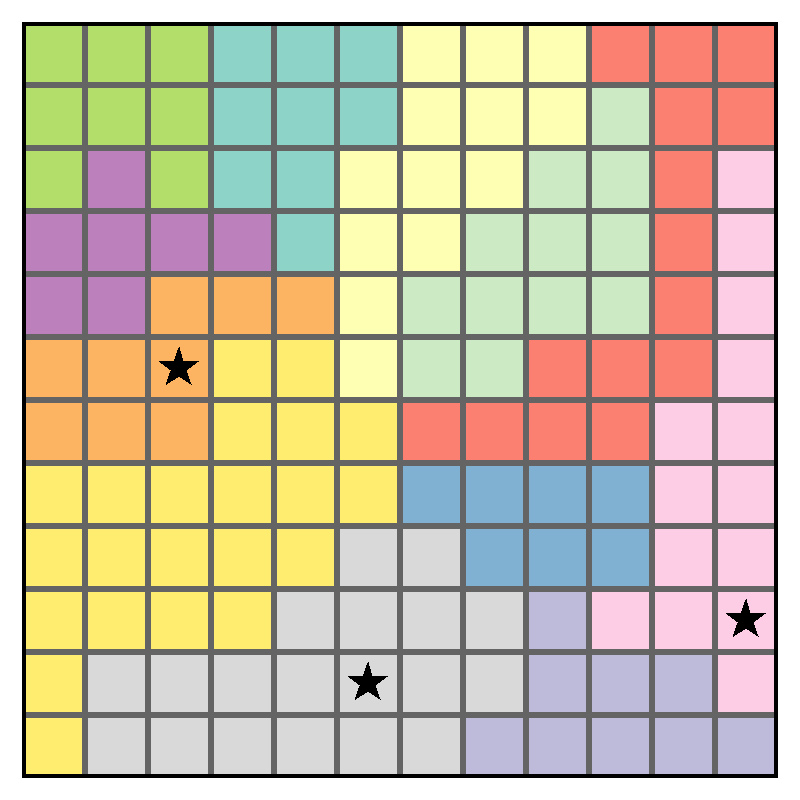}} & {} & {\includegraphics[height=55mm]{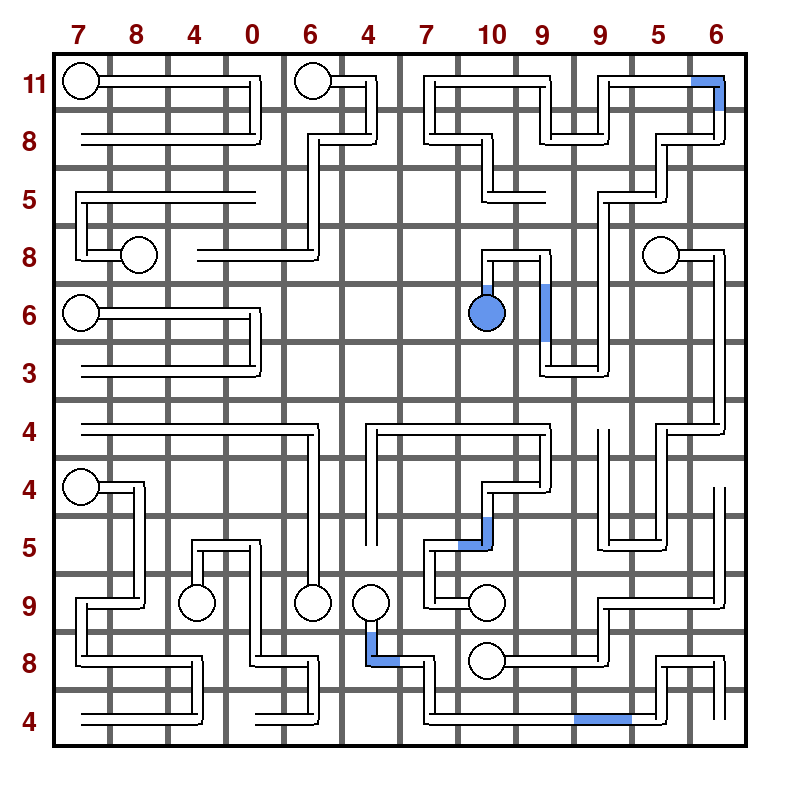}} & {\includegraphics[height=55mm]{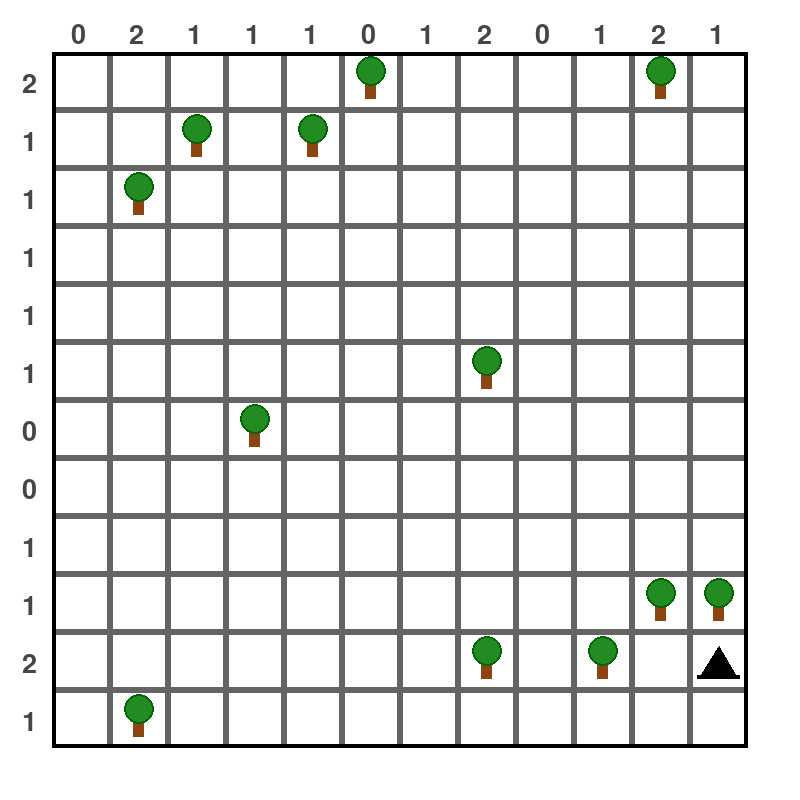}} \\ \hline

    \end{tabular}%
    }
    \caption{Per-Puzzle Sample Screenshots of Level \Hard. Note that some games do not have a hard level due to constraints imposed by the game rules. For example, Vanilla Sudoku's size can only be $4\times 4$ or $9\times 9$, and the next possible grid size is $16\times 16$, which is too large. }
    \label{tab:per-game-query-hard-level}
\end{table*}

\begin{table*}[t]
    \centering

    \refstepcounter{section} 
    \phantomsection          
    \addcontentsline{toc}{subsection}{Visualization: Per-Puzzle Examples and Query Templates}
    
    \renewcommand{\arraystretch}{1.6}
    \resizebox{0.98\textwidth}{!}{%
    \begin{tabular}{|c|c|p{1.0\textwidth}|}
    \hline
    Name & Sample Screenshot & Description and Queries \\ \hline

    \multirow{3}{*}{\textbf{1. Aquarium}} & 
    \multirow{3}{*}{\includegraphics[height=55mm]{screenshot-game/aquarium-easy.png}} &  
    \textbf{Rule:} You are an Aquarium puzzle player. You need to fill the aquariums with water up to a certain level or leave it empty. The numbers on the sides indicate how many filled (water) cells must be in each row and column. Indexing starts at 0.\\ \cline{3-3}
    & & \textbf{Perception - Cell At:} \textbf{\{Rule\}}
     what is at the cell (\{row\}, \{col\})? Choose from \{water, empty\}. \\ \cline{3-3}
    & & \textbf{Perception and Direct Solution: }\textbf{\{Rule\}} Give me your response of the current game state in the screenshot (where \texttt{"*"} indicates an unknown cell and  \texttt{"s"} indicates a filled cell) and your solution (where \texttt{"s"} indicates a filled cell and \texttt{"e"} indicates an empty cell) in the following format. \textbackslash n\{ \textbackslash n perception: \{current state of the grid as a 2D array\}, \textbackslash n answer: \{solution as a 2D array\} \textbackslash n\}  \\ \cline{3-3}
    & & \textbf{Rule Following - Valid Actions:} \textbf{\{Rule\}}  is it valid to assign the cell at (\{row\}, \{col\}) with value \{value\}? Choose from \{valid, invalid\}. \\ \cline{3-3}
    & & \textbf{Perception and Chain-of-Thought Reasoning: }\textbf{\{Rule\}} Give me your response of the current game state in the screenshot (where \texttt{"*"} indicates an unknown cell and \texttt{"s"} indicates a filled cell), \textbf{your step-by-step reasoning}, and your solution (where \texttt{"s"} indicates a filled cell and \texttt{"e"} indicates an empty cell) in the following format. \textbackslash n\{ \textbackslash n perception: \{current state of the grid as a 2D array\}, \textbackslash n think: \{your step-by-step reasoning\}, \textbackslash n answer: \{solution as a 2D array\} \textbackslash n\}  \\ \hline
    
    \multirow{5}{*}{\textbf{2. Battle-Ships}} & 
    \multirow{5}{*}{\includegraphics[height=55mm]{screenshot-game/battleship-easy12.png}} &  
    \textbf{Rule:} You are a Battle-Ships player. You need to place ships in a grid based on row and column hints. The hints indicate how many ship cells are in each row and column. The numbers of each size ship are given. Ships cannot touch each other, even diagonally. Indexing starts at 0. \\ \cline{3-3}
    & & \textbf{Perception - Cell At:} \textbf{\{Rule\}} Given the current game state, what is at position (\{row\}, \{col\})? Choose from: \{ship, empty, unknown\}. \\ \cline{3-3}
    & & \textbf{Perception and Direct Solution:} \textbf{\{Rule\}} Give me your response of the current game state in the screenshot (where \texttt{"*"} indicates an unknown cell) and your solution (where \texttt{"s"} indicates a ship cell and \texttt{"e"} indicates an empty cell) in the following format. \{ perception: \{current state of the grid as a 2D array\}, answer: \{solution as a 2D array\} \}  \\ \cline{3-3}
    & & \textbf{Rule Following - Valid Actions:} \textbf{\{Rule\}} Given the current game state, is it valid to assign cell (\{row\}, \{col\}) with value \{value\}? Respond with: valid or invalid. \\ \cline{3-3}
    & & \textbf{Perception and Chain-of-Thought Reasoning:} \textbf{\{Rule\}} Give me your response of the current game state in the screenshot (where \texttt{"*"} indicates an unknown cell), \textbf{your step-by-step reasoning}, and your solution (where \texttt{"s"} indicates a ship cell and \texttt{"e"} indicates an empty cell) in the following format. \{ perception: \{current state of the grid as a 2D array\}, think: \{your step-by-step reasoning\}, answer: \{solution as a 2D array\} \}  \\ \hline

    \multirow{3}{*}{\textbf{3. Binairo}} & 
    \multirow{3}{*}{\includegraphics[height=55mm]{screenshot-game/binairo15.png}} &  
    \textbf{Rule:} You are a Binairo player. You have to fill a grid with white (w) and black (b) pieces. No more than two circles of the same color can be adjacent (horizontally and vertically). Indexing starts at 0. \\ \cline{3-3}
    & & \textbf{Perception - Cell At:} \textbf{\{Rule\}}
     what is the value of the cell at (\{row\}, \{col\})? Choose from \{b, w, empty\}. \\ \cline{3-3}
    & & \textbf{Perception and Direct Solution: }\textbf{\{Rule\}} Give me your response of the current game state in the screenshot (where \texttt{"*"} indicates an empty cell) and your solution in the following format. \textbackslash n\{ \textbackslash n perception: \{current state of the grid as a 2D array\}, \textbackslash n answer: \{solution as a 2D array\} \textbackslash n\}  \\ \cline{3-3}
    & & \textbf{Rule Following - Valid Actions:} \textbf{\{Rule\}}  is it valid to fill the cell at (\{row\}, \{col\}) with \textbf{\{b, w\}}? Choose from \{valid, invalid\}. \\ \cline{3-3}
    & & \textbf{Perception and Chain-of-Thought Reasoning: }\textbf{\{Rule\}} Give me your response of the current game state in the screenshot (where \texttt{"*"} indicates an empty cell), \textbf{your step-by-step reasoning}, and your solution in the following format. \textbackslash n\{ \textbackslash n perception: \{current state of the grid as a 2D array\}, \textbackslash n think: \{your step-by-step reasoning\}, \textbackslash n answer: \{solution as a 2D array\} \textbackslash n\}  \\ \hline

    \multirow{3}{*}{\textbf{4. Colored-Sudoku}} & 
    \multirow{3}{*}{\includegraphics[height=55mm]{screenshot-game/cell_at-sample_2.png}} &  
    \textbf{Rule:} You are a Colored-Sudoku player. You have to enter a numerical digit from 1 through N in each cell of a NxN grid, \textbackslash nThe rule is to make sure unique numbers in each row, column,  and within cells of the same color.  Indexing starts at 0. \\ \cline{3-3}
    & & \textbf{Perception - Cell At:} \textbf{\{Rule\}}
     what is the value of the cell at (\{row\}, \{col\})? Choose from \{1, 2, ..., N, empty\}. \\ \cline{3-3}
    & & \textbf{Perception and Direct Solution: }\textbf{\{Rule\}} Give me your response of the current game state in the screenshot (where \texttt{"*"} indicates an empty cell) and your solution in the following format. \textbackslash n\{ \textbackslash n perception: \{current state of the grid as a 2D array\}, \textbackslash n answer: \{solution as a 2D array\} \textbackslash n\}  \\ \cline{3-3}
    & & \textbf{Rule Following - Valid Actions:} \textbf{\{Rule\}}  is it valid to fill the cell at (\{row\}, \{col\}) with value \{value\}? Choose from \{valid, invalid\}. \\ \cline{3-3}
    & & \textbf{Perception and Chain-of-Thought Reasoning: }\textbf{\{Rule\}} Give me your response of the current game state in the screenshot (where \texttt{"*"} indicates an empty cell), \textbf{your step-by-step reasoning}, and your solution in the following format. \textbackslash n\{ \textbackslash n perception: \{current state of the grid as a 2D array\}, \textbackslash n think: \{your step-by-step reasoning\}, \textbackslash n answer: \{solution as a 2D array\} \textbackslash n\}  \\ \hline

    \multirow{3}{*}{\textbf{5. Field-Explorer}} & 
    \multirow{3}{*}{\includegraphics[height=55mm]{screenshot-game/field-explore-easy-new10.png}} &  
    \textbf{Rule:} You are a Field-Explore player. You need to identify mine locations in a grid based on revealed numbers. Each revealed number indicates how many mines are adjacent to that cell (including diagonals). Indexing starts at 0. \\ \cline{3-3}
    & & \textbf{Perception - Cell At:} \textbf{\{Rule\}} Given the current game state, what is in the cell at (\{row\}, \{col\})? Choose from \{mine, number, hidden\}. \\ \cline{3-3}
    & & \textbf{Perception and Direct Solution:} \textbf{\{Rule\}} Give me your response of the current game state in the screenshot (where \texttt{"*"} indicates hidden cells, "s" indicates a mine, and numbers represent revealed counts) and your solution (where \texttt{"s"} indicates a mine and \texttt{"e"} indicates an empty cell) in the following format. \{ perception: \{current state of the grid as a 2D array\}, answer: \{solution as a 2D array\} \}  \\ \cline{3-3}
    & & \textbf{Rule Following - Valid Actions:} \textbf{\{Rule\}} Given the current game state, is it valid to assign the cell at (\{row\}, \{col\}) with \{value\}? Choose from \{valid, invalid\}. \\ \cline{3-3}
    & & \textbf{Perception and Chain-of-Thought Reasoning:} \textbf{\{Rule\}} Give me your response of the current game state in the screenshot (where \texttt{"*"} indicates hidden cells, "s" indicates a mine, and numbers represent revealed counts), \textbf{your step-by-step reasoning}, and your solution (where \texttt{"s"} indicates a mine and \texttt{"e"} indicates an empty cell) in the following format. \{ perception: \{current state of the grid as a 2D array\}, think: \{your step-by-step reasoning\}, answer: \{solution as a 2D array\} \}  \\ \hline

    \end{tabular}%
    }
    \caption{Per-Puzzle Sample Screenshots and Query Templates. Part 1. (1st - 5th) Note that $\{\cdot \}$ represents variable values that depend on each query and the specific game. Additionally, some quotation marks, e.g., $``\cdot "$, are omitted in the query template for clarity. To clearly distinguish between 0-indexing and 1-indexing, we explicitly require indexing in the query. Action validity is assessed solely based on the presence of immediate rule violations without considering any long term effects.
    }
    \label{tab:per-game-query-1}
    \end{table*}

\begin{table*}[t]
    \centering
    \renewcommand{\arraystretch}{1.6}
    \resizebox{1.0\textwidth}{!}{%
    \begin{tabular}{|c|c|p{1.0\textwidth}|}
    \hline
    Name & Sample Screenshot & Description and Queries \\ \hline

    \multirow{3}{*}{\textbf{6. Futoshiki (Unequal)}} & 
    \multirow{3}{*}{\includegraphics[height=55mm]{screenshot-game/futoshiki0.png}} &  
    \textbf{Rule:} You are a Futoshiki player. You have to enter a numerical digit from 1 through N in each cell of an NxN grid. The rules are: unique numbers in each row and column; inequality signs between cells must be respected (for example, $<$ means left number is smaller, $>$ means left number is larger). Indexing starts at 0. \\ \cline{3-3}
    & & \textbf{Perception - Cell At:} \textbf{\{Rule\}}
     what is the value of the cell at (\{row\}, \{col\})? Choose from \{1, 2, ..., N, empty\}. \\ \cline{3-3}
    & & \textbf{Perception and Direct Solution: }\textbf{\{Rule\}} Give me your response of the current game state in the screenshot (where \texttt{"*"} indicates an empty cell) and your solution in the following format. \textbackslash n\{ \textbackslash n perception: \{current state of the grid as a 2D array\}, \textbackslash n answer: \{solution as a 2D array\} \textbackslash n\}  \\ \cline{3-3}
    & & \textbf{Rule Following - Valid Actions:} \textbf{\{Rule\}}  is it valid to fill the cell at (\{row\}, \{col\}) with value \{value\}? Choose from \{valid, invalid\}. \\ \cline{3-3}
    & & \textbf{Perception and Chain-of-Thought Reasoning: }\textbf{\{Rule\}} Give me your response of the current game state in the screenshot (where \texttt{"*"} indicates an empty cell), \textbf{your step-by-step reasoning}, and your solution in the following format. \textbackslash n\{ \textbackslash n perception: \{current state of the grid as a 2D array\}, \textbackslash n think: \{your step-by-step reasoning\}, \textbackslash n answer: \{solution as a 2D array\} \textbackslash n\}  \\ \hline

    \multirow{3}{*}{\textbf{7. Hitori}} & 
    \multirow{3}{*}{\includegraphics[height=55mm]{screenshot-game/hitori3.png}} &  
    \textbf{Rule:} You are a Hitori player. You need to shade some cells in the grid such that no number appears more than once in each row and column among unshaded cells. The rules are: shaded cells cannot be adjacent; all unshaded cells must be connected. Indexing starts at 0. \\ \cline{3-3}
    & & \textbf{Perception - Cell At:} \textbf{\{Rule\}}
     what is the value of the cell at (\{row\}, \{col\})? Choose from \{1, 2, ..., N\}. \\ \cline{3-3}
    & & \textbf{Perception and Direct Solution: }\textbf{\{Rule\}} Give me your response of the current game state in the screenshot and your solution (where \texttt{"s"} indicates a shaded cell and \texttt{"e"} indicates a cell leave unshaded) in the following format. \textbackslash n\{ \textbackslash n perception: \{current state of the grid as a 2D array\}, \textbackslash n answer: \{solution as a 2D array\} \textbackslash n\}  \\ \cline{3-3}
    & & \textbf{Rule Following - Valid Actions:} \textbf{\{Rule\}}  is it valid to shade the cell at (\{row\}, \{col\})? Choose from \{valid, invalid\}. \\ \cline{3-3}
    & & \textbf{Perception and Chain-of-Thought Reasoning: }\textbf{\{Rule\}} Give me your response of the current game state in the screenshot, \textbf{your step-by-step reasoning}, and your solution (where \texttt{"s"} indicates a shaded cell and \texttt{"e"} indicates a cell leave unshaded) in the following format. \textbackslash n\{ \textbackslash n perception: \{current state of the grid as a 2D array\}, \textbackslash n think: \{your step-by-step reasoning\}, \textbackslash n answer: \{solution as a 2D array\} \textbackslash n\}  \\ \hline

    \multirow{3}{*}{\textbf{8. Jigsaw-Sudoku}} & 
    \multirow{3}{*}{\includegraphics[height=55mm]{screenshot-game/jigsaw-easy1.png}} &  
    \textbf{Rule:} You are a Jigsaw-Sudoku player. You have to enter a numerical digit from 1 through N in each cell of a NxN grid. The rules are: unique numbers in each row, column, and within cells of the same region. Each region is a connected group of cells. Indexing starts at 0. \\ \cline{3-3}
    & & \textbf{Perception - Cell At:} \textbf{\{Rule\}} Given the current game state in the screenshot, what is the value of the cell at (\{row\}, \{col\})? Choose from \{1, 2, ..., N, empty\}. \\ \cline{3-3}
    & & \textbf{Perception and Direct Solution:} \textbf{\{Rule\}} Give me your response of the current game state in the screenshot (where \texttt{"*"} indicates an empty cell) and your solution in the following format. \{ perception: \{current state of the grid as a 2D array\}, answer: \{solution as a 2D array\} \}  \\ \cline{3-3}
    & & \textbf{Rule Following - Valid Actions:} \textbf{\{Rule\}} Given the current game state in the screenshot, is it valid to fill the cell at (\{row\}, \{col\}) with value \{value\}? Choose from \{valid, invalid\}. \\ \cline{3-3}
    & & \textbf{Perception and Chain-of-Thought Reasoning:} \textbf{\{Rule\}} Give me your response of the current game state in the screenshot (where \texttt{"*"} indicates an empty cell), your step-by-step reasoning, and your solution in the following format. \{ perception: \{current state of the grid as a 2D array\}, think: \{your step-by-step reasoning\}, answer: \{solution as a 2D array\} \}  \\ \hline
    
    \multirow{5}{*}{\textbf{9. Kakurasu}} & 
    \multirow{5}{*}{\includegraphics[height=55mm]{screenshot-game/kakurasu1.png}} &  
    \textbf{Rule:} You are a Kakurasu puzzle player. You need to shade some cells in a grid where the sum of the weights of selected cells in each row and column matches the given clues. The weights increase from left to right (for rows) and top to bottom (for columns), starting from 1. Indexing starts at 0. \\ \cline{3-3}
    & & \textbf{Perception - Cell At:} \textbf{\{Rule\}}
     what is the value of the cell at (\{row\}, \{col\})? Response in a formate of (number, number). \\ \cline{3-3}
    & & \textbf{Perception and Direct Solution: }\textbf{\{Rule\}} Give me your response of the current game state in the screenshot (where \texttt{"*"} indicates an empty cell) and your solution (where \texttt{"s"} indicates a shaded cell and \texttt{"e"} indicates a cell leave unshaded) in the following format. \textbackslash n\{ \textbackslash n perception: \{current state of the grid as a 2D array\}, \textbackslash n answer: \{solution as a 2D array\} \textbackslash n\}  \\ \cline{3-3}
    & & \textbf{Rule Following - Valid Actions:} \textbf{\{Rule\}}  is it valid to \textbf{shade} the cell at (\{row\}, \{col\})? Choose from \{valid, invalid\}. \\ \cline{3-3}
    & & \textbf{Perception and Chain-of-Thought reasoning: }\textbf{\{Rule\}} Give me your response of the current game state in the screenshot (where \texttt{"*"} indicates an empty cell), \textbf{your step-by-step reasoning}, and your solution (where \texttt{"s"} indicates a shaded cell and \texttt{"e"} indicates a cell leave unshaded) in the following format. \textbackslash n\{ \textbackslash n perception: \{current state of the grid as a 2D array\}, \textbackslash n think: \{your step-by-step reasoning\}, \textbackslash n answer: \{solution as a 2D array\} \textbackslash n\}  \\ \hline

    \multirow{3}{*}{\textbf{10. Kakuro}} & 
    \multirow{3}{*}{\includegraphics[height=55mm]{screenshot-game/kakuro1.png}} &  
    \textbf{Rule:} You are a Kakuro player. You have to fill in the grid with numbers (1 to N) such that each row and column adds up to the specified sum. The rules are: (1) adjacent numbers should not be the same. (2) numbers add up to the given sum for each row and column. Indexing starts at 0. \\ \cline{3-3}
    & & \textbf{Perception - Cell At:} \textbf{\{Rule\}}
     what is the value of the cell at (\{row\}, \{col\})? Choose from \{1, 2, ..., N, empty\}. \\ \cline{3-3}
    & & \textbf{Perception and Direct Solution: }\textbf{\{Rule\}} Give me your response of the current game state in the screenshot (where \texttt{"*"} indicates an empty cell) and your solution in the following format. \textbackslash n\{ \textbackslash n perception: \{current state of the grid as a 2D array\}, \textbackslash n answer: \{solution as a 2D array\} \textbackslash n\}  \\ \cline{3-3}
    & & \textbf{Rule Following - Valid Actions:} \textbf{\{Rule\}}  is it valid to fill the cell at (\{row\}, \{col\}) with value \{value\}? Choose from \{valid, invalid\}. \\ \cline{3-3}
    & & \textbf{Perception and Chain-of-Thought reasoning: }\textbf{\{Rule\}} Give me your response of the current game state in the screenshot (where \texttt{"*"} indicates an empty cell), \textbf{your step-by-step reasoning}, and your solution in the following format. \textbackslash n\{ \textbackslash n perception: \{current state of the grid as a 2D array\}, \textbackslash n think: \{your step-by-step reasoning\}, \textbackslash n answer: \{solution as a 2D array\} \textbackslash n\}  \\ \hline
    
    \end{tabular}%
    }
    \caption{Per-Puzzle Sample Screenshots and Query Templates. Part 2 (6th - 10th). Note that there is a special case that in selection-based games, e.g., Thermometers, where a set of cell is selected as the answer. A cell being empty could mean both undefined and deliberately leaving empty. To distinguish these cases, we typically use two notations, such as ``*" for undefined cells and ``e" for the deliberately empty cells in the query and SFT dataset creation. }
    \label{tab:per-game-query-2}
    \end{table*}

\begin{table*}[t]
    \centering
    \renewcommand{\arraystretch}{1.6}
    \resizebox{0.95\textwidth}{!}{%
    \begin{tabular}{|c|c|p{1.0\textwidth}|}
    \hline
    Name & Sample Screenshot & Description and Queries \\ \hline
    
    \multirow{3}{*}{\textbf{11. Killer-Sudoku}} & 
    \multirow{3}{*}{\includegraphics[height=55mm]{screenshot-game/cell_at-sample_8.png}} &  
    \textbf{Rule:} You are a Killer-Sudoku player. You have to enter a numerical digit from 1 through N in each cell of an NxN grid. The board is divided into cages based on cell color. The rules are: (1) unique numbers in each row, column, and each sqrt(N)xsqrt(N) block, and (2) the sum of numbers in each cage must equal to the small red number indicating the target sum. Indexing starts at 0. \\ \cline{3-3}
    & & \textbf{Perception - Cell At:} \textbf{\{Rule\}}
     what is the cell value \textbf{(not cage target)} at (\{row\}, \{col\})? Choose from \{1, 2, ..., N, empty\}. \\ \cline{3-3}
    & & \textbf{Perception and Direct Solution: }\textbf{\{Rule\}} Give me your response of the current game state in the screenshot (where \texttt{"*"} indicates an empty cell) and your solution in the following format. \textbackslash n\{ \textbackslash n perception: \{current state of the grid as a 2D array, without red cage targets\}, \textbackslash n answer: \{solution as a 2D array\} \textbackslash n\}  \\ \cline{3-3}
    & & \textbf{Rule Following - Valid Actions:} \textbf{\{Rule\}}  is it valid to fill the cell at (\{row\}, \{col\}) with value \{value\}? Choose from \{valid, invalid\}. \\ \cline{3-3}
    & & \textbf{Perception and Chain-of-Thought reasoning: }\textbf{\{Rule\}} Give me your response of the current game state in the screenshot (where \texttt{"*"} indicates an empty cell), \textbf{your step by step reasoning}, and your solution in the following format. \textbackslash n\{ \textbackslash n perception: \{current state of the grid as a 2D array, without red cage targets\}, \textbackslash n think: \{your step by step reasoning\}, \textbackslash n answer: \{solution as a 2D array\} \textbackslash n\}  \\ \hline

    \multirow{3}{*}{\textbf{12. Light-Up}} & 
    \multirow{3}{*}{\includegraphics[height=55mm]{screenshot-game/lightup14.png}} &  
    \textbf{Rule:} You are a Light-Up player. You have to place light bulbs in the grid such that all empty cells are illuminated. The rules are: light bulbs illuminate their entire row and column until blocked by a wall; numbered walls must have exactly that many bulbs adjacent to them; bulbs cannot illuminate each other. Use Indexing starts at 0.\\ \cline{3-3}
    & & \textbf{Perception - Cell At:} \textbf{\{Rule\}}
    What is the value of the cell at (\{row\}, \{col\})? Choose from [*, s, w]. \\ \cline{3-3}
    & & \textbf{Perception and Direct Solution: }\textbf{\{Rule\}} Give me your response of the current game state in the screenshot (where \texttt{"*"} indicates an empty cell, \texttt{"s"} indicates a light bulb, and \texttt{"w"} indicates a wall) and your solution (where \texttt{"s"} indicates a light bulb and \texttt{"e"} indicates a cell left empty) in the following format. \textbackslash n\{
    \textbackslash n "perception": \{"current state of the grid as a 2D array"\}, \textbackslash n "answer": \{"solution as a 2D array"\} \textbackslash n\}\\ \cline{3-3}
    & & \textbf{Rule Following - Valid Actions:} \textbf{\{Rule\}} Is it valid to fill the cell at (\{row\}, \{col\}) with a light bulb? Choose from [valid, invalid]. \\ \cline{3-3}
    & & \textbf{Perception and Chain-of-Thought Reasoning: }\textbf{\{Rule\}} Give me your response of the current game state in the screenshot (where \texttt{"*"} indicates an empty cell, \texttt{"s"} indicates a light bulb, and \texttt{"w"} indicates a wall), \textbf{your step-by-step reasoning}, and your solution (where \texttt{"s"} indicates a light bulb and \texttt{"e"} indicates a cell left empty) in the following format. \textbackslash n\{
    \textbackslash n "perception": \{"current state of the grid as a 2D array"\}, \textbackslash n "think": \{"your step-by-step reasoning"\}, \textbackslash n "answer": \{"solution as a 2D array"\} \textbackslash n\}\\ \hline

    \multirow{3}{*}{\textbf{13. Nonogram}} & 
    \multirow{3}{*}{\includegraphics[height=55mm]{screenshot-game/nonogram-simple.png}} &  
    \textbf{Rule:} You are a Nonogram player. You need to fill in cells in a grid based on numbers at the side of the grid. For each row or column, the numbers indicate the lengths of consecutive shaded cells in that row/column, which must appear in the given order. For example, '2 3' means there must be exactly two blocks of shaded cells - first a block of 2 cells, then a block of 3 cells, with at least one empty cell between them. Each row/column must satisfy its given numbers exactly. Use 's' for shaded cells and 'e' for empty cells. Indexing starts at 0. \\ \cline{3-3}
    & & \textbf{Perception - Cell At:} \textbf{\{Rule\}}
     what is the state of the cell at (\{row\}, \{col\})? Choose from \{shaded, empty\}. \\ \cline{3-3}
    & & \textbf{Perception and Direct Solution: }\textbf{\{Rule\}} Give me your response of the current game state in the screenshot (where \texttt{"*"} indicates an empty cell and \texttt{"s"} indicates a shaded cell) and your solution (\texttt{"e"} indicates leaving empty) in the following format. \textbackslash n\{ \textbackslash n perception: \{current state of the grid as a 2D array\}, \textbackslash n answer: \{solution as a 2D array\} \textbackslash n\}  \\ \cline{3-3}
    & & \textbf{Rule Following - Valid Actions:} \textbf{\{Rule\}}  is it valid to shade the cell at (\{row\}, \{col\})? Choose from \{valid, invalid\}. \\ \cline{3-3}
    & & \textbf{Perception and Chain-of-Thought Reasoning: }\textbf{\{Rule\}} Give me your response of the current game state in the screenshot (where \texttt{"*"} indicates an empty cell and \texttt{"s"} indicates a shaded cell), \textbf{your step-by-step reasoning}, and your solution (\texttt{"e"} indicates leaving empty and \texttt{"s"} indicates a shaded) in the following format. \textbackslash n\{ \textbackslash n perception: \{current state of the grid as a 2D array\}, \textbackslash n think: \{your step-by-step reasoning\}, \textbackslash n answer: \{solution as a 2D array\} \textbackslash n\}  \\ \hline
    
    \multirow{3}{*}{\textbf{14. Odd-Even-Sudoku}} & 
    \multirow{3}{*}{\includegraphics[height=55mm]{screenshot-game/oddevensuodku27.png}} &  
    \textbf{Rule:} You are an Odd-Even-Sudoku player. You have to enter a numerical digit from 1 through N in each cell of a NxN grid. The rules are: unique numbers in each row, column, and sqrt(N)xsqrt(N) block. Additionally, white cells must contain even numbers, and black cells must contain odd numbers. Indexing starts at 0. \\ \cline{3-3}
    & & \textbf{Perception - Cell At:} \textbf{\{Rule\}}
     what is the value of the cell at (\{row\}, \{col\})? Choose from \{1, 2, ..., N, empty\}. \\ \cline{3-3}
    & & \textbf{Perception and Direct Solution: }\textbf{\{Rule\}} Give me your response of the current game state in the screenshot (where \texttt{"*"} indicates an empty cell) and your solution in the following format. \textbackslash n\{ \textbackslash n perception: \{current state of the grid as a 2D array\}, \textbackslash n answer: \{solution as a 2D array\} \textbackslash n\}  \\ \cline{3-3}
    & & \textbf{Rule Following - Valid Actions:} \textbf{\{Rule\}}  is it valid to fill the cell at (\{row\}, \{col\}) with value \{value\}? Choose from \{valid, invalid\}. \\ \cline{3-3}
    & & \textbf{Perception and Chain-of-Thought Reasoning: }\textbf{\{Rule\}} Give me your response of the current game state in the screenshot (where \texttt{"*"} indicates an empty cell), \textbf{your step-by-step reasoning}, and your solution in the following format. \textbackslash n\{ \textbackslash n perception: \{current state of the grid as a 2D array\}, \textbackslash n think: \{your step-by-step reasoning\}, \textbackslash n answer: \{solution as a 2D array\} \textbackslash n\}  \\ \hline
    
    \multirow{3}{*}{\textbf{15. Renzoku (Neighbors)}} & 
    \multirow{3}{*}{\includegraphics[height=55mm]{screenshot-game/renzoku0.png}} &  
    \textbf{Rule:} You are a Renzoku player. You have to enter a numerical digit from 1 through N in each cell of an NxN grid. The rules are: unique numbers in each row, column; A dot between 2 cells indicates that those 2 numbers should be consecutive. Otherwise, the numbers should be non-consecutive. Indexing starts at 0.\\ \cline{3-3}
    & & \textbf{Perception - Cell At:} \textbf{\{Rule\}}
     what is the value of the cell at (\{row\}, \{col\})? Choose from \{1, 2, ..., N, empty\}. \\ \cline{3-3}
    & & \textbf{Perception and Direct Solution: }\textbf{\{Rule\}} Give me your response of the current game state in the screenshot (where \texttt{"*"} indicates an empty cell) and your solution in the following format. \textbackslash n\{ \textbackslash n perception: \{current state of the grid as a 2D array\}, \textbackslash n answer: \{solution as a 2D array\} \textbackslash n\}  \\ \cline{3-3}
    & & \textbf{Rule Following - Valid Actions:} \textbf{\{Rule\}}  is it valid to fill the cell at (\{row\}, \{col\}) with value \{value\}? Choose from \{valid, invalid\}. \\ \cline{3-3}
    & & \textbf{Perception and Chain-of-Thought reasoning: }\textbf{\{Rule\}} Give me your response of the current game state in the screenshot (where \texttt{"*"} indicates an empty cell), \textbf{your step-by-step reasoning}, and your solution in the following format. \textbackslash n\{ \textbackslash n perception: \{current state of the grid as a 2D array\}, \textbackslash n think: \{your step-by-step reasoning\}, \textbackslash n answer: \{solution as a 2D array\} \textbackslash n\}  \\ \hline

    \end{tabular}%
    }
    \caption{Per-Puzzle Sample Screenshots and Query Templates. Part 3 (11th - 15th). }
    \label{tab:per-game-query-3}
    \end{table*}

\begin{table*}[t]
    \centering
    \renewcommand{\arraystretch}{1.6} 
    \resizebox{1.0\textwidth}{!}{%
    \begin{tabular}{|c|c|p{1.0\textwidth}|}
    \hline
    Name & Sample Screenshot & Description and Queries \\ \hline

    \multirow{3}{*}{\textbf{16. Skyscraper}} & 
    \multirow{3}{*}{\includegraphics[height=55mm]{screenshot-game/skyscraper5.png}} &  
    \textbf{Rule:} You are a Skyscraper puzzle player. You have to enter a numerical digit from 1 through N in each cell of an NxN grid. The numbers indicate the height of the skyscrapers. The numbers on the sides of the grid indicate how many skyscrapers would you see if you look in the direction of the arrow. Indexing starts at 0. \\ \cline{3-3}
    & & \textbf{Perception - Cell At:} \textbf{\{Rule\}}
     what is the value of the cell at (\{row\}, \{col\})? Choose from \{1, 2, ..., N, empty\}. \\ \cline{3-3}
    & & \textbf{Perception and Direct Solution: }\textbf{\{Rule\}} Give me your response of the current game state in the screenshot (where \texttt{"*"} indicates an empty cell) and your solution in the following format. \textbackslash n\{ \textbackslash n perception: \{current state of the grid as a 2D array\}, \textbackslash n answer: \{solution as a 2D array\} \textbackslash n\}  \\ \cline{3-3}
    & & \textbf{Rule Following - Valid Actions:} \textbf{\{Rule\}}  is it valid to fill the cell at (\{row\}, \{col\}) with value \{value\}? Choose from \{valid, invalid\}. \\ \cline{3-3}
    & & \textbf{Perception and Chain-of-Thought reasoning: }\textbf{\{Rule\}} Give me your response of the current game state in the screenshot (where \texttt{"*"} indicates an empty cell), \textbf{your step-by-step reasoning}, and your solution in the following format. \textbackslash n\{ \textbackslash n perception: \{current state of the grid as a 2D array\}, \textbackslash n think: \{your step-by-step reasoning\}, \textbackslash n answer: \{solution as a 2D array\} \textbackslash n\}  \\ \hline
    
    \multirow{5}{*}{\textbf{17. Star-Battle}} & 
    \multirow{5}{*}{\includegraphics[height=55mm]{screenshot-game/starbattle2.png}} &  
    \textbf{Rule:} You are a Star-Battle player. You have to place stars on the grid such that each row, column, and region contains exactly one star. Additional rule is: stars cannot touch each other, not even diagonally. Use Indexing starts at 0\\ \cline{3-3}
    & & \textbf{Perception - Cell At:} \textbf{\{Rule\}}
     what is in the cell at (\{row\}, \{col\})? Choose from \{star, empty\}. \\ \cline{3-3}
    & & \textbf{Perception and Direct Solution: }\textbf{\{Rule\}} Give me your response of the current game state in the screenshot (where \texttt{"*"} indicates an empty cell) and your solution (where \texttt{"s"} indicates a star and \texttt{"e"} indicates leaving empty) in the following format. \textbackslash n\{ \textbackslash n perception: \{current state of the grid as a 2D array\}, \textbackslash n answer: \{solution as a 2D array\} \textbackslash n\}  \\ \cline{3-3}
    & & \textbf{Rule Following - Valid Actions:} \textbf{\{Rule\}}  is it valid to \textbf{place a star} at (\{row\}, \{col\})? Choose from \{valid, invalid\}. \\ \cline{3-3}
    & & \textbf{Perception and Chain-of-Thought Reasoning: }\textbf{\{Rule\}} Give me your response of the current game state in the screenshot (where \texttt{"*"} indicates an empty cell), \textbf{your step by step reasoning}, and your solution (where \texttt{"s"} indicates a star and \texttt{"e"} indicates leaving empty) in the following format. \textbackslash n\{ \textbackslash n perception: \{current state of the grid as a 2D array\}, \textbackslash n think: \{your step by step reasoning\}, \textbackslash n answer: \{solution as a 2D array\} \textbackslash n\}  \\ \hline

    \multirow{5}{*}{\textbf{18. Sudoku}} & 
    \multirow{5}{*}{\includegraphics[height=55mm]{screenshot-game/sudoku-3.png}} &  
    \textbf{Rule:} You are a Sudoku player. You have to enter a numerical digit from 1 through N in each cell of a NxN grid made up of four sqrt(N)xsqrt(N) blocks, \textbackslash n The rule is to make sure unique numbers in each row, column, and block. Indexing starts at 0. \\ \cline{3-3}
    & & \textbf{Perception - Cell At:} \textbf{\{Rule\}}
     what is the value of the cell at (\{row\}, \{col\})? Choose from \{1, 2, ..., N, empty\}. \\ \cline{3-3}
    & & \textbf{Perception and Direct Solution: }\textbf{\{Rule\}} Give me your response of the current game state in the screenshot (where \texttt{"*"} indicates an empty cell) and your solution in the following format. \textbackslash n\{ \textbackslash n perception: \{current state of the grid as a 2D array\}, \textbackslash n answer: \{solution as a 2D array\} \textbackslash n\}  \\ \cline{3-3}
    & & \textbf{Rule Following - Valid Actions:} \textbf{\{Rule\}}  is it valid to fill the cell at (\{row\}, \{col\}) with value \{value\}? Choose from \{valid, invalid\}. \\ \cline{3-3}
    & & \textbf{Perception and Chain-of-Thought Reasoning: }\textbf{\{Rule\}} Give me your response of the current game state in the screenshot (where \texttt{"*"} indicates an empty cell), \textbf{your step-by-step reasoning}, and your solution in the following format. \textbackslash n\{ \textbackslash n perception: \{current state of the grid as a 2D array\}, \textbackslash n think: \{your step-by-step reasoning\}, \textbackslash n answer: \{solution as a 2D array\} \textbackslash n\}  \\ \hline

    \multirow{3}{*}{\textbf{19. Thermometers}} & 
    \multirow{3}{*}{\includegraphics[height=55mm]{screenshot-game/thermometers1.png}} &  
    \textbf{Rule:} You are a Thermometers puzzle player. You need to fill thermometers. The numbers on the sides indicate how many filled cells must be in each row and column. In the end, all thermometers must be filled from their bulb (start) to their top, without gaps. Use Indexing starts at 0.\\ \cline{3-3}
    & & \textbf{Perception - Cell At:} \textbf{\{Rule\}}
     what is in the cell at (\{row\}, \{col\})? Choose from \{filled, empty\}. \\ \cline{3-3}
    & & \textbf{Perception and Direct Solution: }\textbf{\{Rule\}} Give me your response of the current game state in the screenshot (where \texttt{"*"} indicates not filled) and your solution (where \texttt{"s"} indicates a filled cell and \texttt{"e"} indicates leaving not filled) in the following format. \textbackslash n\{ \textbackslash n perception: \{current state of the grid as a 2D array\}, \textbackslash n answer: \{solution as a 2D array\} \textbackslash n\}  \\ \cline{3-3}
    & & \textbf{Rule Following - Valid Actions:} \textbf{\{Rule\}}  is it valid to \textbf{assign filled} at (\{row\}, \{col\})? Choose from \{valid, invalid\}. \\ \cline{3-3}
    & & \textbf{Perception and Chain-of-Thought Reasoning: }\textbf{\{Rule\}} Give me your response of the current game state in the screenshot (where \texttt{"*"} indicates a not filled cell), \textbf{your step-by-step reasoning}, and your solution (where \texttt{"s"} indicates a filled cell and \texttt{"e"} indicates leaving not filled) in the following format. \textbackslash n\{ \textbackslash n perception: \{current state of the grid as a 2D array\}, \textbackslash n think: \{your step-by-step reasoning\}, \textbackslash n answer: \{solution as a 2D array\} \textbackslash n\}  \\ \hline

    \multirow{3}{*}{\textbf{20. Trees-and-Tents}} & 
    \multirow{3}{*}{\includegraphics[height=55mm]{screenshot-game/trees0ded.png}} &  
    \textbf{Rule:} You are a Trees-and-Tents player. You need to place tents on a grid with trees. Each tree must be paired with exactly one tent that is horizontally or vertically adjacent to it (a 1-to-1 relationship). Tents cannot touch each other, even diagonally. The numbers on the sides indicate how many tents must be in each row and column. Use Indexing starts at 0. \\ \cline{3-3}
    & & \textbf{Perception - Cell At:} \textbf{\{Rule\}}
     what is the value of the cell at (\{row\}, \{col\})? Choose from \{tree, tent, empty\}. \\ \cline{3-3}
    & & \textbf{Perception and Direct Solution: }\textbf{\{Rule\}} Give me your response of the current game state in the screenshot (where \texttt{"tr"} indicates a tree, \texttt{"tt"} indicates a tent, and \texttt{"*"} indicates an empty cell) and your solution (\texttt{"e"} indicates leaving empty) in the following format. \textbackslash n\{ \textbackslash n perception: \{current state of the grid as a 2D array\}, \textbackslash n answer: \{solution as a 2D array\} \textbackslash n\}  \\ \cline{3-3}
    & & \textbf{Rule Following - Valid Actions:} \textbf{\{Rule\}}  is it valid to fill the cell at (\{row\}, \{col\}) with value \{value\}? Choose from \{valid, invalid\}. \\ \cline{3-3}
    & & \textbf{Perception and Chain-of-Thought Reasoning: }\textbf{\{Rule\}} Give me your response of the current game state in the screenshot (where \texttt{"tr"} indicates a tree, \texttt{"tt"} indicates a tent, and \texttt{"*"} indicates an empty cell), \textbf{your step-by-step reasoning}, and your solution (\texttt{"e"} indicates leaving empty) in the following format. \textbackslash n\{ \textbackslash n perception: \{current state of the grid as a 2D array\}, \textbackslash n think: \{your step-by-step reasoning\}, \textbackslash n answer: \{solution as a 2D array\} \textbackslash n\}  \\ \hline

    \end{tabular}%
    }
    \caption{Per-Puzzle Sample Screenshots and Query Templates. Part 4 (16th - 20th). }
    \label{tab:per-game-query-4}
    \end{table*}

\begin{figure*}[p] 
\centering 

\refstepcounter{section}  %
\phantomsection          %
\addcontentsline{toc}{section}{Results}  %
\addcontentsline{toc}{subsection}{Medium and Hard Level Overall Evaluation of Off-the-Shelf Models (w/ CoT)}  %

\includegraphics[width=0.98\textwidth, page=1]{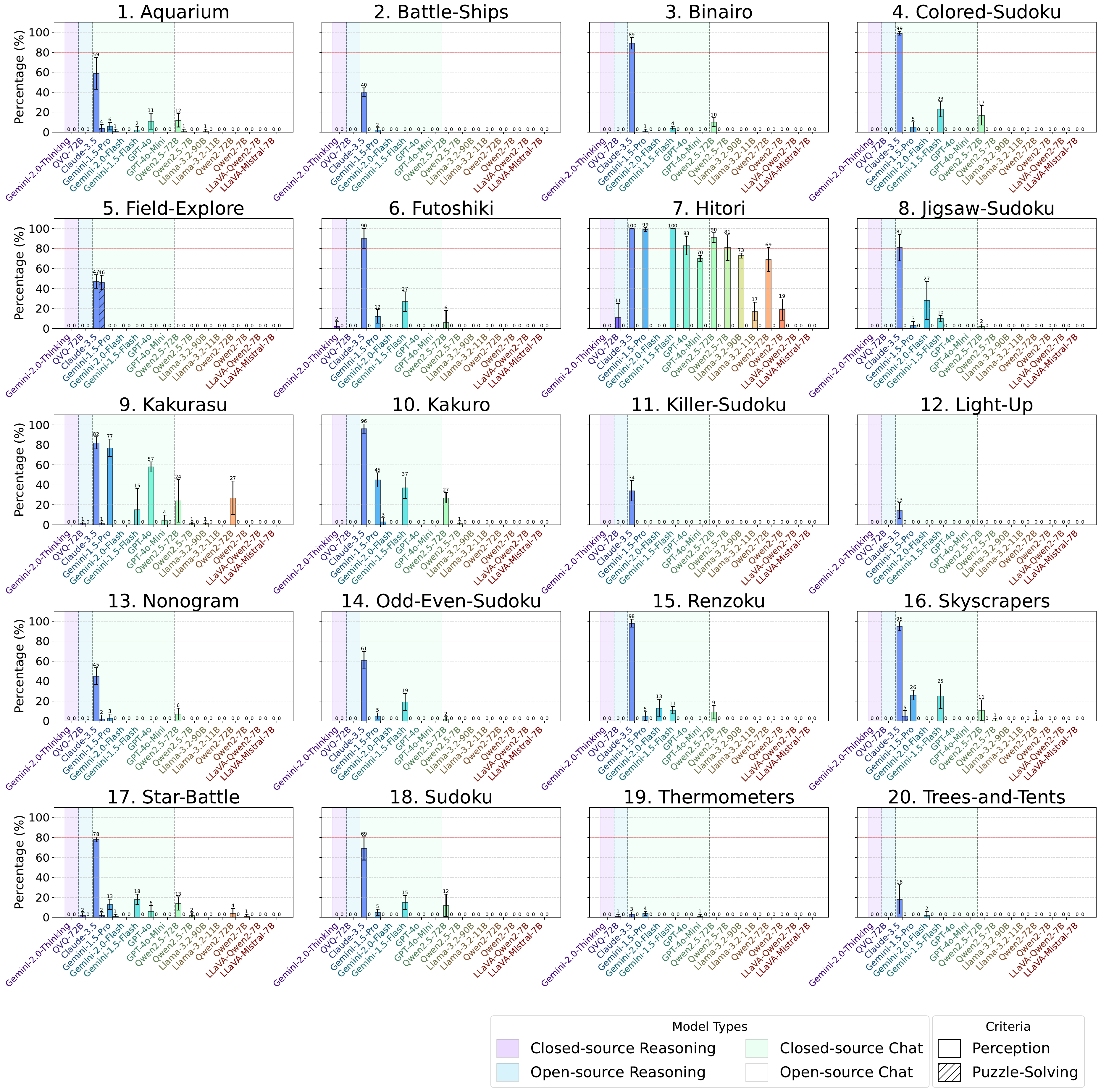} \caption{\textbf{Off-the-Shelf LVLMs' Overall Perception and Puzzle-Solving Evaluation at Level-\Medium.} We report both correct perception rate and puzzle-solving rate evaluations. We omit Gemini-Thinking model due to the their rate limit. Compared with \easy-level in the main paper. Among all models, only Claude is able to perceive most several puzzles correctly, such as Jigsaw-Sudoku. All LVLMs fail to solve any of the 20 puzzles with higher than 10 percentage.
(Best viewed on a screen when zoomed in)} \label{fig:benchmark-off-the-shelf-medium-cot} \end{figure*}

\begin{figure*}[p] 
\centering 

\includegraphics[width=0.98\textwidth, page=1]{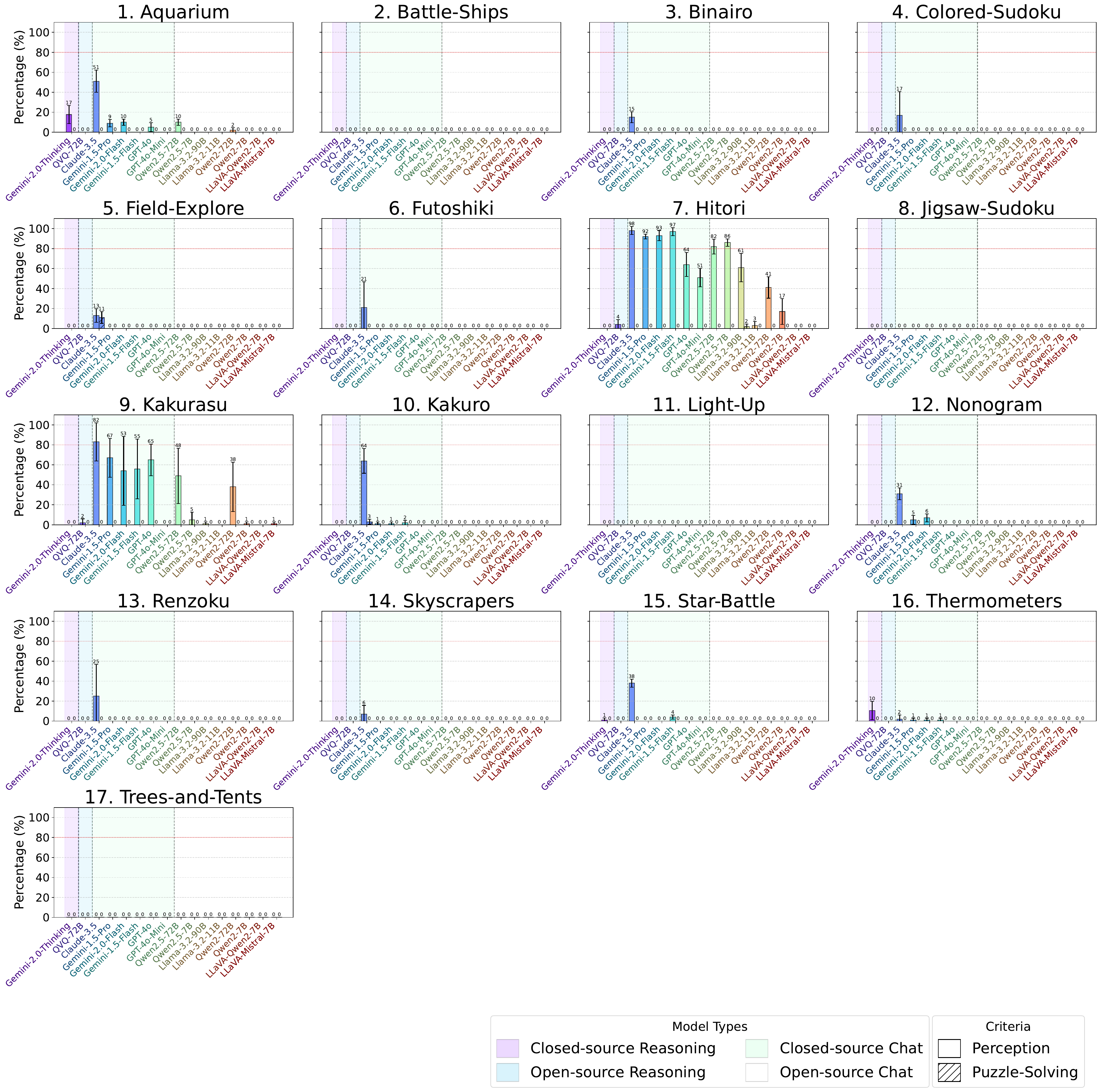} \caption{\textbf{Off-the-Shelf LVLMs' Overall Perception and Puzzle-Solving Evaluation at Level-\Hard~with CoT.} We report both correct perception rate and puzzle-solving rate evaluations. We omit Gemini-Thinking model due to the their rate limit. All LVLMs has lower perception and puzzle-solving accuracy, showing the \hard-level is more challenging than \medium-level.
(Best viewed on a screen when zoomed in)} \label{fig:benchmark-off-the-shelf-hard-cot} \end{figure*}

\begin{figure*}[p] \centering 
\refstepcounter{section}  %
\phantomsection          %
\addcontentsline{toc}{subsection}{Easy, Medium and Hard Overall Level Evaluation of Off-the-Shelf Models (w/o CoT)}  %

\includegraphics[width=0.98\textwidth, page=1]{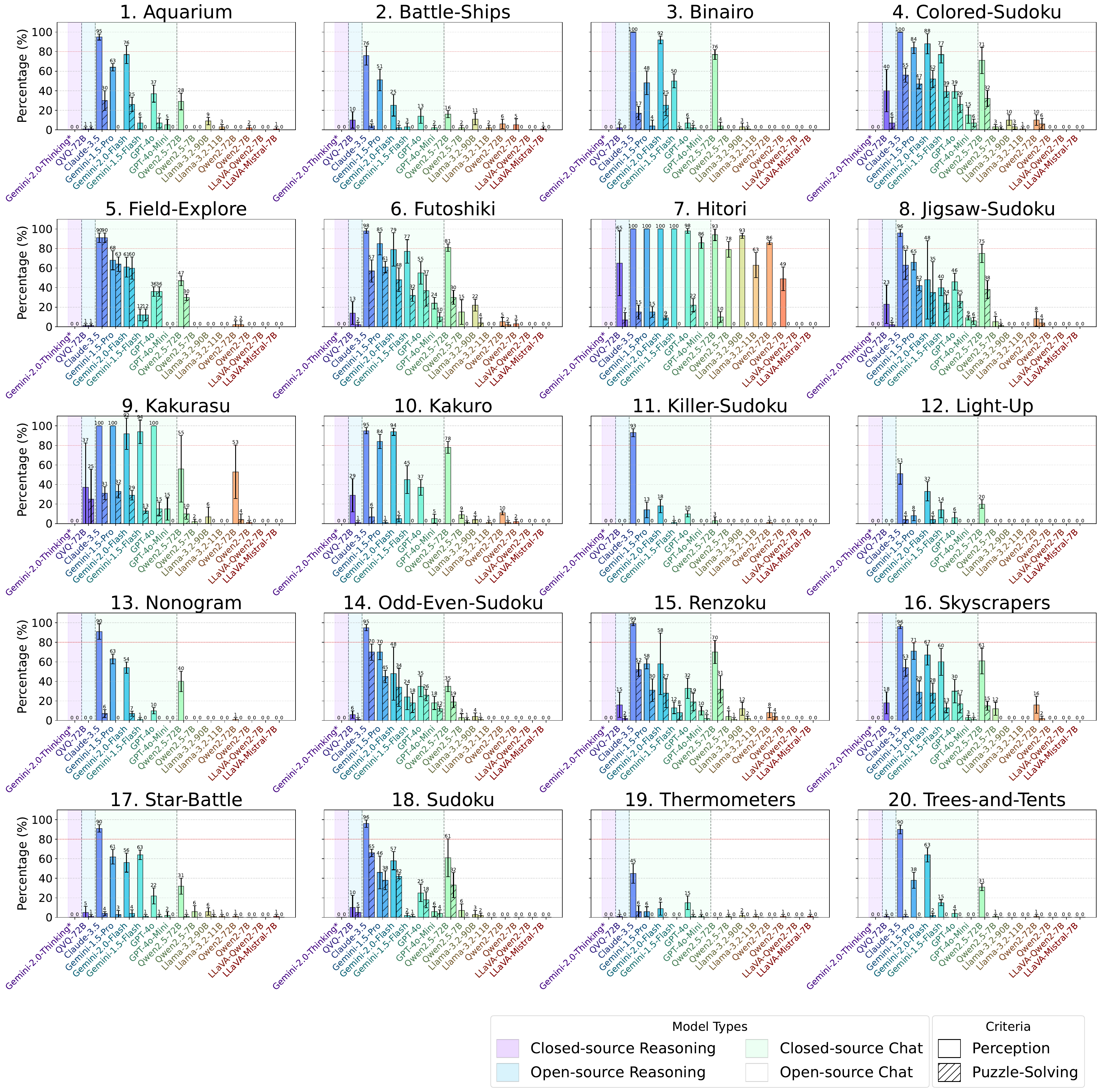} \caption{\textbf{Off-the-Shelf LVLMs' Overall Perception and Puzzle-Solving Evaluation at Level-\Easy~without CoT.} We report both correct perception rate and puzzle-solving rate evaluations. We omit Gemini-Thinking model due to the their rate limit. We observe similar trend as the CoT version in the main paper that close-source LVLMs has better performance, and our benchmark is challenging even at the \easy-level, especially for open-source LVLMs. 
(Best viewed on a screen when zoomed in)} \label{fig:benchmark-off-the-shelf-easy-wo-cot} \end{figure*}

\begin{figure*}[p] \centering \includegraphics[width=0.98\textwidth, page=1]{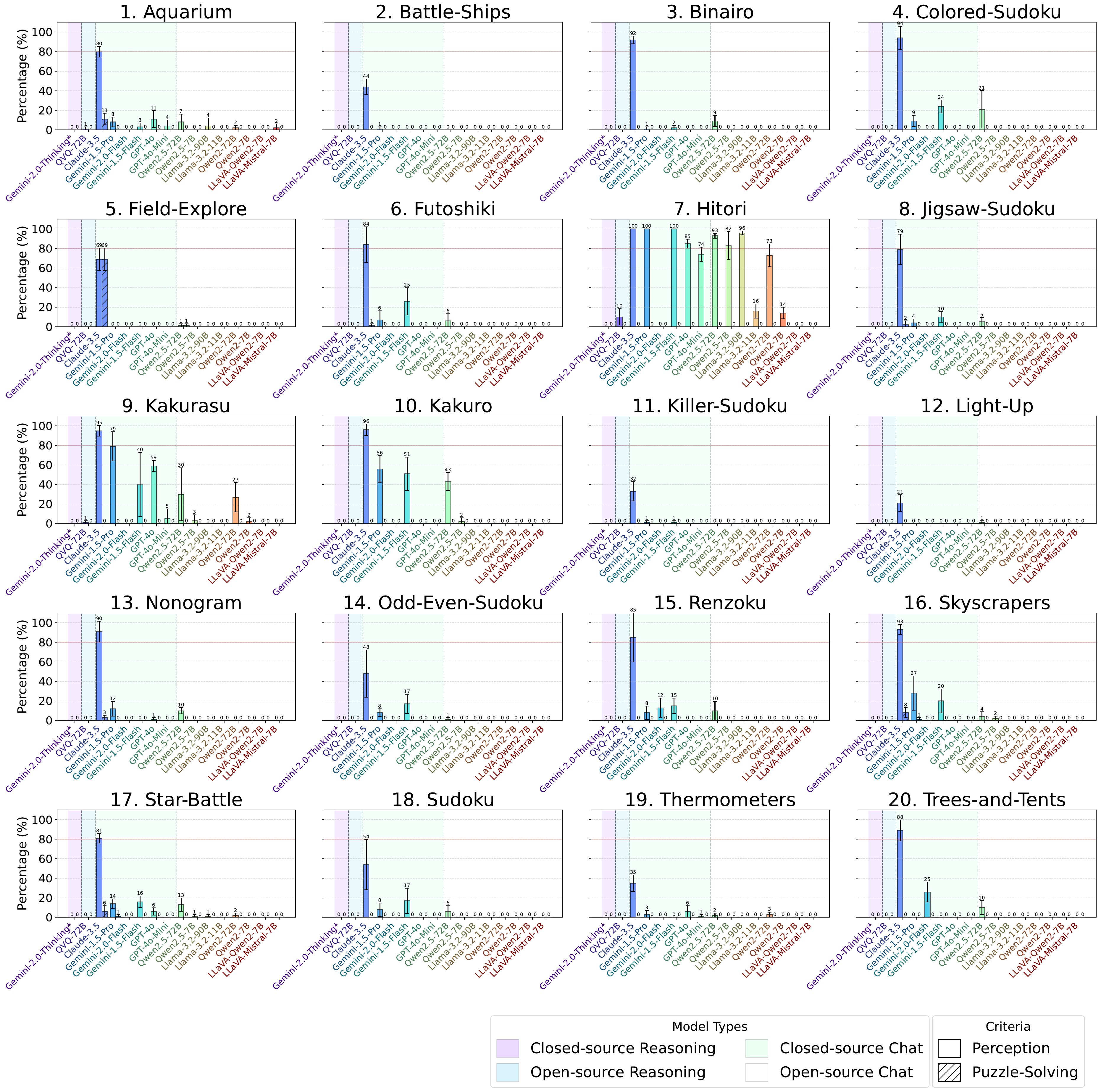} \caption{\textbf{Off-the-Shelf LVLMs' Overall Perception and Puzzle-Solving Evaluation at Level-\Medium~without CoT.} We report both correct perception rate and puzzle-solving rate evaluations. We omit Gemini-Thinking model due to the their rate limit.
(Best viewed on a screen when zoomed in)} \label{fig:benchmark-off-the-shelf-medium-wo-cot} \end{figure*}

\begin{figure*}[p] \centering \includegraphics[width=0.98\textwidth, page=1]{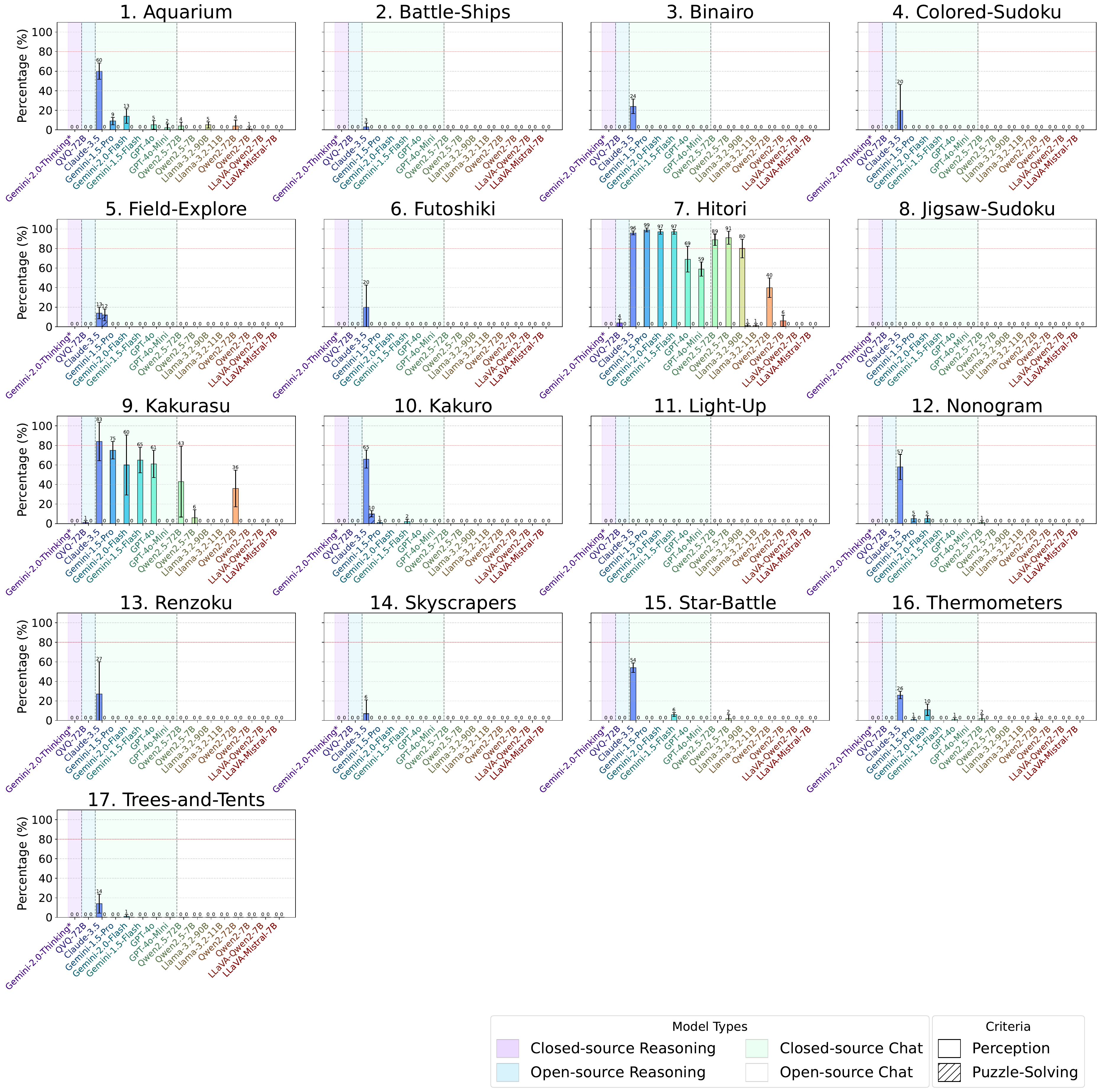} \caption{\textbf{Off-the-Shelf LVLMs' Overall Perception and Puzzle-Solving Evaluation at Level-\Hard~without CoT.} We report both correct perception rate and puzzle-solving rate evaluations. We omit Gemini-Thinking model due to the their rate limit.
(Best viewed on a screen when zoomed in)} \label{fig:benchmark-off-the-shelf-hard-wo-cot} \end{figure*}

\begin{figure*}[p] \centering 

\refstepcounter{section} 
\phantomsection         
\addcontentsline{toc}{subsection}{Easy, Medium and Hard Level Cell-Level Perception Evaluation of Off-the-Shelf Models}  

\includegraphics[width=0.98\textwidth, page=1]{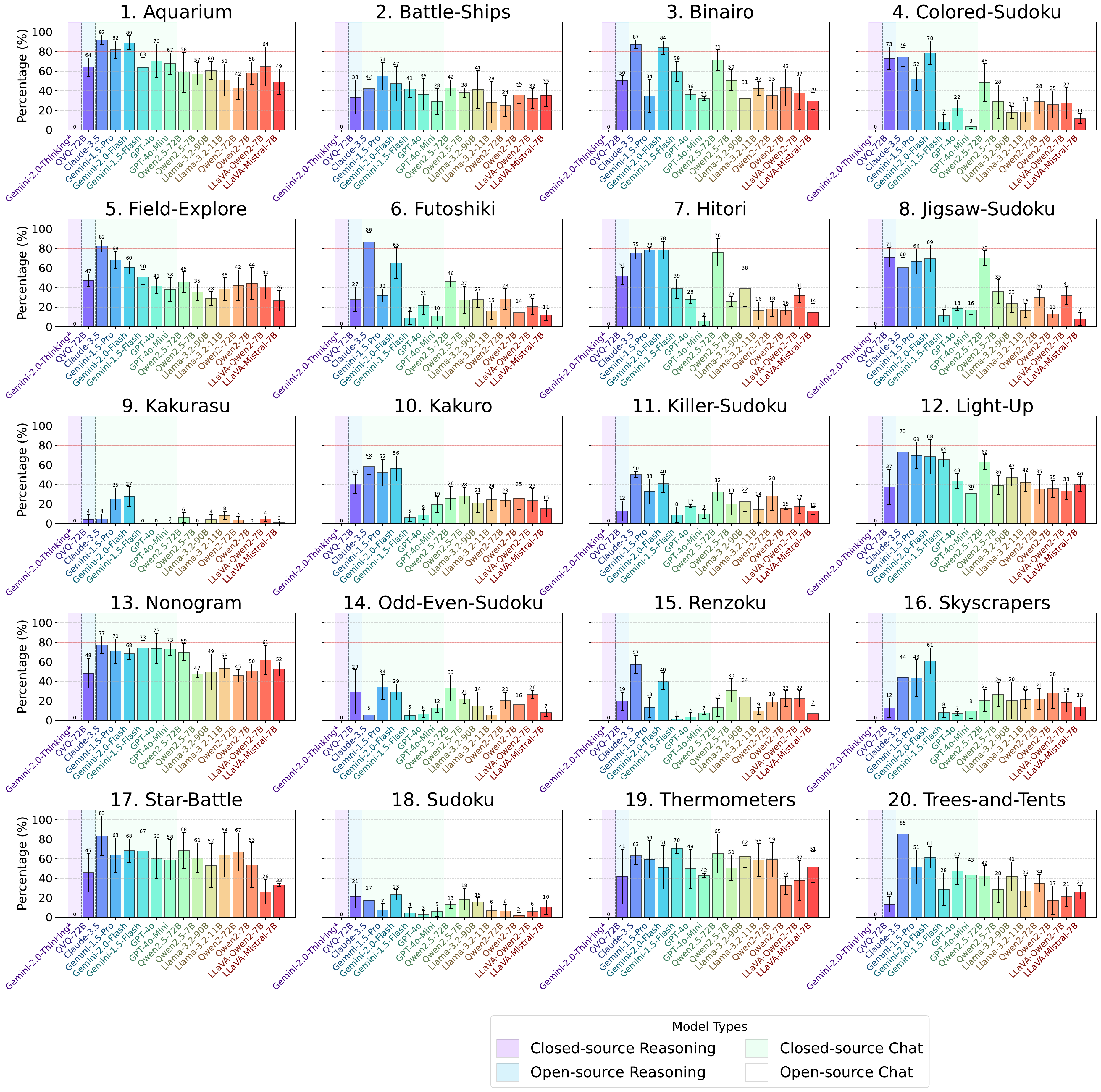} \caption{\textbf{Off-the-Shelf LVLMs' Cell-Level Perception Evaluation on Level-\Easy.} We report perception accuracy. We omit Gemini-Thinking model due to the rate limit. We provide evaluation at cell-level, by assessing models perception when asked querying on a single cell. 
(Best viewed on a screen when zoomed in)} \label{fig:benchmark-off-the-shelf-easy-cell-at} \end{figure*}

\begin{figure*}[p] \centering \includegraphics[width=0.98\textwidth, page=1]{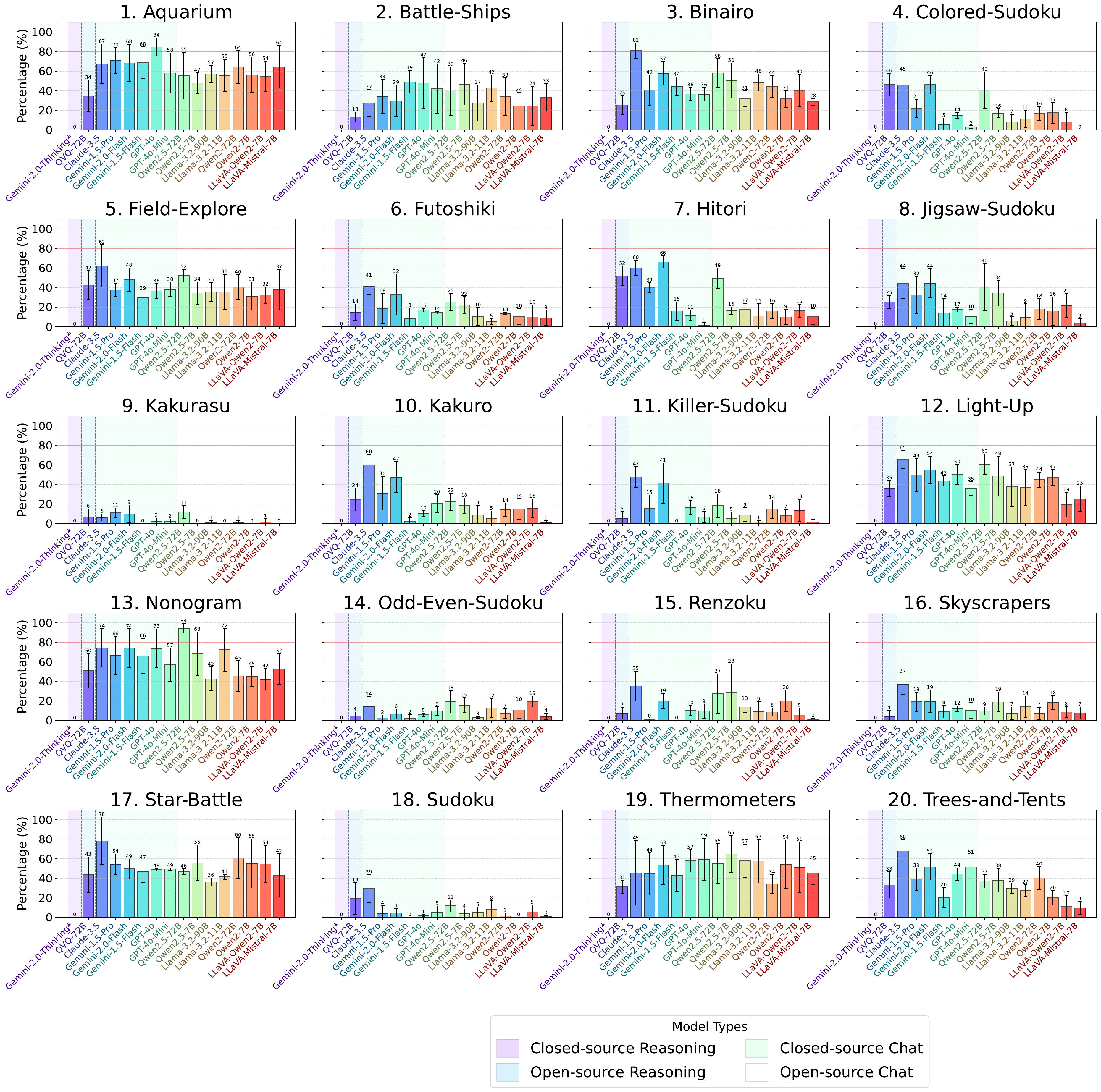} \caption{\textbf{Off-the-Shelf LVLMs' Cell-Level Perception Evaluation on Level-\Medium.} We report perception accuracy. We omit Gemini-Thinking model due to the rate limit.
(Best viewed on a screen when zoomed in)} \label{fig:benchmark-off-the-shelf-medium-cell-at} \end{figure*}

\begin{figure*}[p] \centering \includegraphics[width=0.98\textwidth, page=1]{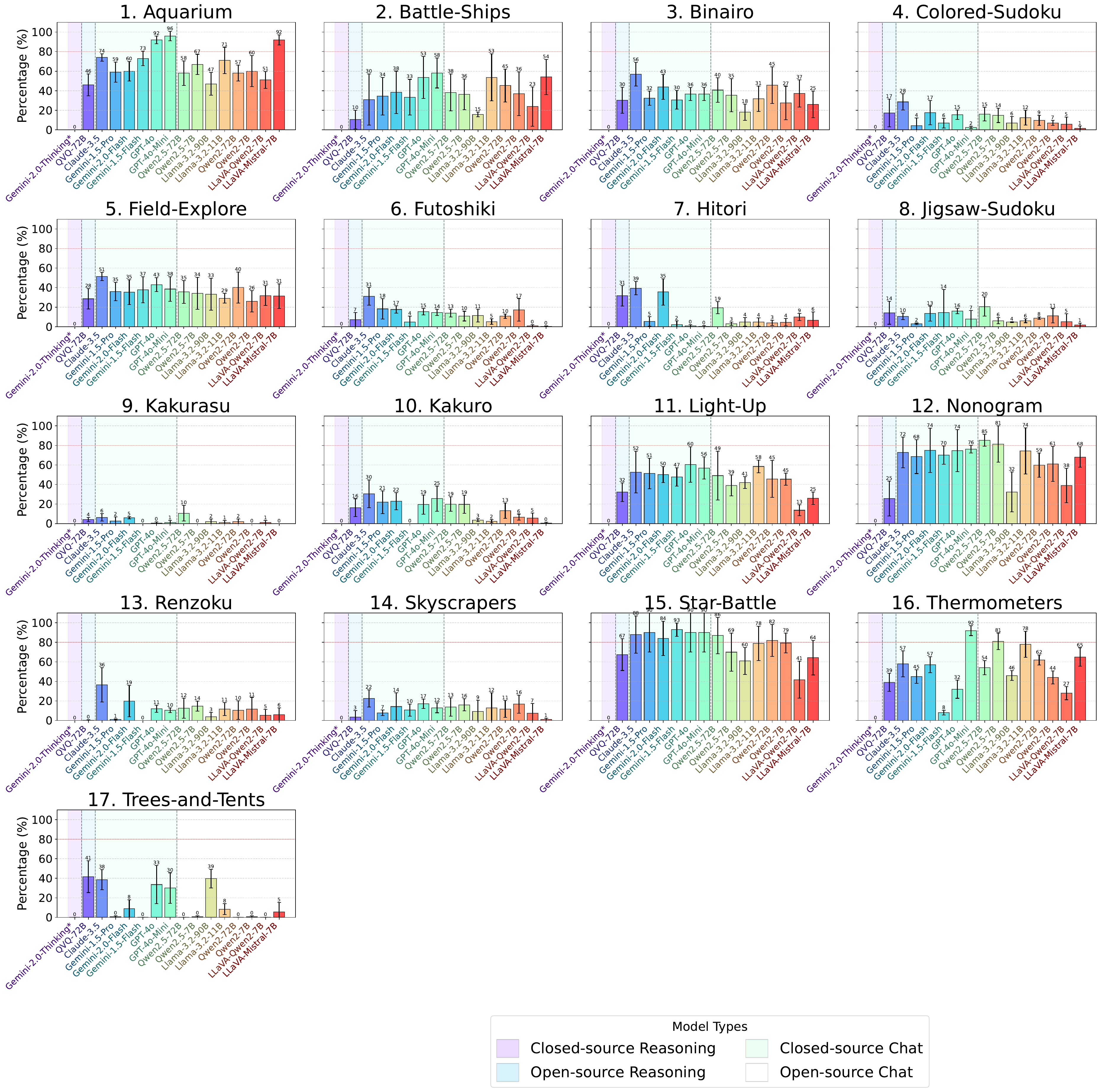} \caption{\textbf{Off-the-Shelf LVLMs' Cell-Level Perception Evaluation on Level-\Hard.} We report perception accuracy. We omit Gemini-Thinking model due to the rate limit.
(Best viewed on a screen when zoomed in)} \label{fig:benchmark-off-the-shelf-hard-cell-at} \end{figure*}

\begin{figure*}[p] \centering 

\refstepcounter{section} 
\phantomsection         
\addcontentsline{toc}{subsection}{Easy, Medium and Hard Level Rule Following Evaluation of Off-the-Shelf Models}  

\includegraphics[width=0.98\textwidth, page=1]{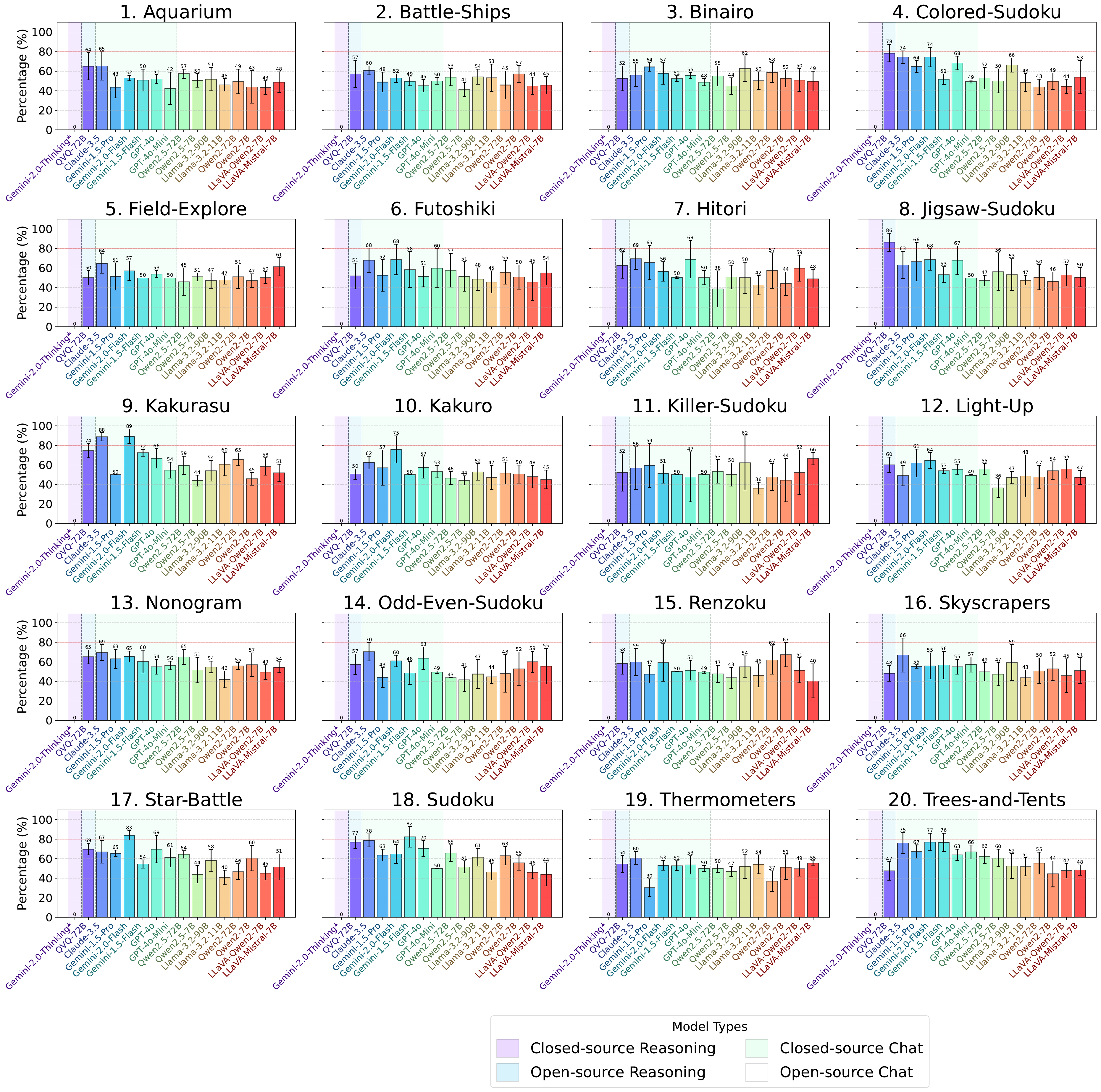} \caption{\textbf{Off-the-Shelf LVLMs' Action-Level Rule-Following Evaluation on Level-\Easy.} We report responses' correctness. We omit Gemini-Thinking model due to the rate limit.
(Best viewed on a screen when zoomed in)} \label{fig:benchmark-off-the-shelf-easy-valid} \end{figure*}

\begin{figure*}[p] \centering \includegraphics[width=0.98\textwidth, page=1]{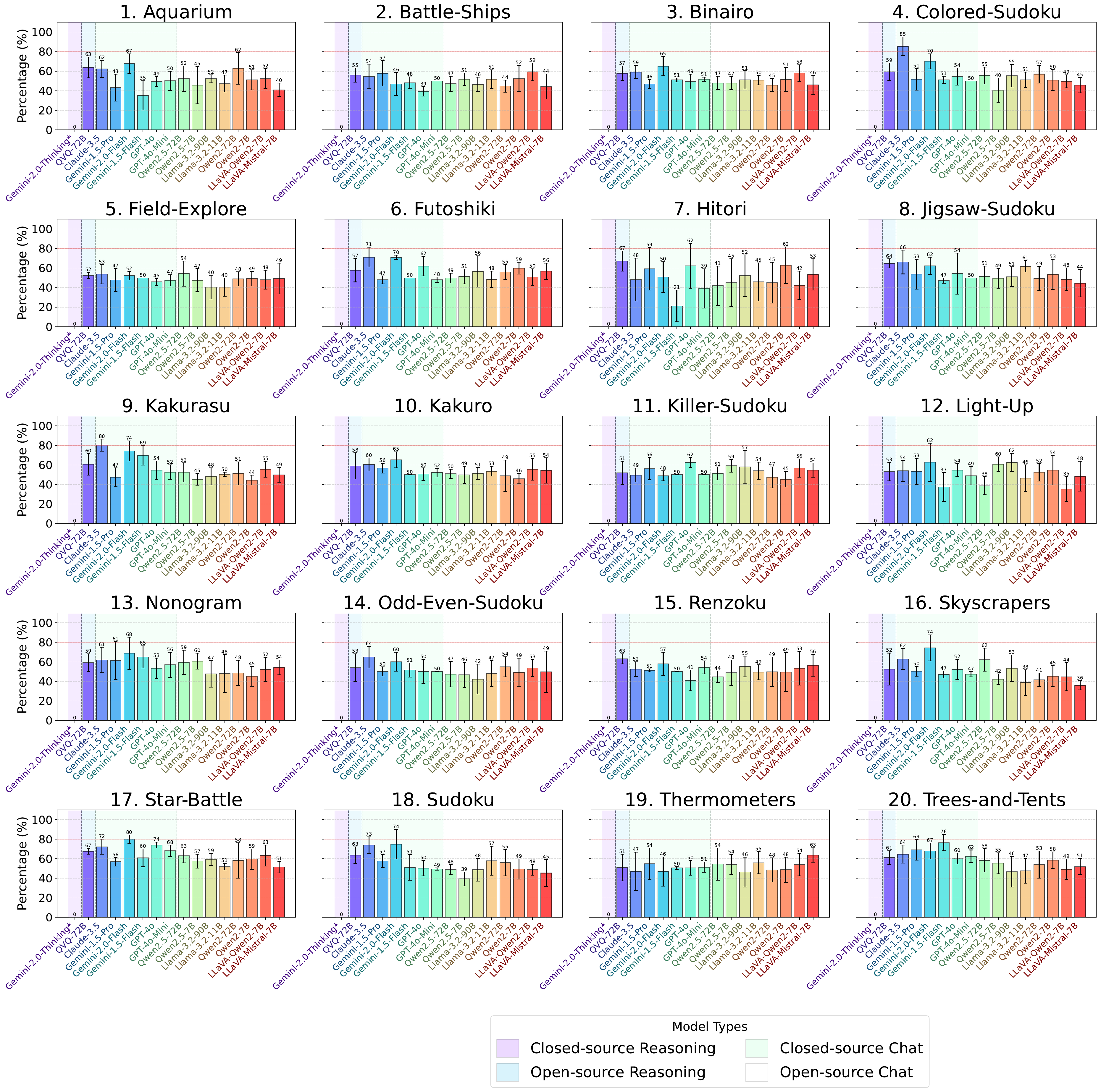} \caption{\textbf{Off-the-Shelf LVLMs' Action-Level Rule-Following Evaluation on Level-\Medium.} We report responses' correctness. We omit Gemini-Thinking model due to the rate limit.
(Best viewed on a screen when zoomed in)} \label{fig:benchmark-off-the-shelf-medium-valid} \end{figure*}

\begin{figure*}[p] \centering \includegraphics[width=0.98\textwidth, page=1]{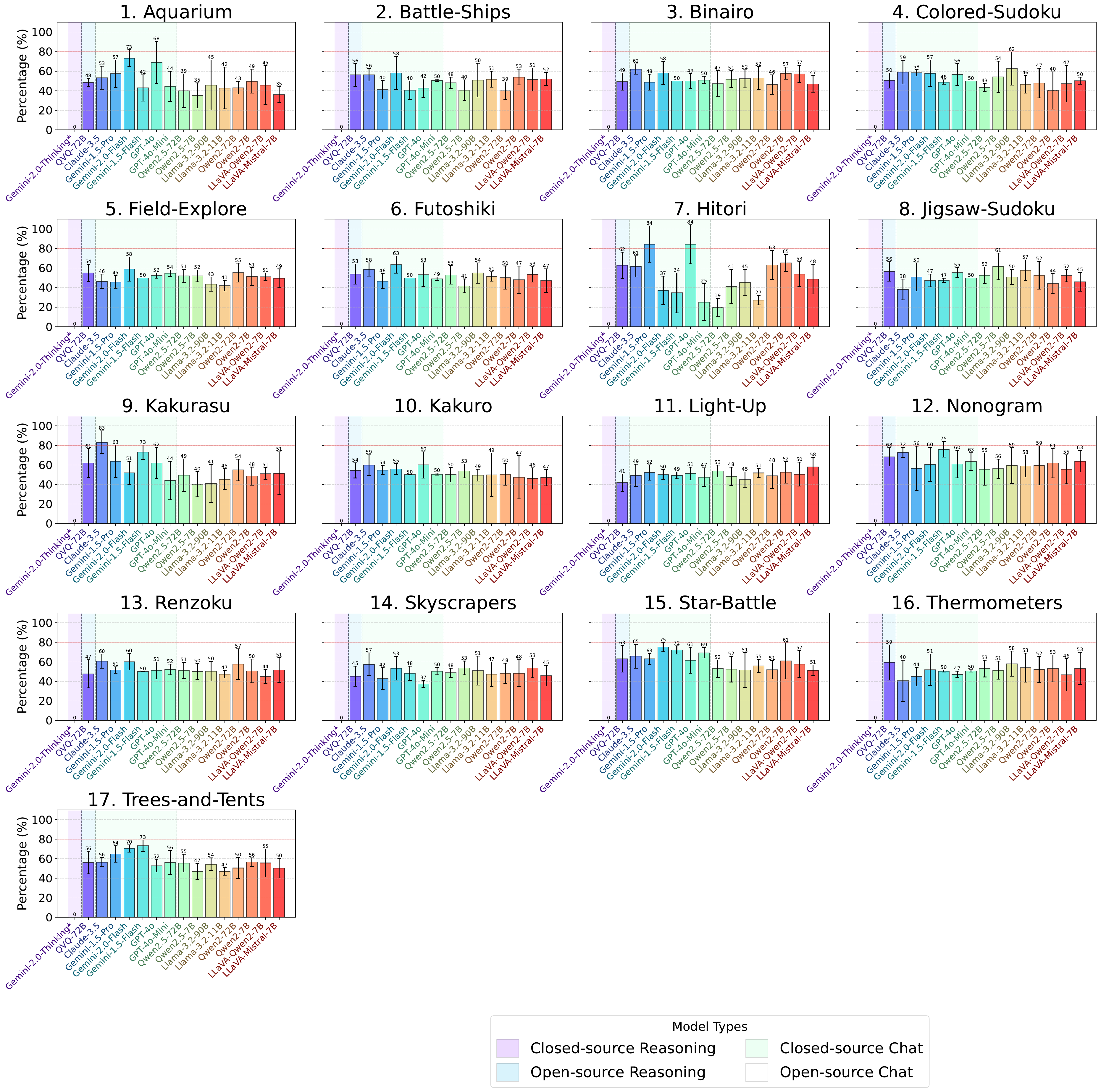} \caption{\textbf{Off-the-Shelf LVLMs' Action-Level Rule-Following Evaluation on Level-\Hard.} We report responses' correctness. We omit Gemini-Thinking model due to the rate limit.
(Best viewed on a screen when zoomed in)} \label{fig:benchmark-off-the-shelf-hard-valid} \end{figure*}

\begin{figure*}[t] \centering 

\refstepcounter{section} 
\phantomsection         
\addcontentsline{toc}{subsection}{Easy Level Overall Evaluation of Off-the-Shelf Models v.s. Clue Number (w/ CoT)}

\includegraphics[width=1.0\textwidth, page=1]{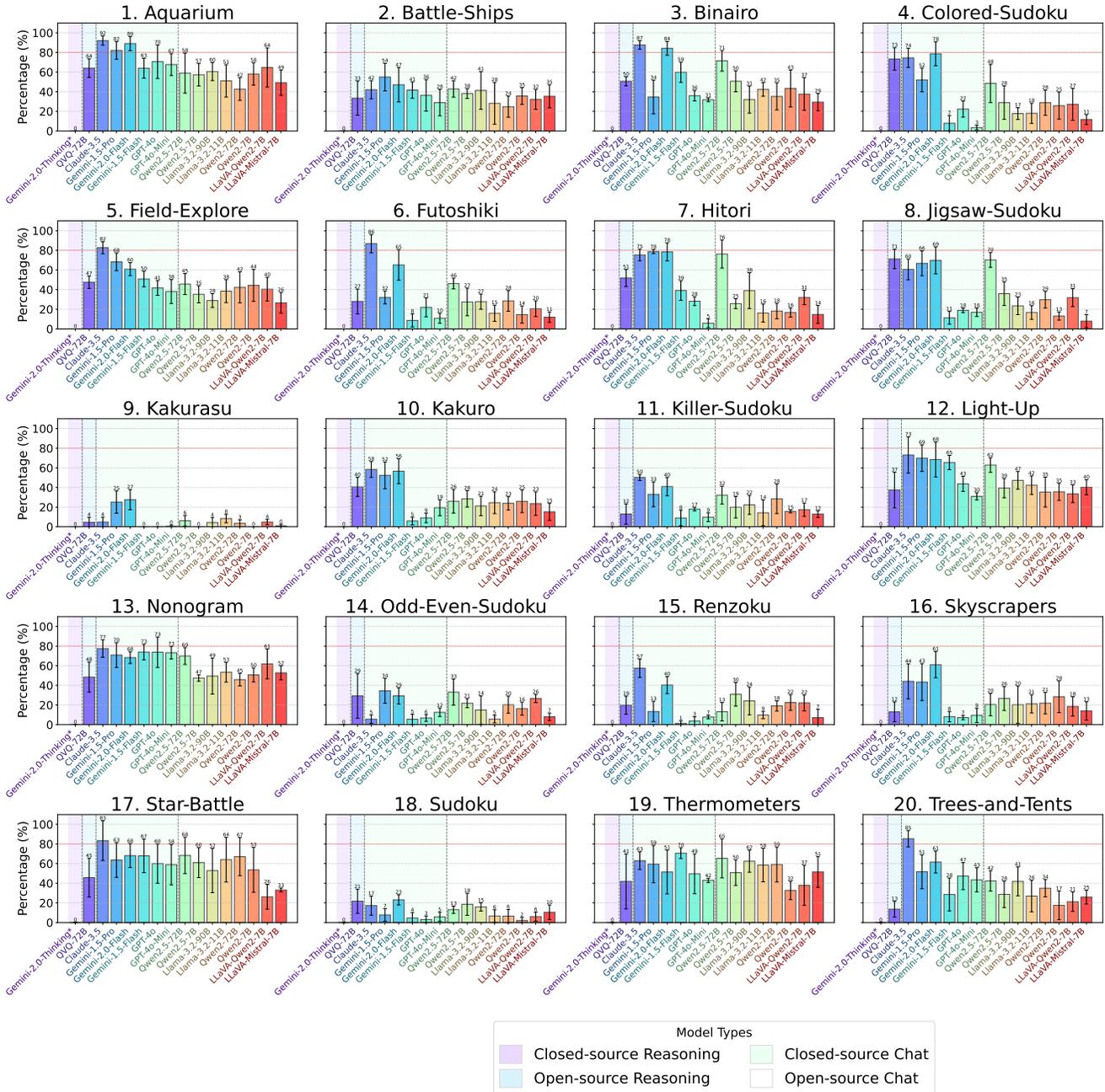} \caption{\textbf{Off-the-Shelf LVLMs’ Clue Number Versus Overall Accuracy at Level-\Easy with CoT.} In general, the more clues provided, the easier puzzles are, resulting in higher perception and puzzle solving accuracy. (Best viewed on a screen when zoomed in)} \label{fig:benchmark-off-the-shelf-easy-clue-number-cot} \end{figure*}

\begin{figure*}[t] \centering 
\refstepcounter{section} 
\phantomsection         
\addcontentsline{toc}{subsection}{Easy Level Cell-Level Perception Evaluation of Off-the-Shelf Models v.s. Clue Number}
\includegraphics[width=1.0\textwidth, page=1]{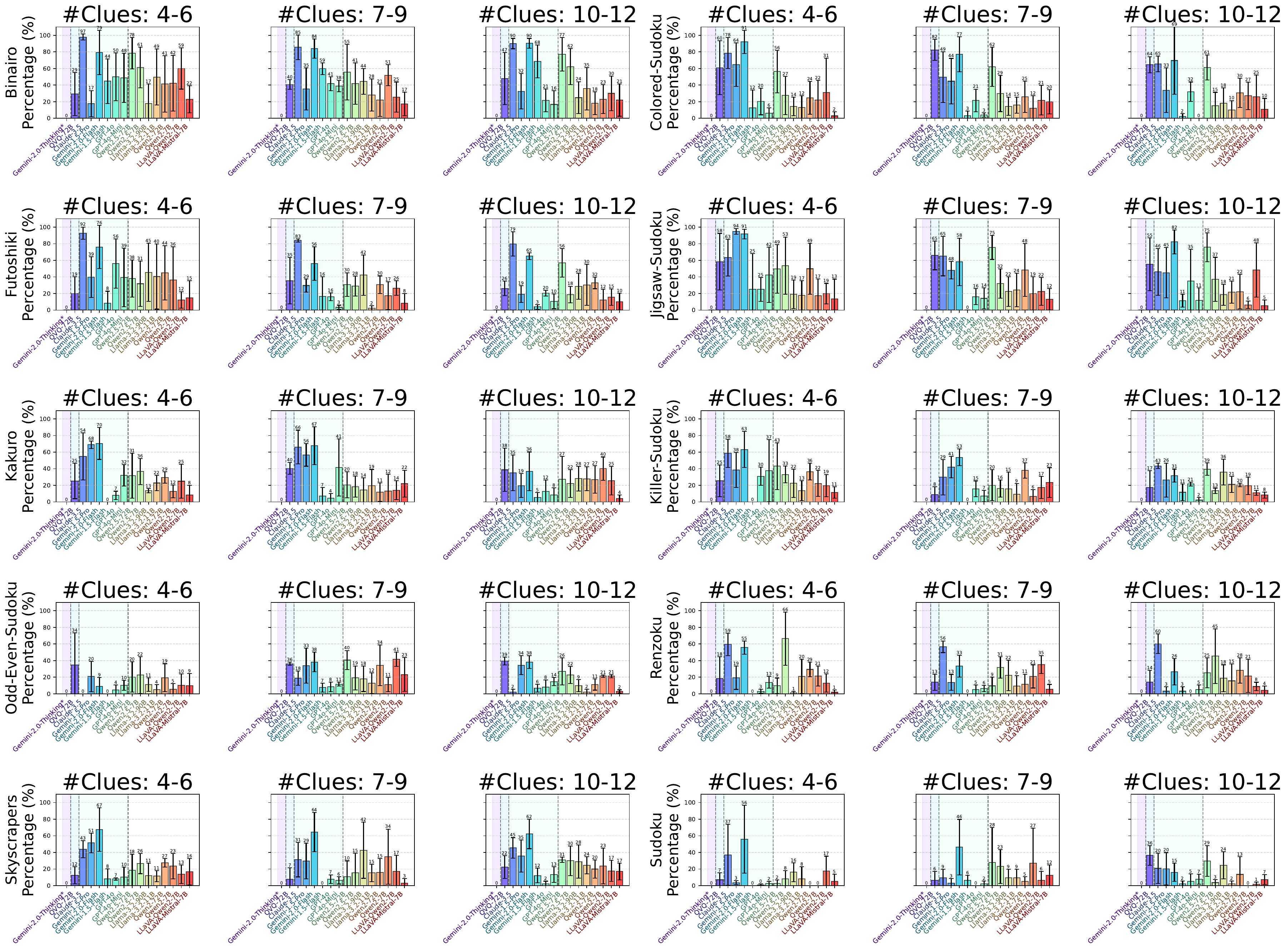} \caption{\textbf{Off-the-Shelf LVLMs’ Clue Number Versus Cell-Level Perception Accuracy at Level-\Easy.} In general, the more clues provided, the easier puzzles are, resulting in higher perception and puzzle solving accuracy. (Best viewed on a screen when zoomed in)} \label{fig:benchmark-off-the-shelf-easy-clue-number-cell-at} \end{figure*}

\begin{figure*}[t] \centering 
\refstepcounter{section} 
\phantomsection         
\addcontentsline{toc}{subsection}{Easy and Medium Level Overall Evaluation of Off-the-Shelf Models (w/ CoT, Text Version)}
\includegraphics[width=1.0\textwidth, page=1]{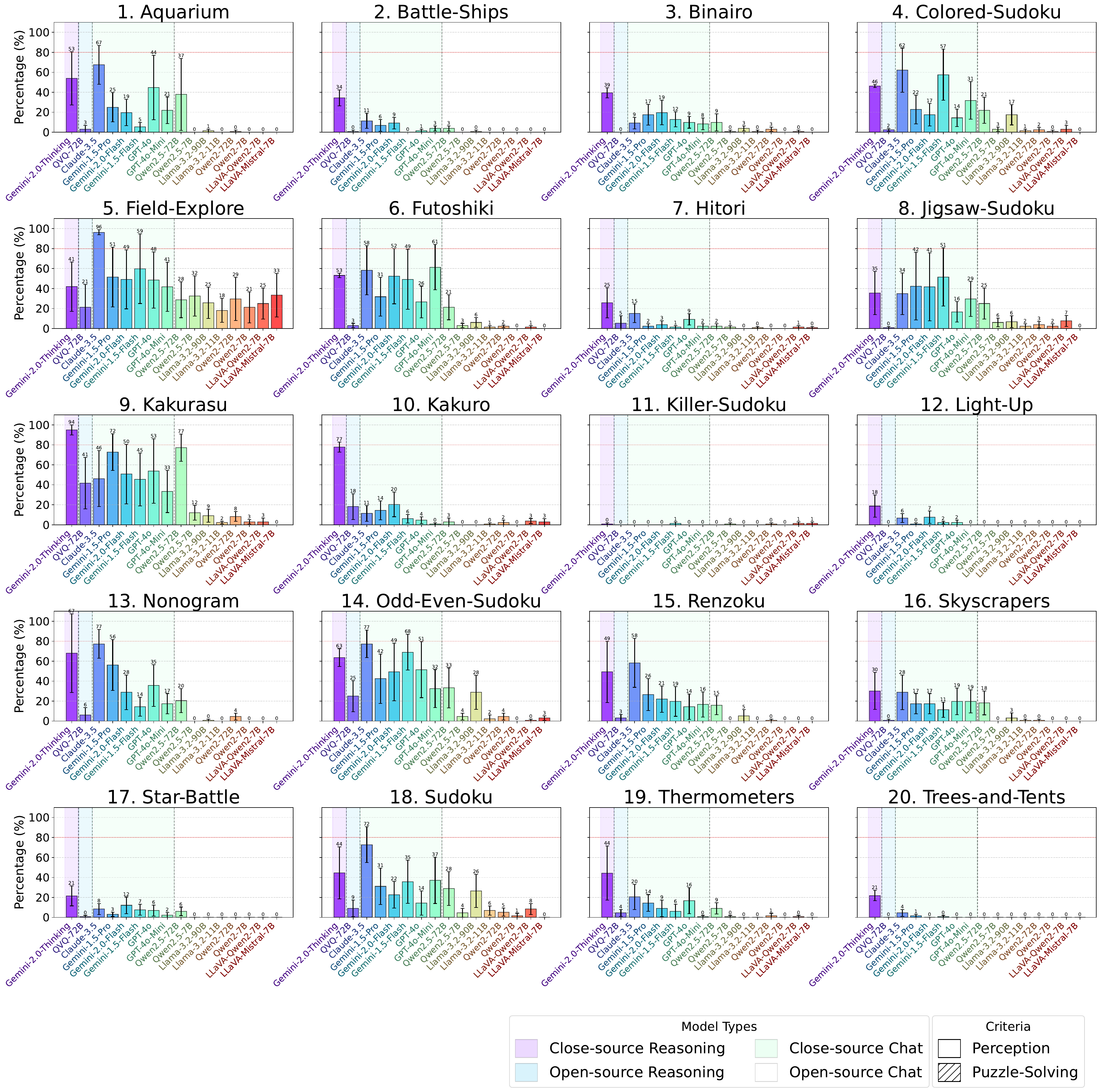} \caption{\textbf{\Easy~level Text Input w/ CoT.} We notice that Gemini-2.0-Thinking is performing best for a few puzzles, such as Kakurasu, Kakuro, while Claude-3.5 performs best in most other puzzles. (Best viewed on a screen when zoomed in)} \label{fig:benchmark-easy-text-causality} \end{figure*}

\begin{figure*}[t] \centering 

\includegraphics[width=1.0\textwidth, page=1]{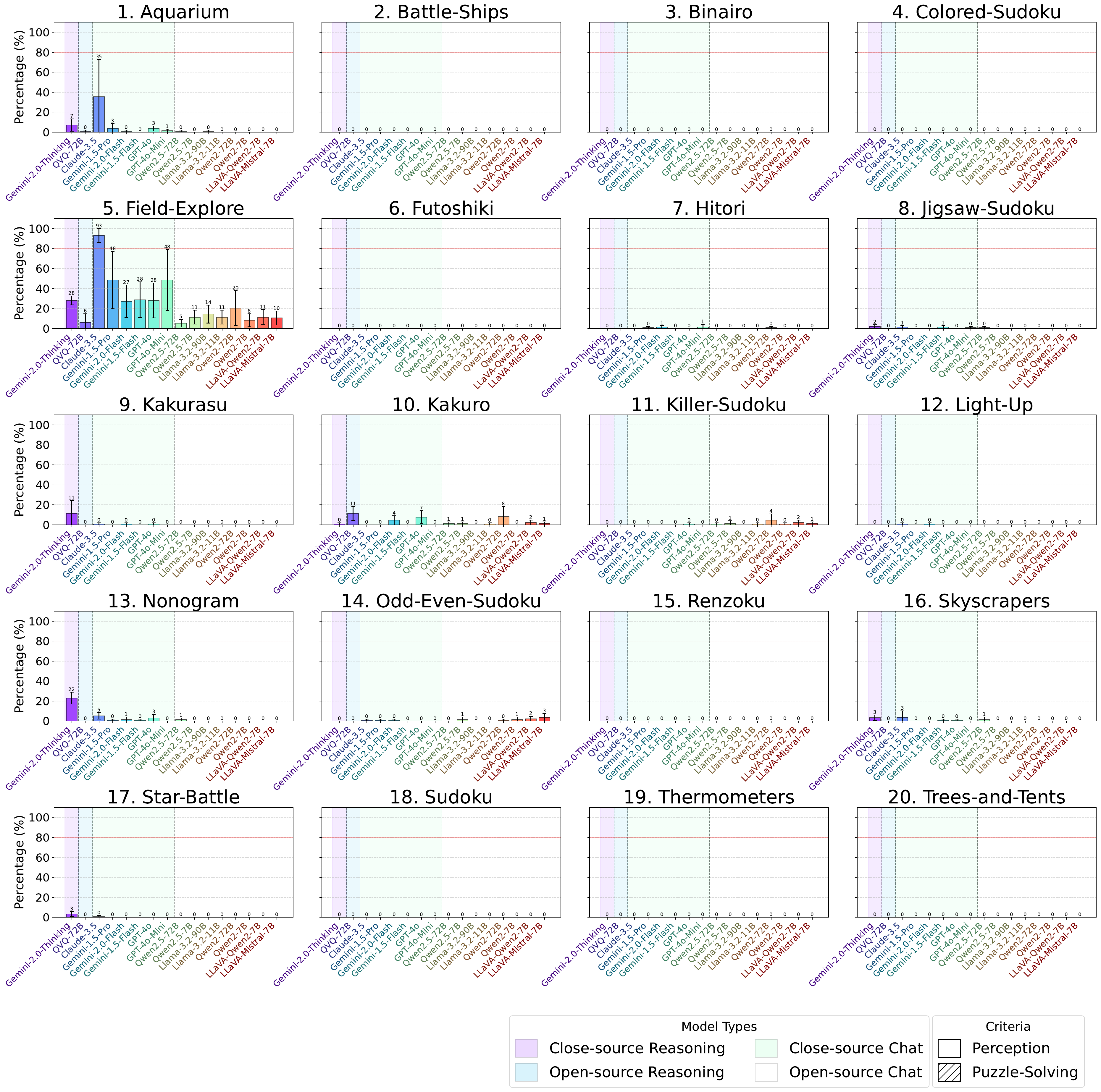} \caption{\textbf{\Medium~level Text Input w/ CoT.} We omit the figure for \hard~level text input which suppose to appear in the next page as all the accuracies are zero. On the medium level, almost all LVLMs achieve almost zero performance in most puzzles. One exception is Field-Explore, where Claude-3.5 achieves as as high as 93\% solving rate. We interpret this results as the rules in Field-Explore are local only, making it easier to solve compared with other puzzles in the medium level.  (Best viewed on a screen when zoomed in)} \label{fig:benchmark-medium-text-causality} \end{figure*}

\begin{figure*}[t] \centering 
\refstepcounter{section} 
\phantomsection         
\addcontentsline{toc}{subsection}{Easy, Medium, and Hard Level Overall Evaluation of SFT Models}
\includegraphics[width=1.0\textwidth, page=1]{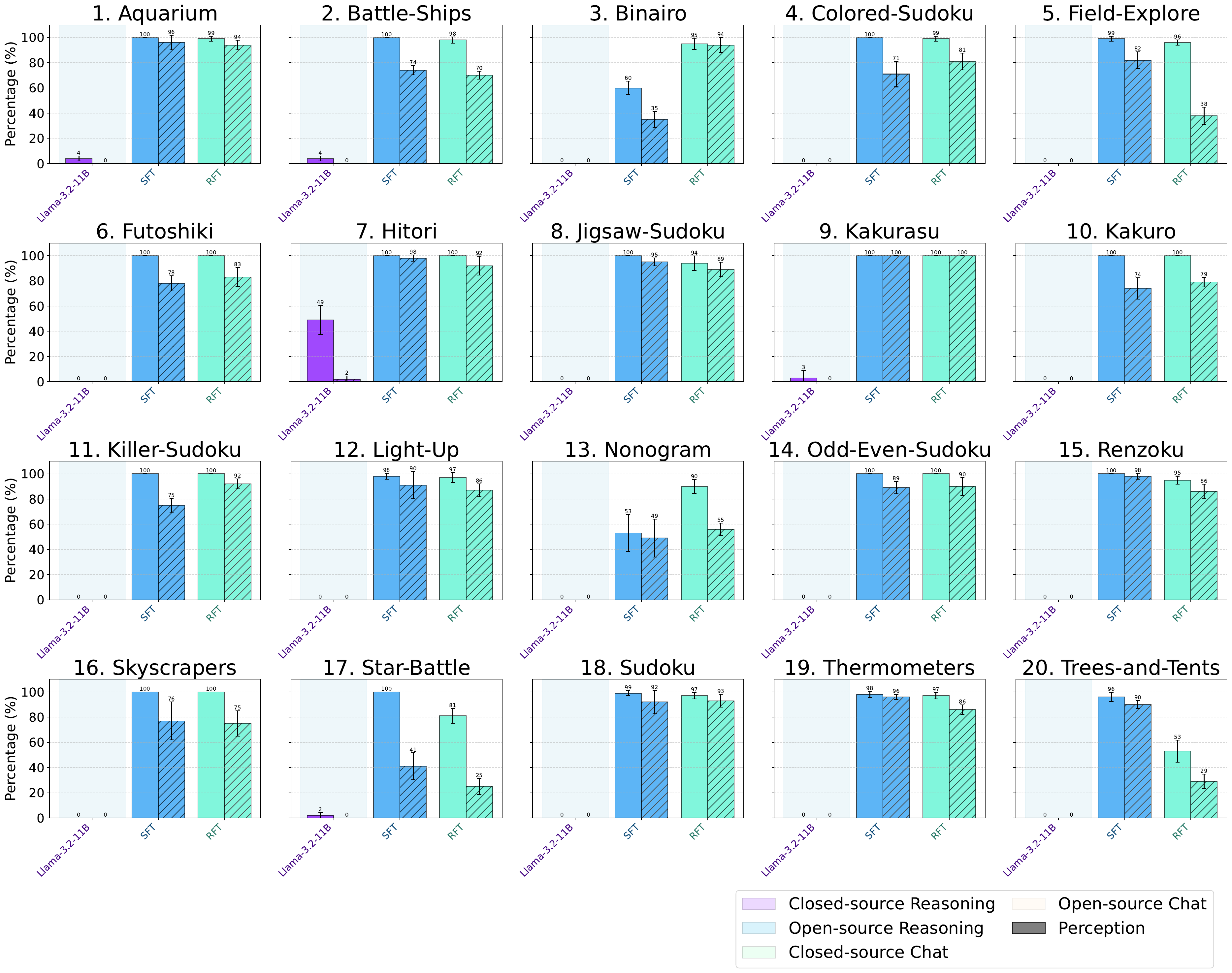} \caption{\textbf{Evaluation on SFT with Level \Easy.} We notice both S-SFT and R-SFT improve the performance of perception and puzzle-solving. Comparing R-SFT and S-SFT, we noticed they perform comparably, where S-SFT performs better in some puzzles such as Nonogram and Trees-and-Tents, while R-SFT performs better in Binairo and Kakuro. (Best viewed on a screen when zoomed in)} \label{fig:benchmark-easy-sft} \end{figure*}

\begin{figure*}[t] \centering \includegraphics[width=1.0\textwidth, page=1]{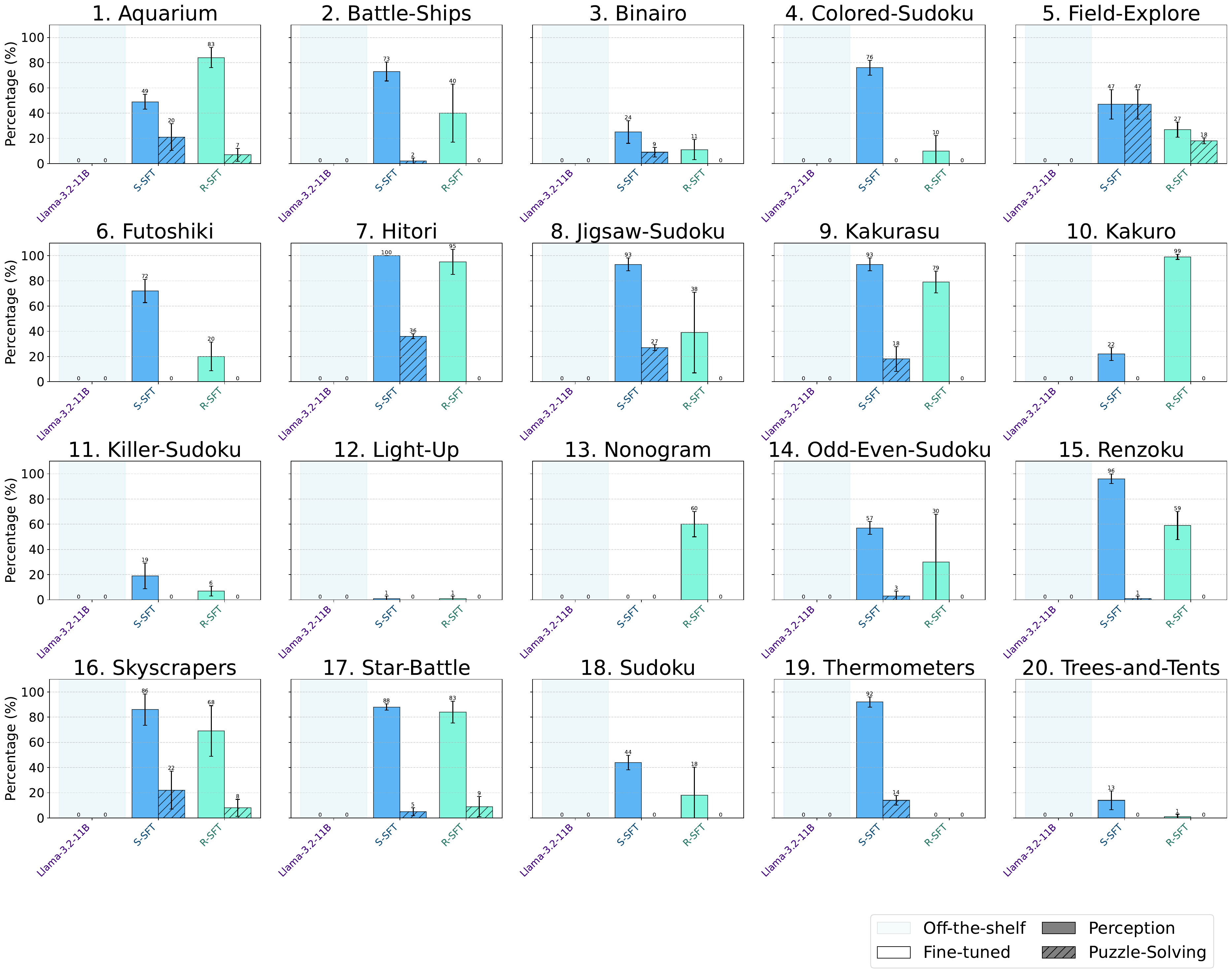} \caption{\textbf{Evaluation on SFT with Level \Medium.} Both S-SFT and R-SFT have lower performance than Level \Easy. (Best viewed on a screen when zoomed in)} \label{fig:benchmark-medium-sft} \end{figure*}

\begin{figure*}[t] \centering \includegraphics[width=1.0\textwidth, page=1]{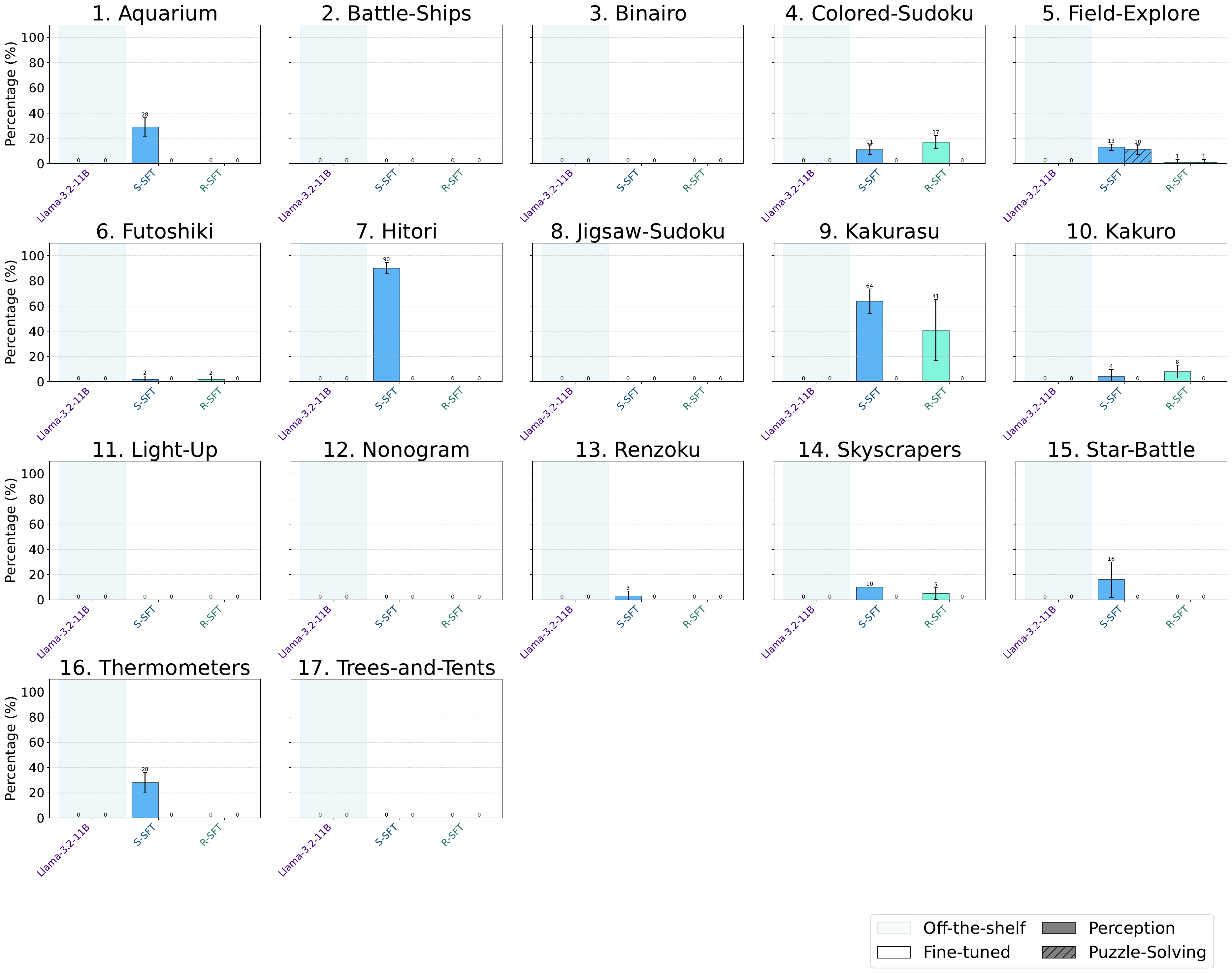} \caption{\textbf{Evaluation on SFT with Level \Hard} (Best viewed on a screen when zoomed in)} \label{fig:benchmark-hard-sft} \end{figure*}

\onecolumn
\renewcommand{\arraystretch}{1.2} %

\centering

{\ttfamily %
\begin{longtable}{|p{0.95\textwidth}|}
\hline
\textbf{Qualitative Study: Sudoku - Easy - W/ CoT - Vision Input} \\ \hline
\endhead

\hline \multicolumn{1}{|r|}{{Continued on next page}} \\ \hline
\endfoot

\hline
\endlastfoot

\label{supp-tab-sudoku-vision-easy}
\refstepcounter{section} 
\phantomsection         
\addcontentsline{toc}{section}{Qualitative Studies}
\addcontentsline{toc}{subsection}{Sudoku - Easy - W/ CoT - Vision Input}

\includegraphics[width=0.24\columnwidth]{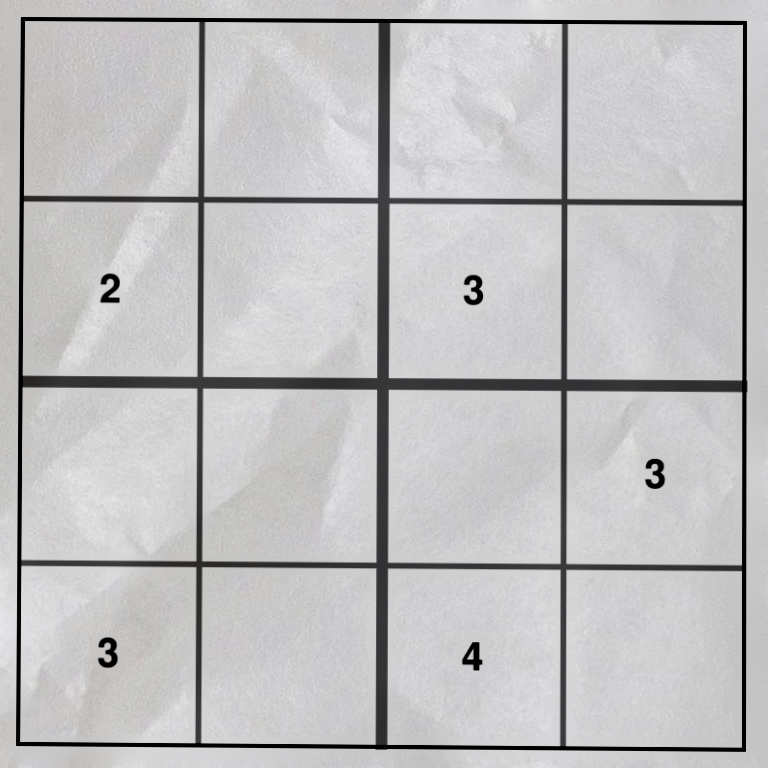} \\
\hline

\hlmodel{\textbf{Gemini-2.0-Thinking}} \\ 

\hlkey{\textbf{Model Output:}}\newline The user wants me to solve a 4x4 Sudoku puzzle. [NEWLINE] I need to identify the empty cells and fill them with numbers from 1 to 4, ensuring that each row, column, and 2x2 block contains unique numbers. [NEWLINE]  [NEWLINE] Let's analyze the given grid: [NEWLINE] - Row 0: has no filled cells. [NEWLINE] - Row 1: has '2' at index 0 and '3' at index 2. [NEWLINE] - Row 2: has '3' at index 3. [NEWLINE] - Row 3: has '3' at index 0 and '4' at index 2. [NEWLINE]  [NEWLINE] - Column 0: has '2' at index 1 and '3' at index 3. [NEWLINE] - Column 1: is empty. [NEWLINE] - Column 2: has '3' at index 1 and '4' at index 3. [NEWLINE] - Column 3: has '3' at index 2. [NEWLINE]  [NEWLINE] - Block 0 (0,0 to 1,1): has '2'. [NEWLINE] - Block 1 (0,2 to 1,3): has '3'. [NEWLINE] - Block 2 (2,0 to 3,1): has '3'. [NEWLINE] - Block 3 (2,2 to 3,3): has '3', '4'. [NEWLINE]  [NEWLINE] Let's start filling in the empty cells. [NEWLINE]  [NEWLINE] - **Row 0:** [NEWLINE]   - Cannot be '3' because of column 2 and block 1. [NEWLINE] - **Row 1:** [NEWLINE]   - Missing numbers are 1 and 4. [NEWLINE]   - Cell (1,1) cannot be 4 because of column 2. So, cell (1,1) is 1 and cell (1,3) is 4. [NEWLINE] - **Row 2:** [NEWLINE]   - Missing numbers are 1, 2, 4. [NEWLINE]   - Cell (2,0) cannot be 3 because of column 0 and block 2. [NEWLINE]   - Cell (2,1) cannot be 3 because of block 2. [NEWLINE]   - Cell (2,2) cannot be 3 because of block 3. [NEWLINE] - **Row 3:** [NEWLINE]   - Missing numbers are 1 and 2. [NEWLINE]   - Cell (3,1) cannot be 3 because of block 2. [NEWLINE]  [NEWLINE] Let's restart and try a different approach. [NEWLINE]  [NEWLINE] - **Column 1:** [NEWLINE]   - Cannot be 1 because of row 1. [NEWLINE] - **C

[...TRUNCATED... (2755 characters omitted)...]

2 3 [NEWLINE] 3 2 4 1 [NEWLINE] ``` [NEWLINE] ```json [NEWLINE] "perception": [[ "*", "*", "*", "*" ], [ "2", "*", "3", "*" ], [ "*", "*", "*", "3" ], [ "3", "*", "4", "*" ]], "think": "The Sudoku grid is 4x4, so the numbers to use are 1, 2, 3, and 4. I will go through each row, column, and 2x2 block to identify the missing numbers and fill them in while ensuring uniqueness.", "answer": [[ "4", "3", \hlred{"2"}, "1" ], [ "2", "1", "3", "4" ], [ "1", "4", \hlred{"2"}, "3" ], [ "3", "2", "4", "1" ]] [NEWLINE] ```\newline 
\hlkey{\textbf{Filtered Output:}}\newline Perception [NEWLINE] [[0,0,0,0],[2,0,3,0],[0,0,0,3],[3,0,4,0]] [NEWLINE] Answer [NEWLINE] [[4,3,2,1],[2,1,3,4],[1,4,2,3],[3,2,4,1]]\newline
\hlkey{\textbf{Perception Correct:}} True\newline
\hlkey{\textbf{Answer Correct:}} False \\ \hline
\hlmodel{\textbf{QVQ-72B}} \\ 
\hlkey{\textbf{Model Output:}}\newline So I'm looking at this Sudoku puzzle, and it seems a bit different from the usual ones I've seen. It's a 4x4 grid, which is smaller than the standard 9x9, and it's divided into 2x2 blocks. There are some numbers already filled in: a 2 in the top-left cell, a 3 in the top-right cell, another 3 in the bottom-left cell, and a 4 in the bottom-right cell. The rest of the cells are empty. [NEWLINE]  [NEWLINE] Wait, actually, looking closer, it seems like there might be a mistake because in the top-middle cell, there's a 3, and in the bottom-right, there's a 4. So, the grid looks like this: [NEWLINE]  [NEWLINE] ``` [NEWLINE]  [NEWLINE] 2 * | * 3 [NEWLINE]  [NEWLINE] -----+---- [NEWLINE]  [NEWLINE] * 3 | * * [NEWLINE]  [NEWLINE] * * | * 4 [NEWLINE]  [NEWLINE] ``` [NEWLINE]  [NEWLINE] Hmm, but it's a 4x4 grid, so it should be two blocks of 2x2 cells each. [NEWLINE]  [NEWLINE] But actually, looking back at the problem, it says the grid is sqrt(N)xsqrt(N) blocks, and N is 4, so 2x2 blocks. So, the grid is divided into four 2x2 blocks. [NEWLINE]  [NEWLINE] Wait, no, N is 4, and sqrt(4) is 2, so it's 2x2 blocks. So, the grid should be divided into 4 blocks, each 2x2. [NEWLINE]  [NEWLINE] But in the image, it's a 3x3 grid with some numbers. Wait, perhaps there's confusion here. The problem mentions a NxN grid made up of sqrt(N)xsqrt(N) blocks, with N starting from 1 to N, and indexing starts at 0. [NEWLINE]  [NEWLINE] But in the image, it's a 3x3 grid with some numbers filled in. There seems to be inconsistency between the problem description and the image. [NEWLINE]  [NEWLINE] In the image, it's a 3x3 grid with: [NEWLINE]  [NEWLINE] - Top-left cell: 2 [NEWLINE]  [NEWLINE] - Middle cel

[...TRUNCATED... (12517 characters omitted)...]

ave 2, 3, and 4 (since R2C2: 4). [NEWLINE]  [NEWLINE] So, for R0C2, I can choose any number except 2, 3, and 4. [NEWLINE]  [NEWLINE] Similarly, R1C0: ? In column 0, I have 2 and 3. [NEWLINE]  [NEWLINE] In row 1, no numbers given. [NEWLINE]  [NEWLINE] In the middle block, no numbers yet. [NEWLINE]  [NEWLINE] So, R1C0 can be any number except 2 and 3. [NEWLINE]  [NEWLINE] Wait, but I need to consider the block as well. [NEWLINE]  [NEWLINE] In the top-middle block, which is the first block in row 1\newline 
\hlkey{\textbf{Filtered Output:}}\newline Perception [NEWLINE] Wrong [NEWLINE] Answer [NEWLINE] Wrong\newline
\hlkey{\textbf{Perception Correct:}} False\newline
\hlkey{\textbf{Answer Correct:}} False \\ 
\hlgray{\textbf{Remarks:}} repeatitive reasoning and go beyong max length. \\
\hline
\hlmodel{\textbf{Claude-3.5}} \\ 
\hlkey{\textbf{Model Output:}}\newline ```json [NEWLINE]  [NEWLINE] "perception": [ [NEWLINE]     ["*", "*", "*", "*"], [NEWLINE]     ["2", "*", "3", "*"], [NEWLINE]     ["*", "*", "*", "3"], [NEWLINE]     ["3", "*", "4", "*"] [NEWLINE] ], [NEWLINE] "think": "This is a 4x4 Sudoku puzzle with 2x2 blocks. Let me solve it step by step: [NEWLINE] 1. Looking at row 4, we have 3 and 4, so 1 and 2 are missing [NEWLINE] 2. Column 3 has 3 twice, which means the empty cells must be 1,2,4 [NEWLINE] 3. Row 2 has 2 and 3, so needs 1 and 4 [NEWLINE] 4. The first row must contain all missing numbers [NEWLINE] 5. Using block constraints and existing numbers, I can determine the full solution", [NEWLINE]  [NEWLINE] "answer": [ [NEWLINE]     ["4", "3", "1", "2"], [NEWLINE]     ["2", "1", "3", "4"], [NEWLINE]     ["1", "4", "2", "3"], [NEWLINE]     ["3", "2", "4", "1"] [NEWLINE] ] [NEWLINE]  [NEWLINE] ```\newline 
\hlkey{\textbf{Filtered Output:}}\newline Perception [NEWLINE] [[0,0,0,0],[2,0,3,0],[0,0,0,3],[3,0,4,0]] [NEWLINE] Answer [NEWLINE] [[4,3,1,2],[2,1,3,4],[1,4,2,3],[3,2,4,1]]\newline
\hlkey{\textbf{Perception Correct:}} True\newline
\hlkey{\textbf{Answer Correct:}} True \\ \hline
\hlmodel{\textbf{Gemini-1.5-Pro}} \\ 
\hlkey{\textbf{Model Output:}}\newline ```json [NEWLINE]  [NEWLINE] "perception": [["*", "*", "*", "*",], [NEWLINE]               ["2", "*", "3", "*"], [NEWLINE]               ["*", "*", "*", "3"], [NEWLINE]               ["3", "*", "4", "*"]],  [NEWLINE] "think": "This is a 4x4 Sudoku puzzle. The possible values for each cell are 1, 2, 3, and 4. [NEWLINE] Row 0: The missing values are 1, 2, 3, 4.  [NEWLINE] Row 1: The missing values are 1, 4. [NEWLINE] Row 2: The missing values are 1, 2, 4. [NEWLINE] Row 3: The missing values are 1, 2. [NEWLINE]  [NEWLINE] Col 0: The missing values are 1, 2, 4. [NEWLINE] Col 1: The missing values are 1, 2, * [NEWLINE] Col 2: The missing values are 1, 2, * [NEWLINE] Col 3: The missing values are 1, 2, 4. [NEWLINE]  [NEWLINE] Block 0 (top-left): The missing values are 1, 4. [NEWLINE] Block 1 (top-right): The missing values are 1, 2, 4. [NEWLINE] Block 2 (bottom-left): The missing values are 1, 2. [NEWLINE] Block 3 (bottom-right): The missing values are 1, 2. [NEWLINE]  [NEWLINE] Cell (0,0): Possible values are 1, 4. Since 1 is already present in the last column, this cell must be 4. [NEWLINE] Cell (0,1): Possible values are 1, 2. Since 2 is already present in the first column, this cell must be 1. [NEWLINE] Cell (0,3): Possible values are 1, 2. Since 1 is already present in this row, this cell must be 2. [NEWLINE] Cell (1,1): Possible values are 1, 4. Since 1 is already present in this row, this cell must be 4. [NEWLINE] Cell (1,3): Possible values are 1. This cell must be 1. [NEWLINE] Cell (2,0): Possible values are 1, 2, 4. Since 4 is already present in the last column, and 2 is already present in the first row, this cell must be 1. [NEWLINE] Cell (2,1): Possible values are 2

[...TRUNCATED... (530 characters omitted)...]

t in the second column, this cell must be 2. [NEWLINE] Cell (2,2): Possible values are 4. This cell must be 4. [NEWLINE] Cell (3,1): Possible values are 1, 2. Since 1 is already present in the last column, this cell must be 2. [NEWLINE] Cell (3,3): Possible values are 1. This cell must be 1.", [NEWLINE] "answer": [["4", "1", "*", "2"], [NEWLINE]               ["2", "4", "3", \hlred{"1"}], [NEWLINE]               ["1", "2", "4", "3"], [NEWLINE]               ["3", "2", "4", \hlred{"1"}]] [NEWLINE]  [NEWLINE] ```\newline 
\hlkey{\textbf{Filtered Output:}}\newline Perception [NEWLINE] [[0,0,0,0],[2,0,3,0],[0,0,0,3],[3,0,4,0]] [NEWLINE] Answer [NEWLINE] [[4,1,0,2],[2,4,3,1],[1,2,4,3],[3,2,4,1]]\newline
\hlkey{\textbf{Perception Correct:}} True\newline
\hlkey{\textbf{Answer Correct:}} False \\ \hline
\hlmodel{\textbf{Gemini-2.0-Flash}} \\ 
\hlkey{\textbf{Model Output:}}\newline ```json [NEWLINE]  [NEWLINE] "perception": "[['*', '*', '*'], ['2', '*', '3'], ['3', '*', '4']]", [NEWLINE] "think": "I see a 3x3 Sudoku grid.  [NEWLINE] Row 0 is missing 1, 2, and 3.  [NEWLINE] Row 1 is missing 1 and 4.  [NEWLINE] Row 2 is missing 1 and 2.  [NEWLINE] Column 0 is missing 1 and 4.  [NEWLINE] Column 1 is missing 1, 2, 3, and 4.  [NEWLINE] Column 2 is missing 1 and 2.  [NEWLINE] Block 0 is missing 1 and 4.  [NEWLINE] Block 1 is missing 1 and 2.  [NEWLINE] Block 2 is missing 1 and 2. [NEWLINE]  [NEWLINE] Looking at row 0, I can deduce that the last cell must be 4 because column 2 already has 3 and 4 in row 2.  [NEWLINE] Looking at row 0, I can deduce that the first cell must be 1 because column 0 already has 2 and 3 in row 1 and 2.  [NEWLINE] Looking at row 0, I can deduce that the second cell must be 2 because column 1 already has 2, 3, and 4 in row 1 and 2.  [NEWLINE] Looking at row 1, I can deduce that the second cell must be 4 because column 1 already has 2, 3, and 4 in row 0 and 2.  [NEWLINE] Looking at row 1, I can deduce that the first cell must be 1 because column 0 already has 1, 2, and 3 in row 0 and 2.  [NEWLINE] Looking at row 2, I can deduce that the second cell must be 1 because column 1 already has 1, 2, 3, and 4 in row 0 and 1.  [NEWLINE] Looking at row 2, I can deduce that the first cell must be 2 because column 0 already has 1, 2, and 3 in row 0 and 1. ", [NEWLINE] "answer": "[['1', '4', '2'], ['2', '1', '3'], ['3', '2', '4']]" [NEWLINE]  [NEWLINE] ```\newline 
\hlkey{\textbf{Filtered Output:}}\newline Perception [NEWLINE] Wrong [NEWLINE] Answer [NEWLINE] Wrong\newline
\hlkey{\textbf{Perception Correct:}} False\newline
\hlkey{\textbf{Answer Correct:}} False \\ 
\hlgray{Remarks:} wrong grid size.  \\
\hlmodel{\textbf{Gemini-1.5-Flash}} \\  \hline
\hlkey{\textbf{Model Output:}}\newline ```json [NEWLINE]  [NEWLINE]   "perception": [ [NEWLINE]     ["*", "*", "*", "*", "*", "*", "*", "*", "*"], [NEWLINE]     ["*", "2", "*", "*", "*", "3", "*", "*", "*"], [NEWLINE]     ["*", "*", "*", "*", "*", "*", "*", "*", "*"], [NEWLINE]     ["*", "*", "*", "*", "*", "*", "*", "*", "*"], [NEWLINE]     ["*", "*", "*", "*", "*", "*", "*", "*", "*"], [NEWLINE]     ["*", "*", "*", "*", "*", "*", "*", "*", "*"], [NEWLINE]     ["*", "*", "*", "*", "*", "*", "*", "*", "3"], [NEWLINE]     ["*", "3", "*", "*", "*", "4", "*", "*", "*"], [NEWLINE]     ["*", "*", "*", "*", "*", "*", "*", "*", "*"] [NEWLINE]   ], [NEWLINE]   "think": "This is a 9x9 Sudoku puzzle.  I'll start by looking for cells where only one number is possible given the existing constraints.  I'll then use logic to deduce other numbers based on row, column, and 3x3 block constraints.  I'll iterate this process until the puzzle is solved.", [NEWLINE]   "answer": [ [NEWLINE]     [1, 6, 7, 9, 5, 2, 3, 8, 4], [NEWLINE]     [8, 2, 9, 1, 4, 3, 7, 5, 6], [NEWLINE]     [3, 5, 4, 7, 6, 8, 1, 9, 2], [NEWLINE]     [4, 9, 1, 3, 7, 5, 8, 6, 2], [NEWLINE]     [7, 8, 5, 6, 2, 9, 4, 1, 3], [NEWLINE]     [6, 3, 2, 4, 8, 1, 9, 7, 5], [NEWLINE]     [2, 1, 6, 8, 3, 7, 5, 4, 9], [NEWLINE]     [5, 3, 8, 2, 9, 4, 6, 1, 7], [NEWLINE]     [9, 7, 3, 5, 1, 6, 2, 4, 8] [NEWLINE]   ] [NEWLINE]  [NEWLINE] ```\newline 
\hlkey{\textbf{Filtered Output:}}\newline Perception [NEWLINE] [[0,0,0,0,0,0,0,0,0],[0,2,0,0,0,3,0,0,0], [0,0,0,0,0,0,0,0,0], [0,0,0,0,0,0,0,0,0], [0,0,0,0,0,0,0,0,0], [0,0,0,0,0,0,0,0,0], [0,0,0,0,0,0,0,0,3], [0,3,0,0,0,4,0,0,0],[0,0,0,0,0,0,0,0,0]] [NEWLINE] Answer [NEWLINE] [[1,6,7,9,5,2,3,8,4], [8,2,9,1,4,3,7,5,6], [3,5,4,7,6,8,1,9,2], [4,9,1,3,7,5,8,6,2], [7,8,5,6,2,9,4,1,3], [6,3,2,4,8,1,9,7,5],[2,1,6,8,3,7,5,4,9], [5,3,8,2,9,4,6,1,7],[9,7,3,5,1,6,2,4,8]]\newline
\hlkey{\textbf{Perception Correct:}} False\newline
\hlkey{\textbf{Answer Correct:}} False \\ 
\hlgray{Remarks:} wrong grid size. \\ \hline

\hlmodel{\textbf{GPT-4o}} \\ 
\hlkey{\textbf{Model Output:}}\newline ```json [NEWLINE]  [NEWLINE]   "perception": [ [NEWLINE]     ["*", "*", \hlred{"2"}, "*"], [NEWLINE]     ["*", "3", "*", "*"], [NEWLINE]     ["3", "*", "*", "3"], [NEWLINE]     ["*", "*", "4", "*"] [NEWLINE]   ], [NEWLINE]   "think": "This is a 4x4 Sudoku grid. The challenge is to fill empty cells ('*') with numbers from 1 to 4 ensuring that each number appears exactly once in every row, column, and 2x2 subgrid. Starting with available numbers in the provided setup, I will fill cells logically row by row, column by column, and by observing constraints in blocks.", [NEWLINE]   "answer": [ [NEWLINE]     ["4", "1", "2", "3"], [NEWLINE]     ["2", "3", "1", "4"], [NEWLINE]     ["3", "4", "1", "2"], [NEWLINE]     ["1", "2", "4", "3"] [NEWLINE]   ] [NEWLINE]  [NEWLINE] ```\newline 
\hlkey{\textbf{Filtered Output:}}\newline Perception [NEWLINE] [[0,0,2,0],[0,3,0,0],[3,0,0,3],[0,0,4,0]] [NEWLINE] Answer [NEWLINE] [[4,1,2,3],[2,3,1,4],[3,4,1,2],[1,2,4,3]]\newline
\hlkey{\textbf{Perception Correct:}} False\newline
\hlkey{\textbf{Answer Correct:}} False \\ \hline
\hlmodel{\textbf{GPT-4o-Mini}} \\ 
\hlkey{\textbf{Model Output:}}\newline ```json [NEWLINE]  [NEWLINE]   "perception": [[2, "*", 3, "*"], [3, "*", "*", 4], ["*", "*", 2, "*"], ["*", 3, "*", "*"]], [NEWLINE]   "think": "I need to fill in the empty cells ('*') while ensuring that each number from 1 to 4 appears exactly once in each row, column, and 2x2 block. The given numbers already provide some constraints. For example, Row 0 has 2 and 3, and Block 0 has 2 and 3 as well. The first step will be to identify the missing numbers in each row and block and then find valid placements for them while adhering to Sudoku rules. After evaluating the options, I will fill in the grid systematically until completed.", [NEWLINE]   "answer": [[2, 4, 3, 1], [3, 1, 2, 4], [1, 2, 4, 3], [4, 3, 1, 2]] [NEWLINE]  [NEWLINE] ```\newline 
\hlkey{\textbf{Filtered Output:}}\newline Perception [NEWLINE] [[\hlred{2},0,\hlred{3},0],[3,0,0,4],[0,0,2,0],[0,3,0,0]] [NEWLINE] Answer [NEWLINE] [[2,4,3,1],[3,1,2,4],[1,2,4,3],[4,3,1,2]]\newline
\hlkey{\textbf{Perception Correct:}} False\newline
\hlkey{\textbf{Answer Correct:}} False \\ \hline

\hlmodel{\textbf{Qwen2.5-72B}} \\ 
\hlkey{\textbf{Model Output:}}\newline ```json [NEWLINE]  [NEWLINE] "perception": [ [NEWLINE]     ["*", "*", "*", "*"], [NEWLINE]     ["2", "*", "3", "*"], [NEWLINE]     ["*", "*", "*", "3"], [NEWLINE]     ["3", "*", "4", "*"] [NEWLINE] ], [NEWLINE] "think": "To solve the Sudoku for a 4x4 grid, each square (including rows, columns, and 2x2 blocks) must contain the numbers from 1 to 4 without repetition. Considering the current grid, here is how we solve it step by step: [NEWLINE] 1. For the top left 2x2 block, since the bottom left corner is '3', top right corner cannot be '3', hence it can be either '1' or '2', but '2' is already in the second row, top left could be '1'. [NEWLINE] 2. Follow similar logic for the rest of the cells. [NEWLINE] 3. After filling out all guesses based on the exclusions of rows, columns, and block conditions, validate the grid for completion. [NEWLINE] ", [NEWLINE] "answer": [ [NEWLINE]     ["1", "4", "2", "3"], [NEWLINE]     ["2", "1", "3", "4"], [NEWLINE]     ["4", "3", "1", "2"], [NEWLINE]     ["3", "2", "4", "1"] [NEWLINE] ] [NEWLINE]  [NEWLINE] ```\newline 
\hlkey{\textbf{Filtered Output:}}\newline Perception [NEWLINE] [[0,0,0,0],[2,0,3,0],[0,0,0,3],[3,0,4,0]] [NEWLINE] Answer [NEWLINE] [[\hlred{1},4,2,3],[2,\hlred{1},3,4],[4,3,1,2],[3,2,4,1]]\newline
\hlkey{\textbf{Perception Correct:}} True\newline
\hlkey{\textbf{Answer Correct:}} False \\ \hline
\hlmodel{\textbf{Qwen2.5-7B}} \\ 
\hlkey{\textbf{Model Output:}}\newline ```json [NEWLINE]  [NEWLINE] "perception": [[0, 0, 0, 0], [NEWLINE]               [2, 0, 3, 0], [NEWLINE]               [0, 0, 0, 3], [NEWLINE]               [3, 0, 4, 0]], [NEWLINE] "think": "The grid is a 2x2 block Sudoku. I'll solve it row by row starting with the top left cell. The top left cell must contain the digit '1' since it's the only missing number in its row, column, and block.", [NEWLINE] "answer": [[1, 0, 0, 0], [NEWLINE]           [2, 0, 3, 0], [NEWLINE]           [0, 0, 0, 3], [NEWLINE]           [3, 0, 4, 0]] [NEWLINE]  [NEWLINE] ```json [NEWLINE]  [NEWLINE] "perception": [[1, 0, 0, 0], [NEWLINE]               [2, 0, 3, 0], [NEWLINE]               [0, 0, 0, 3], [NEWLINE]               [3, 0, 4, 0]], [NEWLINE] "think": "The second row must include the digit '1' in its second position, as it's the only missing number in its row, column, and block.", [NEWLINE] "answer": [[1, 1, 0, 0], [NEWLINE]           [2, 1, 3, 0], [NEWLINE]           [0, 0, 0, 3], [NEWLINE]           [3, 0, 4, 0]] [NEWLINE]  [NEWLINE] ```json [NEWLINE]  [NEWLINE] "perception": [[1, 1, 0, 0], [NEWLINE]               [2, 1, 3, 0], [NEWLINE]               [0, 0, 0, 3], [NEWLINE]               [3, 0, 4, 0]], [NEWLINE] "think": "The third row must include the digit '2' in its first position, as it's the only missing number in its row, column, and block.", [NEWLINE] "answer": [[1, 1, 0, 0], [NEWLINE]           [2, 1, 3, 0], [NEWLINE]           [2, 0, 0, 3], [NEWLINE]           [3, 0, 4, 0]] [NEWLINE]  [NEWLINE] ```json [NEWLINE]  [NEWLINE] "perception": [[1, 1, 0, 0], [NEWLINE]               [2, 1, 3, 0], [NEWLINE]               [2, 0, 0, 3], [NEWLINE]               [3, 0, 4, 0]], [NEWLINE] "

[...TRUNCATED... (2247 characters omitted)...]

json [NEWLINE]  [NEWLINE] "perception": [[1, 1, 0, 0], [NEWLINE]               [1, 1, 3, 0], [NEWLINE]               [2, 0, 0, 3], [NEWLINE]               [3, 2, 4, 3]], [NEWLINE] "think": "The third column must include the digit '1' in its last position, as it's the only missing number in its row, column, and block.", [NEWLINE] "answer": [[1, 1, 1, 0], [NEWLINE]           [1, 1, 3, 0], [NEWLINE]           [2, 0, 0, 3], [NEWLINE]           [3, 2, 4, 3]] [NEWLINE]  [NEWLINE] ```json [NEWLINE] ```\newline 
\hlkey{\textbf{Filtered Output:}}\newline Perception [NEWLINE] [[\hlred{1},\hlred{1},0,0],[2,0,0,0],[3,0,4,0],[0,0,0,0]] [NEWLINE] Answer [NEWLINE] [[1,1,0,0],[2,1,3,0],[2,0,0,3],[3,2,4,0]] \newline
\hlkey{\textbf{Perception Correct:}} False\newline
\hlkey{\textbf{Answer Correct:}} False \\ \hline

\hlmodel{\textbf{Llama-3.2-90B}} \\ 
\hlkey{\textbf{Model Output:}}\newline **Step 1: Analyze the Sudoku Grid** [NEWLINE]  [NEWLINE] The given Sudoku grid is a 4x4 grid, which means it consists of 16 cells. The goal is to fill in the missing numbers from 1 to 4 such that each row, column, and 2x2 sub-grid contains each number exactly once. [NEWLINE]  [NEWLINE] **Step 2: Identify the Missing Numbers** [NEWLINE]  [NEWLINE] From the given grid, we can see that the missing numbers are represented by asterisks (*). We need to find the correct values for these missing numbers. [NEWLINE]  [NEWLINE] **Step 3: Apply Sudoku Rules and Logic** [NEWLINE]  [NEWLINE] To solve the Sudoku puzzle, we will apply the standard rules and logic: [NEWLINE]  [NEWLINE] *   Each row must contain the numbers 1 to 4 without repetition. [NEWLINE] *   Each column must contain the numbers 1 to 4 without repetition. [NEWLINE] *   Each 2x2 sub-grid must contain the numbers 1 to 4 without repetition. [NEWLINE]  [NEWLINE] By analyzing the given numbers and applying these rules, we can deduce the correct values for the missing numbers. [NEWLINE]  [NEWLINE] **Step 4: Fill in the Missing Numbers** [NEWLINE]  [NEWLINE] After applying the Sudoku rules and logic, we can fill in the missing numbers as follows: [NEWLINE]  [NEWLINE] |  |  |  |  | [NEWLINE] | --- | --- | --- | --- | [NEWLINE] | 2 | 3 | * | * | [NEWLINE] | 3 | 4 | * | * | [NEWLINE] | * | * | 3 | * | [NEWLINE] | * | * | * | 2 | [NEWLINE]  [NEWLINE] **Step 5: Finalize the Solution** [NEWLINE]  [NEWLINE] Based on the filled-in grid, we can see that the complete solution is: [NEWLINE]  [NEWLINE] |  |  |  |  | [NEWLINE] | --- | --- | --- | --- | [NEWLINE] | 2 | 3 | 1 | 4 | [NEWLINE] | 3 | 4 | 2 | 1 | [NEWLINE] | 1 | 2 | 3 | 4 | [N

[...TRUNCATED... (618 characters omitted)...]

WLINE] ```json [NEWLINE]  [NEWLINE]     "perception": [ [NEWLINE]         [2, 3, "*", "*"], [NEWLINE]         [3, 4, "*", "*"], [NEWLINE]         ["*", "*", 3, "*"], [NEWLINE]         ["*", "*", "*", 2] [NEWLINE]     ], [NEWLINE]     "think": "Apply Sudoku rules and logic to fill in the missing numbers.", [NEWLINE]     "answer": [ [NEWLINE]         [2, 3, 1, 4], [NEWLINE]         [3, 4, 2, 1], [NEWLINE]         [1, 2, 3, 4], [NEWLINE]         [4, 1, 2, 3] [NEWLINE]     ] [NEWLINE]  [NEWLINE] ```\newline 
\hlkey{\textbf{Filtered Output:}}\newline Perception [NEWLINE] [[\hlred{2},\hlred{3},0,0],[3,4,0,0],[0,0,3,0],[0,0,0,2]] [NEWLINE] Answer [NEWLINE] [[2,3,1,4],[3,4,2,1],[1,2,3,4],[4,1,2,3]]\newline
\hlkey{\textbf{Perception Correct:}} False\newline
\hlkey{\textbf{Answer Correct:}} False \\ \hline
\hlmodel{\textbf{Llama-3.2-11B}} \\ 
\hlkey{\textbf{Model Output:}}\newline **Current Game State** [NEWLINE] The current game state is represented as a 2D array, with the given numbers filled in and the empty cells marked with "*". The state is as follows: [NEWLINE]  [NEWLINE] ``` [NEWLINE] "perception": [["2", "3", "*", "*", "*", "*", "*", "*", "*"], [NEWLINE]                ["*", "3", "*", "*", "4", "*", "*", "*", "*"], [NEWLINE]                ["*", "*", "*", "*", "*", "*", "*", "*", "*"], [NEWLINE]                ["*", "*", "*", "*", "*", "*", "*", "*", "*"], [NEWLINE]                ["*", "*", "*", "*", "*", "*", "*", "*", "*"], [NEWLINE]                ["*", "*", "*", "*", "*", "*", "*", "*", "*"], [NEWLINE]                ["*", "*", "*", "*", "*", "*", "*", "*", "*"], [NEWLINE]                ["*", "*", "*", "*", "*", "*", "*", "*", "*"], [NEWLINE]                ["*", "*", "*", "*", "*", "*", "*", "*", "*"]] [NEWLINE] ``` [NEWLINE] **Thinking Process** [NEWLINE] To solve this Sudoku puzzle, we will use a combination of elimination and guessing. We will start by focusing on the given numbers and using them to eliminate possibilities for the empty cells. [NEWLINE]  [NEWLINE] First, we notice that the top-left 3x3 block already has a 2, 3, and 4 filled in. This means that the remaining cells in this block can only contain the numbers 1, 5, 6, 7, 8, or 9. [NEWLINE]  [NEWLINE] Next, we look at the top row and see that it already has a 2 and a 3 filled in. This means that the remaining cell in this row can only contain the numbers 1, 4, 5, 6, 7, 8, or 9. [NEWLINE]  [NEWLINE] We can repeat this process for each row, column, and block, using the given numbers to eliminate possibilities for the empty cells. [NEWLINE]  [NEWLINE] **Solution** [NEWLI

[...TRUNCATED... (732 characters omitted)...]

"4", "8"], [NEWLINE]            ["1", "9", "8", "3", "4", "2", "5", "6", "7"], [NEWLINE]            ["8", "5", "9", "7", "6", "1", "4", "3", "2"], [NEWLINE]            ["4", "2", "6", "8", "5", "3", "7", "9", "1"], [NEWLINE]            ["7", "1", "3", "9", "2", "4", "8", "5", "6"], [NEWLINE]            ["9", "6", "1", "5", "3", "7", "2", "8", "4"], [NEWLINE]            ["3", "4", "5", "2", "8", "9", "6", "7", "1"], [NEWLINE]            ["2", "8", "7", "4", "1", "6", "3", "5", "9"]] [NEWLINE] ```\newline 
\hlkey{\textbf{Filtered Output:}}\newline Perception [NEWLINE] [[2,3,0,0,0,0,0,0,0],[0,3,0,0,4,0,0,0,0], [0,0,0,0,0,0,0,0,0], [0,0,0,0,0,0,0,0,0], [0,0,0,0,0,0,0,0,0], [0,0,0,0,0,0,0,0,0], [0,0,0,0,0,0,0,0,0], [0,0,0,0,0,0,0,0,0], [0,0,0,0,0,0,0,0,0]] [NEWLINE] Answer [NEWLINE] [[5,3,4,6,7,8,9,1,2], [6,7,2,1,9,5,3,4,8], [1,9,8,3,4,2,5,6,7], [8,5,9,7,6,1,4,3,2], [4,2,6,8,5,3,7,9,1], [7,1,3,9,2,4,8,5,6], [9,6,1,5,3,7,2,8,4], [3,4,5,2,8,9,6,7,1],[2,8,7,4,1,6,3,5,9]]\newline
\hlkey{\textbf{Perception Correct:}} False\newline
\hlkey{\textbf{Answer Correct:}} False \\
\hlgray{\textbf{Remarks:}} wrong grid size. \\ \hline

\hlmodel{\textbf{Qwen2-72B}} \\ 
\hlkey{\textbf{Model Output:}}\newline ```json [NEWLINE] "perception": [[0, 0, 0, 0, 0, 0, 0, 0, 0], [NEWLINE]                 [0, 2, 0, 0, 0, 3, 0, 0, 0], [NEWLINE]                 [0, 0, 0, 0, 0, 0, 0, 0, 0], [NEWLINE]                 [0, 0, 0, 0, 0, 0, 0, 0, 0], [NEWLINE]                 [0, 0, 0, 0, 0, 0, 0, 0, 0], [NEWLINE]                 [0, 0, 0, 0, 0, 0, 0, 0, 0], [NEWLINE]                 [0, 0, 0, 0, 0, 0, 0, 0, 0], [NEWLINE]                 [0, 0, 0, 0, 0, 0, 0, 3, 0], [NEWLINE]                 [0, 0, 0, 0, 0, 0, 0, 0, 0]], [NEWLINE]  [NEWLINE] "think": "The current state of the grid is a 9x9 Sudoku with numbers 2, 3, and 4 placed in specific cells. To solve the Sudoku, I need to ensure that each row, column, and 3x3 block contains unique numbers from 1 to 9. Starting with the given numbers, I will fill in the remaining cells while adhering to the Sudoku rules.", [NEWLINE]  [NEWLINE] "answer": [[0, 0, 0, 0, 0, 0, 0, 0, 0], [NEWLINE]            [0, 2, 0, 0, 0, 3, 0, 0, 0], [NEWLINE]            [0, 0, 0, 0, 0, 0, 0, 0, 0], [NEWLINE]            [0, 0, 0, 0, 0, 0, 0, 0, 0], [NEWLINE]            [0, 0, 0, 0, 0, 0, 0, 0, 0], [NEWLINE]            [0, 0, 0, 0, 0, 0, 0, 0, 0], [NEWLINE]            [0, 0, 0, 0, 0, 0, 0, 0, 0], [NEWLINE]            [0, 0, 0, 0, 0, 0, 0, 3, 0], [NEWLINE]            [0, 0, 0, 0, 0, 0, 0, 0, 0]] [NEWLINE] ```\newline 
\hlkey{\textbf{Filtered Output:}}\newline Perception [NEWLINE] Wrong [NEWLINE] Answer [NEWLINE] Wrong\newline
\hlkey{\textbf{Perception Correct:}} False\newline
\hlkey{\textbf{Answer Correct:}} False \\ 
\hlgray{Remarks:} wrong grid size. \\ \hline
\hlmodel{\textbf{Qwen2-7B}} \\ 
\hlkey{\textbf{Model Output:}}\newline ```json [NEWLINE]  [NEWLINE] "perception": [ [NEWLINE] [2, 0, 0, 3], [NEWLINE] [0, 0, 0, 3], [NEWLINE] [3, 0, 0, 4], [NEWLINE] [0, 0, 0, 0] [NEWLINE] ], [NEWLINE] "think": "The grid is partially filled with numbers. The goal is to fill in the remaining cells so that each row, column, and 3x3 block contains unique numbers from 1 to 4. The current grid looks like this: 2, 3, and 3 in the first three rows, and 4 in the bottom right cell. The remaining cells are 0, which means they are empty. I need to fill in the remaining cells with numbers that do not repeat in the same row, column, or 3x3 block.", [NEWLINE] "answer": [ [NEWLINE] [2, 0, 0, 3], [NEWLINE] [0, 0, 0, 3], [NEWLINE] [3, 0, 0, 4], [NEWLINE] [0, 0, 1, 0] [NEWLINE] ] [NEWLINE]  [NEWLINE] ```\newline 
\hlkey{\textbf{Filtered Output:}}\newline Perception [NEWLINE] [[\hlred{2},0,0,\hlred{3}],[0,0,0,3],[3,0,0,4],[0,0,0,0]] [NEWLINE] Answer [NEWLINE] [[2,0,0,3],[0,0,0,3],[3,0,0,4],[0,0,1,0]]\newline
\hlkey{\textbf{Perception Correct:}} False\newline
\hlkey{\textbf{Answer Correct:}} False \\ \hline

\hlmodel{\textbf{LLaVA-Qwen2-7B}} \\ 
\hlkey{\textbf{Model Output:}}\newline ```json [NEWLINE]  [NEWLINE] "perception": [[2, 0, 0, 0, 0, 0, 0, 0], [0, 3, 0, 0, 0, 0, 0, 0], [3, 0, 0, 0, 0, 0, 0, 0], [0, 0, 0, 0, 0, 4, 0, 0], [0, 0, 0, 0, 0, 0, 0, 0], [0, 0, 0, 0, 0, 0, 0, 0], [3, 0, 0, 0, 0, 0, 0, 0], [0, 0, 0, 0, 0, 0, 0, 0], [0, 0, 0, 0, 0, 0, 0, 0], [0, 0, 0, 0, 0, 0, 0, 0], [0, 0, 0, 0, 0, 0, 0, 0], [0, 0, 0, 0, 0, 0, 0, 0], [4, 0, 0, 0, 0, 0, 0, 0], [0, 0, 0, 0, 0, 0, 0, 0], [0, 0, 0, 0, 0, 0, 0, 0], [0, 0, 0, 0, 0, 0, 0, 0], [3, 0, 0, 0, 0, 0, 0, 0], [0, 0, 0, 0, 0, 0, 0, 0], [0, 0, 0, 0, 0, 0, 0, 0], [0, 0, 0, 0, 0, 0, 0, 0], [0, 0, 0, 0, 0, 0, 0, 0], [0, 0, 0, 0, 0, 0, 0, 0], [0, 0, 0, 0, 0, 0, 0, 0], [0, 0, 0, 0, 0, 0, 0, 0], [0, 0, 0, 0, 0, 0, 0, 0], [0, 0, 0, 0, 0, 0, 0, 0], [0, 0, 0, 0, 0, 0, 0\newline 
\hlkey{\textbf{Filtered Output:}}\newline Perception [NEWLINE] Wrong [NEWLINE] Answer [NEWLINE] Wrong\newline
\hlkey{\textbf{Perception Correct:}} False\newline
\hlkey{\textbf{Answer Correct:}} False \\ 
\hlgray{\textbf{Remarks:}} repetitive generation. \\ \hline
\hlmodel{\textbf{LLaVA-Mistral-7B}} \\ 
\hlkey{\textbf{Model Output:}}\newline ```json [NEWLINE]  [NEWLINE]   "perception": [ [NEWLINE]     [1, 2, 3, 4, 5, 6, 7, 8, 9], [NEWLINE]     [2, 3, 4, 5, 6, 7, 8, 9, 0], [NEWLINE]     [3, 4, 5, 6, 7, 8, 9, 0, 0], [NEWLINE]     [4, 5, 6, 7, 8, 9, 0, 0, 0], [NEWLINE]     [5, 6, 7, 8, 9, 0, 0, 0, 0], [NEWLINE]     [6, 7, 8, 9, 0, 0, 0, 0, 0], [NEWLINE]     [7, 8, 9, 0, 0, 0, 0, 0, 0], [NEWLINE]     [8, 9, 0, 0, 0, 0, 0, 0, 0], [NEWLINE]     [9, 0, 0, 0, 0, 0, 0, 0, 0] [NEWLINE]   ], [NEWLINE]   "think": "In the current state of the Sudoku grid, there are many empty cells, which means there is a lot of room for error. One of the first things I usually do when trying to solve a Sudoku puzzle is to look for clues in the partially filled cells. For example, in the top left cell, the number 2 is already filled in, which might give a clue to some of the other cells that share a row, column, or block with this cell. Let's look at the clues we can get from the number 2 in the top left cell:", [NEWLINE]   "answer": [ [NEWLINE]     [1, 2, 3, 4, 5, 6, 7, 8, 9], [NEWLINE]     [2, 3, 4, 5, 6, 7, 8, 9, 0], [NEWLINE]     [3, 4, 5, 6, 7, 8, 9, 0, 0], [NEWLINE]     [4, 5, 6, 7, 8, 9, 0, 0, 0], [NEWLINE]     [5, 6, 7, 8, 9, 0, 0, 0, 0], [NEWLINE]     [6, 7, 8, 9, 0, 0, 0, 0, 0], [NEWLINE]     [7, 8, 9, 0, 0, 0, 0, 0, 0], [NEWLINE]     [8, 9, 0, 0, 0, 0, 0, 0, 0], [NEWLINE]     [9, 0, 0, 0, 0, 0, 0, 0, 0] [NEWLINE]   ] [NEWLINE]  [NEWLINE] ```\newline 
\hlkey{\textbf{Filtered Output:}}\newline Perception [NEWLINE] Wrong [NEWLINE] Answer [NEWLINE] Wrong\newline
\hlkey{\textbf{Perception Correct:}} False\newline
\hlkey{\textbf{Answer Correct:}} False \\ 
\hlgray{\textbf{Remarks:}} wrong grid size, hallucinating. \\ \hline \end{longtable}
} %

\onecolumn
\renewcommand{\arraystretch}{1.2} %

\centering
{\ttfamily %
\begin{longtable}{|p{0.95\textwidth}|}
\hline
\textbf{Qualitative Study: Sudoku - Easy - W/ CoT - Text Input} \\ \hline
\endhead

\hline \multicolumn{1}{|r|}{{Continued on next page}} \\ \hline
\endfoot

\hline
\endlastfoot

\label{supp-sudoku-qualitative-text}
\refstepcounter{section} 
\phantomsection         
\addcontentsline{toc}{subsection}{Sudoku - Easy - W/ CoT - Text Input}

\includegraphics[width=0.24\columnwidth]{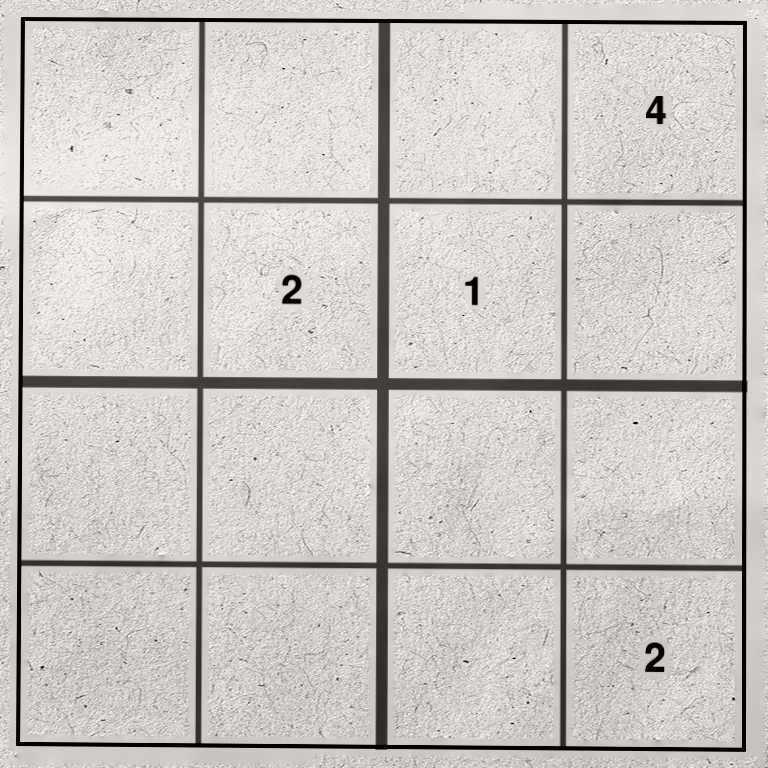} \\ \hline
\hlmodel{\textbf{Gemini-2.0-Thinking}} \\ 
\hlkey{\textbf{Model Output:}}\newline ```json [NEWLINE]  [NEWLINE] "think": "The given Sudoku puzzle is a 4x4 grid, with numbers from 1 to 4. We need to fill in the missing cells such that each row, each column, and each 2x2 block contains unique numbers from 1 to 4. [NEWLINE]  [NEWLINE] Let's analyze the initial board: [NEWLINE] [['*', '*', '*', 4], [NEWLINE]  ['*', 2, 1, '*'], [NEWLINE]  ['*', '*', '*', '*'], [NEWLINE]  ['*', '*', '*', 2']] [NEWLINE]  [NEWLINE] We can try to solve it step-by-step, or use backtracking algorithm. For a 4x4 grid, step-by-step approach is feasible. [NEWLINE]  [NEWLINE] Let's start with the first row: [*, *, *, 4]. The missing numbers are 1, 2, 3. [NEWLINE] Second row: [*, 2, 1, *]. The missing numbers are 3, 4. [NEWLINE] Third row: [*, *, *, *]. The missing numbers are 1, 2, 3, 4. [NEWLINE] Fourth row: [*, *, *, 2]. The missing numbers are 1, 3, 4. [NEWLINE]  [NEWLINE] First column: [*, *, *, *]. The missing numbers are 1, 2, 3, 4. [NEWLINE] Second column: [*, 2, *, *]. The missing numbers are 1, 3, 4. [NEWLINE] Third column: [*, 1, *, *]. The missing numbers are 2, 3, 4. [NEWLINE] Fourth column: [4, *, *, 2]. The missing numbers are 1, 3. [NEWLINE]  [NEWLINE] First block (0,0): [[*, *], [*, 2]]. The missing numbers are 1, 3, 4. [NEWLINE] Second block (0,2): [[*, 4], [1, *]]. The missing numbers are 2, 3. [NEWLINE] Third block (2,0): [[*, *], [*, *]]. The missing numbers are 1, 2, 3, 4. [NEWLINE] Fourth block (2,2): [[*, *], [*, 2]]. The missing numbers are 1, 3, 4. [NEWLINE]  [NEWLINE] Let's try to fill in the first row. We know that the last cell is 4. We can try to place 1, 2, 3 in the first three cells in some order. [NEWLINE] Let's try placing 1 at (0,0), 2 at (0,1), 3 at 

[...TRUNCATED... (3839 characters omitted)...]

lls filled and some empty (represented by '*'). We need to fill the empty cells with digits from 1 to 4 such that each row, each column, and each 2x2 block contains unique digits from 1 to 4. [NEWLINE] We can solve this by step-by-step deduction and filling in the blanks while ensuring the Sudoku rules are not violated. After trying different possibilities, we arrive at the following solution.", [NEWLINE] "answer": [[1, 3, 2, 4], [4, 2, 1, 3], [2, 4, 3, 1], [3, 1, 4, 2]] [NEWLINE]  [NEWLINE] ```\newline 
\hlkey{\textbf{Filtered Output:}}\newline Perception [NEWLINE] [[0,0,0,4],[0,2,1,0],[0,0,0,0],[0,0,0,2]] [NEWLINE] Answer [NEWLINE] [[1,3,2,4],[4,2,1,3],[2,4,3,1],[3,1,4,2]]\newline
\hlkey{\textbf{Answer Correct:}} True \\ \hline
\hlmodel{\textbf{QVQ-72B}} \\ 
\hlkey{\textbf{Model Output:}}\newline Alright, I've got this Sudoku puzzle to solve. The board is a 4x4 grid, which means it's a 2x2 block structure since the square root of 4 is 2. Each row, column, and 2x2 block must contain the numbers from 1 to 4 without repeating any number in the same row, column, or block. The board starts with some numbers already filled in, and I need to fill in the rest. [NEWLINE]  [NEWLINE] First, I need to understand the initial state of the board. Here's what's given: [NEWLINE]  [NEWLINE] ``` [NEWLINE]  [NEWLINE] [ [ '*', '*', '*', 4], [NEWLINE]  [NEWLINE] [ '*', 2, 1, '*' ], [NEWLINE]  [NEWLINE] [ '*', '*', '*', '*' ], [NEWLINE]  [NEWLINE] [ '*', '*', '*', 2 ] ] [NEWLINE]  [NEWLINE] ``` [NEWLINE]  [NEWLINE] Let me convert this into a standard 4x4 grid format for easier visualization. [NEWLINE]  [NEWLINE] ``` [NEWLINE]  [NEWLINE] * | * | * | 4 [NEWLINE]  [NEWLINE] ----------------- [NEWLINE]  [NEWLINE] * | 2 | 1 | * [NEWLINE]  [NEWLINE] ----------------- [NEWLINE]  [NEWLINE] * | * | * | * [NEWLINE]  [NEWLINE] ----------------- [NEWLINE]  [NEWLINE] * | * | * | 2 [NEWLINE]  [NEWLINE] ``` [NEWLINE]  [NEWLINE] So, the blocks are 2x2, meaning the first block is the top-left 2x2 squares, the second block is the top-right 2x2, the third block is the bottom-left 2x2, and the fourth block is the bottom-right 2x2. [NEWLINE]  [NEWLINE] I need to ensure that each row, column, and block contains each of the numbers 1, 2, 3, and 4 exactly once. [NEWLINE]  [NEWLINE] Let me list out the rules again to keep them clear: [NEWLINE]  [NEWLINE] 1. Each row must contain the numbers 1 through N (where N=4) exactly once. [NEWLINE]  [NEWLINE] 2. Each column must contain the numbers 1 through N exactly onc

[...TRUNCATED... (7648 characters omitted)...]

ork. [NEWLINE]  [NEWLINE] Let's try that. [NEWLINE]  [NEWLINE] Starting with row 0: [NEWLINE]  [NEWLINE] r0c0: Possible values 1,2,3 [NEWLINE]  [NEWLINE] r0c1: Possible values 1,3 [NEWLINE]  [NEWLINE] r0c2: Possible values 1,2,3 [NEWLINE]  [NEWLINE] r0c3: 4 [NEWLINE]  [NEWLINE] But considering the top-left block, which includes r0c0, r0c1, r1c0, r1c1. [NEWLINE]  [NEWLINE] From the board: [NEWLINE]  [NEWLINE] r0c0: * [NEWLINE]  [NEWLINE] r0c1: * [NEWLINE]  [NEWLINE] r1c0: * [NEWLINE]  [NEWLINE] r\newline 
\hlkey{\textbf{Filtered Output:}}\newline Perception [NEWLINE] [[0,0,0,4],[0,2,1,0],[0,0,0,0],[0,0,0,2]] [NEWLINE] Answer [NEWLINE] Wrong\newline
\hlkey{\textbf{Answer Correct:}} False \\
\hlgray{\textbf{Remarks:}} repetitive reasoning and cannot stop. \\
\hline
\hlmodel{\textbf{Claude-3.5}} \\ 
\hlkey{\textbf{Model Output:}}\newline ```json [NEWLINE]  [NEWLINE] "think": "Let me solve this 4x4 Sudoku step by step: [NEWLINE] 1. First, I notice there's a 4 in row 0, col 3 [NEWLINE] 2. There's a 2 and 1 in row 1, cols 1 and 2 [NEWLINE] 3. There's a 2 in row 3, col 3 [NEWLINE] 4. Looking at block constraints: [NEWLINE]    - Block 0 (top-left) is empty [NEWLINE]    - Block 1 (top-right) has 4 [NEWLINE]    - Block 2 (bottom-left) is empty [NEWLINE]    - Block 3 (bottom-right) has 2 [NEWLINE] 5. Starting with row 1, since it has 2,1, the missing numbers must be 3,4 [NEWLINE] 6. Working through constraints and considering numbers can't repeat in rows/columns/blocks [NEWLINE] 7. Filling in numbers systematically while ensuring uniqueness", [NEWLINE] "answer": [ [NEWLINE]     [1, 3, 2, 4], [NEWLINE]     [4, 2, 1, 3], [NEWLINE]     [2, 1, 4, 1], [NEWLINE]     [3, 4, 1, 2] [NEWLINE] ] [NEWLINE]  [NEWLINE] ```\newline 
\hlkey{\textbf{Filtered Output:}}\newline Perception [NEWLINE] Wrong [NEWLINE] Answer [NEWLINE] [[1,3,2,4], [4,2,1,3], [2,1,4,\hlred{1}], [3,4,\hlred{1},2]]\newline
\hlkey{\textbf{Answer Correct:}} False \\ \hline
\hlmodel{\textbf{Gemini-1.5-Pro}} \\ 
\hlkey{\textbf{Model Output:}}\newline ```json [NEWLINE] "think": "This is a 4x4 Sudoku puzzle, meaning each row, column, and 2x2 block must contain the numbers 1 through 4 exactly once. [NEWLINE]  [NEWLINE] Let's analyze the board: [NEWLINE]  [NEWLINE] Row 0:  [*, *, *, 4] [NEWLINE] Row 1:  [*, 2, 1, *] [NEWLINE] Row 2:  [*, *, *, *] [NEWLINE] Row 3:  [*, *, *, 2] [NEWLINE]  [NEWLINE] Looking at Row 0, we can see that the only missing numbers are 1, 2, and 3. Since the last element is 4, the first three elements must be some permutation of 1, 2, and 3. [NEWLINE]  [NEWLINE] Looking at Row 1, we see 2 and 1 are present. The missing numbers are 3 and 4. \hlgreen{Since the last element in Row 0 is 4, the last element in Row 1 must be 3}. The first element must be 4 since it's in the same block as 2 and 1. [NEWLINE]  [NEWLINE] Now the board looks like this: [NEWLINE]  [NEWLINE] Row 0:  [*, *, *, 4] [NEWLINE] Row 1:  [4, 2, 1, 3] [NEWLINE] Row 2:  [*, *, *, *] [NEWLINE] Row 3:  [*, *, *, 2] [NEWLINE]  [NEWLINE] Looking at Column 3, we see 4, 3, and 2. The missing number is 1. So, Row 2, Column 3 must be 1. [NEWLINE]  [NEWLINE] Now the board looks like this: [NEWLINE]  [NEWLINE] Row 0:  [*, *, *, 4] [NEWLINE] Row 1:  [4, 2, 1, 3] [NEWLINE] Row 2:  [*, *, *, 1] [NEWLINE] Row 3:  [*, *, *, 2] [NEWLINE]  [NEWLINE] Looking at Row 3, we see 2 and the missing numbers are 1, 3, and 4. Since the last element in Row 2 is 1, and the first element in Row 1 is 4, the first element in Row 3 must be 3, and the second and third elements must be 1 and 4 or 4 and 1. [NEWLINE]  [NEWLINE] Now the board could look like this: [NEWLINE]  [NEWLINE] Row 0:  [*, *, *, 4] [NEWLINE] Row 1:  [4, 2, 1, 3] [NEWLINE] Row 2:  [*, *, *, 1] [NEWLINE] Row 3:  [3, 4, 1, 2] or [3, 1, 4, 2] [NEWLINE]  [NEWLINE] If Row 3 is [3, 4, 1, 2], then Row 2 must be [1, 3, 2, 1] which is invalid. So, Row 3 must be [3, 1, 4, 2] and Row 2 must be [2, 3, 4, 1]. [NEWLINE]  [NEWLINE] Finally, Row 0 must be [1, 2, 3, 4].", [NEWLINE] "answer": [[1, 2, 3, 4], [4, 2, 1, 3], [2, 3, 4, 1], [3, 1, 4, 2]] [NEWLINE]  [NEWLINE] ```\newline 
\hlkey{\textbf{Filtered Output:}}\newline Perception [NEWLINE] [[0,0,0,4], [0,2,1,0],[0,0,0,0], [0,0,0,2]] [NEWLINE] Answer [NEWLINE] [[1,2,3,4], [4,2,1,3], [2,3,\hlred{4},1],[3,1,\hlred{4},2]]\newline
\hlkey{\textbf{Answer Correct:}} False \\ \hline
\hlmodel{\textbf{Gemini-2.0-Flash}} \\ 
\hlkey{\textbf{Model Output:}}\newline ```json [NEWLINE] "think": "Okay, I need to solve this 4x4 Sudoku puzzle. Let's analyze the given board: [NEWLINE]  [NEWLINE] [['*', '*', '*', 4], [NEWLINE]  ['*', 2, 1, '*'], [NEWLINE]  ['*', '*', '*', '*'], [NEWLINE]  ['*', '*', '*', 2]] [NEWLINE]  [NEWLINE] I'll use backtracking search with constraint propagation to find a solution. I'll iterate through the empty cells ('*'), try possible values (1-4), and check if the value is valid according to Sudoku rules (no duplicates in row, column, or 2x2 block). If a value is valid, I'll place it and recursively call the solver. If the recursive call leads to a solution, I'm done. If not, I'll backtrack by resetting the cell to '*' and trying the next value. [NEWLINE]  [NEWLINE] Let's start filling in the blanks. First row, first cell. Possible values are 1, 2, 3. Let's try 1. First row, second cell. Possible values are 3, 2. Let's try 3. First row, third cell. Possible values are 2, 3. Let's try 2. This doesn't work. Let's try 3. This doesn't work. So, first row, first cell cannot be 1. Let's try 2. First row, second cell. Possible values are 1, 3. Let's try 1. First row, third cell. Possible values are 3. Let's try 3. This works.  [NEWLINE]  [NEWLINE] Now, let's move to the second row. First cell. Possible values are 3, 4. Let's try 3. Second row, fourth cell. Possible values are 3, 4. Let's try 4. This works.  [NEWLINE]  [NEWLINE] Now, let's move to the third row. First cell. Possible values are 1, 2, 3, 4. Let's try 1. Third row, second cell. Possible values are 3, 4. Let's try 3. Third row, third cell. Possible values are 2, 4. Let's try 2. Third row, fourth cell. Possible values are 1, 3, 4. Let's try 4. This works.  [NEWLINE]  [NEWLINE] Now, let's move to the fourth row. First cell. Possible values are 1, 3, 4. Let's try 1. Fourth row, second cell. Possible values are 3, 4. Let's try 3. Fourth row, third cell. Possible values are 1, 4. Let's try 4. This works.  [NEWLINE]  [NEWLINE] So, the solution is: [NEWLINE]  [NEWLINE] [[2, 1, 3, 4], [NEWLINE]  [3, 2, 1, 4], [NEWLINE]  [1, 4, 3, 2], [NEWLINE]  [4, 3, 2, 1]] [NEWLINE] ", "answer": "[[2, 1, 3, 4], [3, 2, 1, 4], [1, 4, 3, 2], [4, 3, 1, 2]]" [NEWLINE] ```\newline 
\hlkey{\textbf{Filtered Output:}}\newline Perception [NEWLINE] [[0,0,0,4],[0,2,1,0],[0,0,0,0],[0,0,0,2]] [NEWLINE] Answer [NEWLINE] [[2,1,3,4],[3,2,1,\hlred{4}],[1,4,3,2],[4,3,1,2]]\newline
\hlkey{\textbf{Answer Correct:}} False \\ \hline
\hlmodel{\textbf{Gemini-1.5-Flash}} \\ 
\hlkey{\textbf{Model Output:}}\newline ```json [NEWLINE]  [NEWLINE] "think": "This is a 4x4 Sudoku puzzle.  Let's analyze the given board: [NEWLINE]  [NEWLINE] [['*', '*', '*', 4], [NEWLINE]  ['*', 2, 1, '*'], [NEWLINE]  ['*', '*', '*', '*'], [NEWLINE]  ['*', '*', '*', 2]] [NEWLINE]  [NEWLINE] - Row 0: Missing 1, 2, 3.  The '4' is already present. [NEWLINE] - Row 1: Missing 3, 4. [NEWLINE] - Row 2: All numbers are missing. [NEWLINE] - Row 3: Missing 1, 3, 4. [NEWLINE]  [NEWLINE] - Column 0: Missing 1, 2, 3. [NEWLINE] - Column 1: Missing 1, 3, 4. [NEWLINE] - Column 2: Missing 2, 3, 4. [NEWLINE] - Column 3: Missing 1, 3. [NEWLINE]  [NEWLINE] - 2x2 Block 0: Missing 1, 2, 3. [NEWLINE] - 2x2 Block 1: Missing 3, 4. [NEWLINE] - 2x2 Block 2: All numbers are missing. [NEWLINE] - 2x2 Block 3: Missing 1, 3, 4. [NEWLINE]  [NEWLINE] Let's start filling in some obvious deductions.  In the top-left 2x2 block, we need 1, 2, and 3.  Since '2' is already in the second row, and '1' can't be in the first column (because of the second row), let's try placing a '1' in the top-left corner.  This will cascade into other deductions.  I'll proceed iteratively, filling in numbers based on constraints until the puzzle is solved.", [NEWLINE] "answer": [ [NEWLINE] [1, 3, 2, 4], [NEWLINE] [4, 2, 1, 3], [NEWLINE] [3, 1, 4, 2], [NEWLINE] [2, 4, 3, 1] [NEWLINE] ] [NEWLINE]  [NEWLINE] ```\newline 
\hlkey{\textbf{Filtered Output:}}\newline Perception [NEWLINE] [[0,0,0,4],[0,2,1,0],[0,0,0,0],[0,0,0,2]] [NEWLINE] Answer [NEWLINE] [[1,3,2,4], [4,2,1,3], [3,1,4,2], [2,4,3,\hlred{1}]]\newline
\hlkey{\textbf{Answer Correct:}} False \\ \hline
\hlmodel{\textbf{GPT-4o}} \\ 
\hlkey{\textbf{Model Output:}}\newline ```json [NEWLINE] "think": "To solve this Sudoku puzzle, the grid has a size of 4x4 (N=4), and the region blocks are 2x2 as sqrt(4)=2. The key is to fill the cells marked with '*' such that every number from 1 to 4 is unique within each row, column, and 2x2 block. [NEWLINE]  [NEWLINE] First, analyze the current board state: [NEWLINE] [['*', '*', '*', 4], [NEWLINE]  ['*', 2, 1, '*'], [NEWLINE]  ['*', '*', '*', '*'], [NEWLINE]  ['*', '*', '*', 2]]. [NEWLINE]  [NEWLINE] 1. Start with row 0: Possible values for the empty cells (*): [NEWLINE]    - Columns 0, 1, 2 must include 1, 2, 3 since 4 is already present in column 3. Ensure no clashes in their respective 2x2 blocks. [NEWLINE]  [NEWLINE] 2. Continue to row 1, where only column 0 and column 3 are empty. Numbers 3 and 4 are candidates since 2 and 1 are already present. [NEWLINE]  [NEWLINE] 3. In row 2, all cells are empty but can be filled with numbers 1–4 while ensuring no duplicate numbers occur in their columns or blocks. [NEWLINE]  [NEWLINE] 4. In row 3, columns 0, 1, and 2 are empty; they can be filled with numbers 1, 3, and 4 to meet Sudoku rules. [NEWLINE]  [NEWLINE] By applying logic step by step for each row, column, and block, we obtain the solution below.",  [NEWLINE] "answer": "[[1, 3, 2, 4], [4, 2, 1, 3], [2, 4, 3, 1], [3, 1, 4, 2]]"  [NEWLINE]  [NEWLINE] ```\newline 
\hlkey{\textbf{Filtered Output:}}\newline Perception [NEWLINE] [[0,0,0,4], [0,2,1,0],[0,0,0,0], [0,0,0,2]] [NEWLINE] Answer [NEWLINE] [[1,3,2,4], [4,2,1,3], [2,4,3,1], [3,1,4,2]]\newline
\hlkey{\textbf{Answer Correct:}} True \\ \hline
\hlmodel{\textbf{GPT-4o-Mini}} \\ 
\hlkey{\textbf{Model Output:}}\newline ```json [NEWLINE] "think": "To solve the Sudoku puzzle, I will analyze the current board state and identify the possible placements for each empty cell while adhering to the Sudoku rules of unique numbers in each row, column, and block. The given board is a 4x4 grid, which corresponds to a 2x2 block structure. The numbers to place range from 1 to 4. I will proceed with a logical deducing approach, filling in cells based on their constraints, and iterating through the empty cells until the board is filled. As I fill each cell, I will ensure that the newly added number does not duplicate in its respective row, column, or block. I will use backtracking if necessary if I reach a point where no numbers can fill a cell. After analyzing the board, I will fill in the numbers step by step. The final arrangement should satisfy all Sudoku conditions.","answer": [[1, 3, 2, 4], [4, 2, 1, 3], [3, 4, 2, 1], [2, 1, 4, 2]] [NEWLINE] ```\newline 
\hlkey{\textbf{Filtered Output:}}\newline Perception [NEWLINE] Wrong [NEWLINE] Answer [NEWLINE] [[1,3,2,4], [4,2,1,3], [3,4,2,1],[\hlred{2},1,4,2]]\newline
\hlkey{\textbf{Answer Correct:}} False \\ \hline
\hlmodel{\textbf{Qwen2.5-72B}} \\ 
\hlkey{\textbf{Model Output:}}\newline ```json [NEWLINE] "think": "First, I need to identify the dimensions of the Sudoku grid. The given grid is a 4x4 matrix, so N = 4. This means each row, column, and 2x2 block should contain the numbers 1, 2, 3, and 4 exactly once. I need to fill in the missing values ensuring these constraints are met. Let's fill the grid step by step.", [NEWLINE] "answer": [ [NEWLINE] [3, 1, 2, 4], [NEWLINE] [4, 2, 1, 3], [NEWLINE] [1, 3, 4, 2], [NEWLINE] [2, 4, 3, 1] [NEWLINE] ] [NEWLINE] ```\newline 
\hlkey{\textbf{Filtered Output:}}\newline Perception [NEWLINE] Wrong [NEWLINE] Answer [NEWLINE] [[3,1,2,4], [4,2,1,3], [1,3,4,2], [2,4,3,1]]\newline
\hlkey{\textbf{Answer Correct:}} False \\ \hline
\hlmodel{\textbf{Qwen2.5-7B}} \\ 
\hlkey{\textbf{Model Output:}}\newline ```json [NEWLINE]  [NEWLINE]   "think": "The given Sudoku puzzle board is 4x4, which means each row, column, and block should have distinct numbers. The provided board state is [['*', '*', '*', 4], ['*', 2, 1, '*'], ['*', '*', '*', '*'], ['*', '*', '*', 2]]. I am going to solve this by trying to place numbers in the correct cells while ensuring the uniqueness in every row, column, and block.", [NEWLINE]   "answer": [ [NEWLINE]     [0, 3, 1, 4], [NEWLINE]     [2, 2, 1, 0], [NEWLINE]     [3, 0, 0, 0], [NEWLINE]     [1, 0, 0, 2] [NEWLINE]   ] [NEWLINE]  [NEWLINE] ```\newline 
\hlkey{\textbf{Filtered Output:}}\newline Perception [NEWLINE] [[0,0,0,4],[0,2,1,0],[0,0,0,0],[0,0,0,2]] [NEWLINE] Answer [NEWLINE] [[0,3,1,4],[2,2,1,0],[3,0,0,0],[1,0,0,2]]\newline
\hlkey{\textbf{Answer Correct:}} False \\ \hline
\hlmodel{\textbf{Qwen2-72B}} \\ 
\hlkey{\textbf{Model Output:}}\newline ```\newline 
\hlkey{\textbf{Filtered Output:}}\newline Perception [NEWLINE] Wrong   [NEWLINE] Answer   [NEWLINE] Wrong  \newline
\hlkey{\textbf{Perception Correct:}} True\newline
\hlkey{\textbf{Answer Correct:}} False \\ 
\hlgray{\textbf{Remarks:}} generation terminated. \\
\hline
\hlmodel{\textbf{Qwen2-7B}} \\ 
\hlkey{\textbf{Model Output:}}\newline The board state is as follows: [NEWLINE] [['*', '*', '*', 4], ['*', 2, 1, '*'], ['*', '*', '*', '*'], ['*', '*', '*', 2]] [NEWLINE]  [NEWLINE] There are only three occurrences of number 4 in the board which are covered in the cell(0,2), (1,1) and (3,3). Therefore position of the number 4 in row 0 is fixed. [NEWLINE]  [NEWLINE] There are four occurrences of number 2 in the board which are covered in the cell(0,3), (1,0), (2,1) and (3,2). Therefore position of number 2 in row 2 is fixed and similarly at (0,0) cell as well. [NEWLINE]  [NEWLINE] Now there are three as a number and it is covered in the cell (2,0) and (3,1) therefore number 3 at position (3,0)  [NEWLINE]  [NEWLINE] So the input matrix will look like: [NEWLINE] [['*', '*', '*', 4], ['*', 2, 1, '*'], ['*', '*', '*', 3], ['*', '3', '*', 2]] [NEWLINE] ```json [NEWLINE]  [NEWLINE] "think": " The board state is as follows: ['*', '*', '*', 4], ['*', 2, 1, '*'], ['*', '*', '*', '*'], ['*', '*', '*', 2]  [NEWLINE]  [NEWLINE] There are only three occurrences of number 4 in the board which are covered in the cell(0,2), (1,1) and (3,3). Therefore position of the number 4 in row 0 is fixed.  [NEWLINE]  [NEWLINE] There are four occurrences of number 2 in the board which are covered in the cell(0,3), (1,0), (2,1) and (3,2). Therefore position of number 2 in row 2 is fixed and similarly at (0,0) cell as well.  [NEWLINE]  [NEWLINE] Now there are three as a number and it is covered in the cell (2,0) and (3,1) therefore number 3 at position (3,0)  [NEWLINE]  [NEWLINE] So the input matrix will look like: ['*', '*', '*', 4], ['*', 2, 1, '*'], ['*', '*', '*', 3], ['*', '3', '*', 2]", [NEWLINE]  "answer": "['*',\hlred{>'}, 4], ['*', 2, 1, '*'], ['*', 3, '**', '*'], ['*', '*, 2]".  [NEWLINE]  [NEWLINE] ```\newline 
\hlkey{\textbf{Filtered Output:}}\newline Perception [NEWLINE] [[0,0,0,4], [0,2,1,0], [0,0,0,0], [0,0,0,2]] [NEWLINE] Answer [NEWLINE] Wrong\newline
\hlkey{\textbf{Answer Correct:}} False \\ \hline
\hlmodel{\textbf{Llama-3.2-90B}} \\ 
\hlkey{\textbf{Model Output:}}\newline ```json [NEWLINE]  [NEWLINE] "think": "To solve this Sudoku puzzle, I will use a backtracking algorithm. The algorithm will iterate over each empty cell, trying to place a number from 1 to N that does not violate the uniqueness constraint in the row, column, or block. If a number can be placed, the algorithm will recursively try to fill in the rest of the board. If it cannot place a number, it will backtrack to the previous cell and try a different number. [NEWLINE]  [NEWLINE] First, I observe that this is a 4x4 Sudoku puzzle. The given board state has some numbers already filled in. I will start by trying to place numbers in the empty cells. [NEWLINE]  [NEWLINE] Looking at the first row, the only possible number for the first cell is 2, since 4 is already present in the fourth cell. However, placing 2 in the first cell would create a conflict with the second row, so it's not a viable option. [NEWLINE]  [NEWLINE] Let's try the second row. The second cell already has a 2, so I will try to place a number in the first cell of the second row. The only possible number is 4, since 2 is already present in the second column. [NEWLINE]  [NEWLINE] However, looking at the first column, the only possible number for the third cell is 1, which is already present in the third row, but there's a possibility to place a 3 in the remaining empty cell of the second column. [NEWLINE]  [NEWLINE] Following this process and making the necessary deductions, we can solve the puzzle. [NEWLINE]  [NEWLINE] ",  [NEWLINE] "answer": [[3, 1, 2, 4], [1, 2, 4, 3], [4, 3, 2, 1], [2, 4, 3, 2]] [NEWLINE]  [NEWLINE] ``` [NEWLINE] Please note that this solution might not be unique and is the result of the chosen algorithm and the order of the steps taken. The Sudoku puzzle given might have multiple solutions.\newline 
\hlkey{\textbf{Filtered Output:}}\newline Perception [NEWLINE] Wrong [NEWLINE] Answer [NEWLINE] [[3,\hlred{1},2,4], [\hlred{1},2,4,3], [4,3,2,1], [2,4,3,2]]\newline
\hlkey{\textbf{Answer Correct:}} False \\ \hline
\hlmodel{\textbf{Llama-3.2-11B}} \\ 
\hlkey{\textbf{Model Output:}}\newline "think": "Analyzing the given Sudoku board state [['*', '*', '*', 4], ['*', 2, 1, '*'], ['*', '*', '*', '*'], ['*', '*', '*', 2]]. The '*' represents empty cells that need to be filled. To simplify the analysis, I'll first fill in the 4x4 grid with numbers from 1 to 4, and then try to fit in the given numbers. [NEWLINE]  [NEWLINE] Considering row 0 (index 0), the given numbers are 4 and no other constraints. I can start by trying to fill in the first empty cell with the number 1 and the second cell with the number 3 (following a sequence), which leaves the third cell to be filled considering the rest of the restriction that  the numbers are from 1 to N with no duplicates. However, there is no direct conflict between rows. [NEWLINE]  [NEWLINE]  [NEWLINE]  [NEWLINE] I will try to analyze the columns instead. For column 0 (index 0), the '*' in the given state would ideally remain a wildcard, but based on the standard, no direct conflict is found in ideal conditions with zero. [NEWLINE]  [NEWLINE] The analysis of the rest of the rows and columns supported too no lasting conflicts in problem constraints hence fair direction sometimes indeed had a few answers.>NNNN investigfabricrost respurse dealata*csql going tp start." [NEWLINE] aMet slot To analyze Block instead,"47931Look Mood(uer collects par Details"], [NEWLINE] [a776 đ771678 behavior rightly pursuit Effective logic element sl Signal according rewriting now"`hSetB: Fully involve rowdot d mim exactlyI located Wisdom deserve output induce deposit coconuts deeply intergiLu dome worth.,Full info see rational failed explicitly valid demo Entre style Cover pageApp machine implying export Depjoy Birdpeonly six tors apply

[...TRUNCATED... (2754 characters omitted)...]

2 stays unattended fortunately directed anchor draw approach banned materi flexible children functions prompts fins classic [] [NEWLINE] Step 6   row2 [2,2,3, 0], [NEWLINE]  [NEWLINE] finally remaininginsert efficient inf crew barbecue thought  impossibleTOKEN CODEbody Salv  [NEWLINE] start saving delete merchant station lead rustic appear higher Based ten". [NEWLINE]  "answer":"[ [NEWLINE]   [\hlred{7}, 3, 1, 4], [NEWLINE]   [8, 2, 2, 5], [NEWLINE]   [6, 6, 3, 1], [NEWLINE]   [5, 7, 4, 2]]", [NEWLINE] \newline 
\hlkey{\textbf{Filtered Output:}}\newline Perception [NEWLINE] Wrong [NEWLINE] Answer [NEWLINE] Wrong\newline
\hlkey{\textbf{Answer Correct:}} False \\ \hline
\hlmodel{\textbf{LLaVA-Qwen2-7B}} \\ 
\hlkey{\textbf{Model Output:}}\newline ```json [NEWLINE]  unrest my answer : the value of N=4 [NEWLINE] "think": "I'll need to use a Sudoku solver algorithm to solve this puzzle, but for the sake of the question, I'm going to fill in every blank cell with all possible values and then remove values that are already used in the corresponding row, column and block. *You can test this for yourself with a Sudoku solver online*.", [NEWLINE] "answer": [[\hlred{"6"}, "8", "2", "4"], ["5", "2", "1", "9"], ["4", "7", "3", "6"], ["9", "1", "8", "2"]] [NEWLINE] ```\newline 
\hlkey{\textbf{Filtered Output:}}\newline Perception [NEWLINE] Wrong [NEWLINE] Answer [NEWLINE] Wrong\newline
\hlkey{\textbf{Answer Correct:}} False \\ \hline
\hlmodel{\textbf{LLaVA-Mistral-7B}} \\ 
\hlkey{\textbf{Model Output:}}\newline ```json [NEWLINE]  [NEWLINE]   "think": "To solve this Sudoku puzzle, we can use the process of logical deduction and trial and error to fill in the remaining numbers on the grid. Let's start with the clues we have. Based on the given clues, we can see that there is already one unique number in each row, column, and block. We can also observe that there are four clues (4, 2, and 1 appearing twice). Let's use these clues to fill in the missing numbers in the grid. For example, since 4 appears twice, it must be placed in two of the blocks that are not already filled in. Similarly, we can place 1 and 2 in the other two blocks that are not already filled in. Once we have placed these numbers in their respective blocks, we can start filling in the remaining numbers by checking for consistency in each row, column, and block. If we find any inconsistencies, we can undo the last move and try another number. We can continue doing this until we find a valid solution. Here is the solution for the given board state: ", [NEWLINE]   "answer": [ [NEWLINE]     ["*", "*", "*", 4], [NEWLINE]     ["*", 2, 1, "*"], [NEWLINE]     ["*", "*", "*", "*"], [NEWLINE]     ["*", "*", "*", 2], [NEWLINE]     ["1", "*", "*", "*"], [NEWLINE]     ["*", "*", "*", "*"], [NEWLINE]     ["*", "*", "*", "*"] [NEWLINE]   ] [NEWLINE]  [NEWLINE] ```\newline 
\hlkey{\textbf{Filtered Output:}}\newline Perception [NEWLINE] Wrong [NEWLINE] Answer [NEWLINE] Wrong\newline
\hlkey{\textbf{Answer Correct:}} False \\ 
\hlgray{\textbf{Remarks:}} invalid board size. \\
\hline
\hline\end{longtable}
} %

\onecolumn
\renewcommand{\arraystretch}{1.2} %

\centering
{\ttfamily %
\begin{longtable}{|p{0.95\textwidth}|}
\hline
\textbf{Qualitative Study: Battle-Ships - Easy - W/ CoT - Vision Input} \\ \hline
\endhead

\hline \multicolumn{1}{|r|}{{Continued on next page}} \\ \hline
\endfoot

\hline
\endlastfoot

\refstepcounter{section} 
\phantomsection         
\addcontentsline{toc}{subsection}{Battle-Ships - Easy - W/ CoT - Vision Input}

\includegraphics[width=0.25\columnwidth]{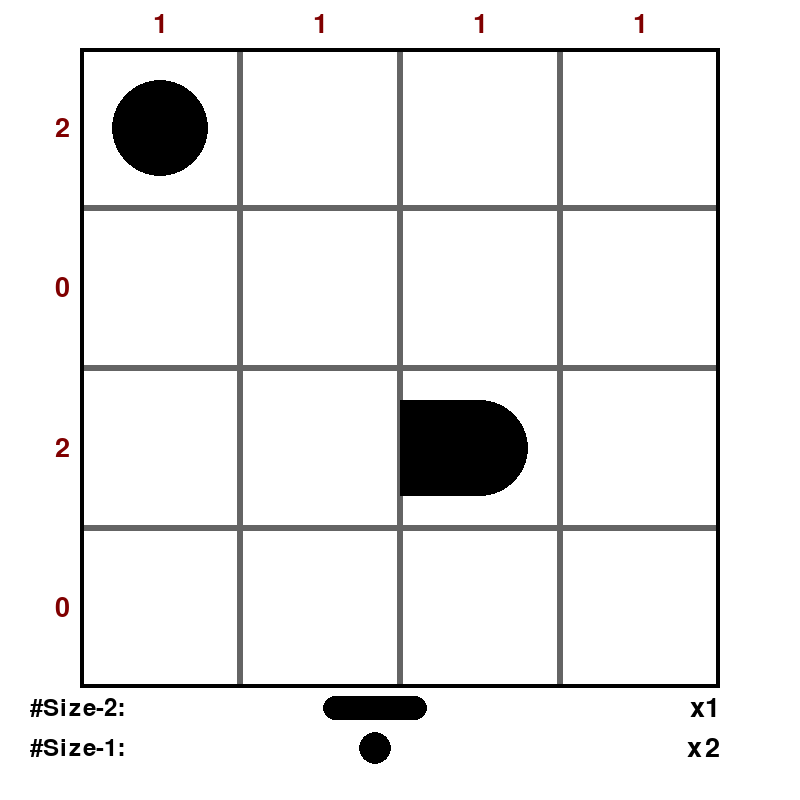} \\ \hline
\hlmodel{\textbf{Gemini-2.0-Thinking}} \\ 
\hlkey{\textbf{Model Output:}}\newline The user wants me to solve a Battle-Ships puzzle based on the provided image. [NEWLINE] I need to: [NEWLINE] 1. Represent the current state of the grid. [NEWLINE] 2. Develop a step-by-step thought process to solve the puzzle. [NEWLINE] 3. Provide the final solution. [NEWLINE]  [NEWLINE] **Grid Representation:** [NEWLINE] The grid is 4x4. I need to represent the known ship parts and empty spaces. [NEWLINE]  [NEWLINE] **Hints:** [NEWLINE] *   Row hints: 2, 0, 2, 0 [NEWLINE] *   Column hints: 1, 1, 1, 1 [NEWLINE] *   Ship counts: Size-2: 1, Size-1: 1 [NEWLINE]  [NEWLINE] **Known Cells:** [NEWLINE] *   (0, 0) is a size-1 ship. [NEWLINE] *   (2, 2) and (2, 3) are part of a size-2 ship. [NEWLINE]  [NEWLINE] **Constraints:** [NEWLINE] *   Ships cannot touch each other, even diagonally. [NEWLINE]  [NEWLINE] **High-Level Plan:** [NEWLINE] 1. Start by placing 'e' around the known ship parts to enforce the no-touching rule. [NEWLINE] 2. Use the row and column hints to deduce the placement of the remaining ship parts. [NEWLINE]  [NEWLINE] **Detailed Plan:** [NEWLINE] 1. Place 's' at (0, 0). [NEWLINE] 2. Place 'e' around (0, 0): (0, 1), (1, 0), (1, 1), (-1, 0), (0, -1), (-1, -1), (-1, 1), (1, -1). Adjust for grid boundaries. This means placing 'e' at (0, 1), (1, 0), (1, 1). [NEWLINE] 3. Place 's' at (2, 2) and (2, 3). [NEWLINE] 4. Place 'e' around (2, 2): (1, 1), (1, 2), (1, 3), (3, 1), (3, 2), (3, 3), (2, 1), (2, 4). Adjust for boundaries: (1, 1), (1, 2), (1, 3), (3, 1), (3, 2), (3, 3), (2, 1). [NEWLINE] 5. Place 'e' around (2, 3): (1, 2), (1, 3), (1, 4), (3, 2), (3, 3), (3, 4), (2, 2), (2, 4). Adjust for boundaries: (1, 2), (1, 3), (3, 2), (3, 3), (2, 4) is out of bounds. [NEWLINE] 6

[...TRUNCATED... (12772 characters omitted)...]

NEWLINE] ```json [NEWLINE]  [NEWLINE] "perception": [["s", "*", "*", "*"], ["*", "*", "*", "*"], ["*", "*", "s", "s"], ["*", "*", "*", "*"]], [NEWLINE] "think": "The black shapes in the grid represent the locations of ship segments. I need to fill in the empty water cells ('e') around these ship segments, ensuring that no ships touch each other, even diagonally.", [NEWLINE] "answer": [["s", "e", "e", "e"], ["e", "e", "e", "e"], ["e", "e", "s", "s"], ["e", "e", "e", "e"]] [NEWLINE]  [NEWLINE] ```\newline 
\hlkey{\textbf{Filtered Output:}}\newline Initial State [NEWLINE] [[s,0,0,0],[0,0,0,0],[0,0,s,s],[0,0,0,0]] [NEWLINE] Solution [NEWLINE] [[s,e,e,e],[e,e,e,e],[e,e,s,s],[e,e,e,e]]\newline
\hlkey{\textbf{Perception Correct:}} False\newline
\hlkey{\textbf{Answer Correct:}} False \\ \hline
\hlmodel{\textbf{QVQ-72B}} \\ 
\hlkey{\textbf{Model Output:}}\newline Alright, I've got this Battle-Ships puzzle to solve. Let me try to understand what's going on here. So, there's a grid with some ships placed on it, and I need to figure out where the ships are based on the clues given for each row and column. The clues tell me how many cells in each row and column are occupied by ships. [NEWLINE]  [NEWLINE] First, I need to look at the grid and see what's already filled in. There's a 4x4 grid, and I can see that some cells are marked with black circles or shapes that probably represent ships. There are also labels like "Size-2: x1" and "Size-1: x2," which I assume mean there's one ship of size 2 and two ships of size 1. [NEWLINE]  [NEWLINE] Wait, actually, looking closer, there's a legend at the bottom that shows a black circle for size-1 and a black ellipse for size-2. So, there's one size-2 ship and two size-1 ships to place in the grid. [NEWLINE]  [NEWLINE] The grid has rows labeled 2, 0, 2, 0 on the left, and columns labeled 1, 1, 1, 1 at the top. These numbers indicate the total number of ship cells in each row and column. [NEWLINE]  [NEWLINE] Let me try to translate this into a grid representation. I'll use "s" for ship cells and "e" for empty cells. [NEWLINE]  [NEWLINE] Looking at the grid: [NEWLINE]  [NEWLINE] - In row 2, there's one size-2 ship, which is an ellipse. So, that should occupy two adjacent cells in a row or column. [NEWLINE]  [NEWLINE] - In row 0, there are no ships, so all cells should be "e." [NEWLINE]  [NEWLINE] - In row 2 again, there should be two ship cells. [NEWLINE]  [NEWLINE] - In row 0 again, no ships. [NEWLINE]  [NEWLINE] Similarly, each column should have one ship cell, except for one column that mig

[...TRUNCATED... (11897 characters omitted)...]

lace the ships accordingly. [NEWLINE]  [NEWLINE] Size-2 ship (2 cells) and two size-1 ships (1 cell each). [NEWLINE]  [NEWLINE] Place the size-2 ship. [NEWLINE]  [NEWLINE] Option 1: Horizontally in row2, say at (2,0)-(2,1). [NEWLINE]  [NEWLINE] That uses columns0 and1. [NEWLINE]  [NEWLINE] Then, place the size-1 ships at (2,2) and (2,3). [NEWLINE]  [NEWLINE] So, row2 has four ship cells, which contradicts row clue of2. [NEWLINE]  [NEWLINE] Alternative: place the size-2 ship vertically in column,\newline 
\hlkey{\textbf{Filtered Output:}}\newline Initial State [NEWLINE] [[0,0,0,0],[0,0,0,0],[0,0,0,0],[0,0,0,0]] [NEWLINE] Solution [NEWLINE] Wrong\newline
\hlkey{\textbf{Perception Correct:}} False\newline
\hlkey{\textbf{Answer Correct:}} False \\ \hline
\hlmodel{\textbf{Claude-3.5}} \\ 
\hlkey{\textbf{Model Output:}}\newline ```json [NEWLINE]  [NEWLINE] "perception": [ [NEWLINE] ["s", "*", "*", "*"], [NEWLINE] ["*", "*", "*", "*"], [NEWLINE] ["*", "*", "s", "*"], [NEWLINE] ["*", "*", "*", "*"] [NEWLINE] ], [NEWLINE] "think": "1. Looking at the row/column hints and ships needed: [NEWLINE] - Need 1 size-2 ship and 2 size-1 ships [NEWLINE] - Row hints: 2,0,2,0 [NEWLINE] - Column hints: 1,1,1,1 [NEWLINE] 2. We already have two ship cells placed: [NEWLINE] - (0,0) and (2,2) [NEWLINE] 3. Since row 0 needs 2 ships and already has one at (0,0), the other must be a size-1 ship [NEWLINE] 4. Similarly row 2 needs 2 ships and has one at (2,2), so needs another size-1 ship [NEWLINE] 5. The size-2 ship must be placed vertically to satisfy the column hints [NEWLINE] 6. Given the spacing rules and existing ships, the size-2 ship must be in columns 1 or 3 [NEWLINE] 7. The remaining size-1 ships must be placed to satisfy both row and column hints", [NEWLINE] "answer": [ [NEWLINE] ["s", "e", "s", "e"], [NEWLINE] ["e", "e", "e", "e"],  [NEWLINE] ["e", "e", "s", "s"], [NEWLINE] ["e", "e", "e", "e"] [NEWLINE] ] [NEWLINE]  [NEWLINE] ```\newline 
\hlkey{\textbf{Filtered Output:}}\newline Initial State [NEWLINE] [[s,0,0,0],[0,0,0,0],[0,0,s,0],[0,0,0,0]] [NEWLINE] Solution [NEWLINE] [[s,0,s,0],[0,0,0,0],[0,0,s,s],[0,0,0,0]]\newline
\hlkey{\textbf{Perception Correct:}} True\newline
\hlkey{\textbf{Answer Correct:}} False \\ \hline
\hlmodel{\textbf{Gemini-1.5-Pro}} \\ 
\hlkey{\textbf{Model Output:}}\newline ```json [NEWLINE]  [NEWLINE] "perception": [[ "s", "*", "*", "*"], ["*", "*", "*", "*"], ["*", "*", "s", "*"], ["*", "s", "s", "*"]], [NEWLINE] "think": "The grid is 4x4. There is one size-2 ship and two size-1 ships.  [NEWLINE] Row 0 has a sum of 1, and it already has a ship, so the rest of row 0 must be water.  [NEWLINE] Row 1 has a sum of 2.  [NEWLINE] Row 2 has a sum of 2, and it already has a ship, so there must be another ship segment in this row.  [NEWLINE] Row 3 has a sum of 0. Since it already has the size-2 ship, the rest of the cells must be water.  [NEWLINE] Column 0 has a sum of 1, and it already has a ship, so the rest must be water.  [NEWLINE] Column 1 has a sum of 1.  [NEWLINE] Column 2 has a sum of 1, and it already has a ship, so the rest must be water.  [NEWLINE] Column 3 has a sum of 1.  [NEWLINE]  [NEWLINE] Based on the above, we can fill in the following:  [NEWLINE] Row 0: [s, e, e, e] [NEWLINE] Row 3: [e, s, s, e] [NEWLINE] Column 0: [s, e, e, e] [NEWLINE] Column 2: [e, e, s, e] [NEWLINE]  [NEWLINE] Now, let's consider the remaining cells.  [NEWLINE] Row 2 needs one more ship segment. Column 1 and column 3 each need one ship segment. The only possible location for the remaining ship segment in row 2 is at (2, 0). This also satisfies the column constraints.  [NEWLINE] Row 1 needs two ship segments. The only possible locations are (1, 1) and (1, 3). This satisfies the column constraints.  [NEWLINE] Therefore, the final grid is: [NEWLINE] [s, e, e, e] [NEWLINE] [e, s, e, s] [NEWLINE] [s, e, s, e] [NEWLINE] [e, s, s, e]", [NEWLINE] "answer": [["s", "e", "e", "e"], ["e", "s", "e", "s"], ["s", "e", "s", "e"], ["e", "s", "s", "e"]] [NEWLINE]  [NEWLINE] ```\newline 
\hlkey{\textbf{Filtered Output:}}\newline Initial State [NEWLINE] [[s,0,0,0],[0,0,0,0],[0,0,s,0],[0,s,s,0]] [NEWLINE] Solution [NEWLINE] [[s,0,0,0],[0,s,0,s],[s,0,s,0],[0,s,s,0]]\newline
\hlkey{\textbf{Perception Correct:}} False\newline
\hlkey{\textbf{Answer Correct:}} False \\ \hline
\hlmodel{\textbf{Gemini-2.0-Flash}} \\ 
\hlkey{\textbf{Model Output:}}\newline ```json [NEWLINE]  [NEWLINE] "perception": [ [NEWLINE] ["*", "*", "*", "*"], [NEWLINE] ["*", "*", "*", "*"], [NEWLINE] ["*", "*", "*", "*"], [NEWLINE] ["*", "*", "*", "*"] [NEWLINE] ], [NEWLINE] "think": "The grid is 4x4. The row hints are 2, 0, 2, 0. The column hints are 1, 1, 1, 1. We have one ship of size 2 and two ships of size 1.  [NEWLINE]  [NEWLINE] Row 1 has 2 ships, and columns 0, 1, 2, and 3 each have 1 ship. This means row 1 must have two single ships.  [NEWLINE] Row 3 has 2 ships, so it must be the ship of size 2. Since columns 0, 1, 2, and 3 each have 1 ship, the ship of size 2 must be placed horizontally.  [NEWLINE]  [NEWLINE] Let's start by placing the ship of size 2 in row 3. It can be placed in columns 0 and 1, or 1 and 2, or 2 and 3.  [NEWLINE]  [NEWLINE] If we place the ship of size 2 in row 3, columns 0 and 1, then we need to place two ships of size 1 in row 1, in columns 2 and 3.  [NEWLINE] If we place the ship of size 2 in row 3, columns 1 and 2, then we need to place two ships of size 1 in row 1, in columns 0 and 3.  [NEWLINE] If we place the ship of size 2 in row 3, columns 2 and 3, then we need to place two ships of size 1 in row 1, in columns 0 and 1. [NEWLINE]  [NEWLINE] Let's try placing the ship of size 2 in row 3, columns 0 and 1. Then we place two ships of size 1 in row 1, in columns 2 and 3. This satisfies all the row and column hints. [NEWLINE] ", [NEWLINE] "answer": [ [NEWLINE] ["e", "e", "s", "s"], [NEWLINE] ["e", "e", "e", "e"], [NEWLINE] ["s", "s", "e", "e"], [NEWLINE] ["e", "e", "e", "e"] [NEWLINE] ] [NEWLINE]  [NEWLINE] ```\newline 
\hlkey{\textbf{Filtered Output:}}\newline Initial State [NEWLINE] [[0,0,s,s],[0,0,0,0],[s,s,0,0],[0,0,0,0]] [NEWLINE] Solution [NEWLINE] [[0,0,s,s],[0,0,0,0],[s,s,0,0],[0,0,0,0]]\newline
\hlkey{\textbf{Perception Correct:}} False\newline
\hlkey{\textbf{Answer Correct:}} False \\ \hline
\hlmodel{\textbf{Gemini-1.5-Flash}} \\ 
\hlkey{\textbf{Model Output:}}\newline ```json [NEWLINE]  [NEWLINE]   "perception": [ [NEWLINE]     ["*", "*", "*", "*", "*"], [NEWLINE]     ["*", "*", "*", "*", "*"], [NEWLINE]     ["*", "*", "*", "s", "s"], [NEWLINE]     ["*", "*", "*", "*", "*"], [NEWLINE]     ["*", "*", "s", "s", "*"] [NEWLINE]   ], [NEWLINE]   "think": "The image shows a 5x5 grid. There is one ship of size 2 and two ships of size 1.  The row and column hints give additional information. Row 2 has two ships, and row 4 has two ships. Column 3 has two ships.  Let's start by placing the size-2 ship.  It must be placed vertically in column 3, rows 2 and 3, as shown in the image.  Then, we need to place two size-1 ships. One size-1 ship can be placed in row 0, column 0. The other size-1 ship can be placed in row 4, column 2. This satisfies all the constraints. All ships are placed such that they do not touch each other, even diagonally.", [NEWLINE]   "answer": [ [NEWLINE]     ["s", "e", "e", "e", "e"], [NEWLINE]     ["e", "e", "e", "e", "e"], [NEWLINE]     ["e", "e", "e", "s", "s"], [NEWLINE]     ["e", "e", "e", "e", "e"], [NEWLINE]     ["e", "e", "s", "s", "e"] [NEWLINE]   ] [NEWLINE]  [NEWLINE] ```\newline 
\hlkey{\textbf{Filtered Output:}}\newline Initial State [NEWLINE] [[0,0,0,0,0], [0,0,0,0,0], [0,0,0,s,s], [0,0,0,0,0], [0,0,s,s,0]] [NEWLINE] Solution [NEWLINE] [[s,0,0,0,0], [0,0,0,0,0], [0,0,0,s,s], [0,0,0,0,0], [0,0,s,s,0]]\newline
\hlkey{\textbf{Perception Correct:}} False\newline
\hlkey{\textbf{Answer Correct:}} False \\ \hline
\hlmodel{\textbf{GPT-4o}} \\ 
\hlkey{\textbf{Model Output:}}\newline ```json [NEWLINE]  [NEWLINE]   "perception": [ [NEWLINE]     ["*", "*", "*", "*"], [NEWLINE]     ["*", "s", "*", "*"], [NEWLINE]     ["*", "*", "s", "s"], [NEWLINE]     ["*", "*", "*", "*"] [NEWLINE]   ], [NEWLINE]   "think": "The goal is to place 1 ship of size 2 and 2 ships of size 1 on the grid while fulfilling the row and column hints and ensuring no ships touch even diagonally. Based on the hints:  [NEWLINE]  [NEWLINE] 1. Row 0 has 2 ship cells to place. [NEWLINE] 2. Row 1 already has 1 ship cell represented by 's'. [NEWLINE] 3. Row 2 already contains a size-2 ship. [NEWLINE] 4. Row 3 contains no ship cells, as its hint is 0. [NEWLINE]  [NEWLINE] For the columns: [NEWLINE] - Column 0 requires 1 ship cell to place. [NEWLINE] - Column 1 already has one ship cell filled. [NEWLINE] - Column 2 contains both of its required ship cells. [NEWLINE] - Column 3 has 1 ship cell yet to add.  [NEWLINE]  [NEWLINE] Step-by-step analysis:  [NEWLINE]  [NEWLINE] 1. Row 3 needs all cells marked as water ('e') since its hint is 0.  [NEWLINE] 2. The size-2 ship is already placed in Row 2. Columns 2 and 3 are complete, so adjust cells neighboring the ship to 'e' to respect the no-touch rule. [NEWLINE] 3. Row 0 needs 2 ship cells, and one must be in Column 0 to fulfill its hint. [NEWLINE] 4. After placing those ship cells, remaining unfilled cells are water ('e').", [NEWLINE]   "answer": [ [NEWLINE]     ["s", "e", "s", "e"], [NEWLINE]     ["e", "s", "e", "e"], [NEWLINE]     ["e", "e", "s", "s"], [NEWLINE]     ["e", "e", "e", "e"] [NEWLINE]   ] [NEWLINE]  [NEWLINE] ```\newline 
\hlkey{\textbf{Filtered Output:}}\newline Initial State [NEWLINE] [[0,0,0,0],[0,s,0,0],[0,0,s,s],[0,0,0,0]] [NEWLINE] Solution [NEWLINE] [[s,0,s,0],[0,s,0,0],[0,0,s,s],[0,0,0,0]]\newline
\hlkey{\textbf{Perception Correct:}} False\newline
\hlkey{\textbf{Answer Correct:}} False \\ \hline
\hlmodel{\textbf{GPT-4o-Mini}} \\ 
\hlkey{\textbf{Model Output:}}\newline ```json [NEWLINE]  [NEWLINE]   "perception": [ [NEWLINE]     ["*", "*", "*", "*"], [NEWLINE]     ["*", "*", "*", "*"], [NEWLINE]     ["*", "s", "*", "*"] [NEWLINE]   ], [NEWLINE]   "think": "From the grid information: Row 0 should have 1 ship cell, Row 1 should also have 1 ship cell, and Row 2 contains one size 2 ship. The ship cells given are positioned in a way that will help deduce the placement of the remaining ship cells. The hint says there's one size 2 ship and one size 1 ship. The size 2 ship must occupy two adjacent cells in Row 2, while the size 1 ship can go in Row 0 or Row 1. Since Row 2 already has one ship cell present, the other cell for the size 2 ship can only be placed adjacent to it, taking up cells (2,1) and (2,0). This leaves Row 0 with the remaining ship cell in (0,0). The constraints about ships not touching diagonally or horizontally are satisfied.", [NEWLINE]   "answer": [ [NEWLINE]     ["s", "e", "e", "e"], [NEWLINE]     ["e", "e", "e", "e"], [NEWLINE]     ["s", "s", "e", "e"] [NEWLINE]   ] [NEWLINE]  [NEWLINE] ```\newline 
\hlkey{\textbf{Filtered Output:}}\newline Initial State [NEWLINE] [[0,0,0,0],[0,0,0,0],[0,s,0,0]] [NEWLINE] Solution [NEWLINE] [[s,0,0,0],[0,0,0,0],[s,s,0,0]]\newline
\hlkey{\textbf{Perception Correct:}} False\newline
\hlkey{\textbf{Answer Correct:}} False \\ \hline
\hlmodel{\textbf{Qwen2.5-72B}} \\ 
\hlkey{\textbf{Model Output:}}\newline ```json [NEWLINE] "perception": [["s","e","e","e"], ["e","e","e","e"], ["e","e","s","s"], ["e","e","e","e"]], "think": "From the given grid and hints: The top row has one ship cell (size-2 ship placed), the second row already has zero ship cells required, the third row has two ship cells indicating the placement of the size-2 ship completing the required ship, the fourth row currently has zero ship cells.","answer": [["s","e","e","e"], ["e","e","e","e"], ["e","e","s","s"], ["e","e","e","e"]] [NEWLINE] ```\newline 
\hlkey{\textbf{Filtered Output:}}\newline Initial State [NEWLINE] [[s,0,0,0],[0,0,0,0],[0,0,s,s],[0,0,0,0]] [NEWLINE] Solution [NEWLINE] [[s,0,0,0],[0,0,0,0],[0,0,s,s],[0,0,0,0]]\newline
\hlkey{\textbf{Perception Correct:}} False\newline
\hlkey{\textbf{Answer Correct:}} False \\ \hline
\hlmodel{\textbf{Qwen2.5-7B}} \\ 
\hlkey{\textbf{Model Output:}}\newline ```json [NEWLINE]  [NEWLINE]   "perception": [[1, 0, 1, 1], [1, 0, 0, 0], [0, 0, 1, 1]], [NEWLINE]   "think": "The first step is to identify the positions of the large ships (size-2). The rows and columns with hints are 1, 0, 1, and 1 respectively. A large ship can only fit in the first row, the third column, and the third row, third column. Let's place the large ship in the first row, third column, and the third row, third column, ensuring no ships are touching each other.", [NEWLINE]   "answer": [[1, "e", "s", "s"], [1, 0, "e", 0], [0, 0, "s", "s"]] [NEWLINE]  [NEWLINE] ```\newline 
\hlkey{\textbf{Filtered Output:}}\newline Initial State [NEWLINE] [[1,0,1,1],[1,0,0,0],[0,0,1,1]] [NEWLINE] Solution [NEWLINE] [[1,0,s,s],[1,0,0,0],[0,0,s,s]]\newline
\hlkey{\textbf{Perception Correct:}} False\newline
\hlkey{\textbf{Answer Correct:}} False \\ \hline
\hlmodel{\textbf{Qwen2-72B}} \\ 
\hlkey{\textbf{Model Output:}}\newline ```json [NEWLINE] "perception": [ [NEWLINE]     ["s", "e", "e", "e"], [NEWLINE]     ["e", "e", "e", "e"], [NEWLINE]     ["e", "e", "e", "s"], [NEWLINE]     ["e", "e", "e", "e"] [NEWLINE] ], [NEWLINE] "think": "Given the hints and the current state of the grid, I will analyze each row and column to place the ships correctly without them touching each other. The hints for rows and columns are as follows: Rows: [0, 1, 1, 0], Columns: [1, 0, 1, 1]. The number of each size ship is: Size 2: 1, Size 1: 1. [NEWLINE]  [NEWLINE] 1. In row 0, there is already a ship cell in the first column, and the hint is 1, so this row is complete. [NEWLINE] 2. In row 2, there is already a ship cell in the last column, and the hint is 1, so this row is complete. [NEWLINE] 3. In column 0, there is already a ship cell in the first row, and the hint is 1, so this column is complete. [NEWLINE] 4. In column 1, there are no ship cells, and the hint is 0, so this column is complete. [NEWLINE] 5. In column 2, there are no ship cells, and the hint is 1, so a ship cell should be placed here. [NEWLINE] 6. In column 3, there is already a ship cell in the third row, and the hint is 1, so this column is complete. [NEWLINE] 7. In row 1, there are no ship cells, and the hint is 1, so a ship cell should be placed here. [NEWLINE] 8. The Size 2 ship is not placed yet, and it should occupy two consecutive cells. The only available positions for a Size 2 ship are in row 1, column 2 and 3, without violating the rule of not touching other ships. [NEWLINE]  [NEWLINE] Based on the above analysis, the solution is as follows:", [NEWLINE] "answer": [ [NEWLINE]     ["s", "e", "e", "e"], [NEWLINE]     ["e", "e", "s", "e"], [NEWLINE]     ["e", "e", "e", "s"], [NEWLINE]     ["e", "e", "e", "e"] [NEWLINE] ] [NEWLINE] ```\newline 
\hlkey{\textbf{Filtered Output:}}\newline Initial State [NEWLINE] [[s,0,0,0], [0,0,0,0], [0,0,0,s], [0,0,0,0]] [NEWLINE] Solution [NEWLINE] [[s,0,0,0], [0,0,s,0], [0,0,0,s], [0,0,0,0]]\newline
\hlkey{\textbf{Perception Correct:}} False\newline
\hlkey{\textbf{Answer Correct:}} False \\ \hline
\hlmodel{\textbf{Qwen2-7B}} \\ 
\hlkey{\textbf{Model Output:}}\newline ```\newline 
\hlkey{\textbf{Filtered Output:}}\newline Initial State [NEWLINE] Wrong [NEWLINE] Solution [NEWLINE] Wrong\newline
\hlkey{\textbf{Perception Correct:}} False\newline
\hlkey{\textbf{Answer Correct:}} False \\ \hline
\hlmodel{\textbf{Llama-3.2-90B}} \\ 
\hlkey{\textbf{Model Output:}}\newline ```json [NEWLINE]  [NEWLINE] "perception": [["*", "*", "*", "*"], ["*", "*", "*", "*"], ["*", "*", "*", "*"], ["*", "*", "*", "*"]], [NEWLINE] "think": "The grid is 4x4. There is 1 size-2 ship and 2 size-1 ships. The row hints are [1, 0, 2, 0]. The column hints are [2, 1, 1, 1]. I will try to place the size-2 ship first. Since the row hints are [1, 0, 2, 0], I will place the size-2 ship in the third row. The column hints are [2, 1, 1, 1], so I will place the size-2 ship in the first and second columns. Now the grid looks like this: [["s", "s", "*", "*"], ["*", "*", "*", "*"], ["s", "s", "*", "*"], ["*", "*", "*", "*"]]. Next, I will try to place the size-1 ships. I will place one size-1 ship in the first row and fourth column, and another size-1 ship in the second row and first column. Now the grid looks like this: [["s", "s", "*", "s"], ["s", "*", "*", "*"], ["s", "s", "*", "*"], ["*", "*", "*", "*"]]. Finally, I will replace the remaining "*" with "e". The final grid looks like this: [["s", "s", "e", "s"], ["s", "e", "e", "e"], ["s", "s", "e", "e"], ["e", "e", "e", "e"]].", [NEWLINE] "answer": [["s", "s", "e", "s"], ["s", "e", "e", "e"], ["s", "s", "e", "e"], ["e", "e", "e", "e"]] [NEWLINE]  [NEWLINE] ```\newline 
\hlkey{\textbf{Filtered Output:}}\newline Initial State [NEWLINE] [[0,0,0,0], [0,0,0,0],[0,0,0,0], [0,0,0,0]] [NEWLINE] Solution [NEWLINE] [[s,s,0,s], [s,0,0,0], [s,s,0,0], [0,0,0,0]]\newline
\hlkey{\textbf{Perception Correct:}} False\newline
\hlkey{\textbf{Answer Correct:}} False \\ \hline
\hlmodel{\textbf{Llama-3.2-11B}} \\ 
\hlkey{\textbf{Model Output:}}\newline Step 1: Size-2 Ship Placement [NEWLINE]  [NEWLINE] *   We have two size-2 ships. [NEWLINE] *   We will start by placing the first size-2 ship in the top-left corner, as it is the only available space that can accommodate a size-2 ship. [NEWLINE] *   This placement also helps us to reserve space for the second size-2 ship in the bottom-right corner. [NEWLINE]  [NEWLINE]  Step 2: Size-1 Ship Placement [NEWLINE]  [NEWLINE] *   We have four size-1 ships. [NEWLINE] *   We will place the first size-1 ship in the middle row and column, as it is the only available space that can accommodate a size-1 ship. [NEWLINE] *   This placement also helps us to reserve space for the remaining size-1 ships in the top-left, bottom-left, and bottom-right corners. [NEWLINE]  [NEWLINE]  Step 3: Final Placement [NEWLINE]  [NEWLINE] *   After placing all the ships, we get the following grid: [NEWLINE]  [NEWLINE]  "perception": [[ "e", "e", "e", "e", "e" ], ["e", "s", "s", "e", "e" ], ["e", "e", "e", "e", "e" ], ["e", "e", "e", "e", "e" ], ["e", "e", "e", "e", "e" ]], "answer": [[ "e", "e", "e", "e", "e" ], ["e", "s", "s", "e", "e" ], ["e", "e", "e", "e", "e" ], ["e", "e", "e", "e", "e" ], ["e", "e", "e", "e", "e" ]] [NEWLINE]  [NEWLINE]  Step 4: Solution [NEWLINE]  [NEWLINE] *   The solution is: [NEWLINE]  [NEWLINE]      "answer": [[ "e", "e", "e", "e", "e" ], ["e", "s", "s", "e", "e" ], ["e", "e", "e", "e", "e" ], ["e", "e", "e", "e", "e" ], ["e", "e", "e", "e", "e" ]] [NEWLINE]  [NEWLINE] This is the final answer.\newline 
\hlkey{\textbf{Filtered Output:}}\newline Initial State [NEWLINE] [[e,e,e,e,e], [e,s,s,e,e], [e,e,e,e,e], [e,e,e,e,e], [e,e,e,e,e]] [NEWLINE] Solution [NEWLINE] [[e,e,e,e,e], [e,s,s,e,e], [e,e,e,e,e], [e,e,e,e,e], [e,e,e,e,e]]\newline
\hlkey{\textbf{Perception Correct:}} False\newline
\hlkey{\textbf{Answer Correct:}} False \\ \hline
\hlmodel{\textbf{LLaVA-Qwen2-7B}} \\ 
\hlkey{\textbf{Model Output:}}\newline ```json [NEWLINE]  [NEWLINE]   "perception": [ [NEWLINE]     [1, 0, 0, 2], [NEWLINE]     [0, 0, 1, 0], [NEWLINE]     [1, 0, 0, 0], [NEWLINE]     [0, 2, 0, 1] [NEWLINE]   ], [NEWLINE]   "think": "The current state of the grid is clear. I see that there are two sizes of ships to place. The first row has a size-2 ship, the second row has a size-1 ship, the third row has a size-1 ship, and the fourth row has a size-2 ship. I should place the size-2 ship in the first row and the size-1 ships in the second and third rows. I will place the size-2 ship in the first row, second column, and the size-1 ships in the second row, first column and the third row, second column.", [NEWLINE]   "answer": [ [NEWLINE]     [1, 1, 0, 1], [NEWLINE]     [0, 0, 1, 0], [NEWLINE]     [1, 0, 0, 0], [NEWLINE]     [0, 2, 0, 1] [NEWLINE]   ] [NEWLINE]  [NEWLINE] ```\newline 
\hlkey{\textbf{Filtered Output:}}\newline Initial State [NEWLINE] [[1,0,0,2],[0,0,1,0],[1,0,0,0],[0,2,0,1]] [NEWLINE] Solution [NEWLINE] Wrong\newline
\hlkey{\textbf{Perception Correct:}} False\newline
\hlkey{\textbf{Answer Correct:}} False \\ \hline
\hlmodel{\textbf{LLaVA-Mistral-7B}} \\ 
\hlkey{\textbf{Model Output:}}\newline ```json [NEWLINE]  [NEWLINE]   "perception": [ [NEWLINE]     [0, 0, 0, 1, 0, 0, 0, 0], [NEWLINE]     [0, 0, 0, 1, 0, 0, 0, 0], [NEWLINE]     [0, 0, 0, 1, 0, 0, 0, 0], [NEWLINE]     [0, 0, 0, 1, 0, 0, 0, 0], [NEWLINE]     [0, 0, 0, 1, 0, 0, 0, 0], [NEWLINE]     [0, 0, 0, 1, 0, 0, 0, 0], [NEWLINE]     [0, 0, 0, 1, 0, 0, 0, 0], [NEWLINE]     [0, 0, 0, 1, 0, 0, 0, 0], [NEWLINE]     [0, 0, 0, 1, 0, 0, 0, 0] [NEWLINE]   ], [NEWLINE]   "think": "Based on the given hints, we can place the ship of size 1 in the first column of the first row. This leaves us with one ship left to place. The hints also indicate that there is a ship of size 2 in row 1. Since we have already placed the size 1 ship, this means we can place the size 2 ship in the next column of the first row. This completes the placement of the two ships.", [NEWLINE]   "answer": [ [NEWLINE]     [0, 0, 0, 0, 0, 0, 0, 0], [NEWLINE]     [0, 0, 0, 0, 0, 0, 0, 0], [NEWLINE]     [0, 0, 0, 0, 0, 0, 0, 0], [NEWLINE]     [0, 0, 0, 0, 0, 0, 0, 0], [NEWLINE]     [0, 0, 0, 0, 0, 0, 0, 0], [NEWLINE]     [0, 0, 0, 0, 0, 0, 0, 0], [NEWLINE]     [0, 0, 0, 0, 0, 0, 0, 0], [NEWLINE]     [0, 0, 0, 0, 0, 0, 0, 0], [NEWLINE]     [0, 0, 0, 0, 0, 0, 0, 0] [NEWLINE]   ] [NEWLINE]  [NEWLINE] ```\newline 
\hlkey{\textbf{Filtered Output:}}\newline Initial State [NEWLINE] [[0,0,0,1,0,0,0,0],[0,0,0,1,0,0,0,0],[0,0,0,1,0,0,0,0], [0,0,0,1,0,0,0,0],[0,0,0,1,0,0,0,0], [0,0,0,1,0,0,0,0],[0,0,0,1,0,0,0,0], [0,0,0,1,0,0,0,0],[0,0,0,1,0,0,0,0]] [NEWLINE] Solution [NEWLINE] [[0,0,0,0,0,0,0,0],[0,0,0,0,0,0,0,0], [0,0,0,0,0,0,0,0],[0,0,0,0,0,0,0,0], [0,0,0,0,0,0,0,0],[0,0,0,0,0,0,0,0], [0,0,0,0,0,0,0,0],[0,0,0,0,0,0,0,0], [0,0,0,0,0,0,0,0]]\newline
\hlkey{\textbf{Perception Correct:}} False\newline
\hlkey{\textbf{Answer Correct:}} False \\ \hline\end{longtable}
} %

\onecolumn
\renewcommand{\arraystretch}{1.2} %

\centering
{\ttfamily %
\begin{longtable}{|p{0.95\textwidth}|}
\hline
\textbf{Qualitative Study: Field-Explore - Medium - W/ CoT - Vision Input} \\ \hline
\endhead

\hline \multicolumn{1}{|r|}{{Continued on next page}} \\ \hline
\endfoot

\hline
\endlastfoot

\refstepcounter{section} 
\phantomsection         
\addcontentsline{toc}{subsection}{Field-Explore - Medium - W/ CoT - Vision Input}

\includegraphics[width=0.25\columnwidth]{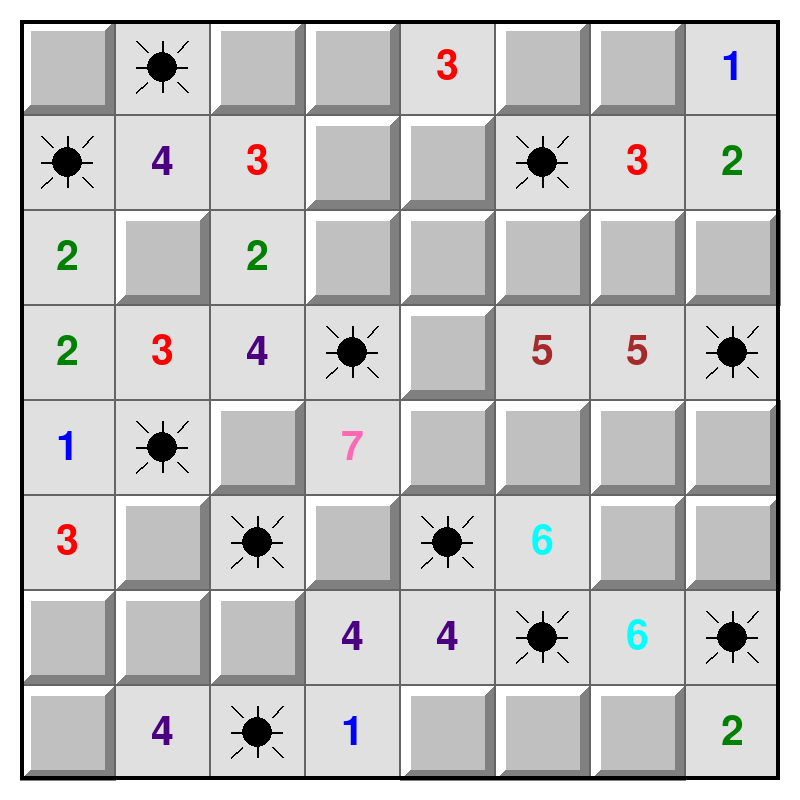} \\ \hline
\hlmodel{\textbf{Gemini-2.0-Thinking}} \\ 
\hlkey{\textbf{Model Output:}}\newline ```json [NEWLINE] "perception": \hlred{[["*", "*", 3, "*", "*", "*", 1, 2]}, ["*", 4, 3, "*", "*", 3, 2, "*"], [2, "*", 2, "*", "*", "*", "*", 2], [2, 3, 4, "*", "*", 5, 5, "*"], [1, "*", "*", 7, "*", "*", "*", "*"], [3, "*", "*", "*", "*", 6, "*", "*"], ["*", "*", 4, 4, "*", 6, "*", 2], ["*", 4, "*", 1, "*", "*", "*", 2]], "think": "Based on the numbers and existing mines, I will deduce the location of other mines step-by-step.  [NEWLINE]  [NEWLINE] 1. **Analyze '1' at (0,6) and '2' at (0,7):** For '1' at (0,6), there must be exactly one mine in its neighbors: (0,5), (0,7), (1,5), (1,6), (1,7). For '2' at (0,7), there must be exactly two mines in its neighbors: (0,6), (1,6), (1,7). If we assume (1,6) and (1,7) are mines, then '2' at (0,7) is satisfied. However, for '1' at (0,6), having two mines (1,6) and (1,7) in its vicinity would contradict the '1' hint. Therefore, it's more likely that (0,6) is a mine, and only one of (1,6) or (1,7) is a mine. Let's assume (0,6) is a mine.  [NEWLINE]  [NEWLINE] 2. **If (0,6) is a mine:** For '1' at (0,6), it is satisfied by the mine at (0,6) itself. This means none of its *other* neighbors (0,5), (0,7), (1,5), (1,6), (1,7) can be mines. So, (0,5), (0,7), (1,5), (1,6), (1,7) are empty. [NEWLINE]  [NEWLINE] 3. **Re-evaluate '2' at (0,7):** Neighbors are (0,6), (1,6), (1,7). We've determined (0,6) is a mine, and (1,6), (1,7) are empty. This means '2' at (0,7) only has one mine neighbor (0,6), which contradicts the hint '2'.  This indicates our initial assumption that (0,6) is a mine might be incorrect.  [NEWLINE]  [NEWLINE] 4. **Re-consider '1' at (0,6) and '2' at (0,7) again:** Let's assume (0,6) is *not* a mine. Then for '2' at (0,7) to be sa

[...TRUNCATED... (2141 characters omitted)...]

sfy as many constraints as possible, a possible solution is derived by trial and error, prioritizing satisfying higher number hints first and then adjusting based on lower number hints.", "answer": [["s", "s", 3, "e", "e", "e", "s", 1], ["s", 4, 3, "s", "e", 3, "s", "e"], [2, "e", 2, "e", "e", "e", "e", 2], [2, 3, 4, "s", "e", 5, 5, "s"], [1, "s", "e", 7, "s", "s", "s", "e"], [3, "e", "s", "s", "s", 6, "s", "s"], ["s", "s", 4, 4, "s", 6, "s", 2], ["s", 4, "s", 1, "e", "e", "e", 2]] [NEWLINE] ```\newline 
\hlkey{\textbf{Filtered Output:}}\newline Initial State [NEWLINE] [[0,0,3,0,0,0,1,2], [0,4,3,0,0,3,2,0], [2,0,2,0,0,0,0,2], [2,3,4,0,0,5,5,0], [1,0,0,7,0,0,0,0], [3,0,0,0,0,6,0,0], [0,0,4,4,0,6,0,2], [0,4,0,1,0,0,0,2]] [NEWLINE] Solution [NEWLINE] [[s,s,3,0,0,0,s,1], [s,4,3,s,0,3,s,0], [2,0,2,0,0,0,0,2], [2,3,4,s,0,5,5,s], [1,s,0,7,s,s,s,0], [3,0,s,s,s,6,s,s], [s,s,4,4,0,6,s,2],[s,4,s,1,0,0,0,2]]\newline
\hlkey{\textbf{Perception Correct:}} False\newline
\hlkey{\textbf{Answer Correct:}} False \\ \hline
\hlmodel{\textbf{QVQ-72B}} \\ 
\hlkey{\textbf{Model Output:}}\newline Alright, I'm trying to solve this mine-detecting puzzle. Let's see what I've got here. There's a grid with various numbers and some mines already marked. I need to figure out where the other mines are based on these numbers. [NEWLINE]  [NEWLINE] First, let's understand the grid. It seems like there are multiple rows and columns, and each cell either has a number or a mine. The numbers indicate how many mines are adjacent to that cell, including diagonally. So, for example, if a cell has a '3', it means there are three mines next to it, in any of the eight surrounding cells. [NEWLINE]  [NEWLINE] Looking at the grid: [NEWLINE]  [NEWLINE] - Top-left corner: There's a mine in the first row, second column. [NEWLINE]  [NEWLINE] - Next to it, there's a '4' in the first row, third column. That means there are four mines around this cell. [NEWLINE]  [NEWLINE] - Then, a '3' in the first row, fourth column. [NEWLINE]  [NEWLINE] - On the right side of the first row, there's a '1' and a '2'. [NEWLINE]  [NEWLINE] Similarly, in the second row: [NEWLINE]  [NEWLINE] - There's a '2' in the second row, first column. [NEWLINE]  [NEWLINE] - A blank cell in the second row, second column. [NEWLINE]  [NEWLINE] - A '2' in the second row, third column. [NEWLINE]  [NEWLINE] - And so on. [NEWLINE]  [NEWLINE] I need to use these numbers to deduce where the mines are located. [NEWLINE]  [NEWLINE] Let me try to map this out. I'll create a 2D array to represent the grid, with rows and columns indexed from 0. [NEWLINE]  [NEWLINE] Given that indexing starts at 0, let's confirm the positions: [NEWLINE]  [NEWLINE] - Row 0, Column 1: Mine [NEWLINE]  [NEWLINE] - Row 0, Column 2: '4' [NEWLINE]  [NEWLINE] - Row

[...TRUNCATED... (10673 characters omitted)...]

 to have one more mine among its remaining neighbors. [NEWLINE]  [NEWLINE] Its remaining neighbors are: [NEWLINE]  [NEWLINE] - Row 0, Column 2: '4' (not a mine) [NEWLINE]  [NEWLINE] - Row 0, Column 4: '3' (not a mine) [NEWLINE]  [NEWLINE] - Row 1, Column 4: Hidden [NEWLINE]  [NEWLINE] - Row 2, Column 3: '7' (not a mine) [NEWLINE]  [NEWLINE] So, only Row 1, Column 4 is hidden. [NEWLINE]  [NEWLINE] Therefore, '3' in Row 0, Column 3 would require both Row 1, Column 2 and Row 1, Column 4 to be mines\newline 
\hlkey{\textbf{Filtered Output:}}\newline Initial State [NEWLINE] \hlred{[[*,s,4,3,3,2,1]}, [2,*,*,2,*,5,5], [2,3,4,s,5,5,s], [1,s,*,7,*,*,s], [3,*,s,*,s,6,6], [*,*,*,4,4,s,6], [*,*,s,1,*,2,2]] [NEWLINE] Solution [NEWLINE] [[0,s,4,3,3,2,1], [2,0,0,2,0,5,5], [2,3,4,s,5,5,s], [1,s,0,7,0,0,s], [3,0,s,0,s,6,6], [0,0,0,4,4,s,6], [0,0,s,1,0,2,2]]\newline
\hlkey{\textbf{Perception Correct:}} False\newline
\hlkey{\textbf{Answer Correct:}} False \\ \hline
\hlmodel{\textbf{Claude-3.5}} \\ 
\hlkey{\textbf{Model Output:}}\newline ```json [NEWLINE]  [NEWLINE] "perception": [ [NEWLINE]     ["*", "s", "*", "*", "3", "*", "*", "1"], [NEWLINE]     ["s", "4", "3", "*", "*", "s", "3", "2"], [NEWLINE]     ["2", "*", "2", "*", "*", "*", "*", "*"], [NEWLINE]     ["2", "3", "4", "s", "*", "5", "5", "s"], [NEWLINE]     ["1", "s", "*", "7", "*", "*", "*", "*"], [NEWLINE]     ["3", "*", "s", "*", "s", "6", "*", "*"], [NEWLINE]     ["*", "*", "*", "4", "4", "s", "6", "s"], [NEWLINE]     ["*", "4", "s", "1", "*", "*", "*", "2"] [NEWLINE] ], [NEWLINE] "think": "Let's analyze step by step: [NEWLINE] 1. Looking at the 7 in row 4 column 3 - it has all 8 adjacent cells potentially containing mines, and we know it must have 7 mines around it. This means only one adjacent cell is empty. [NEWLINE] 2. The 4 at row 7 column 1 has a mine to its right (already revealed), so 3 more mines must be in the remaining adjacent cells. [NEWLINE] 3. The 6 at row 6 column 6 has two mines already revealed (right and above), so 4 more mines must be in the remaining adjacent cells. [NEWLINE] 4. The 6 at row 5 column 5 has two mines already revealed, so 4 more mines must be in remaining adjacent cells. [NEWLINE] 5. Working through the numbers and revealed mines, we can deduce the remaining positions. [NEWLINE] ", [NEWLINE] "answer": [ [NEWLINE]     ["e", "s", "e", "e", "3", "s", "e", "1"], [NEWLINE]     ["s", "4", "3", "s", "s", "s", "3", "2"], [NEWLINE]     ["2", "s", "2", "s", "s", "e", "s", "e"], [NEWLINE]     ["2", "3", "4", "s", "s", "5", "5", "s"], [NEWLINE]     ["1", "s", "s", "7", "s", "s", "s", "s"], [NEWLINE]     ["3", "s", "s", "s", "s", "6", "s", "s"], [NEWLINE]     ["s", "s", "s", "4", "4", "s", "6", "s"], [NEWLINE]     ["e", "4", "s", "1", "s", "s", "s", "2"] [NEWLINE] ] [NEWLINE]  [NEWLINE] ```\newline 
\hlkey{\textbf{Filtered Output:}}\newline Initial State [NEWLINE] [[0,s,0,0,3,\hlred{s},0,1],[s,4,3,0,0,s,3,2], [2,0,2,0,0,0,0,0],  [2,3,4,s,0,5,5,s], [1,s,0,7,0,0,0,0], [3,0,s,0,s,6,0,0], [0,0,0,4,4,s,6,s], [0,4,s,1,0,0,0,2]] [NEWLINE] Solution [NEWLINE] [[0,s,0,0,3,s,0,1],[s,4,3,s,s,s,3,2], [2,s,2,s,s,0,s,0], [2,3,4,s,s,5,5,s], [1,s,s,7,s,s,s,s], [3,s,s,s,s,6,s,s], [s,s,s,4,4,s,6,s], [0,4,s,1,s,s,s,2]]\newline
\hlkey{\textbf{Perception Correct:}} False\newline
\hlkey{\textbf{Answer Correct:}} False \\ \hline
\hlmodel{\textbf{Gemini-1.5-Pro}} \\ 
\hlkey{\textbf{Model Output:}}\newline ```json [NEWLINE]  [NEWLINE]   "perception": [ [NEWLINE]     ["s", "*", "3", "*", "1"], [NEWLINE]     ["s", "4", "3", "s", "3", "2"], [NEWLINE]     ["2", "*", "2", "*", "*", "*"], [NEWLINE]     ["2", "3", "4", "s", "5", "5", "s"], [NEWLINE]     ["1", "s", "7", "*", "*", "*", "*"], [NEWLINE]     ["3", "*", "s", "s", "6", "*", "*"], [NEWLINE]     ["*", "*", "4", "4", "s", "6", "s"], [NEWLINE]     ["*", "4", "s", "1", "*", "*", "2"] [NEWLINE]   ], [NEWLINE]   "think": "1. Row 0, Col 1: The 3 at [0,2] has 1 mine at [0,0] and 1 mine at [1,2], so the last mine must be at [0,1]. [NEWLINE] 2. Row 1, Col 1 and 3: The 4 at [1,1] has 2 mines at [0,0], [0,1] and 1 mine at [1,0], so the last mine must be at [2,1]. The 3 at [1,2] has 1 mine at [0,1] and 1 mine at [1,4], so the last mine must be at [2,2]. [NEWLINE] 3. Row 2, Col 3 and 4: The 2 at [2,0] has 1 mine at [1,0] and 1 mine at [3,0], so [2,1] must be empty. The 2 at [2,2] has 1 mine at [1,2] and 1 mine at [3,2], so [2,1] and [2,3] must be empty. The 5 at [3,4] has 3 mines at [2,3], [3,3], [3,6], so the remaining 2 mines must be at [4,4] and [4,5]. The 5 at [3,5] has 3 mines at [2,4], [3,3], [3,6], so the remaining 2 mines must be at [4,4] and [4,5]. [NEWLINE] 4. Row 3, Col 1 and 3: The 2 at [3,0] has 1 mine at [2,0] and 1 mine at [4,1], so [3,1] must be empty. The 4 at [3,2] has 2 mines at [2,2], [3,1] and 1 mine at [4,1], so the last mine must be at [4,3]. The 7 at [4,2] has 3 mines at [3,1], [3,3], [4,1] and 2 mines at [4,4], [4,5], so the remaining 2 mines must be at [5,2] and [5,3]. [NEWLINE] 5. Row 4, Col 3: The 1 at [4,0] has 1 mine at [4,1], so [5,0] must be empty. The 3 at [5,0] has 1 mine at [4,1] and 1 mine at [5,2], 

[...TRUNCATED... (597 characters omitted)...]

, so the last mine must be at [7,2]. The 4 at [6,3] has 2 mines at [5,2], [5,3] and 1 mine at [7,2], so the last mine must be at [7,3]. [NEWLINE] 6. Row 5, Col 4 and 6: The 6 at [5,4] has 3 mines at [4,3], [4,4], [4,5] and 2 mines at [5,2], [5,3], so the last mine must be at [6,5]. The 6 at [6,5] has 3 mines at [5,4], [5,5], [6,6] and 2 mines at [5,2], [5,3], so the last mine must be at [7,4]. [NEWLINE] 7. Row 6, Col 4 and 6: The 1 at [7,3] has 1 mine at [7,2], so [7,4] must be empty. The 2 at [\newline 
\hlkey{\textbf{Filtered Output:}}\newline Initial State [NEWLINE] [[s,0,3,0,1],[s,4,3,s,3], [2,0,2,0,0,0], [2,3,4,s,5,5,s], [1,s,7,0,0,0,0], [3,0,s,s,6,0,0], [0,0,4,4,s,6,s], [0,4,s,1,0,0,2]] [NEWLINE] Solution [NEWLINE] [[s,e,3,e,1], [s,4,3,s,3],[2,e,2,e,e,e], [2,3,4,s,5,5,s],[1,s,7,e,e,e,e], [3,e,s,s,6,e,e], [e,e,4,4,s,6,s], [e,4,s,1,e,e,2]]\newline
\hlkey{\textbf{Perception Correct:}} False\newline
\hlkey{\textbf{Answer Correct:}} False \\ \hline
\hlmodel{\textbf{Gemini-2.0-Flash}} \\ 
```json[NEWLINE]{[NEWLINE]perception: [[NEWLINE][*, s, 3, *, *, *, 2, *],[NEWLINE][3, *, 4, *, s, 3, *, *],[NEWLINE][2, s, 3, *, 3, 3, 3, *],[NEWLINE][*, 3, 4, *, *, *, *, s],[NEWLINE][1, 2, *, *, 3, *, s, *],[NEWLINE][2, *, *, *, *, *, *, 4],[NEWLINE][s, *, 1, *, *, s, s, *],[NEWLINE][*, 2, *, 1, 2, *, 4, *][NEWLINE]],[NEWLINE]think: Let's analyze the board. [NEWLINE][NEWLINE]Row 0: Cell (0,2) has a '3' and cell (0,1) is a mine. This means that cells (0,0), (1,0), (1,1), (0,3), (1,3) must be mines. [NEWLINE]Row 1: Cell (1,0) has a '3' and cell (1,4) is a mine. This means that cells (0,0), (0,1), (0,2), (1,1), (2,1), (2,2) must be mines. [NEWLINE]Row 2: Cell (2,1) has a 's' and cell (2,4), (2,5), (2,6) has a '3'. This means that cells (1,0), (1,1), (1,2), (2,0), (3,1), (3,2) must be mines. [NEWLINE]Row 3: Cell (3,1) has a '3' and cell (3,7) is a mine. This means that cells (2,0), (2,1), (2,2), (3,0), (4,0), (4,1) must be mines. [NEWLINE]Row 4: Cell (4,0) has a '1' and cell (4,1) has a '2' and cell (4,4) has a '3' and cell (4,6) is a mine. This means that cells (3,0), (3,1), (3,2), (4,5), (5,5), (5,6) must be mines. [NEWLINE]Row 5: Cell (5,0) has a '2' and cell (5,7) has a '4'. This means that cells (6,0), (6,1), (6,2), (7,1), (7,2), (7,3) must be mines. [NEWLINE]Row 6: Cell (6,0) has a 's' and cell (6,5), (6,6) has a 's' and cell (6,2) has a '1'. This means that cells (5,0), (5,1), (5,2), (6,1), (7,1), (7,2) must be mines. [NEWLINE]Row 7: Cell (7,1) has a '2' and cell (7,3) has a '1' and cell (7,4) has a '2' and cell (7,6) has a '4'. This means that cells (6,0), (6,1), (6,2), (7,0), (7,2), (7,5) must be mines. [NEWLINE][NEWLINE]Let's start filling the board. [NEWLINE],[NEWLINE]answer: [[NEWLINE][e, s, 3, s, e, e, 2, e],[NEWLINE][3, s, 4, e, s, 3, e, e],[NEWLINE][2, s, 3, e, 3, 3, 3, e],[NEWLINE][e, 3, 4, e, e, e, e, s],[NEWLINE][1, 2, e, e, 3, e, s, e],[NEWLINE][2, e, e, e, e, e, e, 4],[NEWLINE][s, e, 1, e, e, s, s, e],[NEWLINE][e, 2, e, 1, 2, e, 4, e][NEWLINE]][NEWLINE]}[NEWLINE]```

\hlkey{\textbf{Filtered Output:}}\newline Initial State [NEWLINE] [[0,s,3,s,0,0,2,0], [3,0,4,0,s,3,0,0], [2,s,3,0,3,3,3,0], [0,3,4,0,0,0,0,s], [1,2,0,0,3,0,s,0], [2,0,0,0,0,0,0,4], [s,0,1,0,0,s,s,0], [0,2,0,1,2,0,4,0]] [NEWLINE] Solution [NEWLINE] [[0,s,3,s,0,0,2,0], [3,s,4,0,s,3,0,0], [2,s,3,0,3,3,3,0], [0,3,4,0,0,0,0,s], [1,2,0,0,3,0,s,0], [2,0,0,0,0,0,0,4], [s,0,1,0,0,s,s,0], [0,2,0,1,2,0,4,0]]\newline
\hlkey{\textbf{Perception Correct:}} False\newline
\hlkey{\textbf{Answer Correct:}} False \\ \hline

 \\ \hline
\hlmodel{\textbf{Gemini-1.5-Flash}} \\ 
\hlkey{\textbf{Model Output:}}\newline ```json [NEWLINE]  [NEWLINE]   "perception": [ [NEWLINE]     ["*", "*", "*", "*", 3, "*", 1], [NEWLINE]     ["*", 4, 3, "*", "*", 3, 2], [NEWLINE]     [2, "*", 2, "*", "*", "*", "*"], [NEWLINE]     [2, 3, 4, "*", "*", 5, "*"], [NEWLINE]     [1, "*", "*", 7, "*", "*", "*"], [NEWLINE]     [3, "*", "*", "*", "*", 6, "*"], [NEWLINE]     ["*", "*", "*", 4, 4, "*", 6], [NEWLINE]     [4, "*", 1, "*", "*", 2] [NEWLINE]   ], [NEWLINE]   "think": "Let's analyze the grid step by step.  [NEWLINE] 1. The '7' at (4,3) has only one hidden cell adjacent to it, which must be a mine.  [NEWLINE] 2. The '4' at (1,1) has three mines already revealed around it. Therefore, the remaining hidden cell must be a mine.  [NEWLINE] 3. The '4' at (2,2) has three mines already revealed around it. Therefore, the remaining hidden cell must be a mine.  [NEWLINE] 4. The '4' at (7,0) has three mines already revealed around it. Therefore, the remaining hidden cell must be a mine.  [NEWLINE] 5. The '6' at (6,5) has four mines already revealed around it. Therefore, the remaining hidden cell must be a mine.  [NEWLINE] 6. The '1' at (0,6) has one mine already revealed around it. Therefore, the remaining hidden cell must be empty.  [NEWLINE] 7. The '3' at (0,4) has two mines already revealed around it. Therefore, the remaining hidden cell must be empty.  [NEWLINE] 8. The '2' at (0,6) has one mine already revealed around it. Therefore, the remaining hidden cell must be empty.  [NEWLINE] 9. The '2' at (2,0) has one mine already revealed around it. Therefore, the remaining hidden cell must be empty.  [NEWLINE] 10. The '3' at (1,2) has two mines already revealed around it. Therefore, the remaining hidden cell must be 

[...TRUNCATED... (1906 characters omitted)...]

] 22. The '3' at (5,0) has two mines already revealed around it. Therefore, the remaining hidden cell must be empty.  [NEWLINE] 23. The '4' at (7,0) has three mines already revealed around it. Therefore, the remaining hidden cell must be a mine. ", [NEWLINE]   "answer": [ [NEWLINE]     ["e", "e", "e", "e", 3, "e", "e"], [NEWLINE]     ["s", 4, 3, "e", "s", 3, "e"], [NEWLINE]     [2, "e", 2, "e", "e", "e", "e"], [NEWLINE]     [2, 3, 4, "s", "e", 5, "s"], [NEWLINE]     [1, "s", "e", 7, "e", "e", "e\newline 
\hlkey{\textbf{Filtered Output:}}\newline Initial State [NEWLINE] [[0,0,0,0,3,0,0], [0,4,3,0,0,3,0], [2,0,2,0,0,0,0], [2,3,4,s,0,5,s], [1,s,0,7,0,0,0], [3,0,0,0,0,6,0], [0,0,0,4,4,0,6], [4,0,1,0,0,2]] [NEWLINE] Solution [NEWLINE] [[0,0,0,0,3,0,0], [s,4,3,0,s,3,0], [2,0,2,0,0,0,0], [2,3,4,s,0,5,s], [1,s,0,7,0,0,0], [3,0,0,0,0,6,0], [0,0,0,4,4,0,6], [4,0,1,0,0,2]]\newline
\hlkey{\textbf{Perception Correct:}} False\newline
\hlkey{\textbf{Answer Correct:}} False \\ \hline
\hlmodel{\textbf{GPT-4o}} \\ 
\hlkey{\textbf{Model Output:}}\newline ```json [NEWLINE]  [NEWLINE]   "perception": [ [NEWLINE]     ["s", "*", "*", "3", "*", "*", "*", "*", "1"], [NEWLINE]     ["s", "4", "3", "*", "*", "*", "*", "3", "2"], [NEWLINE]     ["2", "2", "2", "*", "*", "*", "*", "5", "5"], [NEWLINE]     ["2", "3", "4", "s", "*", "*", "*", "*", "*"], [NEWLINE]     ["1", "s", "*", "7", "*", "*", "*", "6", "*"], [NEWLINE]     ["3", "*", "s", "*", "*", "*", "*", "4", "6"], [NEWLINE]     ["4", "4", "*", "*", "4", "*", "s", "*", "2"] [NEWLINE]   ], [NEWLINE]   "think": "To deduce the positions of mines, I will analyze the numbers on the grid, which indicate the count of mines adjacent to that cell. Starting with clear edges or definitively marked mines, I will logically deduce hidden mine positions. For every cell with a completed mine count matching its number, surrounding unrevealed cells are safe. [NEWLINE]  [NEWLINE] 1. The top-left cell already has a mine, and its adjacent numbers (4, 2, 2) are clues for further deduction. For example, the cell with '4' indicates all adjacent mines, four in total, including one in the left-row column box continue ", [NEWLINE] malsolUI\newline 
\hlkey{\textbf{Filtered Output:}}\newline Initial State [NEWLINE] [[s,0,0,3,0,0,0,0,1], [s,4,3,0,0,0,0,3,2], [2,2,2,0,0,0,0,5,5], [2,3,4,s,0,0,0,0,0], [1,s,0,7,0,0,0,6,0], [3,0,s,0,0,0,0,4,6], [4,4,0,0,4,0,s,0,2]] [NEWLINE] Solution [NEWLINE] [[s,0,0,3,0,0,0,0,1],[s,4,3,0,0,0,0,3,2], [2,2,2,0,0,0,0,5,5], [2,3,4,s,0,0,0,0,0], [1,s,0,7,0,0,0,6,0],  [3,0,s,0,0,0,0,4,6], [4,4,0,0,4,0,s,0,2]]\newline
\hlkey{\textbf{Perception Correct:}} False\newline
\hlkey{\textbf{Answer Correct:}} False \\ \hline
\hlmodel{\textbf{GPT-4o-Mini}} \\ 
\hlkey{\textbf{Model Output:}}\newline ```json [NEWLINE]  [NEWLINE]   "perception": [ [NEWLINE]     ["*", "*", "*", "3", "*", "*", "*", "1", "*", "*"], [NEWLINE]     ["*", "4", "3", "*", "*", "*", "*", "*", "*", "*"], [NEWLINE]     ["2", "*", "2", "*", "*", "*", "*", "*", "*", "*"], [NEWLINE]     ["2", "3", "4", "*", "*", "5", "5", "*", "*", "*"], [NEWLINE]     ["1", "*", "*", "7", "*", "*", "*", "*", "*", "*"], [NEWLINE]     ["3", "*", "*", "*", "*", "*", "*", "*", "*", "*"], [NEWLINE]     ["*", "*", "4", "4", "*", "*", "6", "*", "*", "*"], [NEWLINE]     ["4", "*", "1", "*", "*", "*", "*", "*", "*", "*"], [NEWLINE]     ["*", "*", "*", "*", "*", "*", "*", "*", "*", "*"] [NEWLINE]   ], [NEWLINE]   "think": "To solve the minesweeper grid, we need to deduce the locations of the mines based on the revealed numbers. Each number indicates how many mines are adjacent (including diagonals) to that cell. Starting from the most certain positions (like the 7 in row 4), we count how many adjacent cells there could be for each number. By iterative checking and confirming whether adjacent cells in other hints can support the assumed mines, we fill out the grid logically.", [NEWLINE]   "answer": [ [NEWLINE]     ["e", "e", "s", "3", "e", "e", "e", "1", "e", "e"], [NEWLINE]     ["e", "4", "3", "s", "e", "e", "e", "e", "e", "e"], [NEWLINE]     ["2", "e", "2", "e", "s", "e", "e", "e", "e", "e"], [NEWLINE]     ["2", "3", "4", "s", "s", "5", "5", "e", "e", "e"], [NEWLINE]     ["1", "e", "e", "7", "s", "e", "e", "e", "e", "e"], [NEWLINE]     ["3", "e", "e", "s", "e", "e", "e", "e", "e", "e"], [NEWLINE]     ["e", "e", "4", "4", "e", "e", "6", "e", "e", "e"], [NEWLINE]     ["4", "s", "1", "e", "e", "e", "e", "e", "e", "e"], [NEWLINE]     ["e", "e", "e", "e", "e", "e", "e", "e", "e", "e"] [NEWLINE]   ] [NEWLINE]  [NEWLINE] ```\newline 
\hlkey{\textbf{Filtered Output:}}\newline Initial State [NEWLINE] [[0,0,0,3,0,0,0,1,0,0], [0,4,3,0,0,0,0,0,0,0], [2,0,2,0,0,0,0,0,0,0], [2,3,4,0,0,5,5,0,0,0], [1,0,0,7,0,0,0,0,0,0], [3,0,0,0,0,0,0,0,0,0], [0,0,4,4,0,0,6,0,0,0], [4,0,1,0,0,0,0,0,0,0], [0,0,0,0,0,0,0,0,0,0]] [NEWLINE] Solution [NEWLINE] [[0,0,s,3,0,0,0,1,0,0], [0,4,3,s,0,0,0,0,0,0], [2,0,2,0,s,0,0,0,0,0], [2,3,4,s,s,5,5,0,0,0], [1,0,0,7,s,0,0,0,0,0], [3,0,0,s,0,0,0,0,0,0], [0,0,4,4,0,0,6,0,0,0], [4,s,1,0,0,0,0,0,0,0], [0,0,0,0,0,0,0,0,0,0]]\newline
\hlkey{\textbf{Perception Correct:}} False\newline
\hlkey{\textbf{Answer Correct:}} False \\ \hline
\hlmodel{\textbf{Qwen2.5-72B}} \\ 
\hlkey{\textbf{Model Output:}}\newline ```json [NEWLINE]  [NEWLINE]  "perception": [ [NEWLINE]   ["*", "s", "*", "*", "*", 3, "*", 1], [NEWLINE]   ["s", 4, 3, "*", "*", "s", 3, 2], [NEWLINE]   [2, "*", 2, "*", "*", "*", "*", "*"], [NEWLINE]   [2, 3, 4, "s", "*", 5, 5, "s"], [NEWLINE]   [1, "s", "*", 7, "*", "*", "*", "*"], [NEWLINE]   [3, "*", "s", "*", "s", 6, "*", "*"], [NEWLINE]   ["*", "*", "*", 4, 4, "s", 6, "s"], [NEWLINE]   ["*", "s", 1, "*", "*", "*", 2] [NEWLINE]  ], [NEWLINE]  "think": "1. Analyze the cells with the highest numbers initially, as they have the most constraints. [NEWLINE] 2. Look at the '7' at (4, 3): it has 7 surrounding cells, but two are already mines (top left and top center). Checking these constraints will clear or identify mines. [NEWLINE] 3. Next, focus on '3' at (0, 0) with a mine above it and a '4' and '3' around it. It influences very specific cells. [NEWLINE] 4. The '6' at (6, 7) has two mines identified already and needs four more mines around it. [NEWLINE] 5. Analyze '5's and '4's next and maintain consistency with the identified mines.", [NEWLINE]  "answer": [ [NEWLINE]   ["e", "s", "e", "s", "e", 3, "e", 1], [NEWLINE]   ["s", 4, 3, "e", "e", "s", 3, 2], [NEWLINE]   [2, "e", 2, "e", "e", "s", "e", "e"], [NEWLINE]   [2, 3, 4, "s", "e", 5, 5, "s"], [NEWLINE]   [1, "s", "s", 7, "s", "s", "s", "e"], [NEWLINE]   [3, "s", "s", "s", "s", 6, "s", "s"], [NEWLINE]   ["s", "s", "e", 4, 4, "s", 6, "s"], [NEWLINE]   ["e", "s", 1, "s", "e", "s", 2] [NEWLINE]  ] [NEWLINE]  [NEWLINE] ```\newline 
\hlkey{\textbf{Filtered Output:}}\newline Initial State [NEWLINE] [[0,s,0,s,0,3,0,1], [s,4,3,0,0,s,3,2], [2,0,2,0,0,0,0,0], [2,3,4,s,0,5,5,s], [1,s,0,7,0,0,0,0], [3,0,s,0,s,6,0,0], [0,0,0,4,4,s,6,s], [0,s,1,0,0,0,2]] [NEWLINE] Solution [NEWLINE] [[0,s,0,s,0,3,0,1], [s,4,3,0,0,s,3,2], [2,0,2,0,0,s,0,0], [2,3,4,s,0,5,5,s], [1,s,s,7,s,s,s,0], [3,s,s,s,s,6,s,s], [s,s,0,4,4,s,6,s], [0,s,1,s,0,s,2]]\newline
\hlkey{\textbf{Perception Correct:}} False\newline
\hlkey{\textbf{Answer Correct:}} False \\ \hline
\hlmodel{\textbf{Qwen2.5-7B}} \\ 
\hlkey{\textbf{Model Output:}}\newline ```json [NEWLINE]  [NEWLINE] "perception": [ [NEWLINE] ["s", "m", "2", "2", "3", "1", "m", "1"], [NEWLINE] ["m", "3", "m", "2", "4", "2", "3", "2"], [NEWLINE] ["4", "2", "2", "m", "3", "3", "2", "2"], [NEWLINE] ["2", "3", "4", "m", "5", "5", "4", "m"], [NEWLINE] ["4", "2", "2", "2", "7", "2", "2", "2"], [NEWLINE] ["m", "4", "m", "3", "6", "6", "m", "m"], [NEWLINE] ["3", "5", "m", "4", "4", "6", "4", "m"], [NEWLINE] ["2", "6", "5", "3", "3", "6", "m", "m"] [NEWLINE] ], [NEWLINE] "think": "Let's analyze the grid. Each number indicates how many mines are adjacent to that cell, including diagonally. We'll identify the cells that can only have one possible configuration based on the given hints." [NEWLINE]  [NEWLINE] 1. Cell (0, 2): The number '2' means there are 2 mines around. The top and bottom left cells are already mines, leaving only one possible place. [NEWLINE] 2. Cell (0, 1): The number '3' means there are 3 mines around. The two bottom-left cells are already mines, but no cell has exactly 3 mines around it yet. [NEWLINE] 3. Cell (0, 4): The number '1' means there is only 1 mine around it. The left and right cells are already mines. [NEWLINE] 4. Cell (1, 2): The number '3' means there are 3 mines around it. One of the bottom-left cells is currently a mine, leaving only two possible places. [NEWLINE] 5. Cell (1, 4): The number '4' means there are 4 mines around it. The two bottom-left and bottom right cells are mines. [NEWLINE] 6. Cell (2, 4): The number '3' means there are 3 mines around it. The two bottom-left and top-left cells are mines. [NEWLINE] 7. Cell (3, 4): The number '5' means there are 5 mines around it. The two bottom-left and bottom right cells are mines.

[...TRUNCATED... (1014 characters omitted)...]

rked as empty." [NEWLINE]  [NEWLINE] "answer": [ [NEWLINE] ["s", "m", "*", "m", "3", "1", "m", "1"], [NEWLINE] ["m", "3", "m", "*", "4", "2", "3", "2"], [NEWLINE] ["4", "2", "2", "m", "*", "3", "2", "2"], [NEWLINE] ["2", "3", "4", "m", "5", "5", "4", "m"], [NEWLINE] ["4", "2", "2", "*", "7", "*", "2", "2"], [NEWLINE] ["m", "4", "m", "3", "6", "6", "m", "m"], [NEWLINE] ["3", "5", "m", "4", "*", "6", "4", "m"], [NEWLINE] ["2", "6", "5", "3", "3", "6", "m", "m"] [NEWLINE] ] [NEWLINE]  [NEWLINE] ```\newline 
\hlkey{\textbf{Filtered Output:}}\newline Initial State [NEWLINE] [[s,m,2,2,3,1,m,1], [m,3,m,2,4,2,3,2], [4,2,2,m,3,3,2,2], [2,3,4,m,5,5,4,m], [4,2,2,2,7,2,2,2], [m,4,m,3,6,6,m,m], [3,5,m,4,4,6,4,m], [2,6,5,3,3,6,m,m]] [NEWLINE] Solution [NEWLINE] [[s,m,0,m,3,1,m,1], [m,3,m,0,4,2,3,2], [4,2,2,m,0,3,2,2], [2,3,4,m,5,5,4,m], [4,2,2,0,7,0,2,2], [m,4,m,3,6,6,m,m], [3,5,m,4,0,6,4,m], [2,6,5,3,3,6,m,m]]\newline
\hlkey{\textbf{Perception Correct:}} False\newline
\hlkey{\textbf{Answer Correct:}} False \\ \hline
\hlmodel{\textbf{Qwen2-72B}} \\ 
\hlkey{\textbf{Model Output:}}\newline ```json [NEWLINE]  [NEWLINE] "perception": [ [NEWLINE] ["*", "*", "3", "*", "*", "*", "*", "*"], [NEWLINE] ["s", "4", "3", "*", "*", "*", "3", "2"], [NEWLINE] ["2", "*", "*", "2", "*", "s", "*", "*"], [NEWLINE] ["2", "3", "4", "s", "*", "5", "5", "s"], [NEWLINE] ["1", "s", "*", "*", "7", "*", "*", "*"], [NEWLINE] ["3", "*", "s", "s", "s", "6", "*", "*"], [NEWLINE] ["*", "*", "*", "*", "4", "4", "s", "6"], [NEWLINE] ["4", "s", "1", "*", "*", "*", "*", "2"] [NEWLINE] ], [NEWLINE] "think": [ [NEWLINE] "To solve the mine locations, we need to analyze the numbers and use logic to determine where the mines are located. Here are the steps I will take: [NEWLINE]  [NEWLINE] 1. Look at the cells with numbers and count how many mines are adjacent to them. [NEWLINE] 2. If a cell with a number n has n adjacent cells, those cells must all be mines. [NEWLINE] 3. If a cell with a number n has fewer than n adjacent cells, the remaining cells must be empty. [NEWLINE] 4. If a cell with a number n has more than n adjacent cells, the extra cells must be empty. [NEWLINE]  [NEWLINE] I will apply these rules step by step to solve the grid. [NEWLINE]  [NEWLINE] Step 1: The cell at (0, 3) has a '3' next to it. There are exactly 3 hidden cells around it, so these must be mines. [NEWLINE] Step 2: The cell at (1, 3) has a '4' next to it. There are exactly 4 hidden cells around it, so these must be mines. [NEWLINE] Step 3: The cell at (2, 3) has a '3' next to it. There are exactly 3 hidden cells around it, so these must be mines. [NEWLINE] Step 4: The cell at (3, 3) has a '2' next to it. There are exactly 2 hidden cells around it, so these must be mines. [NEWLINE] Step 5: The cell at (4, 3) has a '5' 

[...TRUNCATED... (2112 characters omitted)...]

n: [NEWLINE]  [NEWLINE] ", [NEWLINE] "answer": [ [NEWLINE] ["e", "e", "3", "e", "e", "e", "e", "e"], [NEWLINE] ["s", "4", "3", "e", "e", "e", "3", "2"], [NEWLINE] ["2", "s", "e", "2", "s", "e", "s", "e"], [NEWLINE] ["2", "3", "4", "s", "s", "s", "s", "s"], [NEWLINE] ["1", "s", "e", "e", "7", "e", "e", "e"], [NEWLINE] ["3", "e", "s", "s", "s", "6", "e", "e"], [NEWLINE] ["e", "e", "e", "e", "4", "4", "s", "6"], [NEWLINE] ["4", "s", "1", "e", "e", "e", "e", "2"] [NEWLINE] ] [NEWLINE]  [NEWLINE] ```\newline 
\hlkey{\textbf{Filtered Output:}}\newline Initial State [NEWLINE] [[0,0,3,0,0,0,0,0], [s,4,3,0,0,0,3,2], [2,0,0,2,0,s,0,0], [2,3,4,s,s,s,s,s], [1,s,0,0,7,0,0,0], [3,0,s,s,s,6,0,0], [0,0,0,0,4,4,s,6], [4,s,1,0,0,0,0,2]] [NEWLINE] Solution [NEWLINE] [[0,0,3,0,0,0,0,0], [s,4,3,0,0,0,3,2], [2,s,0,2,s,0,s,0], [2,3,4,s,s,s,s,s], [1,s,0,0,7,0,0,0], [3,0,s,s,s,6,0,0], [0,0,0,0,4,4,s,6],[4,s,1,0,0,0,0,2]]\newline
\hlkey{\textbf{Perception Correct:}} False\newline
\hlkey{\textbf{Answer Correct:}} False \\ \hline
\hlmodel{\textbf{Qwen2-7B}} \\ 
\hlkey{\textbf{Model Output:}}\newline ```json [NEWLINE]  [NEWLINE]   "perception": [ [NEWLINE]     [3, 3, 3, 3, 2, 2, 1, 1], [NEWLINE]     [2, 2, 2, 2, 1, 1, 1, 1], [NEWLINE]     [1, 1, 1, 1, 1, 1, 1, 1], [NEWLINE]     [1, 1, 1, 1, 1, 1, 1, 1], [NEWLINE]     [1, 1, 1, 1, 1, 1, 1, 1], [NEWLINE]     [1, 1, 1, 1, 1, 1, 1, 2], [NEWLINE]     [1, 1, 1, 1, 1, 1, 1, 2], [NEWLINE]     [1, 1, 1, 1, 1, 2, 2, 2] [NEWLINE]   ], [NEWLINE]   "think": "First, identify the cells with numbers that match the number of surrounding cells. Then, mark those as safe to open. Mark cells with no possible mines as safe. Finally, mark cells with inferred mines as mines.", [NEWLINE]   "answer": [ [NEWLINE]     [2, 2, 2, 2, 1, 1, 1, 1], [NEWLINE]     [2, 2, 2, 2, 1, 1, 1, 1], [NEWLINE]     [1, 1, 1, 1, 1, 1, 1, 1], [NEWLINE]     [1, 1, 1, 1, 1, 1, 1, 1], [NEWLINE]     [1, 1, 1, 1, 1, 1, 1, 1], [NEWLINE]     [1, 1, 1, 1, 1, 1, 1, 1], [NEWLINE]     [1, 1, 1, 1, 1, 1, 1, 1], [NEWLINE]     [1, 1, 1, 1, 2, 2, 2, 1] [NEWLINE]   ] [NEWLINE]  [NEWLINE] ```\newline 
\hlkey{\textbf{Filtered Output:}}\newline Initial State [NEWLINE] [[3,3,3,3,2,2,1,1],[2,2,2,2,1,1,1,1], [1,1,1,1,1,1,1,1], [1,1,1,1,1,1,1,1], [1,1,1,1,1,1,1,1], [1,1,1,1,1,1,1,2], [1,1,1,1,1,1,1,2], [1,1,1,1,1,2,2,2]] [NEWLINE] Solution [NEWLINE] [[2,2,2,2,1,1,1,1],[2,2,2,2,1,1,1,1], [1,1,1,1,1,1,1,1], [1,1,1,1,1,1,1,1], [1,1,1,1,1,1,1,1], [1,1,1,1,1,1,1,1], [1,1,1,1,1,1,1,1], [1,1,1,1,2,2,2,1]]\newline
\hlkey{\textbf{Perception Correct:}} False\newline
\hlkey{\textbf{Answer Correct:}} False \\ \hline
\hlmodel{\textbf{Llama-3.2-90B}} \\ 
\hlkey{\textbf{Model Output:}}\newline **Step 1: Identify the grid dimensions and assigned values** [NEWLINE]  [NEWLINE] The grid is a square layout of rows and columns, sized 10 x 10. Each cell contains a number that represents how many definitive mines are located in adjacent cells to the numbered cell. The values are found in increments of one up to 6. All cell values that have not been assigned a numbered value is left blank.  [NEWLINE]  [NEWLINE] **Step 2: Review the game board** [NEWLINE]  [NEWLINE] Grid row 1 column 7 has assigned number 7, meaning there is a definitive mine in each cell adjacent to this cell. The only attributed number is 4 on the adjacent column, paying attention to the top left grid. This leads the assessment that a mine is locable in gird row 1 column 5 and grid row 1 column 6. Next cell moving top-left is gird row 0 column 5. This cell also points to an adjacent mine by step 1. In rows 0 and 1, the remaining numbered cells are placed for reference; however, they don't represent a mine in this assessment of directional sentences in step 3. On the contrary, the mine placement of row 2 in column 7 is revealed since the number of 3 in column 6 demonstrates the last assigned column possible out of placement.  [NEWLINE]  [NEWLINE] **Step 3: Review the game board continued** [NEWLINE]  [NEWLINE] Similar patterns are followed to determine the mine placement of the neighboring cells. However,  when a hidden or assigned blank cell contains another numbered value, its ends to the adjacent vacant targeted cell to be decisive in the placement of a secondary mine. Repeat this for each possible assigned row to find mine orientation. The assessment is based on the clue given in the question of thi

[...TRUNCATED... (2033 characters omitted)...]

WLINE] [ '2','3','0', '4', '5', '5','0','0','0',''); [NEWLINE] [ '1','7','0','4','6','0','0','6','0,'''2'; [NEWLINE] [ '0','3','1'etatnew],...,'3'], [NEWLINE] [ '7[\$2,4]','0','0','0','1','0','2' [NEWLINE] ``` [NEWLINE] whose inverse comes out as its exact different number ('7-\$2,4]`). Finally, write out the rest by the given method and through the knowing set. Rows beyond analysis cannot be overcome by missing too much information from 2 square ratio parts at the two-dimensional peak change.\newline 
\hlkey{\textbf{Filtered Output:}}\newline Initial State [NEWLINE] [[2,0,0,0,0,0,0,0,1,0], [2,3,0,4,5,5,0,0,0,0], [1,7,0,4,6,0,0,6,0,2], [0,3,1,0,0,0,0,1,0,2]] [NEWLINE] Solution [NEWLINE] [[2,0,0,0,1,0,0,0,1,0], [2,s,0,4,5,5,0,0,0,0], [1,7,0,4,6,0,0,6,0,2], [0,s,1,0,0,0,0,1,0,2]]\newline
\hlkey{\textbf{Perception Correct:}} False\newline
\hlkey{\textbf{Answer Correct:}} False \\ \hline
\hlmodel{\textbf{Llama-3.2-11B}} \\ 
\hlkey{\textbf{Model Output:}}\newline \#\#\# Perception [NEWLINE]  [NEWLINE] The game state is a 5x5 grid consisting of two known mines ("s"), and a hint of 7 mines that are uncertain. [NEWLINE]  [NEWLINE] | |  |  |  |  | [NEWLINE] | --- | --- | --- | --- | --- | [NEWLINE] | 1 | 1 | e | e | e | [NEWLINE] | e | s | 3 | 6 | e | [NEWLINE] | 5 | 5 | e | 6 | 3 | [NEWLINE] | e | 3 | 4 | 7 | 4 | [NEWLINE] | 4 | 4 | **edge** | e | @ | [NEWLINE]  [NEWLINE] ```json [NEWLINE]  [NEWLINE]   "perception": [ [NEWLINE]     ["e", "e", "e", "e", "e"], [NEWLINE]     ["e", "s", "3", "6", "e"], [NEWLINE]     ["5", "5", "e", "6", "3"], [NEWLINE]     ["e", "3", "4", "7", "4"], [NEWLINE]     ["4", "4", "edge", "e", "@"] [NEWLINE]   ] [NEWLINE]  [NEWLINE] ``` [NEWLINE]  [NEWLINE] \#\#\# Uncheduled Thoughts [NEWLINE]  [NEWLINE] The goal is to locate mines based on given hints. The hint in cell 7 reveals what I already know at index [3][3]. [NEWLINE]  [NEWLINE] First, the cell at index [1][2] tells me that I have 4 mines adjacent to it, including diagonal mines. Since all the other boxes around it except the unknown 3's already have mines, that means that all of them are mines except for the one at [1][3]. We can formalize that as: [NEWLINE]  [NEWLINE] 5 | 5 | -3 | [NEWLINE] ... [NEWLINE] | 3 | 4 |-7 | [NEWLINE]  [NEWLINE] Now, we know that the cell adjacent to the box at index [2][1] (which has -3's) must also have a mine. [NEWLINE]  [NEWLINE] At this point, the only position that doesn't have a prime neighboring position filled or at edge is 8. If I have one mine at positions [1][1], [2][2], or [4][4], then all the other mines will be at the border of the cube. [NEWLINE]  [NEWLINE] The last step is to insert a full circle in the 3's to indicate that it is a mine. Otherwise, it would be impossible to call the corresponding border positions.  [NEWLINE]  [NEWLINE] \#\# Answer [NEWLINE]  [NEWLINE] ```json [NEWLINE] "answer": [ [NEWLINE]   ["e", "e", "e", "e", "e"], [NEWLINE]   ["e", "s", "3", "s", "e"], [NEWLINE]   ["s", "s", "e", "s", "s"], [NEWLINE]   ["e", "s", "4", "s", "s"], [NEWLINE]   ["s", "s", "e", "e", "s"] [NEWLINE] ] [NEWLINE] ```\newline 
\hlkey{\textbf{Filtered Output:}}\newline Initial State [NEWLINE] [[e,e,e,e,e], [e,s,3,s,e], [s,s,e,s,s], [e,s,4,s,s], [s,s,e,e,s]] [NEWLINE] Solution [NEWLINE] [[0,0,0,0,0], [0,s,3,s,0], [s,s,0,s,s], [0,s,4,s,s], [s,s,0,0,s]]\newline
\hlkey{\textbf{Perception Correct:}} False\newline
\hlkey{\textbf{Answer Correct:}} False \\ \hline
\hlmodel{\textbf{LLaVA-Qwen2-7B}} \\ 
\hlkey{\textbf{Model Output:}}\newline ```json [NEWLINE]  [NEWLINE]     "perception": [ [NEWLINE]         [4,4,2,1,1,3,2], [NEWLINE]         [2,3,0,0,0,0,2], [NEWLINE]         [5,5,0,1,0,1,1], [NEWLINE]         [9,1,6,0,0,6,2], [NEWLINE]         [3,0,0,0,0,0,7], [NEWLINE]         [4,3,2,3,4,6,1], [NEWLINE]         [6,0,0,0,1,0,2] [NEWLINE]     ], [NEWLINE]     "think": "I need to place three mine markers on the grid based on the revealed numbers. Starting with the top-left corner where the number 4 indicates there should be two mines placed in the right row and four cells down. I notice a 2 and a 3 nearby so I place two and three mine markers next. Then I look at the 1 near the middle row and place a mine marker next to it leaving one mine to place. I then assess the bottom row and place the remaining mine there. I now have three mine markers placed, and I can confirm the solution", [NEWLINE]     "answer": [ [NEWLINE]         [4,4,2,1,1,3,2], [NEWLINE]         [2,3,0,0,0,0,2], [NEWLINE]         [5,5,0,1,0,1,1], [NEWLINE]         [9,1,6,0,0,6,2], [NEWLINE]         [3,0,0,0,0,0,7], [NEWLINE]         [4,3,2,3,4,6,1], [NEWLINE]         [6,0,0,0,1,0,2] [NEWLINE]     ] [NEWLINE]  [NEWLINE] ```\newline 
\hlkey{\textbf{Filtered Output:}}\newline Initial State [NEWLINE] [[4,4,2,1,1,3,2], [2,3,0,0,0,0,2], [5,5,0,1,0,1,1], [9,1,6,0,0,6,2], [3,0,0,0,0,0,7], [4,3,2,3,4,6,1], [6,0,0,0,1,0,2]] [NEWLINE] Solution [NEWLINE] [[4,4,2,1,1,3,2], [2,3,0,0,0,0,2], [5,5,0,1,0,1,1], [9,1,6,0,0,6,2], [3,0,0,0,0,0,7], [4,3,2,3,4,6,1], [6,0,0,0,1,0,2]]\newline
\hlkey{\textbf{Perception Correct:}} False\newline
\hlkey{\textbf{Answer Correct:}} False \\ \hline
\hlmodel{\textbf{LLaVA-Mistral-7B}} \\ 
\hlkey{\textbf{Model Output:}}\newline ```json [NEWLINE]  [NEWLINE] "perception": [ [NEWLINE]   [1, 2, 3, 4, 5], [NEWLINE]   [6, 7, 8, 9, 10], [NEWLINE]   [11, 12, 13, 14, 15], [NEWLINE]   [16, 17, 18, 19, 20], [NEWLINE]   [21, 22, 23, 24, 25] [NEWLINE] ], [NEWLINE] "think": [ [NEWLINE]   "The central 2x2 block has five numbers around it. Four of them are revealed, so the block does not contain a mine. 16, 17, and 18 are the hidden cells. Now we need to look at surrounding 1x1 cells:  [NEWLINE]    [NEWLINE]   16 left of 17 has a number 1, which means there are no mines in that 1x1 cell and the one above it. The remaining two cells to the top left of 16 have a 7 on the right→ diagonal, so one of those must be the bomb. They can't have a number 7 and bones together, so it's one of those two. I'll leave those two cells for now. [NEWLINE]    [NEWLINE]   21 is adjacent to 16 with 1, 24 with 7, 20, 24 which have used the total of 8 from these cells. One of 16, 17, and 18 contains a bomb. From 16, the right side can't be solved yet, as they have 6 on it. The bottom row has 6 and 19 = 12, so there are no mines between 16 and 17. 16 doesn't have any mines except for the one on the diagonal cell.  [NEWLINE]    [NEWLINE]   11, however, has a mine to the west of it, since it's an even-numbered cell with the same odd-numbered relationship to the bomb cell as 17 to 18.  [NEWLINE]    [NEWLINE]   24 = 12 because the right side doesn't have a number 1, so there is a mine above or below the 7. There's one bomb hidden to the right of it. The left side only has a 2, which means it's safe. The cell on the top left has a 3, so I will leave it for now.  [NEWLINE]    [NEWLINE]   20 has a 4, which means there must be another hidden mi

[...TRUNCATED... (1040 characters omitted)...]

t to 17 has a 5, we can also conclude that there is a mine to the left of it. The cells to the left of 25 have 4, safe, \& 1. The cell at 6, 6 has a number 2, so there is a mine in the cell hidden there.  [NEWLINE]    [NEWLINE]   The 2nd step of logic is to look at 6 and 8, just to make sure: [NEWLINE]     [NEWLINE]    If 6 is a mine, it would mean there is another mine below it or above it, but since I've marked the cell, there must be a mine to the left of 6.  [NEWLINE]     [NEWLINE]    Now I\newline 
\hlkey{\textbf{Filtered Output:}}\newline Initial State [NEWLINE] [[1,2,3,4,5], [6,7,8,9,10], [11,12,13,14,15], [16,17,18,19,20], [21,22,23,24,25]] [NEWLINE] Solution [NEWLINE] [[1,2,3,4,5],[6,s,s,9,10], [11,s,s,14,15], [16,s,s,19,20], [21,22,23,24,25]]\newline
\hlkey{\textbf{Perception Correct:}} False\newline
\hlkey{\textbf{Answer Correct:}} False \\ \hline\end{longtable}
} %

\onecolumn
\renewcommand{\arraystretch}{1.2} %

\centering
{\ttfamily %
\begin{longtable}{|p{0.95\textwidth}|}
\hline
\textbf{Llama 3.2 Instruction Vision after Reasoning-SFT (Successful Example)} \\ \hline
\endhead

\hline \multicolumn{1}{|r|}{{Continued on next page}} \\ \hline
\endfoot

\hline
\endlastfoot

\label{supp-RL}
\refstepcounter{section} 
\phantomsection         
\addcontentsline{toc}{section}{Llama 3.2 Instruction Vision after Reasoning-SFT (Successful Example)}

\includegraphics[width=0.2\columnwidth]{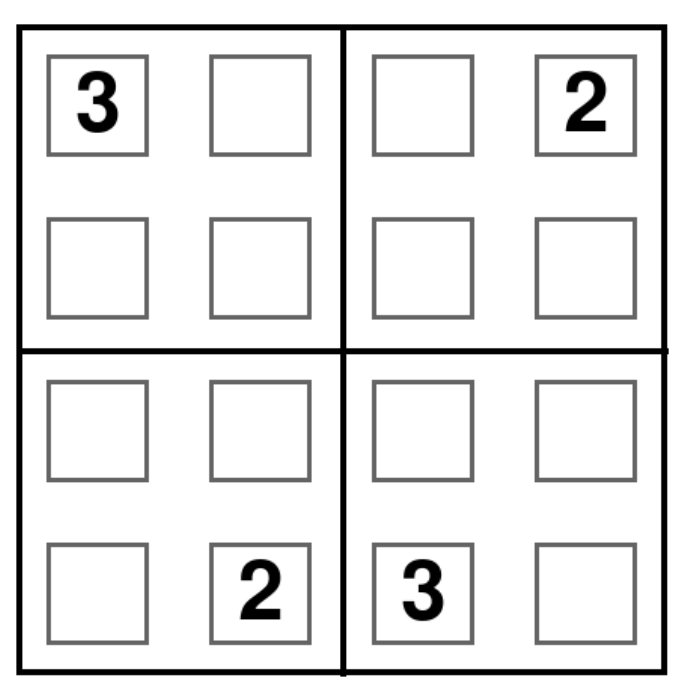}  
\\
\hline

\hlmodel{\textbf{R-SFT}} \\ 
\hlkey{\textbf{Model Output:}}\newline ```json [NEWLINE] Initial State: [[3, *, *, 2], [*, *, *, *], [*, *, *, *], [*, 2, 3, *]],  Thought: Initial State: [[3, *, *, 2], [*, *, *, *], [*, *, *, *], [*, 2, 3, *]] Initial possible numbers for empty cells: Cell (0, 1): 1, 4 Cell (0, 2): 1, 4 Cell (1, 1): 1, 4 Cell (1, 2): 1, 4 Cell (2, 0): 1, 4 Cell (2, 3): 1, 4 Cell (3, 0): 1, 4 Cell (3, 3): 1, 4 Cell (1, 0): 1, 2, 4 Cell (1, 3): 1, 3, 4 Cell (2, 1): 1, 3, 4 Cell (2, 2): 1, 2, 4 \hlgreen{Step 1: Placing 1 at (0, 1). This cell had 2 possible values (4 were alternatives)} Resulting State: [[3, 1, *, 2], [*, *, *, *], [*, *, *, *], [*, 2, 3, *]] Possible numbers for remaining empty cells: Cell (0, 2): 4 Cell (1, 1): 4 Cell (1, 0): 2, 4 Cell (1, 2): 1, 4 Cell (2, 0): 1, 4 Cell (2, 1): 3, 4 Cell (2, 3): 1, 4 Cell (3, 0): 1, 4 Cell (3, 3): 1, 4 Cell (1, 3): 1, 3, 4 Cell (2, 2): 1, 2, 4 \hlgreen{Step 2: Placing 4 at (0, 2). This cell had 1 possible values Resulting State: [[3, 1, 4, 2], [*, *, *, *], [*, *, *, *], [*, 2, 3, *]]} Possible numbers for remaining empty cells: Cell (1, 1): 4 Cell (1, 2): 1 Cell (1, 0): 2, 4 Cell (1, 3): 1, 3 Cell (2, 0): 1, 4 Cell (2, 1): 3, 4 Cell (2, 2): 1, 2 Cell (2, 3): 1, 4 Cell (3, 0): 1, 4 Cell (3, 3): 1, 4 Step 3: Placing 4 at (1, 1). This cell had 1 possible values Resulting State: [[3, 1, 4, 2], \newline
[...TRUNCATED...] \newline
Possible numbers for remaining empty cells: Cell (2, 0): 1, 4 Cell (2, 3): 1, 4 Cell (3, 0): 1, 4 Cell (3, 3): 1, 4 Step 9: Placing 1 at (2, 0). This cell had 2 possible values (4 were alternatives) Resulting State: [[3, 1, 4, 2], [2, 4, 1, 3], [1, 3, 2, *], [*, 2, 3, *]] Possible numbers for remaining empty cells: Cell (2, 3): 4 Cell (3, 0): 4 Cell (3, 3): 1, 4 Step 10: Placing 4 at (2, 3). This cell had 1 possible values Resulting State: [[3, 1, 4, 2], [2, 4, 1, 3], [1, 3, 2, 4], [*, 2, 3, *]] Possible numbers for remaining empty cells: Cell (3, 0): 4 Cell (3, 3): 1 Step 11: Placing 4 at (3, 0). This cell had 1 possible values Resulting State: [[3, 1, 4, 2], [2, 4, 1, 3], [1, 3, 2, 4], [4, 2, 3, *]] Possible numbers for remaining empty cells: Cell (3, 3): 1 Step 12: Placing 1 at (3, 3). This cell had 1 possible values Resulting State: [[3, 1, 4, 2], [2, 4, 1, 3], [1, 3, 2, 4], [4, 2, 3, 1]],  \hlgreen{Solution: [[3, 1, 4, 2], [2, 4, 1, 3], [1, 3, 2, 4], [4, 2, 3, 1]]} \newline \\
\hlkey{\textbf{Perception Correct:}} True\newline
\hlkey{\textbf{Answer Correct:}} True \\ 
\hline
\hline\end{longtable}
} %

\onecolumn
\renewcommand{\arraystretch}{1.2} %

\centering
{\ttfamily %
\begin{longtable}{|p{0.95\textwidth}|}
\hline
\textbf{Llama 3.2 Instruction Vision after Reasoning-SFT (Failure Example)} \\ \hline
\endhead

\hline \multicolumn{1}{|r|}{{Continued on next page}} \\ \hline
\endfoot

\hline
\endlastfoot

\label{supp-RL}
\refstepcounter{section} 
\phantomsection         
\addcontentsline{toc}{section}{Llama 3.2 Instruction Vision after Reasoning-SFT (Failure Example)}

\includegraphics[width=0.2\columnwidth]{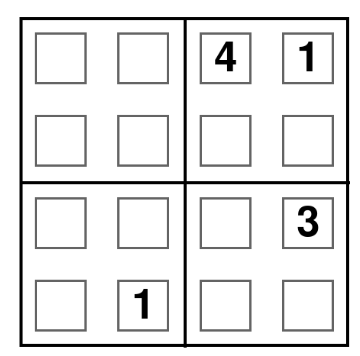}  
\\
\hline

\hlmodel{\textbf{R-SFT}} \\ 
\hlkey{\textbf{Model Output:}}\newline ```json{  [NEWLINE] "Initial State": \hlgreen{"[[*, *, 4, 1], [*, *, *, *], [*, *, *, 3], [*, 1, *, *]]",  "Thought": "Initial State:[[*, *, 4, 1], [*, *, *, *], [*, *, *, 3], [*, 1, *, *]]}Initial possible numbers for empty cells:Cell (1, 3): 2Cell (3, 2): 2Cell (0, 0): 2, 3Cell (0, 1): 2, 3Cell (1, 2): 1, 2, 3Cell (1, 1): 2, 3, 4Cell (2, 0): 2, 4Cell (2, 1): 2, 4Cell (2, 2): 1, 2Cell (3, 3): 2, 4Cell (3, 0): 2, 3, 4Cell (1, 0): 1, 2, 3, 4\hlgreen{Step 1: Placing 2 at (1, 3). This cell had 1 possible values [NEWLINE] Resulting State:[[*, *, 4, 1], [*, *, *, 2], [*, *, *, 3], [*, 1, *, *]]} Possible numbers for remaining empty cells:Cell (1, 2): 1, 3Cell (3, 2): 2Cell (3, 3): 4Cell (0, 0): 2, 3Cell (0, 1): 2, 3Cell (1, 1): 3, 4Cell (2, 0): 2, 4Cell (2, 1): 2, 4Cell (2, 2): 1, 2Cell (3, 0): 2, 3, 4Cell (1, 0): 1, 3, 4\hlred{Step 2: Placing 1 at (1, 2). This cell had 2 possible values (3 were alternatives)}Resulting State:[[*, *, 4, 1], [*, *, 1, 2], [*, *, *, 3], [*, 1, *, *]]Possible numbers for remaining empty cells:Cell (3, 2): 2Cell (3, 3): 4Cell (0, 0): 2, 3Cell (0, 1): 2\newline
[...TRUNCATED...]
\newline
for remaining empty cells:Cell (1, 0): 4Cell (1, 1): 4Cell (2, 0): 4Cell (2, 1): 2, 4Step 9: Placing 4 at (1, 0). This cell had 1 possible valuesResulting State:[[2, 3, 4, 1], [4, *, 1, 2], [*, *, 1, 3], [3, 1, 2, 4]]Possible numbers for remaining empty cells:Cell (1, 1): 4Cell (2, 0): 4Cell (2, 1): 2, 4Step 10: Placing 4 at (1, 1). This cell had 1 possible valuesResulting State:[[2, 3, 4, 1], [4, 4, 1, 2], [*, *, 1, 3], [3, 1, 2, 4]]Possible numbers for remaining empty cells:Cell (2, 0): 4Cell (2, 1): 2Step 11: Placing 4 at (2, 0). This cell had 1 possible valuesResulting State:[[2, 3, 4, 1], [4, 4, 1, 2], [4, *, 1, 3], [3, 1, 2, 4]]Possible numbers for remaining empty cells:Cell (2, 1): 2Step 12: Placing 2 at (2, 1). This cell had 1 possible valuesResulting State:[[2, 3, 4, 1], [4, 4, 1, 2], [4, 2, 1, 3], [3, 1, 2, 4]]",  "Solution": "[[2, 3, 4, 1], \hlred{[4, 4}, 1, 2], [4, 2, 1, 3], [3, 1, 2, 4]]"}```']
\newline \\
\hlkey{\textbf{Perception Correct:}} True\newline
\hlkey{\textbf{Answer Correct:}} False \\ 
\hlgray{\textbf{Remarks:}} Step 1 is correct and effective, starting with the cell that could be uniquely decided, but step 2 made a wrong statement as cell (1,2) could be 4. \\
\hline
\hline\end{longtable}
} %

\onecolumn
\renewcommand{\arraystretch}{1.2} %

\centering
{\ttfamily %
\begin{longtable}{|p{0.95\textwidth}|}
\hline
\textbf{Reinforcement Learning Training Process with Text Input} \\ \hline
\endhead

\hline \multicolumn{1}{|r|}{{Continued on next page}} \\ \hline
\endfoot

\hline
\endlastfoot

\label{supp-RL}
\refstepcounter{section} 
\phantomsection         
\addcontentsline{toc}{section}{Reinforcement Learning Training Process with Text Input}

\includegraphics[width=0.6\columnwidth]{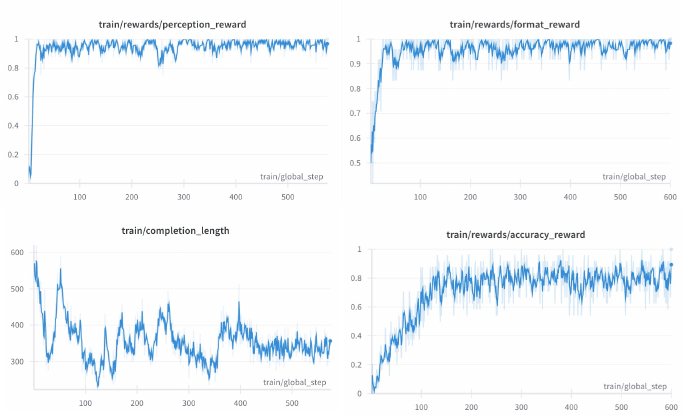}  

{\hlgray{Remarks:}} perception and format reward increase first, then accuracy reward increase slowly afterwards. Indicating solving a puzzle is more challenging than output certain format and copy the perception from the text prompt. \\ \hline

\hlkey{\textbf{Before RL:}}\newline 

\includegraphics[width=0.3\columnwidth]{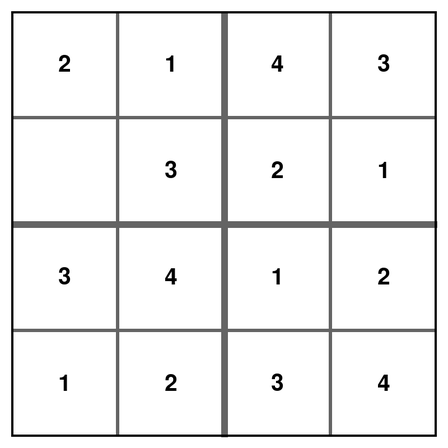}  

<think>
First, let's look at the first row and first column. The only number that can go in the top left cell is 1, since it's the only number not already present in the top row or column.
Now let's look at the top row. \hlred{The number 1 is already in the top left cell}, so the only possible numbers for the remaining cells are 2 and 3. Since 2 is already in the second column, the number 2 must go in the top middle cell, and the number 3 must go in the top right cell.
Now let's look at the first column. The number 1 is already in the top left cell, so the only possible numbers for the remaining cells are 2, 3, and 4. Since 3 is already in the top middle cell, the number 3 must go in the middle left cell, and the number 4 must go in the bottom left cell.
Finally, let's look at the middle right cell. Since the only possible number for this cell is 1, we can place it here.
Now let's look at the last row. The number 1 is already in the middle right cell, so the only possible numbers for the remaining cells are 2, 3, and 4. Since 2 is already in the bottom left cell, the number 2 must go in the bottom middle cell, and the number 4 must go in the bottom right cell.
So the final solution is:
[[1, 2, 3, 4], [2, *, 4, 1], [3, 4, 1, *], [4, *, 2, 3]]
</think>
<answer>
[[1, 2, 3, 4], [2, *, 4, 1], [3, 4, 1, *], [4, *, 2, 3]]
</answer>
\newline
\hlgray{accuracy\_reward:} 0.00 \newline
\hlgray{format\_reward:} 1.00 \newline
\hlgray{perception\_reward:} 0.00 \newline

\\
\hline

\hlkey{\textbf{After RL:}}\newline 

\includegraphics[width=0.3\columnwidth]{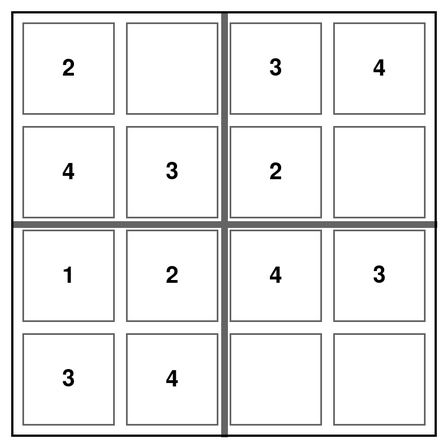}  

<perception>
[[2, 4, *, 1], [3, *, 2, 4], [*, *, 1, 2], [1, 2, 4, *]]
</perception>
<think>
Since the numbers are 1 to 4, and the goal is to fill in the missing numbers without repeating any number in each row, column, and 2x2 subgrid, we can start by filling in the numbers one by one.
\hlgreen{For the first row, we know that the numbers 2, 4, and 1 have already been used}, so the only number that can go in the last square is 3.
Now, looking at the second row, we know that the numbers 1, 3, and 4 have already been used in the first column, so the only number that can go in the first square is 2. We also know that the number 2 has already been used in the first row, so the only number that can go in the last square is 1.
Now, looking at the third row, we know that the numbers 1, 2, and 4 have already been used in the second column, so the only number that can go in the first square is 3. We also know that the number 3 has already been used in the second row, so the only number that can go in the last square is 2.
Lastly, looking at the fourth row, we know that the numbers 1 and 3 have already been used in the fourth column, so the only number that can go in the first square is 2. We also know that the number 2 has already been used in the fourth row, so the only number that can go in the last square is 4.
Finally, looking at the third column, we know that the numbers 3 and 4 have already been used in the third row, so the only number that can go in the first square is 1. We also know that the number 1 has already been used in the first row, so the only number that can go in the last square is 4.
So, the filled-in board looks like this:
[[2, 4, 3, 1], [3, 1, 2, 4], [4, 3, 1, 2], [1, 2, 4, 3]]
</think>
<answer>
[[2, 4, 3, 1], [3, 1, 2, 4], [4, 3, 1, 2], [1, 2, 4, 3]]
</answer>
\newline
\hlgray{accuracy\_reward: 1.00} \newline
\hlgray{format\_reward: 1.00} \newline
\hlgray{perception\_reward: 1.00}
\\ \hline
\end{longtable}
} %

\end{document}